\documentclass{article}
\usepackage[numbers,sort&compress,square]{natbib}

\usepackage{neurips_data_2023}

\nolinenumbers

\usepackage[utf8]{inputenc} % allow utf-8 input
\usepackage[T1]{fontenc}    % use 8-bit T1 fonts
\usepackage{xcolor}
\definecolor{citecolor}{HTML}{0071bc}
\definecolor{citered}{HTML}{8b0000}
\usepackage[pagebackref=true,breaklinks=true,colorlinks,citecolor=citecolor,linkcolor=citered,urlcolor=citered,bookmarks=false]{hyperref}
\usepackage{url}            % simple URL typesetting
\usepackage{booktabs}       % professional-quality tables
\usepackage{amsfonts}       % blackboard math symbols
\usepackage{nicefrac}       % compact symbols for 1/2, etc.
\usepackage{microtype}      % microtypography

\usepackage{graphicx}
\usepackage{float}
\usepackage{amsfonts,amsmath,amssymb,amsthm}
\usepackage{cleveref}
\usepackage{multicol, multirow}
\usepackage{makecell}
\usepackage{subfigure}
\usepackage{adjustbox}
\usepackage{enumitem}
\usepackage{tablefootnote}
\usepackage{wrapfig}

\usepackage{listings}
\definecolor{codegreen}{rgb}{0,0.6,0}
\definecolor{codegray}{rgb}{0.5,0.5,0.5}
\definecolor{codepurple}{rgb}{0.58,0,0.82}
\definecolor{backcolour}{rgb}{0.95,0.95,0.92}
\lstdefinestyle{mystyle}{
    backgroundcolor=\color{backcolour},   
    commentstyle=\color{codegreen},
    keywordstyle=\color{magenta},
    numberstyle=\tiny\color{codegray},
    stringstyle=\color{codepurple},
    basicstyle=\ttfamily\footnotesize,
    breakatwhitespace=false,         
    breaklines=true,                 
    captionpos=b,                    
    keepspaces=true,                 
    numbers=left,                    
    numbersep=5pt,                  
    showspaces=false,                
    showstringspaces=false,
    showtabs=false,                  
    tabsize=2
}
\makeatletter
\newcommand\tabcaption{\def\@captype{table}\caption}
\newcommand\figcaption{\def\@captype{figure}\caption}

\usepackage{minitoc}

\renewcommand \thepart{}
\renewcommand \partname{}

\usepackage{amsmath,amsfonts,bm}

\def\eqref#1{equation~\ref{#1}}
\def\1{\bm{1}}

\def\va{{\bm{a}}}
\def\vb{{\bm{b}}}
\def\vc{{\bm{c}}}

\def\ve{{\bm{e}}}

\def\vg{{\bm{g}}}
\def\vh{{\bm{h}}}

\def\vk{{\bm{k}}}

\def\vm{{\bm{m}}}

\def\vq{{\bm{q}}}
\def\vr{{\bm{r}}}

\def\vv{{\bm{v}}}

\def\vx{{\bm{x}}}

\def\mR{{\bm{R}}}

\def\mX{{\bm{X}}}

\DeclareMathAlphabet{\mathsfit}{\encodingdefault}{\sfdefault}{m}{sl}
\SetMathAlphabet{\mathsfit}{bold}{\encodingdefault}{\sfdefault}{bx}{n}

\makeatletter

\newcommand\xleftrightarrow[2][]{%
  \ext@arrow 9999{\longleftrightarrowfill@}{#1}{#2}}
\newcommand\longleftrightarrowfill@{%
  \arrowfill@\leftarrow\relbar\rightarrow}
\makeatother

\newtheorem{definition}{Definition}

\newcommand{\ie}{\em{i.e.}}
\newcommand{\eg}{\em{e.g.}}
\usepackage{cleveref}
\Crefname{section}{Sec.}{Secs.}
\Crefname{equation}{Eq.}{Eqs.}
\Crefname{figure}{Fig.}{Figs.}
\Crefname{tabular}{Tab.}{Tabs.}

\usepackage{xcolor}
\newcommand{\TotalMethodNum}[1]{16}
\newcommand{\TotalPretrainingMethodNum}[1]{14}
\newcommand{\TotalTaskNum}[1]{46}
\newcommand{\framework}[1]{Geom3D}

\definecolor{ForestGreen}{RGB}{34,139,34}
\definecolor{ForceGreen}{RGB}{31,102,8}

\newcommand{\norm}[1]{\left\lVert#1\right\rVert}

\title{
Symmetry-Informed Geometric Representation for Molecules, Proteins, and Crystalline Materials
}

\author{%
    \fontsize{9}{9}\selectfont
    Shengchao Liu\textsuperscript{1,2},~~~
    Weitao Du\textsuperscript{3},~~~
    Yanjing Li\textsuperscript{4},~~~
    Zhuoxinran Li\textsuperscript{5},~~~
    Zhiling Zheng\textsuperscript{6},~~~
    Chenru Duan\textsuperscript{7},\\
    \fontsize{9}{9}\selectfont
    \textbf{Zhiming Ma\textsuperscript{3}},~~~
    \textbf{Omar Yaghi\textsuperscript{6},}~~~
    \textbf{Anima Anandkumar\textsuperscript{8},}~~~
    \textbf{Christian Borgs\textsuperscript{6},}\\
    \fontsize{9}{9}\selectfont
    \textbf{Jennifer Chayes\textsuperscript{6},}~~~
    \textbf{Hongyu Guo\textsuperscript{9,10},}~~~
    \textbf{Jian Tang\textsuperscript{1,11,12}}\\
\fontsize{9}{9}\selectfont
\textsuperscript{1}Mila - Québec Artificial Intelligence Institute~~~
\textsuperscript{2}Université de Montréal~~~
\\
\fontsize{9}{9}\selectfont
\textsuperscript{3}University of Chinese Academy of Sciences~~~
\textsuperscript{4}Carnegie Mellon University~~~
\\
\fontsize{9}{9}\selectfont
\textsuperscript{5}University of Toronto~~~
\textsuperscript{6}University of California, Berkeley~~~
\textsuperscript{7}Massachusetts Institute of Technology~~~
\\
\fontsize{9}{9}\selectfont
\textsuperscript{8}California Institute of Technology~~~~
\textsuperscript{9}National Research Council Canada~~~~
\textsuperscript{10}University of Ottawa~~~~
\\
\fontsize{9}{9}\selectfont
\textsuperscript{11}HEC Montréal~~~~
\textsuperscript{12}CIFAR AI Chair
}

\begin{document}
\doparttoc % Tell to minitoc to generate a toc for the parts
\faketableofcontents
\maketitle

\begin{abstract}
Artificial intelligence for scientific discovery has recently generated significant interest within the machine learning and scientific communities, particularly in the domains of chemistry, biology, and material discovery. For these scientific problems, molecules serve as the fundamental building blocks, and machine learning has emerged as a highly effective and powerful tool for modeling their geometric structures. Nevertheless, due to the rapidly evolving process of the field and the knowledge gap between science ({\eg}, physics,  chemistry, \& biology) and machine learning communities, a benchmarking study on geometrical representation for such data has not been conducted. To address such an issue, in this paper, we first provide a unified view of the current symmetry-informed geometric methods, classifying them into three main categories: invariance, equivariance with spherical frame basis, and equivariance with vector frame basis. Then we propose a platform, coined Geom3D, which enables benchmarking the effectiveness of geometric strategies. Geom3D contains 16 advanced symmetry-informed geometric representation models and 14 geometric pretraining methods over 46 diverse datasets, including small molecules, proteins, and crystalline materials. We hope that Geom3D can, on the one hand, eliminate barriers for machine learning researchers interested in exploring scientific problems; and, on the other hand, provide valuable guidance for researchers in computational chemistry, structural biology, and materials science, aiding in the informed selection of representation techniques for specific applications.
The source code is available on \href{https://github.com/chao1224/Geom3D}{the GitHub repository}.\looseness=-1
\end{abstract}

\vspace{-1ex}
\section{Introduction}
\vspace{-1ex}
Artificial intelligence (AI) for molecule discovery has recently seen many developments, including small molecular property prediction~\cite{duvenaud2015convolutional,kipf2016semi,velivckovic2017graph,gilmer2017neural,liu2019n,xu2018powerful,corso2020principal,yang2019analyzing,demirel2022attentive,yang2019analyzing,ying2021transformers,rampasek2022GPS,rao2020transformer,rao2021msa}, small molecule design and optimization~\cite{zang2020moflow,brown2019guacamol,jensen2019graph,liu2023graphcg,isert2023structure}, small molecule reaction and retrosynthesis~\cite{shi2020graph,gottipati2020learning,sun2020energy}, protein property prediction~\cite{9477085,zhang2022ontoprotein}, protein folding and inverse folding~\cite{jumper2021highly,meier2021language,hsu2022learning}, protein design~\cite{madani2020progen,dauparas2022robust,gruver2023protein,liu2023text,hie2023efficient}, and crystalline material design~\cite{xie2021crystal,wang2022deep,flam2023language}. One of the most fundamental building blocks for these tasks is the geometric structure of molecules. Exploring effective methods for robust representation learning to leverage such geometric information fully remains an open challenge that interests both machine learning (ML) and science researchers.

To this end, symmetry-informed geometric representation~\cite{atz2021geometric} has emerged as a promising approach. By leveraging physical principles ({\ie}, group theory for depicting symmetric particles) into spatial representation, they facilitate a more robust representation of small molecules, proteins, and crystalline materials. Nevertheless, pursuing geometric learning research is still challenging due to its evolving nature and the knowledge gap between science ({\eg}, physics) and machine learning communities. These factors contribute to a substantial barrier for machine learning researchers to investigate scientific problems and hinder efforts to reproduce results consistently. To overcome this, we introduce \framework{}, a benchmarking of the geometric representation with four advantages, as follows. \footnote{\small In what follows, we may use ``molecule'' to refer to ``small molecule'' for brevity.}\looseness=-1

\begin{figure}[tb!]
\centering
\includegraphics[width=\textwidth]{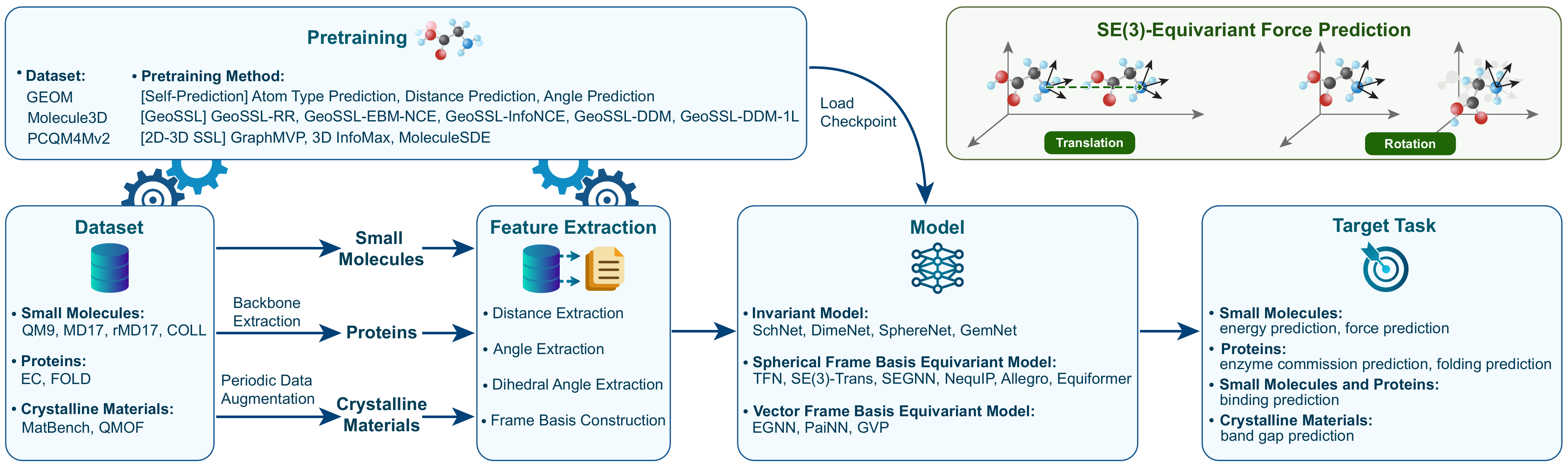}
\vspace{-4ex}
\caption{
\small
Pipeline for \framework{}, including dataset preprocessing, feature extraction, geometric pretraining and representation, and target tasks. We additionally demonstrate the \textcolor{ForceGreen}{\textbf{SE(3)-equivariant force prediction task}}.
}
\label{fig:pipeline}
\vspace{-3.7ex}
\end{figure}

\begin{wrapfigure}[18]{r}{0.5\textwidth}
\vspace{-3ex}
\centering
\includegraphics[width=\linewidth]{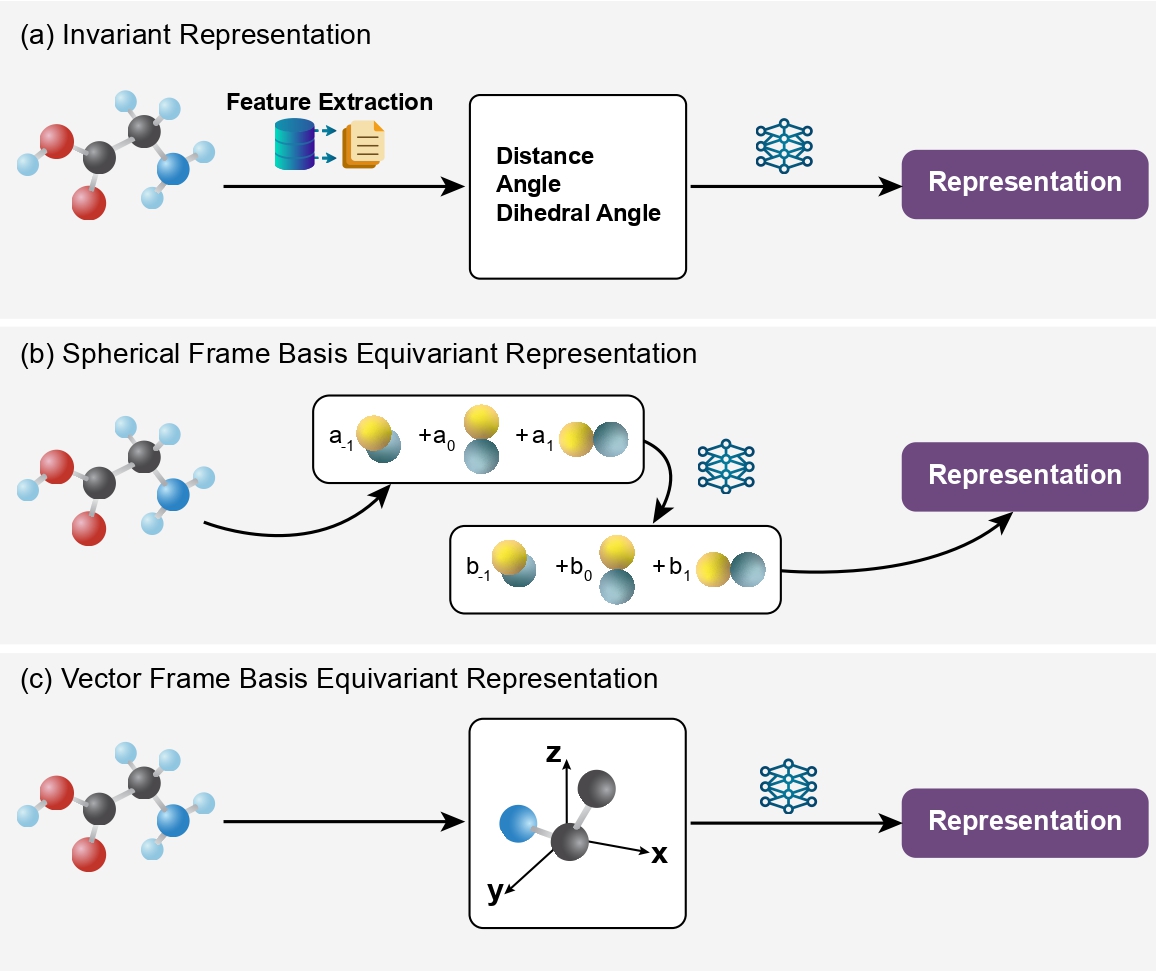}
\vspace{-3.5ex}
\caption{\small Three categories of geometric modules. (a) Invariant models only consider type-0 features. Equivariant models use either (b) spherical harmonics frames or (c) vector frames by projecting the coordinate vectors.} \label{fig:geometric_representation}
\end{wrapfigure}
\textbf{(1) A unified and novel aspect in understanding symmetry-informed geometric models.} The molecule geometry needs to satisfy certain physical constraints regarding the 3D Euclidean space. For instance, the molecules' force needs to be equivariant to translation and rotation (see SE(3)-equivariance in~\Cref{fig:pipeline}). In this work, we classify the geometric methods into three categories: \textit{invariant} model, SE(3)-equivariant model with \textit{spherical frame basis} and \textit{vector frame basis}. The invariant models only consider features that are constant w.r.t. the SE(3) group, while the two families of equivariant models can be further unified  using the \textit{frame basis} to capture equivariant symmetry. An illustration of three categories is in~\Cref{fig:geometric_representation}. Building equivariant models on the \textit{frame basis} provides a novel and unified view of understanding geometric models and paves the way for intriguing more ML researchers to explore scientific problems.

\textbf{(2) A unified platform for various scientific domains.} There exist multiple platforms and tools for molecule discovery, but they are (1) mainly focusing on molecule's 2D graph representation~\cite{zhu2022torchdrug,Ramsundar-et-al-2019,JMLR:v22:21-0343}; (2) using 3D geometry with customized data structures or APIs~\cite{batzner20223,REISER2021100095}; or (3) covering only a few geometric models~\cite{liu2021dig}. Thus, it is necessary to have a platform benchmarking the geometric models, especially for researchers interested in solving scientific problems. In this work, we propose \framework{}, a geometric modeling framework based on PyTorch Geometric (PyG)~\cite{Fey/Lenssen/2019}, one of the most widely-used platforms for graph representation learning. \framework{} benchmarks \TotalMethodNum{} geometric models on solving \TotalTaskNum{} scientific tasks, and these tasks include the three most fundamental molecule types: small molecules, proteins, and crystalline materials. Each of them requires distinct domain-specific preprocessing steps, {\eg}, crystalline materials molecules possess periodic structures and thus need a particular periodic data augmentation. By leveraging such a unified framework, \framework{} serves as a comprehensive benchmarking tool, facilitating effective and consistent analysis components to interpret the existing geometric representation functions in a fair and convenient comparison setting.

\textbf{(3) A framework for a wider range of ML tasks.} The geometric models in \framework{} can serve as a building block for exploring extensive ML tasks, including but not limited to studying the molecule dynamic simulation and scrutinizing the transfer learning effect on molecule geometry. For example, pretraining is an important strategy to quickly transfer knowledge to target tasks, and recent works explore geometric pretraining on 3D conformations (including supervised and self-supervised)~\cite{liu2022molecular,jiao20223d,zaidi2022pre} and multi-modality pretraining on 2D topology and 3D geometry~\cite{liu2022pretraining,liu2023moleculeSDE,fang2021chemrl}. Other transfer learning venues include multi-task learning~\cite{liu2019loss,liu2022structured} and out-of-distribution or domain adaptation~\cite{yao2021functionally,ji2022drugood,yao2023leveraging}, yet no geometry information has been utilized. All of these directions are promising for future exploration, and \framework{} serves as an auxiliary tool to accomplish them. For example, as will be shown in~\Cref{sec:datasets_and_benchmarks}, we leverage \framework{} to effectively evaluate \TotalPretrainingMethodNum{} pretraining methods with benchmarks.\looseness=-1

\textbf{(4) A framework for exploring data preprocessing and optimization tricks.} When comparing different symmetry-informed geometric models, we find that in addition to the model architecture, there are two important factors affecting the performance: the data preprocessing ({\eg}, energy and force rescaling and shift) and optimization methods ({\eg}, learning rate, learning rate schedule, number of epochs, random seeds). In this work, we explore the effect of four preprocessing tricks and around 2-10 optimization hyperparameters for each model and task. In general, we observe that each model may benefit differently in different tasks regarding the preprocessing and optimization tricks. However, data normalization is found to help improve performance hugely in most cases. We believe that \framework{} is an effective tool for exploring and understanding various engineering tricks.

\vspace{-1ex}
\section{Data Structures for Geometric Data} \label{sec:data_structure}
\vspace{-1ex}

\begin{figure}[tb!]
\centering
    \begin{subfigure}[\small An example of 2D topology and 3D geometry.]
    {\includegraphics[width=0.45\linewidth]{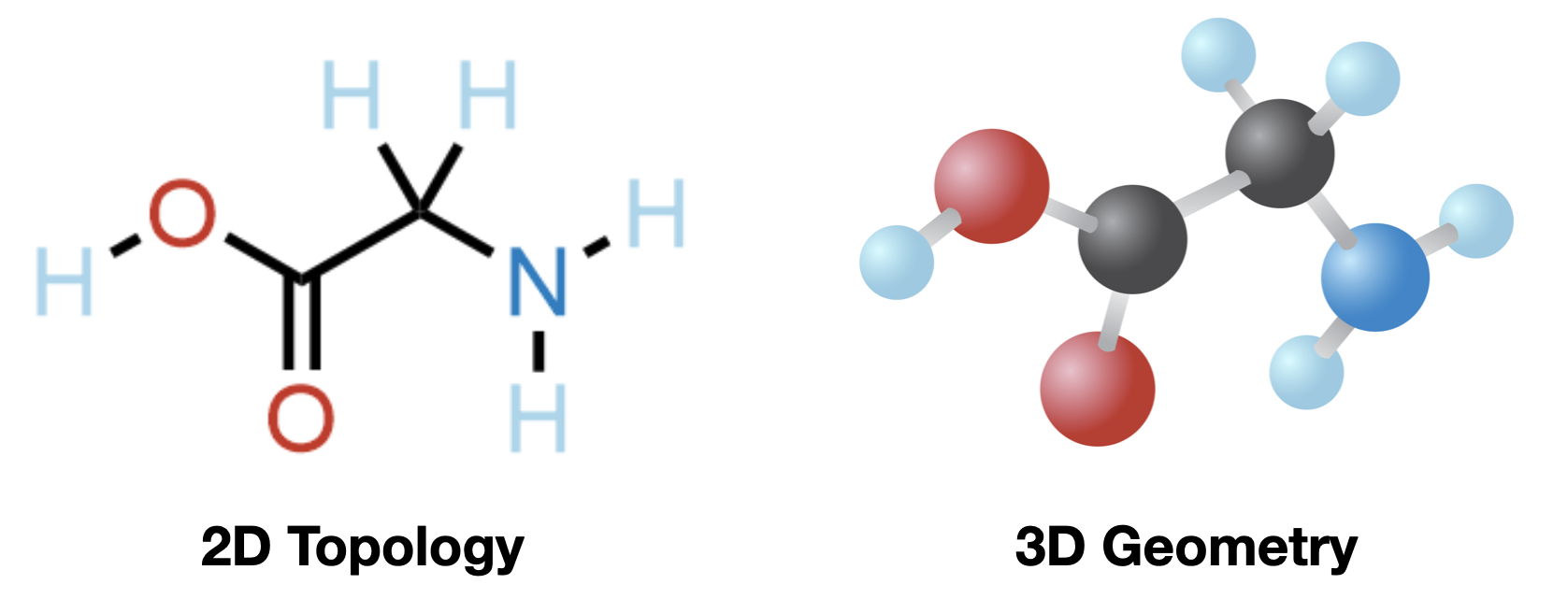}
    \label{fig:small_molecule_visualization}}
    \end{subfigure}
\hfill
    \begin{subfigure}[\small An illustration of the potential energy surface.]
    {\includegraphics[width=0.45\linewidth]{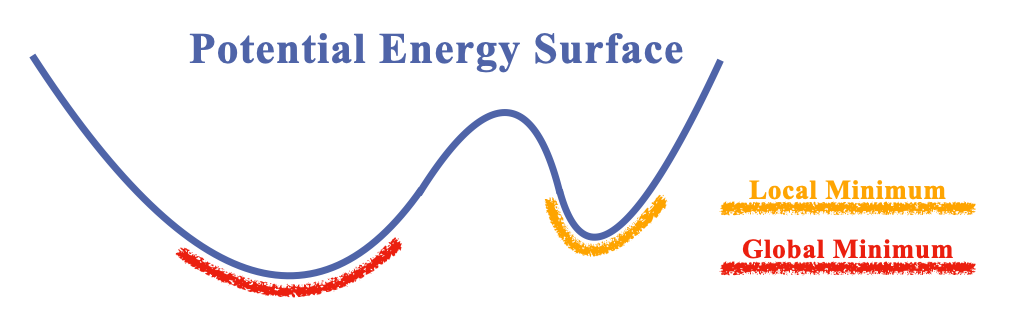}
    \label{fig:PES_visualization}}
    \end{subfigure}\\
\vspace{-2ex}
    \begin{subfigure}[\small An example of protein structure.]
    {\includegraphics[width=0.45\linewidth]{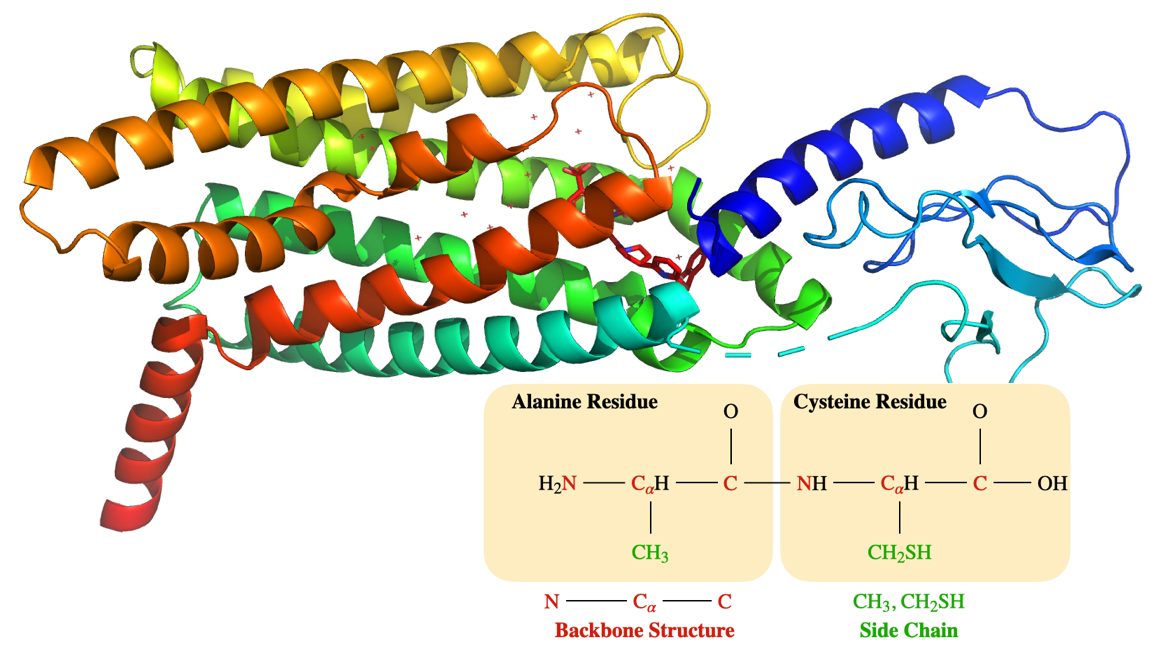}
    \label{fig:protein_visualization}}
    \end{subfigure}
\hfill
    \begin{subfigure}[\small An example of crystalline material.]
    {\includegraphics[width=0.45\linewidth]{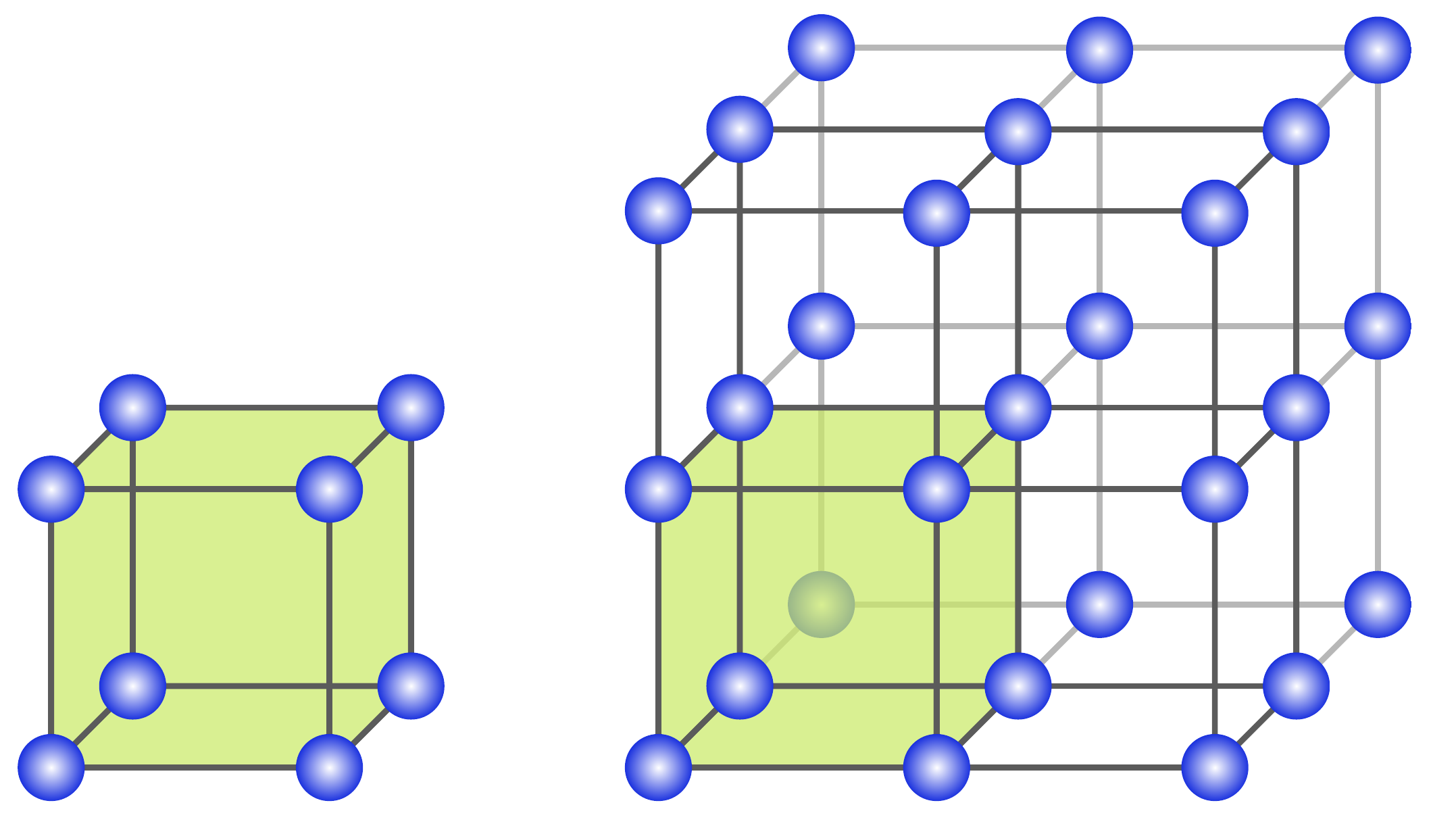}
    \label{fig:material_visualization}}
    \end{subfigure}
\vspace{-2ex}
\caption{
\small
\Cref{fig:small_molecule_visualization} illustrates 2D topology and 3D conformation for molecule \href{https://pubchem.ncbi.nlm.nih.gov/compound/750}{Glycine}.
\Cref{fig:protein_visualization} displays the 3D structure of \href{https://www.rcsb.org/structure/7LCJ}{protein}.
\Cref{fig:material_visualization} shows a simple cubic crystal of the \href{https://www.ccdc.cam.ac.uk/elements/polonium/}{element Po}.
\Cref{fig:PES_visualization} is a demo of PES.\looseness=-1
}
\vspace{-3.3ex}
\end{figure}

\textbf{Small molecule 3D conformation.}
Molecules are sets of points in the 3D Euclidean space, and they move in a dynamic motion, as known as the potential energy surface (PES). The region with the lowest energy corresponds to the most stable state for molecules, and molecules at these positions are called \textbf{conformations}, as illustrated in~\Cref{fig:PES_visualization}. For notation, we mark each 3D molecular graph as $\vg = (\mX, \mR)$, where $\mX$ and $\mR$ are for the atom types and positions, respectively.

\textbf{Crystalline material with periodic structure.} The crystalline materials or extended chemical structures possess a characteristic known as periodicity: their atomic or molecular arrangement repeats in a predictable and consistent pattern across all three spatial dimensions. This is the key aspect that differentiates them from small molecules. In~\Cref{fig:material_visualization}, we show an original unit cell (marked in green) that can repeatedly compose the crystal structure along the lattice. To model such a periodic structure, we adopt the data augmentation from CGCNN~\cite{xie2018crystal}: for each original unit cell, we shift it along the lattice in three dimensions and connect edges within a cutoff value (hyperparameter). For more details on the two augmentation variants, please check~\Cref{sec:data_preprocessing}.

\textbf{Protein with backbone structure.} Protein structures can be classified into four primary levels, {\eg}, the primary structure represents the linear arrangement of \textit{amino acids} within a polypeptide chain, where each amino acid is a small molecule. The geometric models can be naturally adopted to the higher-order structures, and in \framework{}, we consider the tertiary structure, which encompasses the complete three-dimensional organization of a single protein. Regarding the data structure, we consider the tertiary structure: each amino acid has an important \textit{backbone structure} $N-C_\alpha-C$ structure, and the $C_\alpha$ is bonded to the side chain. There are 20 common types of side chains corresponding to 20 amino acids, as illustrated in~\Cref{fig:protein_visualization}. Considering the long-sequence issue in proteins, existing works~\cite{jing2020learning,wang2023learning} mainly model the backbone structures for computational efficiency.

\begin{figure}[tb!]
\centering
\includegraphics[width=\linewidth]{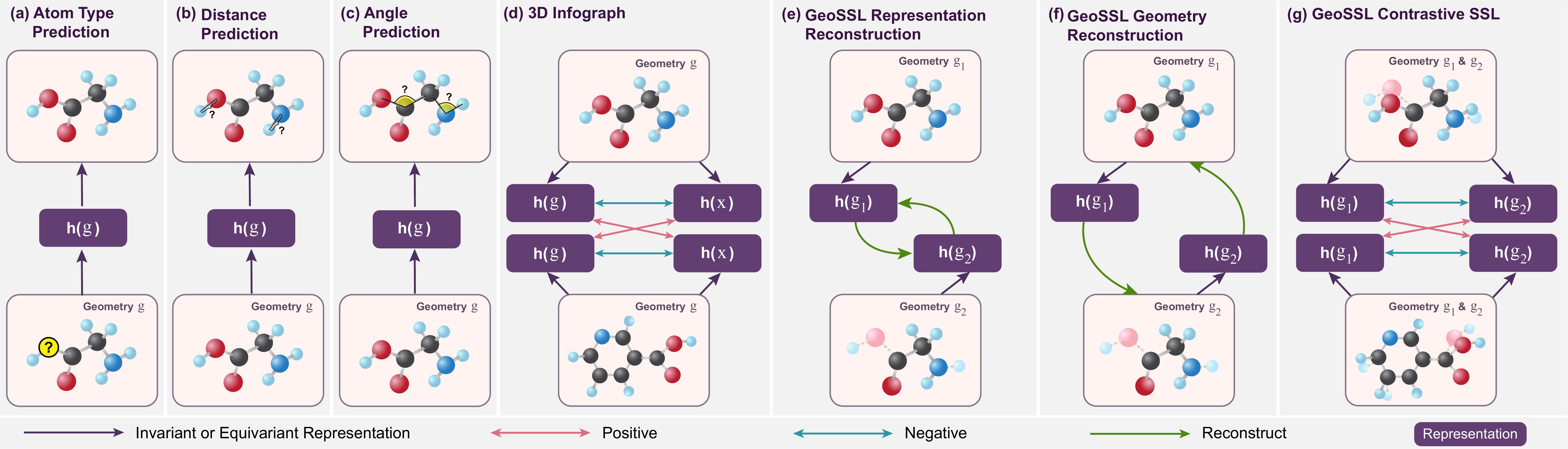}
\vspace{-4ex}
\caption{\small
Pipelines for seven single-modal geometric pretraining methods.
(a-c) conduct self-prediction.
(d) maximizes the MI between nodes and graphs.
(e-g) are GeoSSL, maximizing the MI between views $\vg_1$ and $\vg_2$.
} \label{fig:geometric_pretraining}
\vspace{-2.5ex}
\end{figure}

\vspace{-1ex}
\section{Symmetry-Informed Geometric Representation}
\label{sec:symmetry_informed_geometric_representation}
\vspace{-0.5ex}

%%%%%%%%%%%%%%%%%%%%%%%%%%%%%%%%%%%%%%%%%%%%%%%%%%
\subsection{Group Symmetry and Equivariance}
\vspace{-1ex}
Symmetry means the object remains invariant after certain transformations~\cite{weyl2015symmetry}, and it is everywhere on Earth, such as in animals, plants, and molecules. Formally, the set of all symmetric transformations satisfies the axioms of a group. Therefore, the group theory and its representation theory are common tools to depict such physical symmetry. \textbf{Group} is a set $G$ equipped with a group product $\times$ satisfying:
\begin{equation} \small
(1)~\exists \ve \in G, ~ \va \times \ve = \ve \times \va, \forall \va \in G; \quad
(2)~ \va \times \va^{-1} = \va^{-1} \times \va  = \ve; \quad
(3)~ \va \times (\vb \times \vc) = \va \times \vb \times \vc.
\end{equation}
\textbf{Group representation} is a mapping from the group $G$ to the group of linear transformations of a vector space $X$ with dimension $d$ (see~\cite{zee2016group} for more rigorous definition):
\begin{equation} \small
\rho_X(\cdot) : G \to \mathbb{R}^{d \times d}
\quad\quad \text{s.t.} \quad \rho(\ve) = 1 ~~ \wedge~~ \rho_X(\va) \rho_X(\vb) = \rho_X(\va \times \vb), ~\forall \va, \vb \in G.
\end{equation}
During modeling, the $X$ space can be the input 3D Euclidean space, the equivariant vector space in the intermediate layers, or the output force space. This enables the definition of equivariance as below.\looseness=-1

\textbf{Equivariance} is the property for the geometric modeling function $f: X \to Y$ as:
\begin{equation}\label{eq:equivariance}\small
f(\rho_X(\va) \vx) = \rho_Y(\va) f(\vx), ~~\forall \va \in G, \vx \in X.
\end{equation}
As displayed in~\Cref{fig:pipeline}, for molecule geometric modeling, the property should be rotation-equivariant and translation-equivariant ({\ie}, SE(3)-equivariant). More concretely, $\rho_X(\va)$ and $\rho_Y(\va)$ are the SE(3) group representations on the input ({\eg}, atom coordinates) and output space ({\eg}, force space), respectively. SE(3)-equivariant modeling in~\Cref{eq:equivariance} is essentially saying that the designed deep learning model $f$ is modeling the whole transformation trajectory on the molecule conformations, and the output is the transformed $\hat y$ accordingly. Further, we want to highlight that, in addition to the network architecture or representation function, the input features can also be represented as an equivariant feature mapping from the 3D mesh to $\mathbb{R}^{\tilde d}$~\cite{cohen2016steerable}, where $\tilde d$ depends on input data, {\eg}, $\tilde d$ = 1 (for atom type dimension) + 3 (for atom coordinate dimension) on small molecules. Such features are called steerable features in~\cite{cohen2016steerable,bronstein2021geometric} when only considering the subgroup SO(3)-equivariance.

\textbf{Invariance} is a special type of equivariance, defined as:
\begin{equation}\label{eq:invariance}\small
f(\rho_X(\va) \vx) = f(\vx), ~~\forall \va \in G, \vx \in X,
\end{equation}
with $\rho_Y(\va)$ as the identity $\forall \va \in G$. The group representation helps define the equivariance condition for $f$ to follow. Then, the question boils down to how to design such an equivariant $f$. In the following, we will discuss geometric modelings from a novel and unified perspective using the frame.\looseness=-1

%%%%%%%%%%%%%%%%%%%%%%%%%%%%%%%%%%%%%%%%%%%%%%%%%%
\subsection{Invariant Geometric Representation Learning} \label{sec:invariant_geometric_representation}
\vspace{-1ex}
The simple way of achieving SE(3) group symmetric in molecule geometry is invariant modeling. It means the model considers only the invariant features or type-0 features~\cite{thomas2018tensor} when modeling, and such type-0 features are invariant with respect to rotation and translation. Specifically, several works have been adopting the invariant features for modeling, including but not limited to pairwise distance (SchNet~\cite{schutt2018schnet}), bond angles (DimeNet~\cite{klicpera2020fast}), and torsion angles (SphereNet~\cite{liu2021spherical} and GemNet~\cite{klicpera_gemnet_2021}). Note that the torsion angles are angles between two planes defined by pairwise bonds. We also want to highlight that, from a mathematical perspective, equivariance and invariance can be transformed to each other by the scalarization technique. Please check \cite{hsu2002stochastic} for details.

%%%%%%%%%%%%%%%%%%%%%%%%%%%%%%%%%%%%%%%%%%%%%%%%%%
\subsection{Equivariant Geometric Representation Learning} \label{sec:equivariant_geometric_representation}
\vspace{-1ex}
Invariant modeling only captures the type-0 features. However, equivariant modeling of higher-order particles may bring in extra expressiveness. For example, the elementary particles in high energy physics~\cite{perkins2000introduction} inherit higher order symmetries in the sense of SO(3) representation theory, which makes the equivariant modeling necessary. Such higher-order particles include type-1 features like coordinates and forces in molecular conformation. There are many approaches to design such SE(3)-equivariant model satisfying~\Cref{eq:equivariance}. There are two main venues, as will be discussed below.

\textbf{Spherical Frame Basis.}
This research line utilizes the irreducible representations~\cite{geiger2022e3nn} for building SO(3)-equivariant representations, and the first work is TFN~\cite{thomas2018tensor}. Its main idea is to project the 3D Euclidean coordinates into the spherical harmonics space, which transforms equivariantly according to the irreducible representations of SO(3), and the translation-equivariant can be trivially guaranteed using the relative coordinates. Following this, there have been variants combining it with the attention module (Equiformer~\cite{liao2022equiformer}) or with more expressive network architectures (SEGNN~\cite{brandstetter2021geometric}, Allegro~\cite{musaelian2022learning}).\looseness=-1

\textbf{Vector Frame Basis.} This is an alternative solution by using the vector (in physics) frame basis. It builds the frame in the vector space, and the SO(3)-equivariance can be satisfied with the Gram-Schmidt process. Works along this line for molecule discovery include EGNN~\cite{satorras2021n} and PaiNN~\cite{schutt2021equivariant} for geometric representation, 3D-EMGP~\cite{jiao20223d} and MoleculeSDE~\cite{liu2023moleculeSDE} for geometric pretraining, and ClofNet~\cite{du2022se} for conformation generation. For macromolecules like protein, the equivariant vector frame has been used for protein design (StructTrans~\cite{ingraham2019generative}) and protein folding (AlphaFold2~\cite{jumper2021highly}).

The spherical frame basis can be easily extended to higher-order particles, yet it may suffer from the high computational cost. On the other hand, the vector frame basis is specifically designed for the 3D point clouds; thus, it is more efficient but cannot generalize to higher-order particles. Meanwhile, we would like to acknowledge other equivariant modeling paradigms, including using orbital features~\cite{qiao2020orbnet} and elevating 3D Euclidean space to SE(3) group~\cite{finzi2020generalizing,hutchinson2021lietransformer}. Please check~\Cref{sec:app:other_equivariant_modeling} for details.\looseness=-1

%%%%%%%%%%%%%%%%%%%%%%%%%%%%%%%%%%%%%%%%%%%%%%%%%%
\subsection{Geometric Pretraining}
\vspace{-1ex}
Recent studies have started to explore \textbf{single-modal geometric pretraining} on molecules. The GeoSSL paper~\cite{liu2022molecular} covers a wide range of geometric pretraining algorithms. The type prediction, distance prediction, and angle prediction predict the masked atom type, pairwise distance, and bond angle, respectively. The 3D InfoGraph predicts whether the node- and graph-level 3D representation are for the same molecule. GeoSSL is a novel geometric pretraining paradigm that maximizes the mutual information (MI) between the original conformation $\vg_1$ and augmented conformation $\vg_2$, where $\vg_2$ is obtained by adding small perturbations to $\vg_1$. RR, InfoNCE, and EBM-NCE optimize the objective in the latent representation space, either generative or contrastive. GeoSSL-DDM~\cite{liu2022molecular,zaidi2022pre} optimizes the same objective function using denoising score matching. 3D-EMGP~\cite{jiao2022energy} has the same strategy and utilizes an equivariant module to denoise the 3D noise directly. We illustrate these algorithms in~\Cref{fig:geometric_pretraining}. Another research line is the \textbf{multi-modal pretraining on topology and geometry}. GraphMVP~\cite{liu2022pretraining} first proposes one contrastive objective (EBM-NCE) and one generative objective (VRR) to optimize the MI between the 2D topologies and 3D geometries in the representation space. 3D InfoMax~\cite{stark20223d} is a special case of GraphMVP, with the contrastive part only. MoleculeSDE~\cite{liu2023moleculeSDE} extends GraphMVP by introducing two SDE models for solving the 2D and 3D reconstruction.\looseness=-1

%%%%%%%%%%%%%%%%%%%%%%%%%%%%%%%%%%%%%%%%%%%%%%%%%%
\subsection{Discussion: Reflection-antisymmetric in Geometric Learning}
\vspace{-1ex}
Till now, we have discussed the SE(3)-equivariance, {\ie}, the translation and rotation equivariance. As highlighted in the recent work~\cite{jing2022torsional,liu2023moleculeSDE}, the molecules needlessly satisfy the reflection-equivariant, but instead, they should be reflection-antisymmetric~\cite{liu2023moleculeSDE}. One standard example is that the energy of small molecules is reflection-antisymmetric in a binding system. Each of the two equivariant categories discussed in~\Cref{sec:equivariant_geometric_representation} can solve this problem easily. The spherical frame basis can achieve this by adding the reflection into the Wigner-D matrix~\cite{brandstetter2021geometric}. The vector frame basis can accomplish this using the cross-product during frame construction~\cite{liu2023moleculeSDE}.

\vspace{-1ex}
\section{Geometric Datasets and Benchmarks} \label{sec:datasets_and_benchmarks}
\vspace{-1ex}

In~\Cref{sec:symmetry_informed_geometric_representation}, we introduce a novel aspect for understanding symmetry-informed geometric models. In this section, we discuss utilizing \framework{} framework for benchmarking \TotalMethodNum{} geometric models over \TotalTaskNum{} tasks. For the detailed dataset acquisitions and task specifications ({\eg}, \textit{dataset size}, \textit{splitting}, and \textit{task unit}), please check~\Cref{sec:dataset_acquisition_and_specification_benchmark_hyperparameters}. \framework{} also covers 7 1D models and 10 2D graph neural networks (GNNs) and benchmarks the \TotalPretrainingMethodNum{} pretraining algorithms to learn a robust geometric representation. Additionally, we want to highlight \framework{} enables exploration of important data preprocessing and optimization tricks for performance improvement, as will be introduced next.

\begin{table}[tb!]
\setlength{\tabcolsep}{5pt}
\fontsize{9}{9}\selectfont
\centering
\caption{
\small
Results of 26 models on 12 quantum mechanics prediction tasks in QM9, with 110K for training, 10K for validation, and 11K for testing. The task unit is specified, and the evaluation is the mean absolute error (MAE).\looseness=-1
}
\label{tab:main_result_QM9}
\vspace{-2ex}
\begin{adjustbox}{max width=\textwidth}
\begin{tabular}{ll rrrrrrrrrrrrrrr}
\toprule
\multirow{2}{*}{Featurization} & \multirow{2}{*}{Model} & $\alpha$ $\downarrow$ & $\nabla \mathcal{E}$ $\downarrow$ & $\mathcal{E}_\text{HOMO}$ $\downarrow$ & $\mathcal{E}_\text{LUMO}$ $\downarrow$ & $\mu$ $\downarrow$ & $C_v$ $\downarrow$ & $G$ $\downarrow$ & $H$ $\downarrow$ & $R^2$ $\downarrow$ & $U$ $\downarrow$ & $U_0$ $\downarrow$ & ZPVE $\downarrow$\\
& & $\alpha_0^3$ & $meV$ & $meV$ & $meV$ & $D$ & $\frac{cal}{mol\cdot K}$ & $meV$ & $meV$ & $\alpha_0^2$ & $meV$ & $meV$ & $meV$\\
\midrule
\multirow{3}{*}{1D FPs}
% model: MLP_FPs
& MLP & 2.231 & 196.72 & 131.27 & 164.94 & 0.526 & 0.919 & 2158.64 & 2358.23 & 68.621 & 2340.61 & 2314.77 & 155.921\\
% model: RF_FPs
& RF & 3.801 & 207.02 & 165.72 & 183.04 & 0.534 & 1.485 & 3391.79 & 3729.94 & 94.512 & 3705.75 & 3678.25 & 253.132\\
% model: XGB_FPs
& XGB & 2.748 & 199.71 & 139.88 & 165.43 & 0.516 & 1.062 & 2563.93 & 2804.27 & 82.959 & 2786.28 & 2769.29 & 180.989\\
\midrule
\multirow{2}{*}{1D SMILES}
% model: CNN_SMILES
% 1e-4_CosineAnnealingLR_1000
& CNN & 0.364 & 165.22 & 124.65 & 114.81 & 0.566 & 0.173 & 156.66 & 170.59 & 20.403 & 166.18 & 169.89 & 10.070\\
% model: BERT_SMILES
% 1e-5_CosineAnnealingLR_150
& BERT & 0.313 & 117.50 & 84.93 & 98.88 & 0.446 & 0.176 & 170.01 & 183.43 & 18.002 & 183.84 & 188.60 & 13.410\\
\midrule
\multirow{2}{*}{1D SELFIES}
% model: CNN_SELFIES
% 1e-4_CosineAnnealingLR_1000
& CNN & 0.345 & 157.04 & 115.51 & 113.00 & 0.499 & 0.168 & 136.42 & 146.56 & 20.080 & 143.00 & 140.01 & 10.149\\
% model: BERT_SELFIES
% 1e-5_CosineAnnealingLR_200
& BERT & 0.348 & 123.11 & 91.15 & 90.80 & 0.461 & 0.203 & 168.20 & 187.50 & 19.125 & 204.93 & 195.98 & 17.328\\
\midrule
\multirow{10}{*}{2D Graph}
% model: GCN
% 5e-4_CosineAnnealingLR_1000
& GCN & 1.338 & 145.82 & 96.21 & 106.66 & 0.434 & 0.526 & 1198.12 & 1291.57 & 37.585 & 1281.03 & 1303.39 & 85.103\\
% model: ENN_S2S
% 5e-4_CosineAnnealingLR_1000
& ENN-S2S & 1.401 & 270.59 & 129.18 & 132.84 & 0.577 & 0.760 & 1487.21 & 955.24 & 34.609 & 1800.79 & 1521.32 & 51.226\\
% model: GraphSAGE
% 5e-4_CosineAnnealingLR_1000
& GraphSAGE & 1.601 & 131.45 & 88.78 & 93.21 & 0.402 & 0.544 & 1473.42 & 1617.73 & 38.112 & 1553.01 & 1565.65 & 95.344\\
% model: GAT
% 5e-4_CosineAnnealingLR_1000
& GAT & 1.132 & 135.90 & 94.70 & 98.52 & 0.406 & 0.291 & 911.82 & 991.31 & 26.583 & 1161.29 & 592.67 & 55.061\\
% model: GIN
% 5e-4_CosineAnnealingLR_1000
& GIN & 1.165 & 175.82 & 90.66 & 110.74 & 0.539 & 0.691 & 848.24 & 1090.36 & 35.110 & 1498.23 & 1364.18 & 108.331\\
% model: DMPNN
% 5e-4_CosineAnnealingLR_1000
& D-MPNN & 0.568 & 118.42 & 85.01 & 86.20 & 0.441 & 0.241 & 423.14 & 458.39 & 24.816 & 470.01 & 445.91 & 29.291\\
% model: PNA
% 5e-4_CosineAnnealingLR_1000
& PNA & 0.681 & 148.88 & 88.72 & 97.31 & 0.361 & 0.409 & 664.98 & 692.74 & 23.855 & 616.70 & 694.92 & 57.217\\
% % 5e-4_CosineAnnealingLR_300_256_300
% Graphormer & 2.950 & 85.57 & 56.84 & 55.00 & 0.354 & 1.078 & 2341.64 & 2229.97 & 131.888 & 2531.71 & 2503.99 & 147.902\\
% 5e-4_CosineAnnealingLR_300_256_1000
& Graphormer & 2.836 & 79.27 & 54.24 & 52.42 & 0.330 & 0.080 & 2066.28 & 2546.01 & 131.158 & 2229.88 & 2525.51 & 144.595\\
% 100_300_1e-4_0_6_use_bond_1000
& AWARE & 0.297 & 144.91 & 133.89 & 98.86 & 0.602 & 0.129 & 86.62 & 94.47 & 22.180 & 93.59 & 95.73 & 5.275\\
% model: GPS
% 1e-4_CosineAnnealingLR_1000
& GraphGPS & 0.209 & 75.98 & 54.75 & 54.53 & 0.288 & 0.089 & 528.50 & 693.19 & 12.488 & 296.00 & 411.16 & 49.888\\
\midrule

\multirow{9}{*}{3D Graph}

% 5e-4_CosineAnnealingLR_300_128_1000
& SchNet & 0.060 & 44.13 & 27.64 & 22.55 & 0.028 & 0.031 & 14.19 & 14.05 & 0.133 & 13.93 & 13.27 & 1.749\\

% 5e-4_CosineAnnealingLR_300_5_500
& DimeNet++ & 0.044 & 36.22 & 20.01 & 16.66 & 0.028 & 0.022 & 7.45 & 6.14 & 0.323 & 6.33 & 7.18 & 1.118\\

% 1e-3_CosineAnnealingWarmRestarts_300_100_7_32_4_2_8_96_use_rotation_transform
& SE(3)-Trans & 0.137 & 56.52 & 34.65 & 34.41 & 0.050 & 0.063 & 65.28 & 70.70 & 1.747 & 68.92 & 68.88 & 5.428\\

% 5e-4_CosineAnnealingLR_300_1000
& EGNN & 0.062 & 49.56 & 30.08 & 24.98 & 0.029 & 0.030 & 10.01 & 9.14 & 0.089 & 9.28 & 9.08 & 1.519\\

% 5e-4_CosineAnnealingLR_300_128_1000
& PaiNN & 0.049 & 42.73 & 24.46 & 20.16 & 0.016 & 0.025 & 8.43 & 7.88 & 0.169 & 8.18 & 7.63 & 1.419\\

% 5e-4_CosineAnnealingLR_300_128_1000
& GemNet-T & 0.041 & 35.46 & 17.85 & 15.86 & 0.021 & 0.023 & 7.61 & 7.08 & 0.271 & 6.42 & 5.88 & 1.232\\

% 5e-4_CosineAnnealingLR_300_1000
& SphereNet & 0.047 & 38.93 & 21.45 & 18.25 & 0.027 & 0.025 & 8.16 & 13.68 & 0.288 & 6.77 & 7.43 & 1.295\\

% 1e-4_CosineAnnealingLR_300_500
& SEGNN & 0.048 & 33.61 & 17.66 & 17.01 & 0.021 & 0.026 & 11.60 & 12.45 & 0.404 & 11.29 & 12.20 & 1.590\\

% % 5e-4_CosineAnnealingLR_5.0_128_1000
% & Allegro & 0.097 & 102.44 & 61.86 & 63.17 & 0.176 & 0.032 & 42.08 & 44.96 & 1.977 & 44.64 & 44.43 & 2.949\\

% % 5e-4_CosineAnnealingLR_5.0_128_1000
% & NequIP & 0.066 & 61.94 & 42.00 & 31.64 & 0.036 & 0.028 & 22.08 & 23.36 & 0.415 & 23.23 & 23.02 & 1.899\\

% 5e-4_CosineAnnealingLR_300_128_300
& Equiformer & 0.051 & 33.46 & 17.93 & 16.85 & 0.015 & 0.023 & 14.49 & 14.60 & 0.433 & 14.88 & 13.78 & 2.342\\
\bottomrule
\end{tabular}
\end{adjustbox}
\vspace{-4ex}
\end{table}

%%%%%%%%%%%%%%%%%%%%%%%%%%%%%%%%%%%%%%%%%%%%%%%%%%
\subsection{Small Molecules: QM9}
\vspace{-1ex}
QM9~\cite{ramakrishnan2014quantum} is a dataset consisting of 134K molecules, each with up to 9 heavy atoms. It includes 12 tasks that are related to the quantum properties. For example, U0 and U298 are the internal energies at 0K and 298.15K, respectively, and U298 and G298 are the other two energies that can be calculated from H298. The other 8 tasks are quantum mechanics related to the density functional theory (DFT) process.  On the QM9 dataset, we can easily get the 1D descriptors (Fingerprints/FPs~\cite{rogers2010extended}, SMILES~\cite{weininger1988smiles}, SELFIES~\cite{krenn2020self}), 2D topology, and 3D conformation.  This enables us to build models on each of them respectively: (1) We benchmark 7 models on 1D descriptors, including multi-layer perception (MLP), random forest (RF), XGBoost (SGB), convolution neural networks (CNN), and BERT~\cite{devlin2018bert}. (2) We benchmark 10 2D GNN models on the molecular topology, including GCN~\cite{duvenaud2015convolutional,kipf2016semi}, ENN-S2S~\cite{gilmer2017neural}, GraphSAGE~\cite{hamilton2017inductive}, GAT~\cite{velivckovic2017graph}, GIN~\cite{xu2018powerful}, D-MPNN~\cite{yang2019analyzing}, PNA~\cite{corso2020principal}, Graphormer~\cite{ying2021transformers}, AWARE~\cite{demirel2022attentive}, GraphGPS~\cite{rampasek2022GPS}. (3) We benchmark 9 3D geometric models on the molecular conformation, including SchNet~\cite{schutt2018schnet}, DimeNet++~\cite{klicpera2020fast}, SE(3)-Trans~\cite{fuchs2020se}, EGNN~\cite{satorras2021n}, PaiNN~\cite{schutt2021equivariant}, GemNet-T~\cite{klicpera_gemnet_2021}, SphereNet~\cite{liu2021spherical}, SEGNN~\cite{brandstetter2021geometric}, Equiformer~\cite{liao2022equiformer}. The evaluation metric is the mean absolute error (MAE). The detailed training tricks are in~\Cref{sec:dataset_acquisition_and_specification_benchmark_hyperparameters}.

The results of these 26 models are in~\Cref{tab:main_result_QM9}, and two important insights are below: (1) There is no one universally best geometric model, yet PaiNN, GemNet, and SphereNet perform well in most tasks. However, GemNet-T and SphereNet take up to 5 GPU days per task, and PaiNN takes less than 20 GPU hours. (2) The geometric conformation is important for quantum property prediction. The performance of using 3D conformation is better than all the 1D and 2D models \textit{by orders of magnitudes}.\looseness=-1

\begin{table}[tb!]
\setlength{\tabcolsep}{10pt}
\fontsize{9}{5}\selectfont
\centering
\caption{
\small 
Results on 6 energy ($\frac{kcal}{mol}$) and force ($\frac{kcal}{mol\cdot \text{\r{A}}}$) prediction tasks in MD17 and rMD17 (w/o normalization), and the metric is the mean absolute error (MAE).
The data split and complete results are in~\Cref{sec:dataset_acquisition_and_specification_benchmark_hyperparameters,sec:app:complete_results}.\looseness=-1
}
\label{tab:main_result_MD17_rMD17}
\vspace{-2ex}
\begin{adjustbox}{max width=\textwidth}
\begin{tabular}{ll cccccccccccc}
\toprule
\multirow{2}{*}{Model} & \multirow{2}{*}{\makecell{Energy\\/Force}} & \multicolumn{6}{c}{MD17} & \multicolumn{6}{c}{rMD17} \\
\cmidrule(lr){3-8} \cmidrule(lr){9-14}
& & Aspirin $\downarrow$ & Ethanol $\downarrow$ & Malonaldehyde $\downarrow$ & Naphthalene $\downarrow$ & Salicylic $\downarrow$ & Toluene $\downarrow$
& Aspirin $\downarrow$ & Ethanol $\downarrow$ & Malonaldehyde $\downarrow$ & Naphthalene $\downarrow$ & Salicylic $\downarrow$ & Toluene $\downarrow$
\\
\midrule
\multirow{2}{*}{SchNet}
& Energy & 0.475 & 0.109 & 0.3 & 0.167 & 0.212 & 0.149 & 0.534 & 1.757 & 0.26 & 0.124 & 2.618 & 0.119\\
& Force & 1.203 & 0.386 & 0.794 & 0.587 & 0.826 & 0.568 & 1.243 & 0.449 & 0.862 & 0.587 & 0.878 & 0.574\\
\midrule
\multirow{2}{*}{EGNN}
& Energy & 17.892 & 0.436 & 0.896 & 12.177 & 6.964 & 4.051 & 17.35 & 0.402 & 0.534 & 12.164 & 7.794 & 15.021\\
& Force & 3.042 & 0.924 & 1.566 & 1.136 & 1.177 & 1.202 & 3.825 & 0.989 & 1.334 & 1.183 & 1.571 & 1.165\\
\midrule
\multirow{2}{*}{PaiNN}
& Energy & 27.626 & 0.063 & 0.102 & 0.622 & 0.371 & 0.165 & 30.156 & 1.17 & 0.07 & 5.297 & 5.219 & 0.045\\
& Force & 0.572 & 0.23 & 0.338 & 0.132 & 0.288 & 0.141 & 0.573 & 0.316 & 0.377 & 0.161 & 0.321 & 0.231\\
\midrule
\multirow{2}{*}{GemNet-T}
& Energy & 0.684 & 4.598 & 4.966 & 0.482 & 0.128 & 0.098 & 5.389 & 1.615 & 9.496 & 0.031 & 21.411 & 959.745\\
& Force & 0.558 & 0.219 & 0.433 & 0.212 & 0.326 & 0.174 & 0.555 & 0.233 & 0.337 & 0.154 & 0.371 & 0.4\\
\midrule
\multirow{2}{*}{SphereNet}
& Energy & 0.244 & 1.603 & 1.559 & 0.167 & 0.188 & 0.113 & 0.304 & 0.072 & 0.138 & 0.093 & 0.771 & 20.479\\
& Force & 0.546 & 0.168 & 0.667 & 0.315 & 0.479 & 0.194 & 0.622 & 0.217 & 0.5 & 0.279 & 2.088 & 0.254\\
\midrule
\multirow{2}{*}{SEGNN}
& Energy & 17.774 & 0.151 & 0.247 & 0.655 & 2.173 & 0.624 & 15.721 & 0.13 & 0.182 & 1.11 & 1.494 & 0.814\\
& Force & 9.003 & 0.893 & 1.249 & 0.895 & 2.22 & 1.138 & 8.549 & 0.846 & 1.185 & 0.926 & 2.056 & 1.241\\
\midrule
\multirow{2}{*}{NequIP}
& Energy & 8.333 & 0.971 & 2.293 & 1.032 & 2.952 & 1.303 & 9.618 & 0.936 & 2.313 & 2.089 & 3.302 & 1.306\\
& Force & 23.769 & 5.832 & 12.099 & 5.247 & 14.048 & 6.8 & 22.904 & 6.027 & 12.372 & 5.529 & 15.693 & 7.094\\
\midrule
\multirow{2}{*}{Allegro}
& Energy & 1.138 & 0.258 & 1.33 & 0.824 & 1.114 & 0.441 & 1.366 & 1.002 & 0.417 & 1.756 & 1.035 & 0.437\\
& Force & 3.405 & 1.412 & 4.191 & 3.743 & 4.934 & 1.968 & 3.186 & 2.799 & 2.125 & 3.815 & 4.781 & 2.048\\
\midrule
\multirow{2}{*}{Equiformer}
& Energy & 0.308 & 0.096 & 0.183 & 0.097 & 0.189 & 0.209 & 0.375 & 0.064 & 0.085 & 0.069 & 0.143 & 0.104\\
& Force & 0.286 & 0.142 & 0.23 & 0.068 & 0.2 & 0.08 & 0.305 & 0.162 & 0.24 & 0.07 & 0.218 & 0.077\\
\bottomrule
\end{tabular}
\end{adjustbox}
\vspace{-3.2ex}
\end{table}

\begin{figure}[tb!]
\centering
\includegraphics[width=\textwidth]{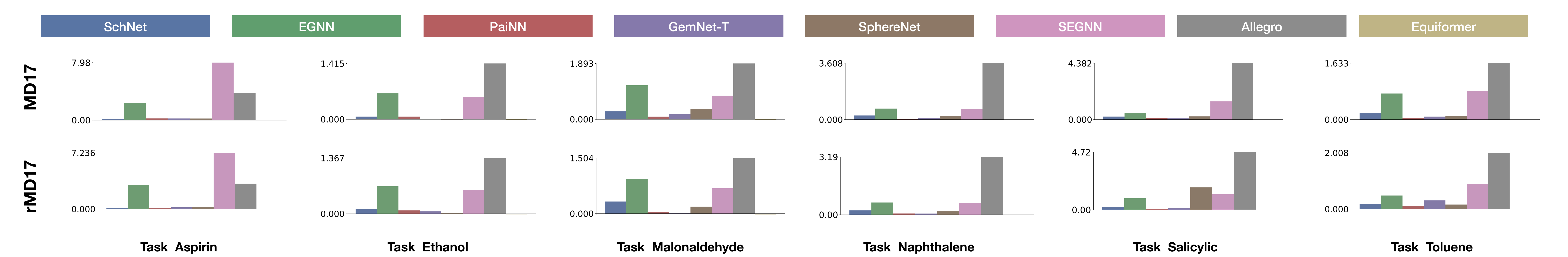}
\vspace{-4.5ex}
\caption{
\small
Ablation study on the effect of data normalization. Here are visualizations on performance differences on 6 tasks and 2 datasets, with MAE(force pred w/o normalization) - MAE(force pred w/ normalization).
}
\label{fig:ablation_main_data_normalization}
\vspace{-3.2ex}
\end{figure}

%%%%%%%%%%%%%%%%%%%%%%%%%%%%%%%%%%%%%%%%%%%%%%%%%%
\subsection{Small Molecules: MD17 and rMD17}
\vspace{-1ex}
MD17~\cite{chmiela2017machine} is a dataset of molecular dynamics simulation. It has 8 tasks corresponding to eight organic molecules, and each task includes the molecule positions along the PES (see~\Cref{fig:PES_visualization}). The goal is to predict each atom's energy and interatomic forces for each molecule's position. We follow the literature~\cite{schutt2018schnet,klicpera2020fast,liu2021spherical,schutt2021equivariant} of using 8 subtasks, 1K for training and 1K for validation, while the test set (from 48K to 991K) is much larger. However, the MD17 dataset contains non-negligible numerical noises~\cite{christensen2020revised}, and it is corrected by the revised MD17 (rMD17) dataset~\cite{ChristensenAndersS2020Otro}. 100K structures were randomly chosen for each task/molecule in MD17, and the single-point force and energy calculations were performed for each structure using the PBE/def2-SVP level of theory. The calculations were conducted with tight SCF convergence and a dense DFT integration grid, significantly minimizing the computational noises. The results on MD17 and rMD17 are in~\Cref{tab:main_result_MD17_rMD17}. We select 12 subtasks for illustration, and more comprehensive results can be found in~\Cref{sec:app:complete_results}. We can observe that, in general, PaiNN and Equiformer perform well on MD17 and rMD17 tasks. We also report \textbf{ablation study on data normalization}. NequIP~\cite{batzner20223} and Allegro~\cite{musaelian2022learning} introduce a normalization trick: multiplying the predicted energy with the mean of ground-truth force (reproduced results in~\Cref{sec:app:ablation_studies}). We plot the performance gap, MAE(w/o normalization) - MAE(w/ normalization), in~\Cref{fig:ablation_main_data_normalization}, and observe most of the gaps are positive, meaning that adding data normalization can lead to generally better performance.\looseness=-1

%%%%%%%%%%%%%%%%%%%%%%%%%%%%%%%%%%%%%%%%%%%%%%%%%%
\subsection{Small Molecules: COLL}
\vspace{-1ex}
\begin{wraptable}[9]{r}{0.45\textwidth}
\setlength{\tabcolsep}{5pt}
\renewcommand\arraystretch{0.55}
\fontsize{8}{9}\selectfont
\vspace{-5.2ex}
\centering
\caption{
\small
Results on energy and force prediction in COLL.
120k for training, 10k for val, 9.48k for test.
The metric is the mean absolute error (MAE).%\looseness=-1
}
\label{tab:main_result_COLL}
\begin{adjustbox}{max width=\textwidth}
\begin{tabular}{l rr}
\toprule
Model & Energy ($eV$) $\downarrow$ & Force ($eV/\text{\r{A}}$) $\downarrow$\\
\midrule
% 5e-4_CosineAnnealingLR_300_1000
SchNet & 0.178 & 0.130\\
% 5e-4_CosineAnnealingLR_300_1000
DimeNet++ & 0.036 & 0.049\\
% 5e-4_CosineAnnealingLR_300_1000
EGNN & 1.808 & 0.234\\
% 5e-4_CosineAnnealingLR_300_1000
PaiNN & 0.030 & 0.052\\
% 5e-4_CosineAnnealingLR_300_500
GemNet-T & 0.017 & 0.028\\
% 1e-4_CosineAnnealingLR_300_50_6_300
SphereNet & 0.032 & 0.047\\
% 1e-4_CosineAnnealingLR_2_300_100
SEGNN & 7.085 & 0.642\\
% 5e-4_CosineAnnealingLR_5_300_200
Equiformer & 0.036 & 0.030\\
\bottomrule
\end{tabular}
\end{adjustbox}
\end{wraptable}
The COLL dataset~\cite{GasteigerJohannes2020FaUD} comprises energy and force data for 140K random snapshots obtained from molecular dynamics simulations of molecular collisions. These simulations were conducted using the semiempirical GFN2-xTB method. To obtain the data, DFT calculations were performed utilizing the revPBE functional and def2-TZVP basis set, which also incorporated D3 dispersion corrections. The task is to predict the energy and force for each atom in each molecule, and we consider 8 most advanced geometric models for benchmarking. The results are in~\Cref{tab:main_result_COLL}, and two invariant models (GemNet and SphereNet) reach more optimal performance.

%%%%%%%%%%%%%%%%%%%%%%%%%%%%%%%%%%%%%%%%%%%%%%%%%%
\subsection{Small Molecules \& Proteins: LBA \& LEP}
\vspace{-0.5ex}
The binding affinity measures the strength of the binding interaction between a small molecule (ligand) to the target protein. In \framework{}, we consider modeling both the ligands and proteins with their 3D structures. During binding, a cavity in a protein can potentially possess suitable properties for binding a small molecule, and it is called a pocket~\cite{stank2016protein}. Due to the large volume of protein, \framework{} follows existing works~\cite{townshend2020atom3d} by only taking the binding pocket instead of the whole protein structure. Specifically, \framework{} models up to 600 atoms for each ligand and protein pair. For the benchmarking, we consider two binding affinity tasks. (1) The first task is ligand binding affinity (LBA)~\cite{wang2004pdbbind}. It is gathered from~\cite{wang2005pdbbind}, and the task is to predict the binding affinity strength between a ligand and a protein pocket. (2) The second task is ligand efficacy prediction (LEP)~\cite{friesner2004glide}. The input is a ligand and both the active and inactive conformers of a protein, and the goal is to classify whether or not the ligand can activate the protein's function. The results on two binding tasks are in~\Cref{tab:main_results_LBA_LEP}, and we can observe that PaiNN, GemNet, and SEGNN are generally outstanding on the two tasks.

\begin{table}[tb!]
\centering
\setlength{\tabcolsep}{20pt}
\fontsize{9}{10}\selectfont
\caption{
\small
Results on 2 binding affinity prediction tasks. We select three evaluation metrics for LBA: the root mean squared error (RMSD), the Pearson correlation ($R_p$) and the Spearman correlation ($R_S$). LEP is a binary classification task, and we use the area under the curve for receiver operating characteristics (ROC) and precision-recall (PR) for evaluation. We run cross-validation with 5 seeds, and the mean and std are reported.
}
\label{tab:main_results_LBA_LEP}
\vspace{-1.5ex}
\begin{adjustbox}{max width=\textwidth}
\begin{tabular}{l c c c c c}
\toprule
\multirow{2}{*}{\makecell{Model}} & \multicolumn{3}{c}{LBA} & \multicolumn{2}{c}{LEP}\\
\cmidrule(lr){2-4}
\cmidrule(lr){5-6}
 & RMSD $\downarrow$ & $R_P$ $\uparrow$ & $R_C$ $\uparrow$ & ROC $\uparrow$ & PR $\uparrow$\\
\midrule
% LBA 1e-4_CosineAnnealingLR_300_64_300
SchNet & 1.521 $\pm$ 0.02 & 0.474 $\pm$ 0.01 & 0.452 $\pm$ 0.01
% LEP 1e-4_CosineAnnealingLR_300_16_300
 & 0.450 $\pm$ 0.03 & 0.379 $\pm$ 0.03\\

% LBA 1e-4_CosineAnnealingLR_300_32_300
DimeNet++ & 1.672 $\pm$ 0.09 & 0.550 $\pm$ 0.01 & 0.556 $\pm$ 0.01
% LEP 1e-4_CosineAnnealingLR_300_4_300
 & 0.590 $\pm$ 0.06 & 0.496 $\pm$ 0.05\\

% LBA 1e-4_CosineAnnealingLR_300_128_300
EGNN & 1.494 $\pm$ 0.04 & 0.503 $\pm$ 0.04 & 0.483 $\pm$ 0.05
% LEP 1e-4_CosineAnnealingLR_300_16_300
 & 0.657 $\pm$ 0.05 & 0.559 $\pm$ 0.05\\
 
% LBA 1e-3_CosineAnnealingLR_300_128_300
PaiNN & 1.434 $\pm$ 0.02 & 0.583 $\pm$ 0.02 & 0.580 $\pm$ 0.02
% LEP 1e-4_CosineAnnealingLR_300_16_300
 & 0.585 $\pm$ 0.02 & 0.432 $\pm$ 0.03\\

% LBA 1e-4_CosineAnnealingLR_300_16_300
GemNet-T & -- & -- & --
% LEP 1e-4_CosineAnnealingLR_128_8_300
 & 0.674 $\pm$ 0.04 & 0.565 $\pm$ 0.05\\

% LBA 1e-4_CosineAnnealingLR_300_3_128_300
SphereNet & 1.581 $\pm$ 0.02 & 0.538 $\pm$ 0.01 & 0.529 $\pm$ 0.01
% LEP 1e-4_CosineAnnealingLR_300_3_8_300
 & 0.523 $\pm$ 0.04 & 0.432 $\pm$ 0.05\\

% LBA 1e-4_CosineAnnealingLR_300_5_16_300
SEGNN & 1.416 $\pm$ 0.03 & 0.566 $\pm$ 0.02 & 0.550 $\pm$ 0.02
% LEP 1e-4_CosineAnnealingLR_300_5_2_300
 & 0.574 $\pm$ 0.03 & 0.485 $\pm$ 0.03\\
\bottomrule
\end{tabular}
\end{adjustbox}
\vspace{-4ex}
\end{table}

%%%%%%%%%%%%%%%%%%%%%%%%%%%%%%%%%%%%%%%%%%%%%%%%%%
\subsection{Proteins: EC and Fold}
\vspace{-1ex}
\begin{wraptable}[7]{r}{0.5\textwidth}
\setlength{\tabcolsep}{8pt}
\renewcommand\arraystretch{1.1}
\vspace{-4ex}
\centering
\caption{\small
Results on EC and Fold classification, and metric is the accuracy. The data splits are in~\Cref{sec:dataset_acquisition_and_specification_benchmark_hyperparameters}.
}
\label{tab:main_results_protein}
\begin{adjustbox}{max width=0.5\textwidth}
\begin{tabular}{l c cccc}
\toprule
\multirow{2}{*}{Model} & \multirow{2}{*}{EC (\%)} & \multicolumn{4}{c}{Fold}\\
\cmidrule(lr){3-6}
& & Fold (\%) & Sup (\%) & Fam (\%) & Avg (\%) \\
\midrule
GVP-GNN~\cite{JingBowen2020LfPS} & 63.936 & 34.819 & 52.711	& 95.047 & 60.859\\
GearNet-IEConv~\cite{HermosillaPedro2020ICaP} & -- & 39.694 & 59.330 & 98.506 & 65.843\\
GearNet~\cite{ZhangZuobai2022PRLb} & 78.836 & 29.109 & 43.062 & 95.991 & 56.054\\
ProNet~\cite{WangLimei2022LHPR} & 84.251 & 52.089 & 69.378 & 98.270 & 73.246\\
CDConv~\cite{fan2023cdconv} & 86.887 & 60.028 & 79.904 & 99.528 & 79.820\\
\bottomrule
\end{tabular}
\end{adjustbox}
\end{wraptable}
An essential aspect of proteins is their ability to serve as bio-catalysts, known as enzymes. Enzyme Commission (EC) number~\cite{J.M.1993Enpb} is a numerical classification scheme that describes the enzyme functionalities. Here we follow a recent work~\cite{HermosillaPedro2020ICaP} in predicting 37K proteins with 384 EC types. Another protein geometric task we consider is protein folding. It is an important biological task in predicting the 3D structures from 1D amino acid sequences. Here we apply the folding pattern classification task~\cite{HouJie2018Ddcn,lin2013hierarchical}, comprising 16K proteins and 1,195 fold patterns. We further consider three testsets (Fold, Superfamily, and Family) based on the sequence and structure similarity~\cite{MurzinAlexeyG.1995SAsc}. The detailed specifications are in~\Cref{sec:dataset_acquisition_and_specification_benchmark_hyperparameters}. The results of 5 models are in~\Cref{tab:main_results_protein}, and CDConv~\cite{fan2023cdconv} outperforms other methods by a large margin.\looseness=-1

%%%%%%%%%%%%%%%%%%%%%%%%%%%%%%%%%%%%%%%%%%%%%%%%%%
\subsection{Crystalline Materials: MatBench and QMOF}
\begin{figure}
\begin{minipage}[c]{0.58\textwidth}
\fontsize{9}{9}\selectfont
\tabcaption{
\small
Results on the 8 tasks from MatBench and 1 task from QMOF (with optimal DA).
The data split and task unit are in~\Cref{sec:dataset_acquisition_and_specification_benchmark_hyperparameters}, and the metric is the mean absolute error (MAE).
}
\label{tab:main_result_material}
% \vspace{-1ex}
\begin{adjustbox}{max width=\textwidth}
\begin{tabular}{l rrrrrrrrrrr}
\toprule
\multirow{3}{*}{Model} & \multicolumn{8}{c}{MatBench} & \multicolumn{1}{c}{QMOF}\\
\cmidrule(lr){2-9} \cmidrule(lr){10-10}
& Per. $E_{\text{form}}$ $\downarrow$ & Dielectric $\downarrow$ & $log_{10} G$ $\downarrow$ & $log_{10} K$ $\downarrow$ & $E_{\text{exfo}}$ $\downarrow$ & Phonons $\downarrow$ & $E_{\text{form}}$ $\downarrow$ & Band Gap $\downarrow$ & Band Gap $\downarrow$\\
& 18,928 & 4,764 & 10,987 & 10,987 & 636 & 1,265 & 132,752 & 106,113 & 20,425\\
\midrule
SchNet & 0.040 & 0.334 & 0.081 & 0.060 & 65.201 & 42.586 & 0.026 & 0.327 & 0.236\\
DimeNet++ & 0.037 & 0.357 & 0.081 & 0.058 & 68.685 & 38.339 & 0.025 & 0.208 & 0.240\\
EGNN & 0.039 & 0.306 & 0.089 & 0.064 & 78.205 & 76.143 & 0.026 & 0.211 & 0.256\\
PaiNN & 0.038 & 0.317 & 0.080 & 0.053 & 67.752 & 44.602 & 0.022 & 0.190 & 0.235\\
GemNet-T & 0.042 & 0.325 & 0.088 & 0.061 & 68.425 & 48.986 & 0.026 & 0.186 & 0.207\\
SphereNet & 0.043 & 0.388 & 0.087 & 0.061 & 72.987 & 36.300 & 0.029 & 0.217 & 0.251\\
SEGNN & 0.046 & 0.360 & 0.087 & 0.059 & 65.052 & 43.638 & 0.047 & 0.330 & 0.330\\
Equiformer & 0.046 & 0.280 & 0.087 & 0.057 & 62.977 & 37.381 & 0.031 & 0.216 & 0.238\\
\bottomrule
\end{tabular}
\end{adjustbox}
\end{minipage}
\hfill
\begin{minipage}[c]{0.4\textwidth}
\centering
\vspace{+1.2ex}
\includegraphics[width=\textwidth]{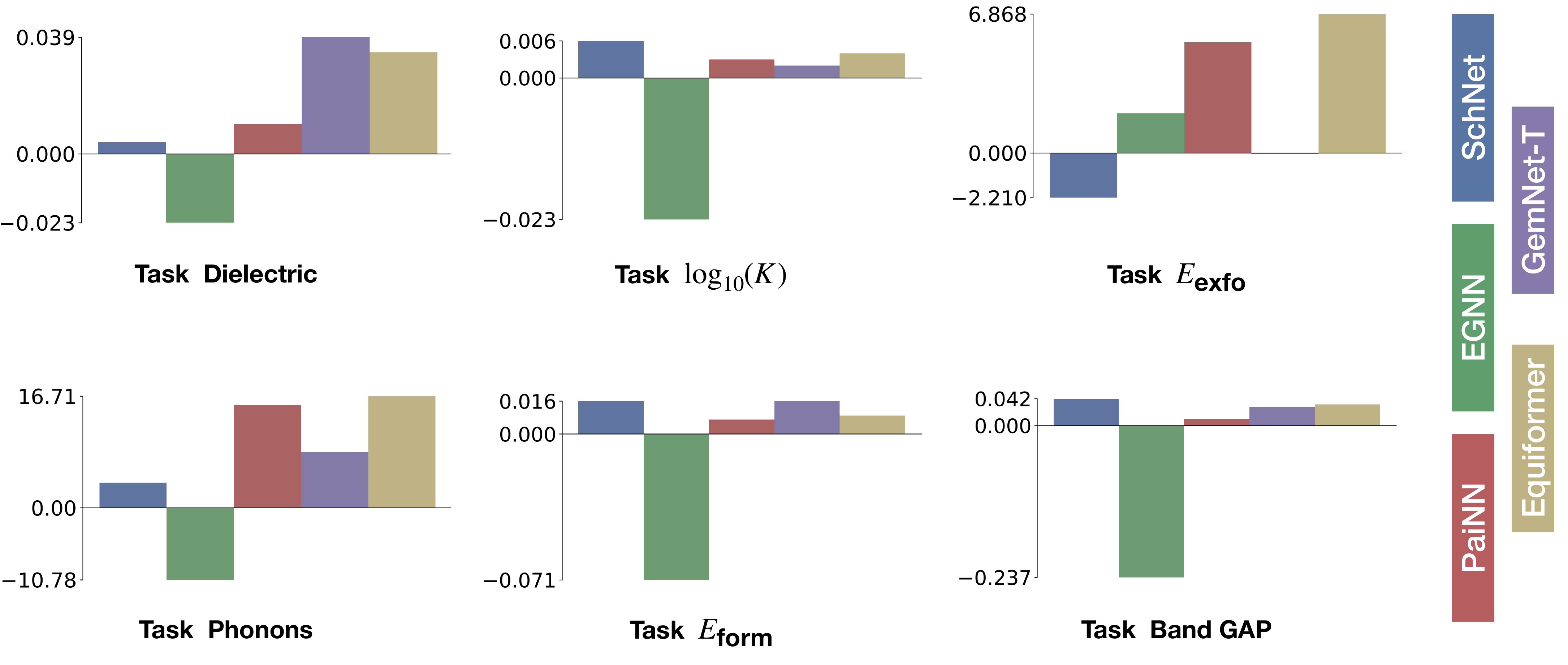}
\vspace{-4ex}
\caption{\small 
\fontsize{7.5}{2}\selectfont
Ablation study on the performance gap with DA: MAE(expanded DA) - MAE(gathered DA).}
\label{fig:ablation_main_DA}
\end{minipage}
\vspace{-1ex}
\end{figure}

\vspace{-1ex}
MatBench~\cite{DunnAlexander2020Bmpp} is created specifically to evaluate the performance of machine learning models in predicting properties of inorganic bulk materials covering mechanical, electronic, and thermodynamic material properties~\cite{DunnAlexander2020Bmpp}. Here we consider 8 regression tasks with crystal structures, including predicting the formation energy (Perovskites, $E_{\text{form}}$), exfoliation energies ($E_{\text{exfo}}$), band gap, shear and bulk modulus ($log_{10} G$ and $log_{10} K$), etc. Please check~\Cref{sec:dataset_acquisition_and_specification_benchmark_hyperparameters} for more details. Quantum MOF (QMOF)~\cite{rosen2021machine} is a dataset of over 20K metal-organic frameworks (MOFs) and coordination polymers derived from DFT. The task is to predict the band gap, the energy gap between the valence band and the conduction band. The results of 8 geometric models on 8 MatBench tasks and 1 QMOF task are in~\Cref{tab:main_result_material}, and we can observe that the performance of all the models is very close; only PaiNN, GemNet, and Equiformer are slightly better. We also conduct \textbf{ablation study on periodic data augmentation.} We note that there are two data augmentation (DA) methods: gathered and expanded. Gathered DA means that we shift the original unit cell along three dimensions, and the translated unit cells will have the \textit{same} node indices as the original unit cell, {\ie}, a multi-edge graph. However, expanded DA will assume the translated unit cells have different node indices from the original unit cell. (A visual demonstration is in~\Cref{sec:data_preprocessing}). We conduct an ablation study on the effect of these two DAs, and we plot MAE(expanded DA) - MAE(gathered DA) on six tasks in~\Cref{fig:ablation_main_DA}. It reveals that for most of the models (except EGNN), using gathered DA can lead to consistently better performance, and thus it is preferred. For more qualitative analysis, please check~\Cref{sec:app:ablation_studies}.

%%%%%%%%%%%%%%%%%%%%%%%%%%%%%%%%%%%%%%%%%%%%%%%%%%
\subsection{Geometric Pretraining on Small Molecules} \label{sec:geometric_pretraining}
\vspace{-1ex}
We run \TotalPretrainingMethodNum{} pretraining algorithms, including one supervised pretraining: the pretraining dataset ({\eg}, PCQM4Mv2~\cite{hu2020ogb}) possess the energy or energy gap label for each conformation, which can be naturally adopted for pretraining. The benchmark results of using SchNet as the backbone model pretrained on PCQM4Mv2 and fine-tuning on QM9 tasks are in~\Cref{tab:main_result_QM9_SchNet_pretraining}. We observe that MoleculeSDE and GeoSSL-DDM utilizing the geometric denoising diffusion models outperform other pretraining methods in most cases. On the other hand, supervised pretraining (pretrained on energy gap $\nabla \mathcal{E}$) reaches outstanding performance on $\nabla \mathcal{E}$ downstream task, yet the generalization to other tasks is modest. Please check~\Cref{sec:app:complete_results} for more pretraining results with different backbone models.\looseness=-1

\begin{table}[tb]
\setlength{\tabcolsep}{5pt}
\fontsize{9}{9}\selectfont
\caption{
\small
Pretraining results on 12 quantum mechanics prediction tasks from QM9, and the backbone model is SchNet. We take 110K for training, 10K for validation, and 11K for testing. The evaluation metric is the mean absolute error (MAE), and the best and the second best results are marked in \underline{\textbf{bold}} and \textbf{bold}, respectively.
}
\label{tab:main_result_QM9_SchNet_pretraining}
\vspace{-1.5ex}
\begin{adjustbox}{max width=\textwidth}
\begin{tabular}{l c c c c c c c c c c c c}
\toprule
Pretraining & $\alpha$ $\downarrow$ & $\nabla \mathcal{E}$ $\downarrow$ & $\mathcal{E}_\text{HOMO}$ $\downarrow$ & $\mathcal{E}_\text{LUMO}$ $\downarrow$ & $\mu$ $\downarrow$ & $C_v$ $\downarrow$ & $G$ $\downarrow$ & $H$ $\downarrow$ & $R^2$ $\downarrow$ & $U$ $\downarrow$ & $U_0$ $\downarrow$ & ZPVE $\downarrow$\\
\midrule

% 5e-4_CosineAnnealingLR_300_128_1000
-- (random init) & 0.060 & 44.13 & 27.64 & 22.55 & 0.028 & 0.031 & 14.19 & 14.05 & 0.133 & 13.93 & 13.27 & 1.749\\
\midrule

% pretrain_Supervised/PCQM4Mv2_schnet/1e-4_CosineAnnealingLR_100
Supervised & 0.062 & 40.31 & 25.57 & 21.69 & 0.030 & 0.030 & 14.36 & 14.68 & 0.308 & 15.21 & 16.13 & 1.638\\
\midrule

% pretrain_ChargePrediction/PCQM4Mv2_schnet/1e-4_CosineAnnealingLR_0.3_100
Type Prediction & 0.073 & 45.38 & 28.76 & 24.83 & 0.036 & 0.032 & 16.66 & 16.28 & 0.275 & 15.56 & 14.66 & 2.094\\

% pretrain_DistancePrediction/PCQM4Mv2_schnet/1e-4_CosineAnnealingLR_100
Distance Prediction & 0.065 & 45.87 & 27.61 & 23.34 & 0.031 & 0.033 & 14.83 & 15.81 & 0.248 & 15.07 & 15.01 & 1.837\\

% pretrain_TorsionAnglePrediction/PCQM4Mv2_schnet/1e-4_CosineAnnealingLR_1e-3_100
Angle Prediction & 0.066 & 48.45 & 29.02 & 24.40 & 0.034 & 0.031 & 14.13 & 13.77 & 0.214 & 13.50 & 13.47 & 1.861\\

% pretrain_3DInfoGraph/PCQM4Mv2_schnet/1e-4_CosineAnnealingLR_100
3D InfoGraph & 0.062 & 45.96 & 29.29 & 24.60 & 0.028 & 0.030 & 13.93 & 13.97 & 0.133 & 13.55 & 13.47 & 1.644\\

% pretrain_RR/PCQM4Mv2_schnet/1e-4_CosineAnnealingLR_0_0.3_0.3_RR_100
GeoSSL-RR & 0.060 & 43.71 & 27.71 & 22.84 & 0.028 & 0.031 & 14.54 & 13.70 & \textbf{0.122} & 13.81 & 13.75 & 1.694\\

% pretrain_InfoNCE/PCQM4Mv2_schnet/1e-4_CosineAnnealingLR_0_0.3_0.3_InfoNCE_100
GeoSSL-InfoNCE & 0.061 & 44.38 & 27.67 & 22.85 & \textbf{0.027} & 0.030 & 13.38 & 13.36 & \underline{\textbf{0.116}} & 13.05 & 13.00 & 1.643\\

% pretrain_EBM_NCE/PCQM4Mv2_schnet/1e-4_CosineAnnealingLR_0_0.3_0.3_EBM_NCE_100
GeoSSL-EBM-NCE & 0.057 & 43.75 & 27.05 & 22.75 & 0.028 & 0.030 & 12.87 & 12.65 & 0.123 & 13.44 & 12.64 & 1.652\\

% pretrain_GraphMVP/PCQM4Mv2_schnet/Vanilla_Contrastive_1e-4_0.3_100
3D InfoMax & 0.057 & 42.09 & 25.90 & 21.60 & 0.028 & 0.030 & 13.73 & 13.62 & 0.141 & 13.81 & 13.30 & 1.670\\

% pretrain_GraphMVP/PCQM4Mv2_schnet/Vanilla_1e-4_0.3_100
GraphMVP & 0.056 & 41.99 & 25.75 & \textbf{21.58} & \textbf{0.027} & \textbf{0.029} & 13.43 & 13.31 & 0.136 & 13.03 & 13.07 & 1.609\\

 % pretrain_DeepMind/PCQM4Mv2_schnet/1e-4_CosineAnnealingLR_0_0.3_0.15_GeoSSL_10_0.01_30_symmetry_0.1_100
GeoSSL-DDM-1L & 0.058 & 42.64 & 26.32 & 21.87 & 0.028 & 0.030 & 12.61 & 12.81 & 0.173 & 12.45 & 12.12 & 1.696\\

% pretrain_GeoSSL/PCQM4Mv2_schnet/1e-4_CosineAnnealingLR_0_0.3_0_GeoSSL_10_0.01_30_symmetry_0.05_100
% GeoSSL & 0.056 & 42.42 & 25.81 & 22.03 & 0.027 & 0.029 & 12.00 & 11.14 & \todo{0.176} & 11.35 & 10.96 & 1.693\\
GeoSSL-DDM & 0.056 & 42.29 & \underline{\textbf{25.61}} & 21.88 & \textbf{0.027} & \textbf{0.029} & \textbf{11.54} & \textbf{11.14} & 0.168 & \textbf{11.06} & \underline{\textbf{10.96}} & 1.660\\

% pretrain_MoleculeSDE_generative/PCQM4Mv2_schnet_SDEModel2Dto3D_01_SDEModel3Dto2D_node_adj_dense/2Dto3D_1_VE_3Dto2D_1_VE_5e-4_0_anneal_2_anneal_0_50
MoleculeSDE (VE) & \textbf{0.056} & \textbf{41.84} & 25.79 & 21.63 & \textbf{0.027} & \textbf{0.029} & \underline{\textbf{11.47}} & \underline{\textbf{10.71}} & 0.233 & \underline{\textbf{11.04}} & \textbf{10.95} & \underline{\textbf{1.474}}\\

% pretrain_MoleculeSDE/PCQM4Mv2_schnet_SDEModel2Dto3D_02_SDEModel3Dto2D_node_adj_dense/2Dto3D_1_VP_3Dto2D_1_VP_CL_EBM_node_dot_prod_1_0.1_0_5e-4_0.3_anneal_0_100
MoleculeSDE (VP) & \underline{\textbf{0.054}} & \underline{\textbf{41.77}} & \textbf{25.74} & \underline{\textbf{21.41}} & \underline{\textbf{0.026}} & \underline{\textbf{0.028}} & 13.07 & 12.05 & 0.151 & 12.54 & 12.04 & \textbf{1.587}\\
\bottomrule
\end{tabular}
\end{adjustbox}
\vspace{-3ex}
\end{table}

\vspace{-1ex}
\section{Conclusion and Future Directions}
\vspace{-1ex}

\framework{} provides a unified view on the SE(3)-equivariant models, together with the implementations. Indeed these can serve as the building blocks to various tasks, such as geometric pretraining (as displayed in~\Cref{sec:geometric_pretraining}) and the conformation generation (ClofNet~\cite{du2022se}, MoleculeSDE~\cite{liu2023moleculeSDE}), paving the way for building more foundational models and solving more challenging tasks.

\textbf{Limitations on models and tasks.}
\framework{} includes 10 2D graph models, \TotalMethodNum{} geometric models, \TotalPretrainingMethodNum{} pretraining methods, and \TotalTaskNum{} diverse tasks. We would also like to acknowledge there exist many more tasks ({\eg}, Atom3D~\cite{townshend2020atom3d}, Molecule3D~\cite{xu2021molecule3d}, OC20~\cite{ocp_dataset}) and more geometric models ({\eg}, OrbNet~\cite{qiao2020orbnet}, MACE~\cite{Batatia2022Design} and LieTransformer~\cite{hutchinson2021lietransformer}). We will continue adding them in the future.

\textbf{Multi-modality as future exploration.}
Recently, there have been quite some explorations on building multi-modal applications on molecules, especially by incorporating textual data~\cite{zeng2022deep,edwards2021text2mol,edwards2022translation,su2022molecular,liu2022multi,liu2023text,liu2023chatgpt,zhao2023gimlet}. However, these works mainly focus on the 1D sequence or 2D topology, and 3D geometry is rarely considered. We believe that \framework{} can support this for future exploration.

\clearpage

\section*{Acknowledgement}
\vspace{-1ex}
The authors would like to thank Zichao Rong, Chengpeng Wang, Jiarui Lu, Farzaneh Heidari, Zuobai Zhang, Limei Wang, and Hanchen Wang for their helpful discussions. This project is supported by the Natural Sciences and Engineering Research Council (NSERC) Discovery Grant, the Canada CIFAR AI Chair Program, collaboration grants between Microsoft Research and Mila, Samsung Electronics Co., Ltd., Amazon Faculty Research Award, Tencent AI Lab Rhino-Bird Gift Fund, and a National Research Council of Canada (NRC) Collaborative R\&D Project. This project was also partially funded by IVADO Fundamental Research Project grant PRF-2019-3583139727.

\small{
\bibliography{reference}
\bibliographystyle{plain}
}
\clearpage

\appendix

\addcontentsline{toc}{section}{Appendix} % Add the appendix text to the document TOC
\renewcommand \thepart{} % make "Part" text invisible
\renewcommand \partname{}
\part{\Large\centering{Appendix}}
 % Start the appendix part
\parttoc % Insert the appendix TOC

\newpage

\section{Data Structure and Data Preprocessing} \label{sec:data_preprocessing}

%%%%%%%%%%%%%%%%%%%%%%%%%%%%%%%%%%%%%%%%%%%%%%%%%%
\subsection{Small Molecules}

In the machine learning and computational chemistry domain, existing works are mainly focusing on the molecule 1D description~\cite{rogers2010extended,weininger1988smiles,krenn2020self} and 2D topology graph~\cite{duvenaud2015convolutional,kipf2016semi,velivckovic2017graph,gilmer2017neural,liu2019n,hamilton2017inductive,xu2018powerful,corso2020principal,yang2019analyzing,demirel2022attentive,yang2019analyzing,ying2021transformers,rampasek2022GPS}. Especially as the 2D graph, where the atoms and bonds are treated as nodes and edges, respectively. To model this graph structure, a message-passing graph neural network model family has been proposed. In~\Cref{tab:main_result_QM9}, we provide a comparison of models on 1D descriptions, 2D topological graphs, and 3D geometric conformations. The observation verifies the necessity of using conformation for quantum property prediction tasks.

%%%%%%%%%%%%%%%%%%%%%%%%%%%%%%%%%%%%%%%%%%%%%%%%%%
\subsection{Proteins}
Protein structures can be classified into four primary levels. The primary structure represents the linear arrangement of amino acids within a polypeptide chain. Secondary structure arises from local interactions between adjacent amino acids, resulting in the formation of recognizable patterns like alpha helices and beta sheets. The tertiary structure encompasses the complete three-dimensional organization of a single protein, involving additional folding and structural modifications beyond the secondary structure. Quaternary structure emerges when multiple polypeptide chains or subunits interact to form a protein complex.

Specifically for geometric modeling, we are now focusing on the protein tertiary structure, which can be constructed based on different structural levels, namely the all-atom level, backbone level, and residue level. We explain the details below, and you can find an illustration in~\Cref{fig:protein_visualization}.
\begin{itemize}[noitemsep,topsep=0pt]
    \item At the all-atom level, the graph nodes represent individual atoms, capturing the fine-grained details of the protein structure.
    \item At the backbone level, the graph nodes correspond to the backbone atoms ($N-C_{\alpha}-C$), omitting the side chain information. This level of abstraction focuses on the essential backbone structure of the protein.
    \item At the residue level, the graph nodes represent amino acid residues. The position of each residue can be represented by the position of its $C_{\alpha}$ atom or calculated as the average position of the backbone atoms within the residue. This level provides a higher-level representation of the protein structure, grouping atoms into residue units.
\end{itemize}

%%%%%%%%%%%%%%%%%%%%%%%%%%%%%%%%%%%%%%%%%%%%%%%%%%
\subsection{Crystalline Materials}
\textbf{Periodic structure.}
The crystalline materials or extended chemical structures possess a characteristic known as periodicity: their atomic or molecular arrangement repeats in a predictable and consistent pattern across all three spatial dimensions. This is the key aspect that differentiates them from small molecules. In~\Cref{fig:material_visualization}, we show an original unit cell (marked in green) that can repeatedly compose the crystal structure along the lattice. To model such a periodic structure, we adopt the data augmentation (DA) from CGCNN~\cite{xie2018crystal}, yet with two variants as explained below.

\begin{figure}[htb!]
\centering
    \begin{subfigure}[DA: gathered.]
    {\includegraphics[width=0.48\linewidth]{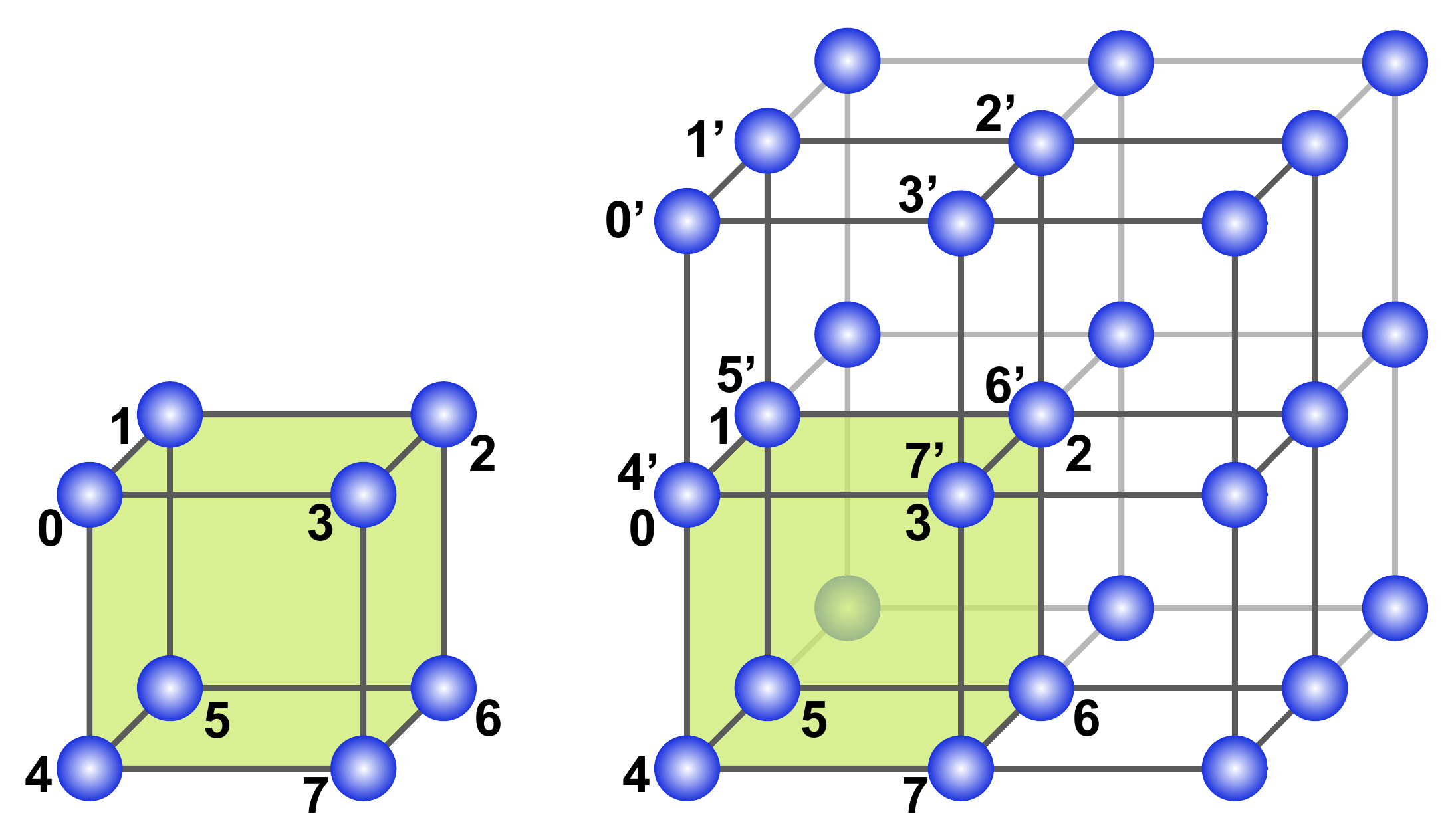}
    \label{fig:material_image_gathered}}
    \end{subfigure}
\hfill
    \begin{subfigure}[DA: expanded.]
    {\includegraphics[width=0.48\linewidth]{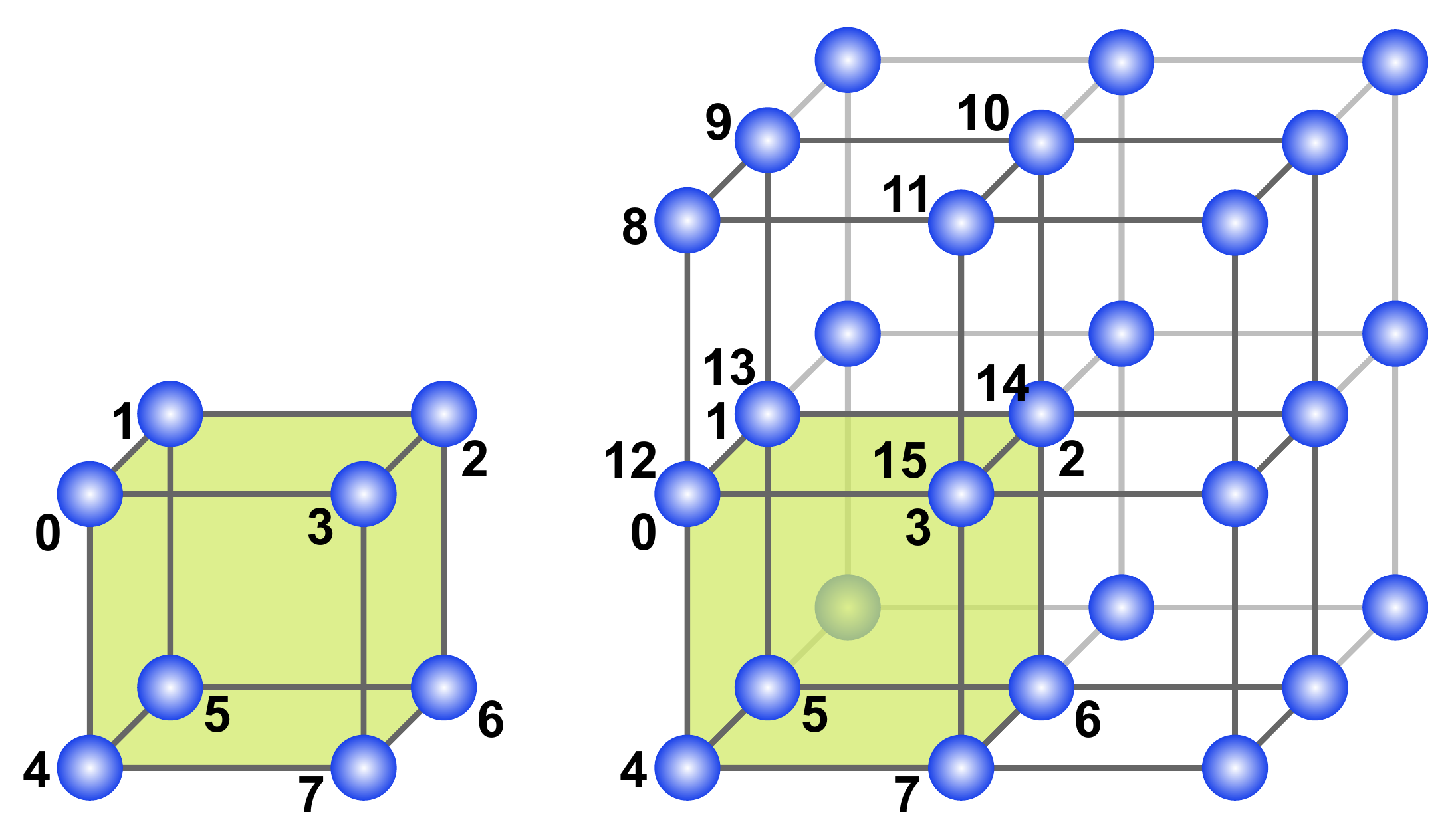}
    \label{fig:material_image_expanded}}
    \end{subfigure}
\vspace{-2ex}
\caption{\small
An illustration for crystalline material data augmentation (DA). Notice that in~\Cref{fig:material_image_gathered}, the shifted unit cells and the original unit cells share the same corresponding node indices; for the demonstration clarity, we mark them with $'$, {\eg}, ) $0$ and $0'$ are the indices for the same nodes.
}
\label{fig:material_DA_illustration}
\vspace{-2.5ex}
\end{figure}

\paragraph{Data augmentation 1: Gathered.} Gathered DA means that we will shift the original unit cell along three dimensions, and the translated unit cells will have the same node indices as the original unit cell. An example is in~\Cref{fig:material_image_gathered}.

\paragraph{Data augmentation 2: Expanded.} Expanded DA refers that we shift the original unit cell in the same way as Gathered, but the translated unit cells have different node indices from the original unit cell. An example is in~\Cref{fig:material_image_expanded}.

Once we have these two augmentations, we have the augmented nodes and corresponding periodic coordinates. The edge connection needs to satisfy three conditions simultaneously:
\begin{itemize}[noitemsep,topsep=0pt]
    \item The pairwise distance should be larger than 0 and no larger than the threshold $\tau$, {\ie}, the distance is within $(0, \tau)$.
    \item At least one of the linked nodes (bonded atoms) belongs to the anchor unit cell.
    \item No self-loop. % will double-check this
\end{itemize}

In specific, we give an example of the two DAs below. We take the same simple cubic crystal in~\Cref{fig:material_DA_illustration} for illustration, and we assume that the edge length in the unit cell is $l$. The threshold for building the edge is $\tau = l$.
\begin{itemize}[noitemsep,topsep=0pt]
    \item Gathered DA. $(0, 1)$ satisfies the conditions; $(0, 3')$ violets the condition; $(0, 4')$ violets the conditions; $(0', 1')$ violets the conditions.
    \item Expanded DA. $(0, 1)$ satisfies the conditions; $(0, 11)$ violets the conditions; $(0, 12)$ violets the condition; $(8, 9)$ violets the conditions.
\end{itemize}

In terms of implementation, this can be easily achieved by calling the pymatgen~\cite{ong2013python} package. Such data augmentation is merely one way of handling the periodic data structure in crystalline materials. There could be more potential ways, and we would like to leave them for future exploration.

\newpage
\section{Dataset Acquisition and Preparation \& Benchmark Hyperparameters} \label{sec:dataset_acquisition_and_specification_benchmark_hyperparameters}

For the dataset download, please check \href{https://github.com/chao1224/Geom3D}{this GitHub repository} for detailed instructions.

%%%%%%%%%%%%%%%%%%%%%%%%%%%%%%%%%%%%%%%%%%%%%%%%%%
\subsection{Small Molecules: QM9}

\textbf{Task specification.}
QM9~\cite{ramakrishnan2014quantum} is a dataset of 134K molecules, consisting of 9 heavy atoms. It includes 12 tasks that are related to the quantum properties. For example, U0 and U298 are the internal energies at 0K and 298.15K, respectively, and H298 and G298 are the other two energies that can be transferred from U298, respectively. The other 8 tasks are quantum mechanics related to the DFT process. 

\textbf{Task unit.}
We list the units for 12 QM9 tasks below.

\begin{table}[h!]
\setlength{\tabcolsep}{5pt}
\fontsize{9}{9}\selectfont
\centering
\caption{\small Units for 12 tasks in QM9.}
\vspace{-1.5ex}
\begin{adjustbox}{max width=\textwidth}
\begin{tabular}{cccccccccccc}
\toprule
$\alpha$ & $\nabla \mathcal{E}$ & $\mathcal{E}_\text{HOMO}$ & $\mathcal{E}_\text{LUMO}$ & $\mu$ & $C_v$ & $G$ & $H$ & $R^2$ & $U$ & $U_0$ & ZPVE\\
\midrule
$\alpha_0^3$ & $meV$ & $meV$ & $meV$ & $D$ & $\frac{cal}{mol\cdot K}$ & $meV$ & $meV$ & $\alpha_0^2$ & $meV$ & $meV$ & $meV$\\
\bottomrule
\end{tabular}
\end{adjustbox}
\end{table}

\textbf{Dataset size and split.}
There are 133,885 molecules in QM9, where 3,054 are filtered out, leading to 130,831 molecules. For data splitting, we use 110K for training, 10K for validation, and 11K for testing.

\textbf{Others.}
Current work is using different optimization strategies and different data splits (in terms of the splitting size). During the benchmark, we find that: (1) The performance on QM9 is very robust to either using (i) 110K for training, 10K for validation, 10,831 for test or using (ii) 100K for training, 13,083 for validation and 17,748 for test. (2) The optimization, especially the learning rate scheduler, is very critical. During the benchmarking, we find that using cosine annealing learning rate schedule~\cite{loshchilov2016sgdr} is generally the most robust.

%%%%%%%%%%%%%%%%%%%%%%%%%%%%%%%%%%%%%%%%%%%%%%%%%%
\subsection{Small Molecules: MD17}
\textbf{Task specification.}
MD17~\cite{chmiela2017machine} is a dataset on molecular dynamics simulation. It includes eight tasks, corresponding to eight organic molecules, and each task includes the molecule positions along the potential energy surface (PES), as shown in~\Cref{fig:PES_visualization}. The goal is to predict the energy-conserving interatomic forces for each atom in each molecule position.

\textbf{Task unit.}
The MD17 aims for energy and force prediction. The unit is $\frac{kcal}{mol}$ for energy and $\frac{kcal}{mol\cdot \text{\r{A}}}$ for force.

\textbf{Dataset size and split.}
We follow the literature~\cite{schutt2018schnet,klicpera2020fast,liu2021spherical,schutt2021equivariant} of using 1K for training and 1K for validation, while the test set (from 48K to 991K) is much larger, and we list them below.
\begin{table}[h!]
\centering
\caption{
\small
Dataset size and splits on MD17.
}
\vspace{-1.5ex}
\label{tab:sec:MD17_statistics}
\begin{adjustbox}{max width=\textwidth}
\begin{tabular}{l r r r r r r r r}
\toprule
Pretraining & Aspirin & Benzene & Ethanol & Malonaldehyde & Naphthalene & Salicylic & Toluene & Uracil \\
\midrule
Train & 1K & 1K & 1K & 1K & 1K & 1K & 1K & 1K\\
Validation & 1K & 1K & 1K & 1K & 1K & 1K & 1K & 1K\\
Test & 209,762 & 47,863 & 553,092 & 991,237 & 324,250 & 318,231 & 440,790 & 131,770\\
\bottomrule
\end{tabular}
\end{adjustbox}
\end{table}

\textbf{Others.}
There are multiple ways to predict the energy, {\eg}, using the SE(3)-equivariant to predict the forces directly. In \framework{}, we first predict the energy for each position; then, we take the gradient w.r.t. the input position. The Python codes are attached below:
\begin{lstlisting}[language=Python]
from torch.autograd import grad

positions = batch.positions # input positions
energy = model_3D(batch) # energy prediction
force = -grad(outputs=energy, inputs=positions) # force prediction
\end{lstlisting}
Notice that this holds for all the force prediction tasks, like rMD17 and COLL, which will be introduced below.

Additionally, in~\Cref{sec:app:ablation_study_data_normalization}, we will discuss the data normalization for MD prediction.

%%%%%%%%%%%%%%%%%%%%%%%%%%%%%%%%%%%%%%%%%%%%%%%%%%
\subsection{Small Molecules: rMD17}
\textbf{Task specification.}
The revised MD17 (rMD17) dataset~\cite{ChristensenAndersS2020Otro} is constructed based on the original MD17 dataset. 100K structures were randomly chosen for each type of molecule present in the MD17 dataset. Subsequently, the single-point force and energy calculations were performed for each of these structures using the PBE/def2-SVP level of theory. The calculations were conducted with tight SCF convergence and a dense DFT integration grid, significantly minimizing noise.

\textbf{Task unit.}
The rMD17 aims for energy and force prediction. The unit is $\frac{kcal}{mol}$ for energy and $\frac{kcal}{mol\cdot \text{\r{A}}}$ for force.

\textbf{Dataset size and split.}
We use 950 for training, 50 for validation, and 1000 for test.

%%%%%%%%%%%%%%%%%%%%%%%%%%%%%%%%%%%%%%%%%%%%%%%%%%
\subsection{Small Molecules: COLL}
\textbf{Task specification.}
COLL dataset~\cite{GasteigerJohannes2020FaUD} is a collection of configurations obtained from molecular dynamics simulations on molecular collisions. Around 140,000 snapshots were randomly taken from the trajectories of the collision, for each of which the energy and force were calculated using density functional theory (DFT). 

\textbf{Task unit.}
The rMD17 aims for energy and force prediction. The unit is $eV$ for energy and $eV/\text{\r{A}}$ for force.

\textbf{Dataset size and split.}
The published COLL dataset has split the whole data into 120,000 training samples, 10,000 validation samples, and 9,480 testing samples.

%%%%%%%%%%%%%%%%%%%%%%%%%%%%%%%%%%%%%%%%%%%%%%%%%%
\subsection{Small Molecules \& Proteins: LBA \& LEP}
\textbf{Task specification.}
Ligand-protein binding is formed between a small molecule (ligand) and a target protein. During the binding process, there is a cavity in a protein that can potentially possess suitable properties for binding a small molecule, called pocket~\cite{stank2016protein}. Due to the large volume of protein, \framework{} follows existing works~\cite{townshend2020atom3d} by only taking the binding pocket, where there are no more than 600 atoms for each molecule and protein pair. For the benchmarking, we consider two binding affinity tasks. (1) The first task is ligand binding affinity (LBA)~\cite{wang2004pdbbind}. It is gathered from~\cite{wang2005pdbbind}, and the task is to predict the binding affinity strength between a small molecule and a protein pocket. (2) The second task is ligand efficacy prediction (LEP)~\cite{friesner2004glide}. We have a molecule bounded to pockets, and the goal is to detect if the same molecule has a higher binding affinity with one pocket compared to the other one.

\textbf{Task unit.}
LBA is to predict $pK = -\log(K)$, where K is the binding affinity in Molar units. LEP has no unit since it is a classification task.

\textbf{Dataset size and split.}
The dataset size and splitting are listed below.

\begin{table}[h!]
\setlength{\tabcolsep}{5pt}
\fontsize{9}{9}\selectfont
\centering
\caption{
\small
Dataset size and splits on LBA \& LEP.
For LBA, we use split-by-sequence-identity-30: we split protein-ligand complexes such that no protein in the test dataset has more than 30\% sequence identity with any protein in the training dataset.
For LEP, we split the complex pairs by protein target.
}
\vspace{-1.5ex}
\begin{adjustbox}{max width=\textwidth}
\begin{tabular}{l r r}
\toprule
Pretraining & LBA & LEP\\
\midrule
Train & 3,507 & 304\\
Validation & 466 & 110\\
Test & 490 & 104\\
Split & split-by-identity-30 & split-by-target\\
\bottomrule
\end{tabular}
\end{adjustbox}
\end{table}

%%%%%%%%%%%%%%%%%%%%%%%%%%%%%%%%%%%%%%%%%%%%%%%%%%
\subsection{Proteins: EC}
\paragraph{Task Specification EC}
The Enzyme Commission(EC) Number is a numerical classification of enzymes according to the catalyzed chemical reactions~\cite{HuQian-Nan2012AoEn}. Therefore, the functions of enzymes and the chemical reaction type they catalyze can be represented by different EC numbers. An example of EC number is $EC 3.1.1.4$: $3$ represents Hydrolases (the first number represents enzyme class); $3.1$ represents Ester Hydrolases (the second number represents enzyme subclass); $3.1.1$ represents Carboxylic-ester Hydrolases (the third number represents enzyme sub-subclass); $3.1.1.4$ represents Phospholipases (the fourth number represents the specific enzyme). The EC dataset was constructed by Hermosilla et al.~\cite{HermosillaPedro2020ICaP} for the protein function prediction task. The enzyme reaction data with Enzyme Committee annotations were originally collected from the SIFTS database~\cite{DanaJoseM2019SuSI}. Then, all the protein chains were clustered using a 50\% similarity threshold. EC numbers that were annotated for at least five clusters were selected and five proteins with less than 100\% similarities were selected from each cluster, annotated by the EC number.

\textbf{Task unit.}
No unit is available since it is a classification task.

\textbf{Dataset size and split.}
EC contains 37,428 protein chains, which were split into 29,215 for training, 2,562 for validation, and 5,651 for testing.

%%%%%%%%%%%%%%%%%%%%%%%%%%%%%%%%%%%%%%%%%%%%%%%%%%
\subsection{Proteins: FOLD}
\textbf{Task specification.}
Proteins can be hierarchically divided into different levels: Family, Superfamily, and Fold based on their sequence similarity, structure similarity, and evolutionary relations~\cite{MurzinAlexeyG.1995SAsc}. Proteins with (1) $\geq$30\% residue identities or (2) lower residue identities but have similar functions are grouped into the same Family. A Superfamily is for families whose proteins have low residue identities but their structural and functional features suggest a possible same evolutionary origin. A Fold is for proteins sharing the same major secondary structures with the same arrangement and topological connections.

Based on the SCOP 1.75 database, all the fold categories can be grouped into seven structural classes with in total of 1195 fold types~\cite{LinChen2013Hcop}: (a) all $\alpha$ proteins (primarily formed by $\alpha-$helices, 284 folds), (b) all $\beta$ proteins (primarily formed by $\beta-$sheets, 174 folds), (c) $\alpha$/$\beta$ proteins ($\alpha$-helices and $\beta$-strands interspersed, 147 folds), (d) $\alpha+\beta$ proteins ($\alpha$-helices and $\beta$-strands segregated, 376 folds), (e) multi-domain proteins (66 folds), (f) membrane and cell surface proteins and peptides (58 folds), and (g) small proteins (90 folds). DeepSF~\cite{HouJie2018Ddcn} proposed a three-level redundancy removal at fold/superfamily/family levels, resulting in three subsets for testing. 
\begin{itemize}[noitemsep,topsep=0pt]
\item \textbf{Fold testing set} Firstly, the proteins are split into Fold-level training set and testing set, where the training set and testing set don’t share the same superfamily.
\item \textbf{Superfamily testing set} Then, the Fold-level training set is split into Superfamily-level training set and testing set, where they don’t share the same family.
\item \textbf{Family testing set} Finally, the Superfamily-level training set is split into Family-level training set and
testing set, where for proteins in the same family, 80\% of them are used for training and 20\% of them are used for testing.
\end{itemize}

\textbf{Task unit.}
No unit is available since they are classification tasks.

\textbf{Dataset size and split.}
FOLD contains 16,292 proteins, and we follow \cite{HouJie2018Ddcn}: 12,312 training samples, 736 validation samples, 3,244 testing samples. The testing samples contain 3 sub testsets: 718 for folding testset, 1,254 for superfamily testset, and 1,272 for family testset.

%%%%%%%%%%%%%%%%%%%%%%%%%%%%%%%%%%%%%%%%%%%%%%%%%%
\subsection{Crystalline Materials: MatBench}
\textbf{Task specification.}
MatBench~\cite{DunnAlexander2020Bmpp} is a test suite for benchmarking 13 machine learning model performances for predicting different material properties. The dataset size for these tasks varies from 312 to 132k. The MatBench dataset has been pre-processed to clean up the task-irrelevant and unphysical-computed data. For benchmarking, we take 8 regression tasks with crystal structure data.
These tasks are~\cite{jain2013commentary,dunn2020benchmarking,de2021robust} Formation energy per Perovskite cell (Per. $E_{\text{form}}$), Refractive index (Dielectric), Shear modulus ($log_{10} G$), Bulk modulus($log_{10} K$), exfoliation energy ($E_{\text{exfo}}$), frequency at last phonon PhDOS peak (Phonons), formation energy ($E_{\text{form}}$), and band gap (Band Gap). Detailed explanations are as below:
\begin{itemize}[noitemsep,topsep=0pt]
    \item Perovskites: predicting formation energy from the crystal structure.
    \item Dielectric: predicting refractive index from the crystal structure.
    \item $log_{10} G$: predicting DFT log10 VRH-average shear modulus from crystal structure.
    \item $log_{10} K$: predicting DFT log10 VRH-average bulk modulus from crystal structure.
    \item $E_{\text{exfo}}$: predicting exfoliation energies from the crystal structure.
    \item Phonons: predicting vibration properties from the crystal structure.
    \item $E_{\text{form}}$: predicting DFT formation energy from the crystal structure.
    \item Band Gap: predicting DFT PBE band gap from the crystal structure.
\end{itemize}

\textbf{Task unit.}
The unit for each task is listed below.

\textbf{Dataset size and split.}
The dataset size for each task is listed above. For benchmarking, we take 60\%-20\%-20\% as training-validation-testing for all 8 tasks.

\begin{table}[h!]
\setlength{\tabcolsep}{5pt}
\fontsize{9}{9}\selectfont
\centering
\caption{
\small
Unit, dataset size, and naming specifications for MatBench.
}
\vspace{-1.5ex}
\begin{adjustbox}{max width=\textwidth}
\begin{tabular}{l rrrrrrrrr}
\toprule
Column in MatBench & Perovskites  & Dielectric  & log gvrh & log kvrh  & jdft2d  & Phonons  & E Form  & Band Gap \\
\midrule
Task Name in~\Cref{tab:main_result_material} & Per. $E_{\text{form}}$ & Dielectric  & $log_{10} G$ & $log_{10} K$ & $E_{\text{exfo}}$  & Phonons  & $E_{\text{form}}$ & Band Gap \\
Size & 18,928 & 4,764 & 10,987 & 10,987 & 636 & 1,265 & 132,752 & 106,113\\
Unit & $eV$ & -- & ~~ $log_{10}$ GPa & ~~ $log_{10}$ GPa & $meV$ & $\text{cm}^{-1}$ & $eV$/atom & $eV$\\
\bottomrule
\end{tabular}
\end{adjustbox}
\end{table}

%%%%%%%%%%%%%%%%%%%%%%%%%%%%%%%%%%%%%%%%%%%%%%%%%%
\subsection{Crystalline Materials: QMOF}
\textbf{Task specification.}
QMOF~\cite{rosen2021machine} is a database containing 20,425 metal–organic frameworks (MOFs) with quantum-chemical properties generated using density functional theory (DFT) calculations. The task is to predict the band gap, the energy gap between the valence band and the conduction band.

\textbf{Task unit.}
The unit for the band gap task is $eV$.

\textbf{Dataset size and split.}
As mentioned above, there are 20,425 MOFs, and we take 80\%-10\%-10\% for training-validation-testing.

\newpage
\section{Group Representation and Equivariance}
Symmetry is everywhere on Earth, such as in animals, plants, and molecules. The group theory is the most expressive tool to depict such physical symmetry. In this section, we would like to go through certain key concepts in group theory.

\textbf{Symmetry} is the collection of all transformations under which an object is invariant. The readers can easily check that these transformations are automatically invertible and form a  group, where the group multiplication is identified with the composition operation of two transformations. From a dynamical system point of view, symmetries are essential for reducing the degree of freedom of a system. For example, Noether's first theorem states that every differentiable symmetry of a physical system with conservative forces has a corresponding conservation law \cite{physics/0503066}. Therefore, symmetries form an important source of inductive bias that can shed light on the design of neural networks for modeling physical systems.

%%%%%%%%%%%%%%%%%%%%%%%%%%%%%%%%%%%%%%%%%%%%%%%%%%
\subsection{Group}
A \textbf{group} is a set $G$ equipped with an operator (group product) $\times$, and they need to follow three rules:
\begin{enumerate}[noitemsep,topsep=0pt]
    \item It contains an identity element $\ve \in G$, s.t. $\va \ve = \ve \va = \va, \forall \va \in G$.
    \item Associativity rule $(\va \vb) \vc = \va(\vb \vc)$.
    \item Each element has an inverse $\va \va^{-1} = \va^{-1}\va = \ve$.
\end{enumerate}

Below we list several well-known groups:
\begin{itemize}[noitemsep,topsep=0pt]
    \item \textbf{O(n)} is an n-dimensional \textbf{orthogonal group} that consists of rotation and reflections.
    \item \textbf{SO(n)} is a \textbf{special orthogonal group} that only consists of rotations.
    \item \textbf{E(n)} is an n-dimensional \textbf{Euclidean group} that consists of rotations, translations, and reflections.
    \item \textbf{SE(n)} is an n-dimensional \textbf{special Euclidean group}, which comprises arbitrary combinations of rotations and translations (no reflections). 
    \item \textbf{Lie Group} is a group whose elements form a differentiable manifold. All the groups above are specific examples of the Lie Group.
\end{itemize}

%%%%%%%%%%%%%%%%%%%%%%%%%%%%%%%%%%%%%%%%%%%%%%%%%%
\subsection{Group Representation and Irreducible Group Representation}

\textbf{Group representation} is a mapping from the group $G$ to the group of linear transformations of a vector space $X$ with dimension $d$ (see~\cite{zee2016group} for more rigorous definition):
\begin{equation} \small
\rho_X(\cdot) : G \to \mathbb{R}^{d \times d}
\quad\quad \text{s.t.} \quad \rho(\ve) = 1 ~~ \wedge~~ \rho_X(\va) \rho_X(\vb) = \rho_X(\va \times \vb), ~\forall \va, \vb \in G.
\end{equation}
During modeling, the $X$ space can be the input 3D Euclidean space, the equivariant vector space in the intermediate layers, or the output force space. This enables the definition of equivariance as in~\Cref{sec:app:equivariance_and_invariance}.

Group representation of SO(3) can be applied to any n-dimensional vector space. If we map SO(3) to the 3D Euclidean space ({\ie}, $n=3$), the group representation has the same formula as the rotation matrix.

\textbf{Irreducible representations of rotations}
The irreducible representations (irreps) of SO(3) are indexed by the integers 0, 1, 2, ..., and we call this index $l$. The $l$-irrep is of dimension $2l+1$. $l=0$ (dimension 1) corresponds to scalars and $l=1$ (dimension 3) corresponds to vectors.

%%%%%%%%%%%%%%%%%%%%%%%%%%%%%%%%%%%%%%%%%%%%%%%%%%
\subsection{Equivariance and Invariance} \label{sec:app:equivariance_and_invariance}

\textbf{Equivariance} is the property for the geometric modeling function $f: X \to Y$, and we want to design a function $f$ that is equivariant as:
\begin{equation}\label{eq:app:equivariance}
f(\rho_X(\va) \vx) = \rho_Y(\va) f(\vx), ~~\forall \va \in G, \vx \in X.
\end{equation}
How to understand this in the molecule discovery scenarios? $\rho_X(g)$ is the group representation on the input space, like atom coordinates; and $\rho_Y(g)$ is the group representation on the output space $Y=f(X)$, {\eg}, the force field space. Equivariance modeling in~\Cref{eq:app:equivariance} is essentially saying that the designed deep learning model $f$ is modeling the whole transformation trajectory ({\eg}, rotation for SO(3)-group) on the molecule conformations, and the output is the transformed $\hat y$ accordingly.

Note that in deep learning, a function with learned parameters can be abstracted as $f: W \times X \to Y$, where $w \in W$ is a choice of learned parameters (or weights). The parameters are scalars, {\ie}, they don't transform under a transformation of E(3)/SE(3). This implies that weights are scalars and are invariant under any choice of coordinate system.

\textbf{Invariance} is a special type of equivariance where 
\begin{equation}\label{eq:app:invariance}
f(\rho_X(\va) \vx) = f(\vx), ~~\forall \va \in G, \vx \in X,
\end{equation}
with $\rho_Y(g)$ as the identity $\forall g \in G$.

Thus, group and group representation help define the equivariance condition for $f$ to follow.  Then, the question turns to how to design such invariant or equivariant $f$.
\begin{itemize}[noitemsep,topsep=0pt]
    \item In~\Cref{sec:invariant_geometric_representation}, we introduced the invariant geometric models.
    \item In~\Cref{sec:equivariant_geometric_representation}, we briefly discussed two main categories of equivariant geometric models: the spherical frame basis model and the vector frame basis model. In the following, we will introduce both in more detail in~\Cref{sec:app:equicariance_with_spherical_frame_basis,sec:app:equicariance_with_vector_frame_basis}, respectively.
\end{itemize}

Through lifting from the original geometric space to its frame bundle (see \cite{hsu2002stochastic} for the precise definition), equivariant operations like covariant derivatives are realized in an invariant way. From a practical perspective, the lifting operation can be alternatively replaced by scalarization by equivariant frames. See \cite{du2022se,2304.04757} for an illustration. 
Therefore, invariance and equivariance are just two equivariant descriptions of characterizing symmetry that can be transformed into each other through frames.

One thing we want to highlight is that convolutional neural networks (CNNs) on images are translation-equivariant on $\mathbb{R}^2$, which demonstrates the power of encoding symmetry into the deep neural network architectures.\looseness=-1

\newpage
%%%%%%%%%%%%%%%%%%%%%%%%%%%%%%%%%%%%%%%%%%%%%%%%%%
\section{Equivariance with Spherical Frame Basis} \label{sec:app:equicariance_with_spherical_frame_basis}
First, we would like to give a high-level idea of this basis:
\begin{itemize}[noitemsep,topsep=0pt]
    \item It introduces the spherical harmonics as the basis and maps all the points into such a space.
    \item The mapping from 3D Euclidean space to the spherical harmonics space satisfies the E(3)/SE(3)-equivariance property as defined~\Cref{eq:app:equivariance}.
    \item Based on such basis, we can design a message-passing framework to learn the desired properties.
\end{itemize}

Then, we would like to refer to Figure 2 in SEGNN \href{https://arxiv.org/abs/2110.02905}{ArXiv version v3}. It nicely illustrates how the equivariance works in the spherical harmonics space.

\paragraph{Spherical Harmonics}
The spherical harmonics are functions from points on the sphere to vectors, or more rigoriously:
\begin{definition}
The spherical harmonics are a family o functions $Y^l$ from the unit sphere to the irrep $D^l$. For each $l=0,1,2...$, the spherical harmonics can be seen as a vector of $2l+1$ functions $Y^l(\vec x) = \big( Y^{l}_{-l}(\vec x), Y^{l}_{-l+1}(\vec x), ..., Y^{l}_{l}(\vec x) \big)$. Each $Y^l$ is equivariant to SO(3) with respect to the irrep of the same order, {\ie},
\begin{equation} \label{eq:app:spherical_harmonics}
\small
Y_m^l(R \vec x) = \sum_{n=-1}^l D^l(R)_{mn} Y_n^l(\vec x),
\end{equation}
where $R$ is any rotation matrix and $D^l$ are the irreducible representation of SO(3). They are normalized $\| Y^l(\vec x) = 1$ when evaluated on the sphere $\| \vec x \| = 1$.
\end{definition}

According to~\Cref{eq:app:equivariance}, \Cref{eq:app:spherical_harmonics} satisfies the equivariance property: the input space $X$ is the 3D Euclidean space, and the output space $Y$ is the Spherical Harmonics space.

Some key points we would like to highlight:
\begin{itemize}[noitemsep,topsep=0pt]
    \item Sphere $\mathbb{S}^2$ is not a group, but it is a homogeneous space of SO(3).
    \item The decomposition into the irreducible group representations makes it steerable.
    \item The parameter $l$ is named the \textbf{rotation order}.
\end{itemize}

\paragraph{Model Design}
With the spherical basis, we can design our own geometric models. Notice that during the modeling process, all the variables are tensors.

For instance, we can take the vector $\vr_j-\vr_i$ as the vector in $Y_m^l (\frac{\vr_j - \vr_i}{\| \vr_j - \vr_i \|})$. As shown in~\Cref{eq:app:spherical_harmonics}, this is rotation-equivariant. And we can easily see $\vr_j-\vr_i$ is translation-equivariant. 

This term can be naturally adopted for the edge embedding under the message passing framework~\cite{gilmer2017neural}, and we can parameterize it with a radial term~\cite{thomas2018tensor} as:
\begin{equation}
\small
\vh_{i,j} = \text{Radial} (\|\vr_j - \vr_i\|) Y_m^l (\frac{\vr_j - \vr_i}{\| \vr_j - \vr_i \|}),
\end{equation}
where the radial function is invariant with the pairwise distance as the input. This is for the message function. Then generally, for the update and aggregate function of node-level tensor $\vv_i$, we have two options:
\begin{equation}
\small
\vv_i = \begin{cases}
    \vv_i + \sum_{j \in \mathcal{N}(i)} \vh_{i,j} + \vv_j\\
    \vv_i + \sum_{j \in \mathcal{N}(i)} \vh_{i,j} \otimes \vv_j,
\end{cases}
\end{equation}
where the update can be done either with plus or multiplication. Note that $\otimes$ is the tensor product, which can be calculated using the \textit{Clebsch-Gordan coefficients}. Please refer to~\cite{geiger2022e3nn} for more details.

\newpage
%%%%%%%%%%%%%%%%%%%%%%%%%%%%%%%%%%%%%%%%%%%%%%%%%%
\section{Equivariance with Vector Frame Basis} \label{sec:app:equicariance_with_vector_frame_basis}
In physics, the vector frame is equivalent to the coordinate system. For example, we may assign a frame to all observers, although different observers may collect different data under different frames, the underlying physics law should be the same. In other words, denote the physics law by $f$, then $f$ should be an equivariant function.

Since there are three orthogonal directions in $\mathbf{R}^3$, a vector frame in $\mathbf{R}^3$ consists of three orthogonal vectors:
$$F = (\ve_1,\ve_2,\ve_3).$$ Once equipped with a vector frame (coordinate system), we can project all geometric quantities to this vector frame. For example, an abstract vector $\vr \in \mathbf{R}^3$ can be written as $\vr = (r_1,r_2,r_3)$ under vector frame $F$, if: $\vr = r_1 \ve_1 + r_2 \ve_2 + r_3 \ve_3.$
An equivariant vector frame further requires the three orthonormal vectors in $(\ve_1,\ve_2,\ve_3)$ to be equivariant. Intuitively, an equivariant vector frame will transform  according to the global rotation or translation of the whole system. Once equipped with an equivariant vector frame, we can project equivariant vectors into this vector frame:
\begin{equation} \label{projection}
\vr = \Tilde{r}_1 \ve_1 + \Tilde{r}_2 \ve_2 + \Tilde{r}_3 \ve_3.    
\end{equation}
We call the process of $\vr \rightarrow \Tilde{r}:= (\Tilde{r}_1,\Tilde{r}_2,\Tilde{r}_3)$ the \textbf{projection} operation. Since $\Tilde{r}_i = \ve_i \cdot \vr_i$ is expressed as an inner product between equivariant vectors, we know that $\Tilde{r}$ consists of scalars. 

To incorporate equivariant frames with graph message passing, we assign an equivariant vector frame to each node/edge. Therefore, we call them the local frames. For example, consider node $i$ and one of its neighbors $j$ with positions $\vx_i$ and $\vx_j$, respectively. The orthonormal equivariant frame $\mathcal{F}_{ij} := (\ve^{ij}_1,\ve^{ij}_2,\ve^{ij}_3)$ in \textbf{Clofnet}~\cite{du2022se} is defined with respect to $\vx_i$ and $\vx_j$ as follows:
\begin{equation}
(\frac{\vx_i - \vx_j}{\norm{\vx_i - \vx_j}}, \frac{\vx_i \times \vx_j}{\norm{\vx_i \times \vx_j}},\frac{\vx_i - \vx_j}{\norm{\vx_i - \vx_j}} \times \frac{\vx_i \times \vx_j}{\norm{\vx_i \times \vx_j}}).
\end{equation}
Note that this frame is translation invariant if the system's mass is set to zero during data prepossessing. On the other hand, \textbf{MoleculeSDE}~\cite{liu2023moleculeSDE} implemented the \textbf{Clofnet}'s output layers for transforming 2D representation into  3D equivariant output. Finally, it's worth mentioning that global frames can be built by pooling local frames. For example, a graph-level equivariant frame is obtained by aggregating node frames and implementing the Gram-Schmidt orthogonalization. However, the Newton dynamics experiments in~\cite{du2022se} demonstrated that the global frame's performance is worse than edge local frames. Therefore,  although edge-, node-, and global- frames are equal in terms of equivariance, the optimization properties of different equivariant frames depend varies according to different scientific datasets.

\newpage
%%%%%%%%%%%%%%%%%%%%%%%%%%%%%%%%%%%%%%%%%%%%%%%%%%
\section{Other Geometric Modeling (Featurization and Lie Group)} \label{sec:app:other_equivariant_modeling}
We also want to acknowledge other equivariant modeling methods.

\textbf{Featurization.}
OrbNet~\cite{qiao2020orbnet} models the atomic orbital, the description of the location and wave-like behavior of electrons in atoms. This possesses a finer-grained featurization level than other methods.
Voxel means that we discretize the 3D Euclidean space into bins, and recent work~\cite{flam2023language} empirically shows that this also applies to geometry learning tasks.

\textbf{Equivariance modeling with Lie group.}
In previous sections, equivariant algorithms are viewed as mappings from a 3D point cloud (which discretizes the 3D Euclidean space) to another 3D point cloud, or as mappings to invariant quantities. From this point of view, the symmetry group $E(3)/ SE(3)$ manifests itself as a group action transforming the Euclidean space.  However, it is worth noting that this action is transitive in the sense that any two points in 3D Euclidean space can be transformed from one to the other through a combination of translation and rotation. In mathematical terms, the 3D Euclidean space is a \textbf{homogeneous space} of the group $E(3)$. Exploiting this observation, LieConv~\cite{finzi2020generalizing} and LieTransformer~\cite{hutchinson2021lietransformer} elevate the 3D point cloud to the $E(3)$ group and perform parameterized group convolution (and attention) operations, ensuring equivariance, to obtain an equivariant embedding on the group $E(3)$. Finally, by projecting the result back to $\mathbf{R}^3$ (taking the quotient), an equivariant map from $\mathbf{R}^3$ to the output space is obtained. The main limitation of Lie group modeling lies in the convolution operation, which often involves high-dimensional integration and requires approximation for most groups. For more in-depth insights into the properties of convolution on groups, we refer readers to \cite{bronstein2021geometric}. Another lifting of $\mathbf{R}^3$ is to lift it to the SO(3) frame bundle, such that the SO(3) group transforms one orthonormal frame to another orthonormal frame transitively. This lifting also inspires the design of \cite{du2022se, 2304.04757}.

\newpage
\section{Expressive Power: from Invariance to Equivariance}

Equivariant neural networks are constructed for equivariant tasks. That is, to approximate an equivariant function. Comparing with ordinary neural networks, a natural question arises:  \emph{Does an equivariant neural network have the universal approximation property whiten the equivariant function class?}. By the novel D-spanning concept~\cite{dym2020universality}, this question is partially answered. The author further proposed two types of equivariant architectures that can enjoy the D-spanning property: 1. the G-equivariant polynomials enhanced TFN; 2. the minimal universal architecture constructed by tensor products. Therefore, at least in terms of universal approximation, an equivariant neural network doesn't necessarily require irreducible representations and Clebsch-Gordan decomposition. The reader can check \cite{du2022se} for how to realize the minimal universal architecture in an invariant way through equivariant frames and tensorized graph neural networks ({\eg}, \cite{1905.04943}). Informally, we conclude that an invariant graph neural network equipped with a powerful message-passing mechanism can achieve 
the universal approximation property. Another proof strategy of the universality of invariant scalars that doesn't rely on theories of tensorized graph neural networks can be found in \cite{2106.06610}.

However, the mainstream GNN is usually based on a 1-hop message passing mechanism (although tensorized graph neural networks have empirically shown competitive performances in molecular tasks) for computational efficiency. For 1-hop message passing mechanisms (including node-based transformers), our previous conclusion no longer holds, and vector (or higher order tensors) updates are necessary for enhancing the expressiveness power. The reader can consult the concrete example from PaiNN~\cite{schutt2021equivariant} to illustrate this point.  

More precisely, We denote the nodes in Figure 1 of \cite{schutt2021equivariant} as $\{a: white, b: blue, c: red, d: white\}$, and we consider whether the message of $b$ and $c$ received from their 1-hop neighbors can discriminate the two different geometric structures. For node $b$, the invariant geometric information we can get from 1-hop neighbors are the relative distance $d_{ab}$ and $d_{bc}$ and their intersection angle $\alpha_1$. Since the relative distances of the two structures remain equal, only the angle information is useful. Similarly, for node $c$, we have the intersection angle $\alpha_2$. Unfortunately, the intersection angles $\alpha_1$ and $\alpha_2$ of the two structures are still the same, and we conclude that invariant features are insufficient for discriminating the two different structures. On the other hand, \cite{schutt2021equivariant} showed that by introducing directional vector features (type-1 equivariant steerable features), we are able to solve the problem in this special case, which proves the superiority of 'equivariance' over 'invariance' within 1-hop message passing mechanisms. Another invariant way of filling in this type of expressiveness gaps systematically is to introduce the information of frame transitions \textbf{FTE}, as was demonstrated in \cite{2304.04757}.

Vector update is just a special case of the more general higher-order tensor updates. To merge general equivariant tensors into our GNN, we can either utilize tensor products of vector frames \cite{2304.04757}, or introduce the concepts of spherical harmonics, which form a complete basis in the sense of irreducible representations of group $SO(3)$ and $O(3)$. However, to express the output of the tensor product between  spherical harmonics as a combination of  spherical harmonics is nontrivial. Fortunately, this procedure has been studied by quantum physicists, which is named after the Clebsch-Gordan decomposition (coefficients) \cite{1907.09930}. Combining these blocks, we can build convolution or attention-based equivariant graph neural networks, see \cite{geiger2022e3nn,liao2022equiformer} for detailed constructions.

\newpage
\section{Architecture for Geometric Representation} \label{sec:architecture}

In this section, we are going to give a brief review of certain advanced geometric models, and a summary of more methods can be found in~\Cref{tab:geometric_modeling_comparison_complete}. Meanwhile, we will keep updating more advanced models.

We include all the hyperparameters in \href{https://github.com/chao1224/Geom3D}{this GitHub repository}. We are sure that we won't be able to tune all the hyperparameters, yet we want to claim that our reported results are reproducible using the hyperparameters listed above. In the future, we appreciate any contribution to do more searching on this.

\begin{table}[h]
\centering
\caption{\small Categorization on geometric methods. For pretraining methods, the categorization is based on the pretraining algorithms and backbone models are not considered.}
\label{tab:geometric_modeling_comparison_complete}
\vspace{-1.5ex}
\begin{adjustbox}{max width=\textwidth}
\begin{tabular}{l l c cc}
\toprule
& \multirow{2}{*}{Model} & \multirow{2}{*}{Invariance} & \multicolumn{2}{c}{Equivariance} \\
\cmidrule(lr){4-5}
& & & Spherical Frame & Vector Frame\\
\midrule
\multirow{17}{*}{Representation}
& SchNet~\cite{schutt2018schnet} & \checkmark \\
& DimeNet~\cite{klicpera2020fast} & \checkmark \\
& SphereNet~\cite{liu2021spherical} & \checkmark \\
& GemNet~\cite{klicpera_gemnet_2021} & \checkmark \\
& IEConv~\cite{HermosillaPedro2020ICaP} & \checkmark\\
& GearNet~\cite{ZhangZuobai2022PRLb} & \checkmark\\
& ProNet~\cite{WangLimei2022LHPR} & \checkmark \\
\cmidrule(lr){2-5}
& TFN~\cite{fuchs2020se} & & \checkmark \\
& SEGNN~\cite{brandstetter2021geometric} & & \checkmark \\
& SE(3)-Trans~\cite{fuchs2020se} & & \checkmark \\
& NequIP~\cite{batzner20223} & & \checkmark \\
& Allegro~\cite{musaelian2022learning} & & \checkmark \\
& Equiformer~\cite{liao2022equiformer} & & \checkmark \\
\cmidrule(lr){2-5}
& EGNN~\cite{satorras2021n} & & & \checkmark\\
& PaiNN~\cite{schutt2021equivariant} & & & \checkmark\\
& GVP-GNN~\cite{JingBowen2020LfPS} & & & \checkmark\\
& CDConv~\cite{fan2023cdconv} & & & \checkmark \\
\midrule
\multirow{9}{*}{Pretraining}
& GraphMVP~\cite{liu2022pretraining} & \checkmark\\
& 3D InfoMax~\cite{stark20223d} & \checkmark\\
& GeoSSL-RR~\cite{liu2023molecular} & \checkmark\\
& GeoSSL-EBM-NCE~\cite{liu2023molecular} & \checkmark\\
& GeoSSL-InfoNCE~\cite{liu2023molecular} & \checkmark\\
& GeoSSL-DDM~\cite{liu2023molecular} & \checkmark\\
& GeoSSL-DDM-1L~\cite{zaidi2022pre} & \checkmark\\
& 3D-EMGP~\cite{jiao20223d} & & & \checkmark \\
& MoleculeSDE~\cite{liu2023moleculeSDE} & & & \checkmark \\
\bottomrule
\end{tabular}
\end{adjustbox}
\end{table}

Generally, all the algorithms can be classified into two categories: SE(3)-invariant and SE(3)-equivariant. Note that, rigorously, SE(3)-invariant is also SE(3)-equivariant. Here we follow the definition in~\cite{geiger2022e3nn}\footnote{Also a video by Tess et al, link is \href{https://www.youtube.com/watch?v=q9EwZsHY1sk}{here}.}:
\begin{itemize}[noitemsep,topsep=0pt]
    \item SE(3)-invariant models only operate on scalars ($l=0$) which interact thought simple scalar multiplication. These scalars include pairwise distance, triplet-wise angle, etc, that will not change under rotation. In other words, the SE(3)-invariant pre-compute the invariant features and throw away the coordinate system.
    \item SE(3)-equivariant models keep the coordinate system and if the coordinate system changes, the outputs change accordingly. These models have been believed to empower larger model capacity~\cite{geiger2022e3nn,batzner20223} with $l>0$ quantities.
\end{itemize}

There are other variants, like the activation functions, the number of layers, normalization layers, etc. In this section, we will stick to the key module, {\ie}, the SE(3)-invariant and SE(3)-equivariant modules for each backbone model.

The aggregation function is the same as:
\begin{equation}
\small
\begin{aligned}
h_{i}' & = \text{aggregate}_{j \in \mathcal{N}(i)}(m_{ij}).
\end{aligned}
\end{equation}
In the following, we will be mainly discussing the message-passing function as below.

%%%%%%%%%%%%%%%%%%%%%%%%%%%%%%%%%%%%%%%%%%%%%%%%%%
\subsection{Invariant Models}
\paragraph{SchNet}
SchNet~\cite{schutt2018schnet} simply handles a molecule by feeding in the pairwise distance and throws them into the message-passing style GNN.
\begin{equation}
\small
\begin{aligned}
m_{ij} & = \text{MLP}(h_j, \textbf{RBF}(d_{ij})).
\end{aligned}
\end{equation}
where $\textbf{RBF}(\cdot)$ is the RBF kernel.

\paragraph{DimeNet}
The directional message passing neural network (DimeNet and DimeNet++)~\cite{klicpera2020directional}. The message passing function in DimeNet is two-hop instead of one-hop. Such message-passing step is similar to directed message-passing neural network (D-MPNN)~\cite{yang2019analyzing}, and it can reduce the redundancy during the message passing process.
\begin{equation} \label{eq:DimeNet_message_passing}
\small
\begin{aligned}
m_{ji}^{l+1} & = \sum_{k \in \mathcal{N}_{j} \textbackslash \{i\}} \text{MLP}(m_{ji}^l, \textbf{RBF}(d_{ji}), \textbf{SBF}(d_{kj}, \alpha_{\angle kij})),
\end{aligned}
\end{equation}
where $\text{SBF}_{ln}(d_{kj}, \alpha_{\angle kij}) = \sqrt{\frac{2}{c^3 j^2_{l+1}(z_{ln})}} j_l(\frac{z_{ln}}{c}d_{kj}) Y_{l}^0(\alpha)$ is the spherical Fourier-Bessel (spherical harmonics) basis, a joint 2D basis for distance $d_{kj}$ and angle $\alpha_{\angle kij}$.

\paragraph{SphereNet}
SphereNet~\cite{liu2021spherical} is an extension of DimeNet by further modeling the dihedral angle. It first adopts the spherical Fourier-Bessel (spherical harmonics) basis for dihedral angle modeling, namely
\begin{equation} \label{eq:SPhereNet_SBF_message_passing}
\small
\begin{aligned}
\text{SBF}(d,\theta,\phi) = j_l(\frac{\beta_{l_n}}{c} d) Y_l^m(\theta,\phi).
\end{aligned}
\end{equation}
In addition, the basic operation of SphereNet is based on the quadruplets: r, s, $q_1$, and these three nodes formulate a reference plan to provide the polar angle to the point $q_2$. However, SphereNet provides an acceleration module, by projecting all the neighborhoods of s, in an anticlockwise direction, and the reference plan for each node $q_i$ is determined by r, s and $q_{i-1}$. Thus, the computational complexity is reduced by one order of magnitude. SphereNet further considers the following for distance and angle modeling:
\begin{equation}
\small
\begin{aligned}
& \text{CBF}_{ln}(d_{kj}, \alpha_{\angle kij}) = \sqrt{\frac{2}{c^3 j^2_{l+1}(z_{ln})}} j_l(\frac{z_{ln}}{c}d_{kj}) Y_{l}^0(\alpha),\quad\quad
& \text{RBF}_{ln}(d_{kj}) = \sqrt{\frac{2}{c}} \frac{\sin (\frac{n\pi}{c}d)}{d}.
\end{aligned}
\end{equation}

\paragraph{GemNet}
GemNet~\cite{klicpera_gemnet_2021} further extends DimeNet and SphereNet. It explicitly models the dihedral angle. Notice that both GemNet and SphereNet are using the SBF for dihedral angle modeling, yet the difference is that GemNet is using edge-based 2-hop information, {\ie}, the torsion angle, while SphereNet is using the edge-based 1-hop information. Thus, GemNet is expected to possess richer information, while the trade-off is the larger computational efficiency (by one order of magnitude): GemNet has complexity of $O(n k^3)$ while SphereNet is $O(n k^2)$.

%%%%%%%%%%%%%%%%%%%%%%%%%%%%%%%%%%%%%%%%%%%%%%%%%%
\subsection{Spherical Frame Basis Equivariant Model}

\paragraph{TFN}
Tensor field network (TFN)~\cite{thomas2018tensor} first introduces using the SE(3)-equivariance group symmetry for modeling the geometric molecule data. As will be introduced later, the translation-equivariance can be easily achieved by considering the relative coordinates, {\ie}, $\vec \vr = \vr_i - \vr_j$. Then the problem is simplified to design an SO(3)-equivariant model. To handle this, TFN first proposes a general \textbf{framework} by using the spherical harmonics as the basis satisfying the following for all $\va \in SO(3)$ and $\hat \vr$:
\begin{equation}
\small
\begin{aligned}
Y_m^{l}(R(\va) \hat \vr) = \sum_{m'=-l}^{l} D_{mm'}^{(l)}(\va) Y_{m'}^{(l)}(\hat \vr),
\end{aligned}
\end{equation}
where $\hat \vr = \vec \vr / \| \vec \vr\|$, and $D^{l}$ is the irreducible representations of SO(3) to $(2l+1) \times (2l+1)$-dim matrices ({\ie}, the Wigner-D matrices). This is one design criterion for SE(3)-equivariant neural networks with the spherical harmonics frame. In specific, to design an SE(3)-equivariant network, we take the following form:
\begin{equation}\label{eq:modeling_01}
\small
\begin{aligned}
F(\vec \vr) = W(\vr) Y(\hat \vr),
\end{aligned}
\end{equation}
where $\vr = \| \vec \vr \|$, $W(\cdot)$ is the learnable function. Thus we are separating the spherical harmonics basis and the radial signal. For modeling, we only need to learn the $W(\cdot)$ on the radial. Then we use the Clebsch-Gordan tensor product for message passing on node $i$, which is:
\begin{equation}\label{eq:modeling_02}
\small
\begin{aligned}
\vv_{i} = \vv_i + \sum_{j \in \mathcal{N}(i)} F(\vec \vr_{ij}) \otimes \vv_{j},
\end{aligned}
\end{equation}
where $\otimes$ is the Clebsch-Gordan tensor product. \textbf{Note} that for brevity and to give the audience a high-level idea of the spherical frame basis modeling, we omit the \textbf{rotation order} and the \textbf{channel index} in~\Cref{eq:modeling_01,eq:modeling_02}. First, we want to acknowledge that the rotation order is the key to conducting the message passing along tensors, and please refer to the original paper for details. Then for the channel or depth of the message passing layers (notation $c$ in the TFN paper), they are important to expand the model capacity.

To sum up, by far, we can observe that TFN only considers the pairwise information ({\ie}, 1-hop neighborhood) for SE(3)-equivariance.

\paragraph{SE(3)-Transformer}
SE(3)-Transformer~\cite{fuchs2020se} extends the TFN by introducing an attention score, {\ie},
\begin{equation}
\small
\begin{aligned}
\vv_{i} = \vv_i + \sum_{j \in \mathcal{N}(i)} \alpha_{ij} F(\vec \vr_{ij}) \otimes \vv_{j},
\end{aligned}
\end{equation}
where $\alpha_{ij}$ is the attention score. 

To calculate the attention score, first, we need to define the following:
\begin{equation}
\small
\begin{aligned}
\vq_i = \bigoplus_{l \ge 0} \sum_{k \ge 0} W_Q^{lk} \vv_i^{k},
\quad\quad\quad
\vk_{ij} = \bigoplus_{l \ge 0} \sum_{k \ge 0} F_K^{lk}(\vr_j - \vr_i) \otimes \vv^k_j,
\end{aligned}
\end{equation}
where $k$ and $l$ correspond to the rotation order of the input and output tensor, $W_Q$ is a learnable linear matrix, $F_K$ follows the same formation as~\Cref{eq:modeling_01}, and $\bigoplus$ is the direct sum. Then we can obtain the attention coefficients with dot product as:
\begin{equation}
\begin{aligned}
&\alpha_{ij} = \frac{\exp(\vq_i^T \vk_{ij})}{\sum_{j' \in \mathcal{N}_i \setminus i} \exp (\vq_i^T \vk_{ij'}) }\\
\end{aligned}
\end{equation}

\paragraph{Equiformer}
SE(3)-Trans adopts the dot product attention, and Equiformer~\cite{liao2022equiformer} extends this with an MLP attention and with higher efficiency.
We also want to mention that during modeling, Equiformer has an option of adding extra atom and bond information, and we set this hyperparameter as False for a fair comparison when comparing with other geometric models.

\paragraph{NequIP}
Neural Equivariant Interatomic Potentials (NequIP)~\cite{batzner20223} is a follow-up of TFN, which mainly focuses on improving the force prediction. Originally, TFN was directly predicting the $l=1$ tensor for the force prediction. In NequIP, the output only includes the $l=1$ tensor, while the force is obtained by taking the gradient with respect to the energy. There are also other minor architecture design updates, such as adding the skip-connection~\cite{he2016deep}. Please refer \cite{batzner20223} for more details.

\paragraph{Allegro}
Allegro~\cite{musaelian2022learning} is a follow-up of NequIP by further modeling a local frame around each atom. In specific, the standard message-passing framework is based on the nodes (or atoms here), while Allegro focuses on the edge-level information.

\paragraph{Difference with Spherical Harmonics in Invariant Modeling}
As you may notice, the invariant models also adopt the spherical harmonics (or spherical Fourier-Bessel), {\eg}, \Cref{eq:DimeNet_message_passing} in DimeNet and \Cref{eq:SPhereNet_SBF_message_passing} in SphereNet and GemNet. However, their usage of the spherical harmonics is different from the spherical frame models discussed in this section.
\begin{itemize}[noitemsep,topsep=0pt]
    \item In invariant models, the spherical harmonics are used for embedding the angle information, either bond angles or dihedral angles. Such angles are type-0 features, and they are invariant w.r.t. the SO(3) group. Note that this embedding is related to quantum mechanics since the spherical harmonics appear as general solutions of the Schrödinger equations.
    \item In the spherical frame models, the spherical harmonics are used to serve as the basis for transforming the relative coordinates into tensors, utilizing the fact that spherical harmonics are equivariant functions with respect to SO(3) group. 
\end{itemize}
Thus, they may follow the same numerical calculation, but their physical meanings are different.

%%%%%%%%%%%%%%%%%%%%%%%%%%%%%%%%%%%%%%%%%%%%%%%%%%
\subsection{Vector Frame Basis Equivariant Model}
From a very high-level view, we can view this as first constructing the tensor and then conducting the message-passing between the type-0 tensor and type-1 tensor.

\paragraph{EGNN}
E(n)-equivariant graph neural network (EGNN)~\cite{satorras2021n} has a very neat design to achieve the E(n)-equivariance property. It constructs the message update function for both the atom positions and atom attributes simultaneously. Concretely, for edge embedding $\ve$, input node embedding $\vh$ and coordinate $\vv = \vr$, the $l$-th layer updates are:
\begin{equation}
\small \label{eq: egnn}
\begin{aligned}
& \vm_{ij} = W_{e} \big( \vh_i^l, \vh_j^l, \| \vv_i^l - \vv_j^l \|, \ve_{ij} \big)\\
& \vv_{i}^{l+1} = \vv_i^l + \sum_{j \ne i} (\vv_i^l - \vv_j^l), W_{v} (\vm_{ij}) \\
& \vm_i = \sum_{j \ne i} \vm_{ij}\\
& h_i^{l+1} = W_{h} (\vh_i^l, \vm_i),
\end{aligned}
\end{equation}
where $W_e, W_v, W_h$ are learnable parameters.
The equivariance can be proved easily and with good efficiency. However, one inherent limitation of EGNN is that it is essentially a global vector frame model and utilizes only one projection (scalarization) dimension, and it does not satisfy the reflection-antisymmetric condition for certain tasks like binding.

\paragraph{PaiNN}
Polarizable atom interaction neural network (PaiNN)~\cite{schutt2021equivariant} utilizes a multi-channel vector aggregation method, which contains more expressive equivariant vector information than \Cref{eq: egnn}. More precisely, each node of PaiNN maintains a multi-channel vector: $\vv_i \in \mathbf{R}^{F \times 3}$, where $F$ denotes the channel number. Comparing with \Cref{eq: egnn}, the $\vv_i \in \mathbf{R}^{1 \times 3}$ of EGNN restricted the expressiveness power. \cite{2304.04757} provides a geometric explanation of the updating method of PaiNN ((9) of \cite{schutt2021equivariant}) by the frame transition functions between local vector frames.

\newpage
\section{Complete Results} \label{sec:app:complete_results}
In the main body, due to space limitations, we cannot provide the results on certain tasks. Here we would like to provide more comprehensive results.

For the results not listed either in the main body or in this section, there are two possible reasons for us to exclude them:
(1) We cannot reproduce them using the reported hyperparameters in the original paper, and we may need to do more hyperparameter tuning as the next steps.
(2) Some models are too large to fit in the GPU memory, even with batch-size=1.

%%%%%%%%%%%%%%%%%%%%%%%%%%%%%%
\subsection{Small Molecules: MD17 and rMD17}
In~\Cref{tab:main_result_MD17_rMD17}, we select 6 subtasks in MD17 and 6 subtasks in rMD17. Next we will show the complete results of MD17 and rMD17 are in~\Cref{tab:main_result_MD17,tab:main_result_rMD17}.

\begin{table}[htb!]
\setlength{\tabcolsep}{5pt}
\fontsize{9}{9}\selectfont
\centering
\caption{
\small
Results on 8 energy ($\frac{kcal}{mol}$) and force ($\frac{kcal}{mol\cdot \text{\r{A}}}$) prediction tasks in MD17.
The evaluation is the mean absolute error.
No data normalization is used.
}
\label{tab:main_result_MD17}
\vspace{-1.5ex}
\begin{adjustbox}{max width=\textwidth}
\begin{tabular}{ll c c c c c c c c}
\toprule
Model & Energy/Force & Aspirin $\downarrow$ & Benzene $\downarrow$ & Ethanol $\downarrow$ & Malonaldehyde $\downarrow$ & Naphthalene $\downarrow$ & Salicylic $\downarrow$ & Toluene $\downarrow$ & Uracil $\downarrow$ \\
\midrule

\multirow{2}{*}{SchNet}
% 5e-4_CosineAnnealingLR_300_5_1000_energy_force_no_normalization
& Energy & 0.475 & 0.117 & 0.109 & 0.300 & 0.167 & 0.212 & 0.149 & 0.170\\
& Force & 1.203 & 0.380 & 0.386 & 0.794 & 0.587 & 0.826 & 0.568 & 0.773\\
\midrule

\multirow{2}{*}{DimeNet++}
% 5e-4_CosineAnnealingLR_300_800_energy_force_no_normalization
& Energy & 4.168 & 0.893 & 1.238 & 1.385 & 1.846 & 2.445 & 1.484 & 1.522\\
& Force & 7.212 & 0.603 & 0.753 & 1.842 & 8.515 & 1.752 & 1.037 & 1.632\\
\midrule

\multirow{2}{*}{EGNN}
% 1e-4_CosineAnnealingLR_300_5_1_1000_energy_force_no_normalization
& Energy & 17.892 & 1.142 & 0.436 & 0.896 & 12.177 & 6.964 & 4.051 & 0.854\\
& Force & 3.042 & 0.736 & 0.924 & 1.566 & 1.136 & 1.177 & 1.202 & 1.367\\
\midrule

\multirow{2}{*}{PaiNN}
% 5e-4_CosineAnnealingLR_300_5_1000_energy_force_no_normalization
& Energy & 27.626 & 0.095 & 0.063 & 0.102 & 0.622 & 0.371 & 0.165 & 0.111\\
& Force & 0.572 & 0.053 & 0.230 & 0.338 & 0.132 & 0.288 & 0.141 & 0.201\\
\midrule

\multirow{2}{*}{GemNet-T}
% 5e-4_CosineAnnealingLR_300_5_1000_energy_force_no_normalization
& Energy & 0.684 & 0.097 & 4.598 & 4.966 & 0.482 & 0.128 & 0.098 & 1.349\\
& Force & 0.558 & 0.089 & 0.219 & 0.433 & 0.212 & 0.326 & 0.174 & 486.612\\
\midrule

\multirow{2}{*}{SphereNet}
% 1e-4_CosineAnnealingLR_300_1_1000_energy_force_no_normalization
& Energy & 0.244 & 0.107 & 1.603 & 1.559 & 0.167 & 0.188 & 0.113 & 7.115\\
& Force & 0.546 & 0.135 & 0.168 & 0.667 & 0.315 & 0.479 & 0.194 & 0.442\\
\midrule

\multirow{2}{*}{SEGNN}
% 1e-4_CosineAnnealingLR_300_5_5_0_800_energy_force_no_normalization
& Energy & 17.774 & 0.086 & 0.151 & 0.247 & 0.655 & 2.173 & 0.624 & 0.259\\
& Force & 9.003 & 0.265 & 0.893 & 1.249 & 0.895 & 2.220 & 1.138 & 0.948\\
\midrule

\multirow{2}{*}{NequIP}
% 1e-3_CosineAnnealingLR__4.0_1000_energy_force_with_normalization
& Energy & 8.333 & 0.355 & 0.971 & 2.293 & 1.032 & 2.952 & 1.303 & 1.266\\
& Force & 23.769 & 2.383 & 5.832 & 12.099 & 5.247 & 14.048 & 6.800 & 8.060\\
\midrule

\multirow{2}{*}{Allegro}
% 1e-3_CosineAnnealingLR_4.0_512_1000_energy_force_no_normalization
& Energy & 1.138 & 0.154 & 0.258 & 1.330 & 0.824 & 1.114 & 0.441 & 0.613\\
& Force & 3.405 & 0.823 & 1.412 & 4.191 & 3.743 & 4.934 & 1.968 & 3.544\\
\midrule

\multirow{2}{*}{Equiformer}
% 1e-4_CosineAnnealingLR_300_5_1_1000_energy_force_no_normalization
& Energy & 0.308 & 0.075 & 0.096 & 0.183 & 0.097 & 0.189 & 0.209 & 0.106\\
& Force & 0.286 & 0.045 & 0.142 & 0.230 & 0.068 & 0.200 & 0.080 & 0.141\\
\bottomrule
\end{tabular}
\end{adjustbox}
\end{table}

\begin{table}[htb!]
\setlength{\tabcolsep}{5pt}
\fontsize{9}{9}\selectfont
\centering
\caption{\small
Results on 10 energy ($\frac{kcal}{mol}$) and force ($\frac{kcal}{mol\cdot \text{\r{A}}}$) prediction tasks in rMD17.
The evaluation is the mean absolute error.
Data normalization is used.
}
\label{tab:main_result_rMD17}
\vspace{-1.5ex}
\begin{adjustbox}{max width=\textwidth}
\begin{tabular}{ll c c c c c c c c c c}
\toprule
Model & Energy/Force & Aspirin $\downarrow$ & Azobenzene $\downarrow$ & Benzene $\downarrow$ & Ethanol $\downarrow$ & Malonaldehyde $\downarrow$ & Naphthalene $\downarrow$ & Paracetamol $\downarrow$ & Salicylic $\downarrow$ & Toluene $\downarrow$ & Uracil $\downarrow$ \\
\midrule

\multirow{2}{*}{SchNet}
% 5e-4_CosineAnnealingLR_300_5_1000_energy_force_no_normalization
& Energy & 0.534 & 1.818 & 0.111 & 1.757 & 0.260 & 0.124 & 8.138 & 2.618 & 0.119 & 7.029\\
& Force & 1.243 & 3.596 & 0.233 & 0.449 & 0.862 & 0.587 & 2.320 & 0.878 & 0.574 & 0.762\\
\midrule

\multirow{2}{*}{DimeNet++}
% 5e-4_CosineAnnealingLR_300_800_energy_force_no_normalization
& Energy & 2.438 & 3.955 & 0.741 & 1.456 & 2.317 & 1.648 & 2.261 & 1.555 & 1.210 & 2.320\\
& Force & 2.009 & 1.243 & 0.340 & 1.213 & 7.029 & 0.629 & 1.047 & 0.934 & 0.921 & 3.181\\
\midrule

\multirow{2}{*}{EGNN}
% 1e-4_CosineAnnealingLR_300_5_1_1000_energy_force_no_normalization
& Energy & 17.350 & 21.333 & 0.315 & 0.402 & 0.534 & 12.164 & 26.902 & 7.794 & 15.021 & 1.669\\
& Force & 3.825 & 2.330 & 0.529 & 0.989 & 1.334 & 1.183 & 2.313 & 1.571 & 1.165 & 1.323\\
\midrule

\multirow{2}{*}{PaiNN}
% 1e-4_CosineAnnealingLR_300_5_1000_energy_force_no_normalization
& Energy & 30.156 & 0.107 & 0.010 & 1.170 & 0.070 & 5.297 & 0.117 & 5.219 & 0.045 & 2.478\\
& Force & 0.573 & 0.326 & 0.032 & 0.316 & 0.377 & 0.161 & 0.440 & 0.321 & 0.231 & 0.235\\
\midrule

\multirow{2}{*}{GemNet-T}
% 5e-4_CosineAnnealingLR_300_5_1000_energy_force_no_normalization
& Energy & 5.389 & 7.770 & 0.007 & 1.615 & 9.496 & 0.031 & 2.173 & 21.411 & 959.745 & 994.036\\
& Force & 0.555 & 0.347 & 0.033 & 0.233 & 0.337 & 0.154 & 0.388 & 0.371 & 0.400 & 1.165\\
\midrule

\multirow{2}{*}{SphereNet}
% 1e-4_CosineAnnealingLR_300_1_1000_energy_force_no_normalization
& Energy & 0.304 & 0.257 & 0.052 & 0.072 & 0.138 & 0.093 & 0.183 & 0.771 & 20.479 & 12.211\\
& Force & 0.622 & 0.532 & 0.076 & 0.217 & 0.500 & 0.279 & 0.482 & 2.088 & 0.254 & 0.959\\
\midrule

\multirow{2}{*}{SEGNN}
% 1e-4_CosineAnnealingLR_300_5_5_0_700_energy_force_no_normalization
& Energy & 15.721 & 3.474 & 0.270 & 0.130 & 0.182 & 1.110 & 4.197 & 1.494 & 0.814 & 1.115\\
& Force & 8.549 & 2.579 & 0.174 & 0.846 & 1.185 & 0.926 & 3.191 & 2.056 & 1.241 & 0.966\\
\midrule

\multirow{2}{*}{NequIP}
% 1e-3_CosineAnnealingLR__4.0_1000_energy_force_no_normalization
& Energy & 9.618 & 1.993 & 3.048 & 0.936 & 2.313 & 2.089 & 5.136 & 3.302 & 1.306 & 1.738\\
& Force & 22.904 & 6.406 & 1.523 & 6.027 & 12.372 & 5.529 & 17.574 & 15.693 & 7.094 & 10.220\\
\midrule

\multirow{2}{*}{Allegro}
% 1e-3_CosineAnnealingLR_4.0_512_1000_energy_force_no_normalization
& Energy & 1.366 & 0.872 & 0.029 & 1.002 & 0.417 & 1.756 & 0.944 & 1.035 & 0.437 & 0.387\\
& Force & 3.186 & 2.763 & 0.237 & 2.799 & 2.125 & 3.815 & 3.081 & 4.781 & 2.048 & 1.939\\
\midrule

\multirow{2}{*}{Equiformer}
% 1e-4_CosineAnnealingLR_300_5_1_1000_energy_force_no_normalization
& Energy & 0.375 & 0.127 & 0.027 & 0.064 & 0.085 & 0.069 & 0.215 & 0.143 & 0.104 & 0.200\\
& Force & 0.305 & 0.132 & 0.020 & 0.162 & 0.240 & 0.070 & 0.258 & 0.218 & 0.077 & 0.149\\
\bottomrule
\end{tabular}
\end{adjustbox}
\end{table}

\clearpage
%%%%%%%%%%%%%%%%%%%%%%%%%%%%%%
\subsection{Geometric Pretraining}
\textbf{Single-modal Pretraining.} Recent studies have started to explore \textbf{single-modal geometric pretraining} on molecules. The GeoSSL paper~\cite{liu2022molecular} covers a wide range of geometric pretraining algorithms. The type prediction, distance prediction, and angle prediction predict the masked atom type, pairwise distance, and bond angle, respectively. The 3D InfoGraph predicts whether the node- and graph-level 3D representation are for the same molecule. GeoSSL is a novel geometric pretraining paradigm that maximizes the mutual information (MI) between the original conformation $\vg_1$ and augmented conformation $\vg_2$, where $\vg_2$ is obtained by adding small perturbations to $\vg_1$. RR, InfoNCE, and EBM-NCE optimize the objective in the latent representation space, either generative or contrastive. GeoSSL-DDM~\cite{liu2022molecular} optimizes the same objective function using denoising score matching. GeoSSL-DDM-1L~\cite{zaidi2022pre} is a special case of GeoSSL-DDM with one layer of denoising. 3D-EMGP~\cite{jiao2022energy} geometric pretraining is specifically built on equivariant models, and the goal is to denoise the 3D coordinates directly using a diffusion model. We illustrate these seven algorithms in~\Cref{fig:geometric_pretraining}.

\begin{figure}[htb!]
\centering
\includegraphics[width=\linewidth]{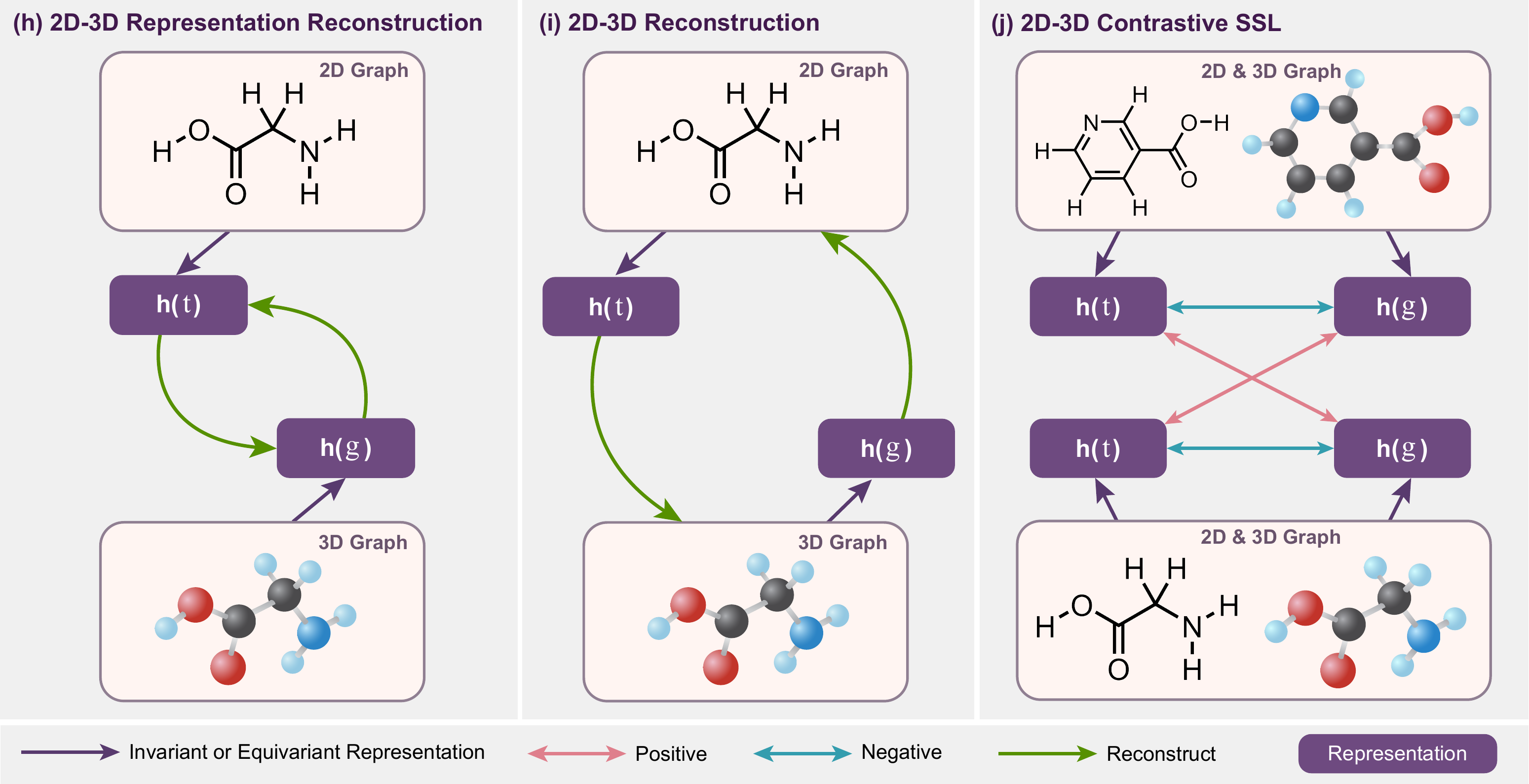}
\vspace{-4ex}
\caption{\small
Pipelines for seven geometric pretraining methods.
} \label{fig:geometric_pretraining_2D_3D_MI}
\end{figure}

\textbf{2D-3D Multi-modal Pretraining.} Another promising direction is the \textbf{multi-modal pretraining on topology and geometry}. GraphMVP~\cite{liu2022pretraining} first proposes one contrastive objective (EBM-NCE) and one generative objective (variational representation reconstruction, VRR) to optimize the mutual information between the 2D and 3D modalities. Specifically, VRR does the 2D and 3D reconstruction in the latent space. 3D InfoMax~\cite{stark20223d} is a special case of GraphMVP, with the contrastive part only. MoleculeSDE~\cite{liu2023moleculeSDE} extends GraphMVP by introducing two SDE models for solving the 2D and 3D reconstruction. An illustration of them is in~\Cref{fig:geometric_pretraining_2D_3D_MI}.

\clearpage
In~\Cref{tab:main_result_QM9_SchNet_pretraining}, we show the pretraining results of using \textbf{SchNet as the backbone} and fine-tuning on QM9. The pretraining results of using \textbf{SchNet as the backbone} and fine-tuning on MD17 are in~\Cref{tab:main_result_MD17_SchNet_pretraining}. The pretraining results of using \textbf{PaiNN as the backbone} and fine-tuning on QM9 and MD17 are in~\Cref{tab:main_result_QM9_PaiNN_pretraining,tab:main_result_MD17_PaiNN_pretraining}. For MD17, as will be discussed in~\Cref{sec:app:ablation_studies}, we do not consider the data normalization trick. Notice that some pretraining results are skipped due to the collapsed performance.

\begin{table}[htb]
\setlength{\tabcolsep}{5pt}
\fontsize{9}{9}\selectfont
\centering
\caption{
\small
Pretraining results on eight force prediction tasks from MD17, and the backbone model is SchNet. We take 1K for training, 1K for validation, and 48K to 991K molecules for the test concerning different tasks. The evaluation is mean absolute error, and the best results are marked in \underline{\textbf{bold}} and \textbf{bold}, respectively.
}
\label{tab:main_result_MD17_SchNet_pretraining}
\vspace{-1.5ex}
\begin{adjustbox}{max width=\textwidth}
\begin{tabular}{l c c c c c c c c}
\toprule
Pretraining & Aspirin $\downarrow$ & Benzene $\downarrow$ & Ethanol $\downarrow$ & Malonaldehyde $\downarrow$ & Naphthalene $\downarrow$ & Salicylic $\downarrow$ & Toluene $\downarrow$ & Uracil $\downarrow$ \\
\midrule

% 5e-4_CosineAnnealingLR_300_5_1000
-- (random init) & 1.203 & 0.380 & 0.386 & 0.794 & 0.587 & 0.826 & 0.568 & 0.773\\

% pretrain_Supervised/PCQM4Mv2_schnet/1e-4_CosineAnnealingLR_100
Supervised & 1.867 & 0.434 & 0.566 & 1.106 & 0.637 & 13.037 & 0.607 & 0.759\\

% pretrain_ChargePrediction/PCQM4Mv2_schnet/1e-4_CosineAnnealingLR_0.15_100
Type Prediction & 1.383 & 0.402 & 0.450 & 0.879 & 0.622 & 1.028 & 0.662 & 0.840\\

% pretrain_DistancePrediction/PCQM4Mv2_schnet/1e-4_CosineAnnealingLR_100
Distance Prediction & 1.427 & 0.396 & 0.434 & 0.818 & 0.793 & 0.952 & 0.509 & 1.567\\

% pretrain_TorsionAnglePrediction/PCQM4Mv2_schnet/1e-4_CosineAnnealingLR_1e-3_100
Angle Prediction & 1.542 & 0.447 & 0.669 & 1.022 & 0.680 & 1.032 & 0.623 & 0.768\\

% pretrain_3DInfoGraph/PCQM4Mv2_schnet/1e-4_CosineAnnealingLR_100
3D InfoGraph & 1.610 & 0.415 & 0.560 & 0.900 & 0.788 & 1.278 & 0.768 & 1.110\\

% pretrain_RR/PCQM4Mv2_schnet/1e-4_CosineAnnealingLR_0_0.3_0.3_RR_100
GeoSSL-RR & 1.215 & 0.393 & 0.514 & 1.092 & 0.596 & 0.847 & 0.570 & 0.711\\

% pretrain_InfoNCE/PCQM4Mv2_schnet/1e-4_CosineAnnealingLR_0_0.3_0_InfoNCE_100
GeoSSL-InfoNCE & 1.132 & 0.395 & 0.466 & 0.888 & 0.542 & 0.831 & 0.554 & 0.664\\

% pretrain_EBM_NCE/PCQM4Mv2_schnet/1e-4_CosineAnnealingLR_0_1_0.3_EBM_NCE_100
GeoSSL-EBM-NCE & 1.251 & 0.373 & 0.457 & 0.829 & 0.512 & 0.990 & 0.560 & 0.742\\

% pretrain_GraphMVP/PCQM4Mv2_schnet/Vanilla_Contrastive_1e-4_0.3_100
3D InfoMax & 1.142 & 0.388 & 0.469 & 0.731 & 0.785 & 0.798 & 0.516 & 0.640\\

% pretrain_GraphMVP/PCQM4Mv2_schnet/Vanilla_1e-4_0.3_100
GraphMVP & 1.126 & 0.377 & 0.430 & 0.726 & 0.498 & 0.740 & 0.508 & 0.620\\

% pretrain_DeepMind/PCQM4Mv2_schnet/1e-4_CosineAnnealingLR_0_0.3_0.15_GeoSSL_10_0.01_30_symmetry_0.1_100
GeoSSL-DDM-1L & 1.364 & 0.391 & 0.432 & 0.830 & 0.599 & 0.817 & 0.628 & 0.607\\

% pretrain_GeoSSL/PCQM4Mv2_schnet/1e-4_CosineAnnealingLR_1_0.3_0_GeoSSL_10_0.01_50_symmetry_0.2_100
GeoSSL-DDM & \underline{\textbf{1.107}} & 0.360 & 0.357 & 0.737 & 0.568 & 0.902 & 0.484 & 0.502\\

% pretrain_MoleculeSDE/PCQM4Mv2_schnet_SDEModel2Dto3D_02_SDEModel3Dto2D_node_adj_dense/2Dto3D_1_VE_3Dto2D_1_VE_CL_EBM_node_dot_prod_1_0.1_0_5e-4_0_anneal_0_100
MoleculeSDE (VE) & \textbf{1.112} & \underline{\textbf{0.304}} & \underline{\textbf{0.282}} & \textbf{0.520} & \textbf{0.455} & \textbf{0.725} & \textbf{0.515} & \underline{\textbf{0.447}}\\

% pretrain_MoleculeSDE/PCQM4Mv2_schnet_SDEModel2Dto3D_01_SDEModel3Dto2D_node_adj_dense/2Dto3D_1_VP_3Dto2D_1_VP_CL_EBM_node_dot_prod_1_0.1_0_5e-4_0_anneal_0_100
MoleculeSDE (VP) & 1.244 & \textbf{0.315} & \textbf{0.338} & \underline{\textbf{0.488}} & \underline{\textbf{0.432}} & \underline{\textbf{0.712}} & \underline{\textbf{0.478}} & \textbf{0.468}\\
\bottomrule
\end{tabular}
\end{adjustbox}
\end{table}

\null

\begin{table}[htb]
\setlength{\tabcolsep}{5pt}
\fontsize{9}{9}\selectfont
\caption{
\small
Pretraining results on 12 quantum mechanics prediction tasks from QM9, and the backbone model is PaiNN. We take 110K for training, 10K for validation, and 11K for testing. The evaluation is mean absolute error, and the best and the second best results are marked in \underline{\textbf{bold}} and \textbf{bold}, respectively.
}
\label{tab:main_result_QM9_PaiNN_pretraining}
\vspace{-1.5ex}
\begin{adjustbox}{max width=\textwidth}
\begin{tabular}{l c c c c c c c c c c c c}
\toprule
Pretraining & $\alpha$ $\downarrow$ & $\nabla \mathcal{E}$ $\downarrow$ & $\mathcal{E}_\text{HOMO}$ $\downarrow$ & $\mathcal{E}_\text{LUMO}$ $\downarrow$ & $\mu$ $\downarrow$ & $C_v$ $\downarrow$ & $G$ $\downarrow$ & $H$ $\downarrow$ & $R^2$ $\downarrow$ & $U$ $\downarrow$ & $U_0$ $\downarrow$ & ZPVE $\downarrow$\\
\midrule

% 5e-4_CosineAnnealingLR_300_128_1000
-- & 0.049 & 42.73 & 24.46 & 20.16 & 0.016 & 0.025 & 8.43 & 7.88 & 0.169 & 8.18 & 7.63 & 1.419\\

% pretrain_Supervised/PCQM4Mv2_painn/1e-4_CosineAnnealingLR_100
Supervised & 0.161 & 64.30 & 23.41 & 19.31 & 0.015 & 0.024 & 9.01 & 9.53 & 0.152 & 16.17 & 9.43 & 1.470\\

% % \textbf{Type Prediction} \\

% pretrain_DistancePrediction/PCQM4Mv2_painn/1e-4_CosineAnnealingLR_100
Distance Prediction & 0.049 & 37.23 & 22.75 & 18.26 & 0.014 & 0.030 & 9.31 & 9.35 & 0.143 & 9.85 & 9.07 & 1.566\\

% pretrain_3DInfoGraph/PCQM4Mv2_painn/1e-4_CosineAnnealingLR_100
3D InfoGraph & 0.047 & 44.25 & 24.06 & 18.54 & 0.015 & 0.052 & 8.81 & 7.97 & 0.143 & 8.68 & 8.08 & 1.416\\

% pretrain_GeoSSL_RR/PCQM4Mv2_painn/1e-4_CosineAnnealingLR_0_0.3_0_RR_100
GeoSSL-RR & 0.046 & 41.20 & 23.93 & 19.36 & 0.016 & 0.025 & 8.32 & 8.17 & 0.174 & 7.99 & 8.20 & 1.438\\

% pretrain_GeoSSL_InfoNCE/PCQM4Mv2_painn/1e-4_CosineAnnealingLR_0_0.3_0_InfoNCE_100
GeoSSL-InfoNCE & 0.045 & 39.29 & 23.23 & 18.40 & 0.015 & 0.024 & 8.34 & 8.37 & \underline{\textbf{0.127}} & 7.45 & 8.34 & 1.356\\

% pretrain_GeoSSL_EBM_NCE/PCQM4Mv2_painn/1e-4_CosineAnnealingLR_0_0.3_0_EBM_NCE_100
GeoSSL-EBM-NCE & 0.045 & 38.87 & 22.71 & 17.89 & 0.014 & 0.082 & 8.28 & 7.35 & 0.130 & 7.85 & 7.68 & 1.338\\

% pretrain_GraphMVP/PCQM4Mv2_painn/Vanilla_Contrastive_1e-4_0.3_100
3D InfoMax & 0.046 & 36.97 & 21.31 & 17.69 & 0.014 & 0.024 & 8.38 & 7.36 & 0.135 & 8.60 & 7.99 & 1.453\\

% pretrain_GraphMVP/PCQM4Mv2_painn/Vanilla_1e-4_0.3_100
GraphMVP & 0.044 & 36.03 & 20.71 & 17.02 & 0.014 & 0.024 & 8.31 & 7.36 & 0.132 & 7.57 & 7.34 & 1.337\\

% pretrain_DeepMind/PCQM4Mv2_painn/1e-4_CosineAnnealingLR_0_0.3_0.15_GeoSSL_10_0.01_30_symmetry_0.1_100
GeoSSL-DDM-1L & 0.045 & 36.13 & 20.59 & 17.26 & 0.014 & 0.024 & 9.45 & 8.43 & 0.128 & 8.88 & 8.16 & 1.380\\

% pretrain_GeoSSL_DDM/PCQM4Mv2_painn/1e-4_CosineAnnealingLR_0_0.3_0_GeoSSL_10_0.01_50_symmetry_0.5_100
GeoSSL-DDM & \textbf{0.043} & 35.55 & 20.57 & \textbf{16.95} & 0.014 & 0.024 & 8.25 & 7.42 & \underline{\textbf{0.127}} & 7.36 & 7.34 & 1.334\\

% pretrain_GeoSSL_3DEMGP/PCQM4Mv2_painn/1e-4_CosineAnnealingLR_0_0.3_0_10_0.01_50_symmetry_1_gaussian_0.1_0_100
3D-EMGP (Gaussian) & 0.277 & 40.56 & 21.25 & 23.99 & 0.014 & 0.039 & 9.16 & 9.14 & 0.340 & 9.31 & 8.59 & 1.433\\

% pretrain_MoleculeSDE/PCQM4Mv2_painn_SDEModel2Dto3D_01_SDEModel3Dto2D_node_adj_dense/2Dto3D_1_VE_3Dto2D_1_VE_CL_EBM_node_dot_prod_1_0.1_0_5e-4_0_anneal_0_100
MoleculeSDE (VE) & 0.044 & \underline{\textbf{34.67}} & \underline{\textbf{20.14}} & 17.05 & \underline{\textbf{0.013}} & \underline{\textbf{0.023}} & \underline{\textbf{7.64}} & \textbf{7.05} & 0.139 & \underline{\textbf{6.88}} & \underline{\textbf{6.79}} & \underline{\textbf{1.273}}\\

% pretrain_MoleculeSDE_generative/PCQM4Mv2_painn_SDEModel2Dto3D_02_SDEModel3Dto2D_node_adj_dense/2Dto3D_1_VP_3Dto2D_1_VP_5e-4_0_anneal_2_50
MoleculeSDE (VP) & \underline{\textbf{0.042}} & \textbf{35.09} & \underline{\textbf{20.14}} & \underline{\textbf{16.78}} & \underline{\textbf{0.013}} & \underline{\textbf{0.023}} & \textbf{8.17} & \underline{\textbf{7.01}} & 0.133 & \textbf{7.30} & \textbf{7.05} & \textbf{1.315}\\

\bottomrule
\end{tabular}
\end{adjustbox}
\vspace{-1ex}
\end{table}

\begin{table}[htb]
\setlength{\tabcolsep}{5pt}
\fontsize{9}{9}\selectfont
\centering
\caption{
\small
Results on eight force prediction tasks from MD17, and the backbone model is PaiNN. We take 1K for training, 1K for validation, and 48K to 991K molecules for the test concerning different tasks. The evaluation is mean absolute error, and the best results are marked in \underline{\textbf{bold}} and \textbf{bold}, respectively.
}
\label{tab:main_result_MD17_PaiNN_pretraining}
\vspace{-1.5ex}
\begin{adjustbox}{max width=\textwidth}
\begin{tabular}{l c c c c c c c c}
\toprule
Pretraining & Aspirin $\downarrow$ & Benzene $\downarrow$ & Ethanol $\downarrow$ & Malonaldehyde $\downarrow$ & Naphthalene $\downarrow$ & Salicylic $\downarrow$ & Toluene $\downarrow$ & Uracil $\downarrow$ \\
\midrule

% 5e-4_CosineAnnealingLR_300_5_1000
% PaiNN-Energy & 27.626 & 0.095 & 0.063 & 0.102 & 0.622 & 0.371 & 0.165 & 0.111\\
-- & 0.572 & 0.053 & 0.230 & 0.338 & 0.132 & 0.288 & 0.141 & 0.201\\

% pretrain_Supervised/PCQM4Mv2_painn/1e-4_CosineAnnealingLR_100
Supervised & 0.509 & 0.056 & 0.181 & 0.330 & -- & 0.284 & 0.163 & --\\

% % pretrain_ChargePrediction/PCQM4Mv2_painn/1e-4_CosineAnnealingLR_0.3_100
% Type Prediction & -- & -- & -- & 0.389 & -- & -- & -- & --\\

% pretrain_DistancePrediction/PCQM4Mv2_painn/1e-4_CosineAnnealingLR_100
Distance Prediction & 0.480 & 0.053 & 0.200 & 0.296 & 0.131 & 0.265 & 0.171 & 0.168\\

% pretrain_3DInfoGraph/PCQM4Mv2_painn/1e-4_CosineAnnealingLR_100
3D InfoGraph & 0.554 & 0.067 & 0.249 & 0.353 & 0.177 & 0.331 & 0.179 & 0.213\\

% pretrain_GeoSSL_RR/PCQM4Mv2_painn/1e-4_CosineAnnealingLR_0_0.3_0_RR_100
GeoSSL-RR & 0.559 & 0.051 & 0.262 & 0.368 & 0.146 & 0.303 & 0.154 & 0.202\\

% pretrain_GeoSSL_InfoNCE/PCQM4Mv2_painn/1e-4_CosineAnnealingLR_0_0.3_0_InfoNCE_100
GeoSSL-InfoNCE & 0.428 & 0.051 & 0.197 & 0.337 & 0.127 & 0.247 & 0.136 & 0.169\\

% pretrain_GeoSSL_EBM_NCE/PCQM4Mv2_painn/1e-4_CosineAnnealingLR_0_0.3_0_EBM_NCE_100
GeoSSL-EBM-NCE & 0.435 & 0.048 & 0.198 & \textbf{0.295} & 0.143 & 0.245 & 0.132 & 0.172\\

% pretrain_GraphMVP/PCQM4Mv2_painn/Vanilla_Contrastive_1e-4_0.3_100
3D InfoMax & 0.479 & 0.052 & 0.220 & 0.344 & 0.138 & 0.267 & 0.155 & 0.174\\

% pretrain_GraphMVP/PCQM4Mv2_painn/Vanilla_1e-4_0.3_100
GraphMVP & 0.465 & 0.050 & 0.205 & 0.316 & \textbf{0.119} & 0.242 & 0.136 & 0.168\\

% pretrain_DeepMind/PCQM4Mv2_painn/1e-4_CosineAnnealingLR_0_0.3_0.15_GeoSSL_10_0.01_30_symmetry_0.1_100
GeoSSL-DDM-1L & 0.436 & 0.048 & 0.209 & 0.320 & \textbf{0.119} & 0.249 & 0.132 & 0.177\\

% pretrain_GeoSSL_DDM/PCQM4Mv2_painn/1e-4_CosineAnnealingLR_0_0.3_0.3_GeoSSL_10_0.01_50_symmetry_0.2_100
GeoSSL-DDM & \textbf{0.427} & 0.047 & \underline{\textbf{0.188}} & 0.313 & 0.120 & \textbf{0.240} & \textbf{0.129} & 0.167\\

% pretrain_GeoSSL_3DEMGP/PCQM4Mv2_painn/1e-4_CosineAnnealingLR_0_0.3_0_10_0.01_50_symmetry_1_gaussian_1_0.2_100
3D-EMGP (Gaussian) & 0.487 & 0.048 & 0.217 & 0.329 & 0.151 & 0.299 & 0.141 & 0.182\\

% pretrain_MoleculeSDE/PCQM4Mv2_painn_SDEModel2Dto3D_01_SDEModel3Dto2D_node_adj_dense/2Dto3D_1_VE_3Dto2D_1_VE_CL_EBM_node_dot_prod_1_0.1_0_5e-4_0_anneal_0_100
MoleculeSDE (VE) & \underline{\textbf{0.421}} & \underline{\textbf{0.043}} & 0.195 & \underline{\textbf{0.284}} & \underline{\textbf{0.105}} & \underline{\textbf{0.236}} & \underline{\textbf{0.123}} & \underline{\textbf{0.158}}\\

% pretrain_MoleculeSDE_generative/PCQM4Mv2_painn_SDEModel2Dto3D_02_SDEModel3Dto2D_node_adj_dense/2Dto3D_1_VP_3Dto2D_1_VP_5e-4_0_anneal_2_50
MoleculeSDE (VP) & 0.443 & \textbf{0.045} & \textbf{0.191} & 0.301 & 0.131 & 0.261 & 0.140 & \textbf{0.159}\\
\bottomrule
\end{tabular}
\end{adjustbox}
\end{table}

\clearpage

\newpage
\section{Ablation Studies} \label{sec:app:ablation_studies}
We have the following challenges in the literature: (1) Different data splits, {\ie}, with different seeds or different train-valid-test sizes. (2) Different running epochs. (3) Different optimizers (SGD, Adam) and learning rate schedulers. (4) Different preprocessors, including data augmentations and normalization strategies. These factors can significantly affect performance, and \framework{} is a useful tool for careful scrutinization.

%%%%%%%%%%%%%%%%%%%%%%%%%%%%%%%%%%%%%%%%%%%%%%%%%%
\subsection{Ablation Studies on the Effect of Latent Dimension $d$} \label{sec:app:ablation_latent_dimension}

\begin{figure}[htb!]
\centering
    \begin{subfigure}[\small Task $\alpha$]
    {\includegraphics[width=0.23\linewidth]{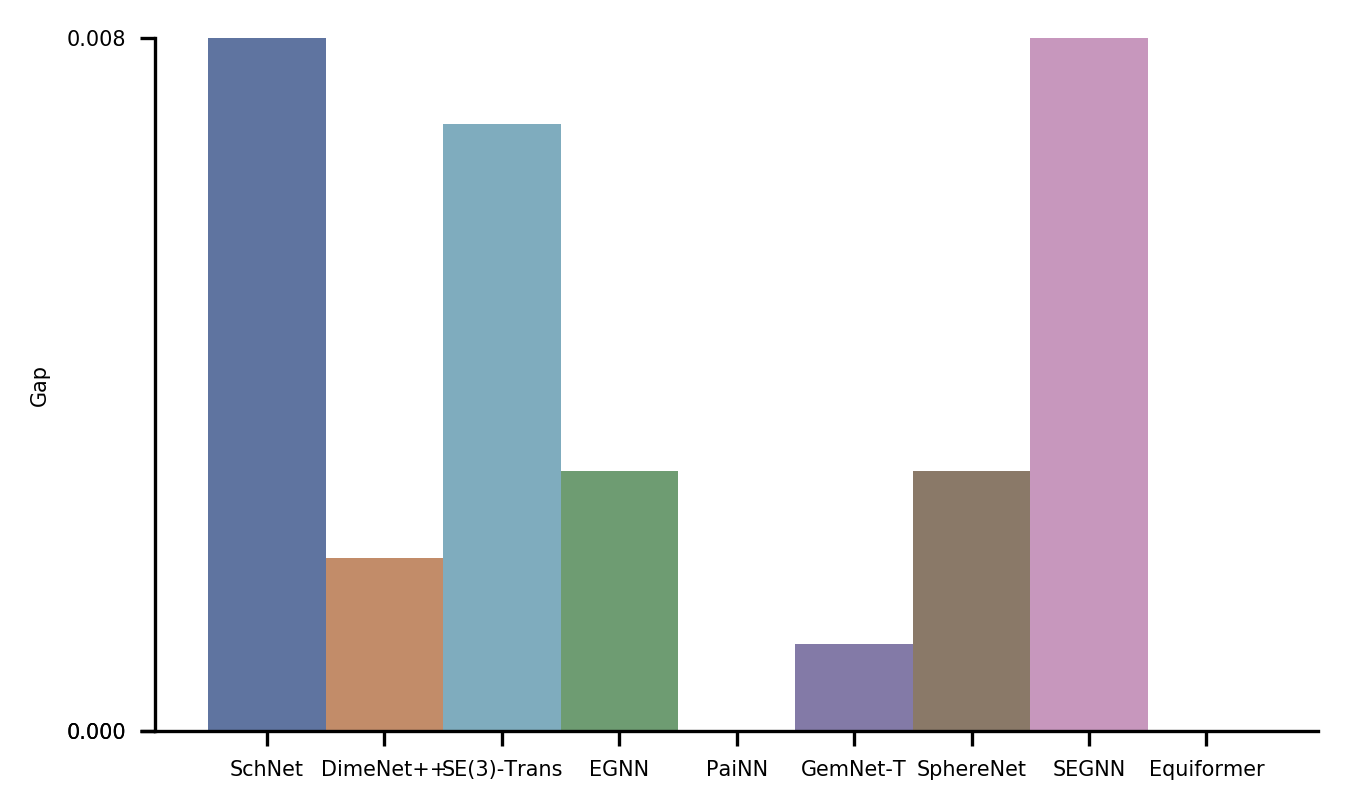}
    }
    \end{subfigure}
\hfill
    \begin{subfigure}[\small Task $\nabla \mathcal{E}$]
    {\includegraphics[width=0.23\linewidth]{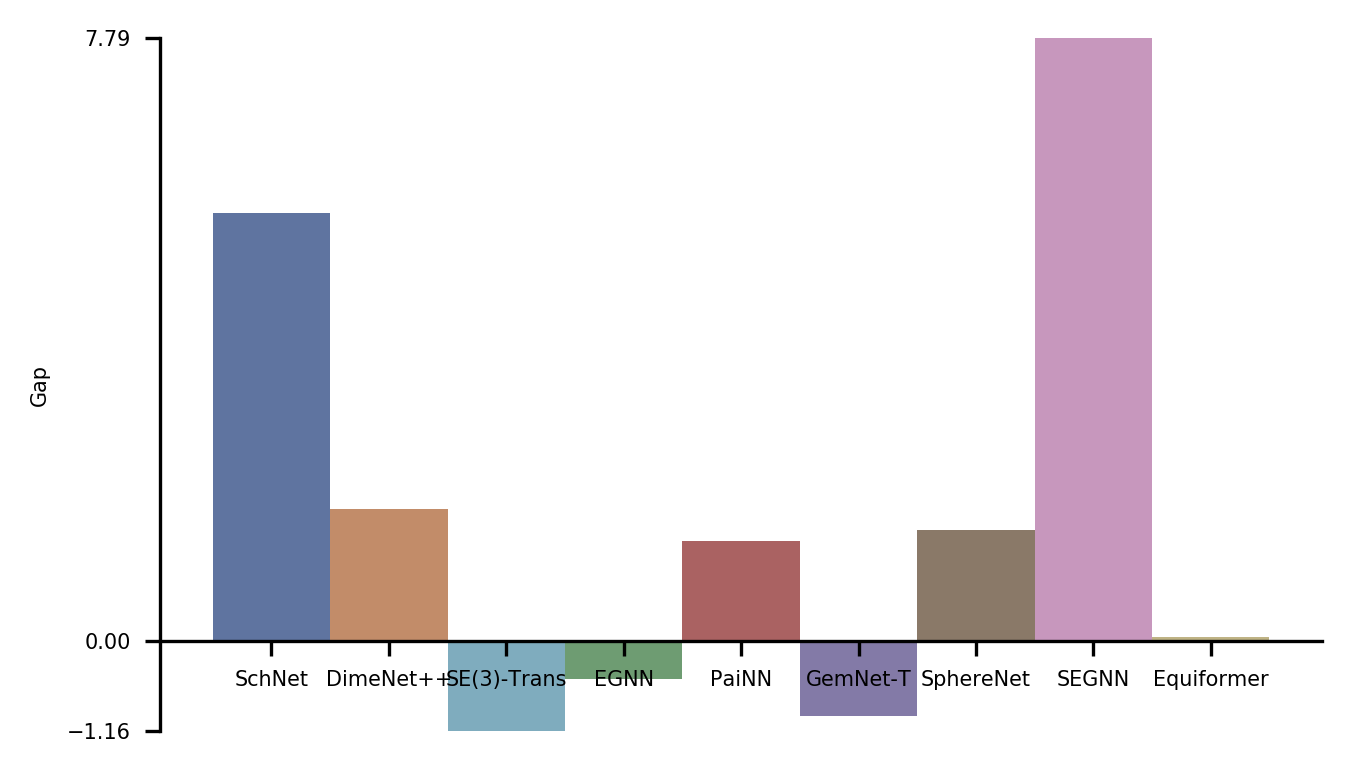}
    }
    \end{subfigure}
\hfill
    \begin{subfigure}[\small Task $\mathcal{E}_\text{HOMO}$]
    {\includegraphics[width=0.23\linewidth]{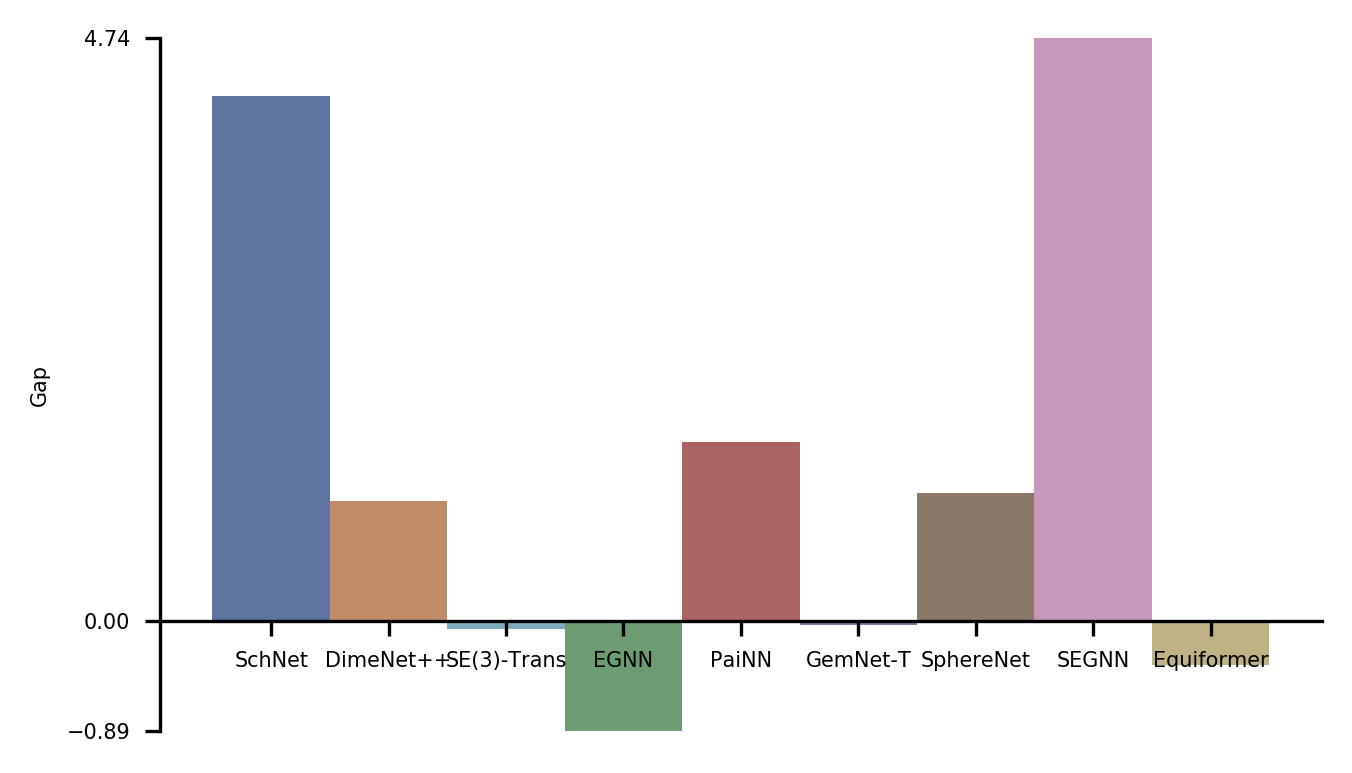}
    }
    \end{subfigure}
\hfill
    \begin{subfigure}[\small Task $\mathcal{E}_\text{LUMO}$]
    {\includegraphics[width=0.23\linewidth]{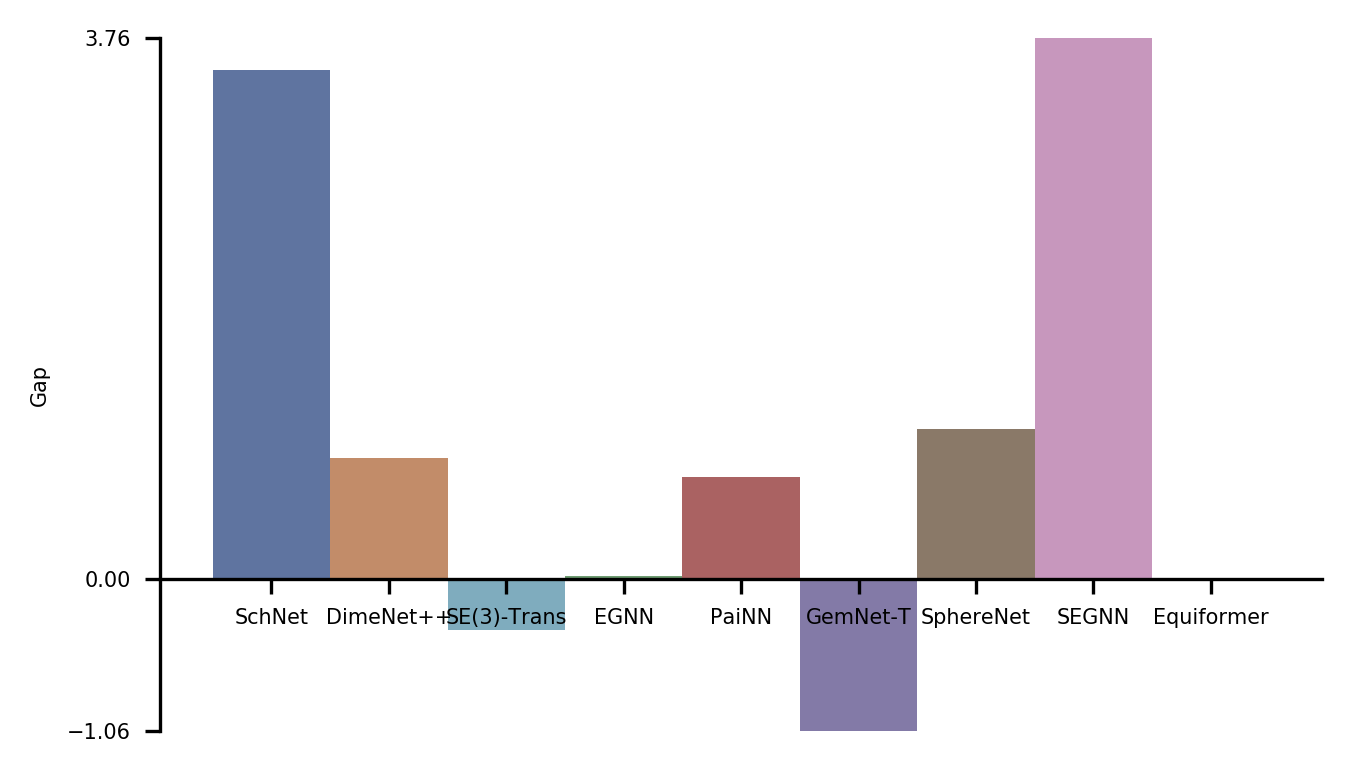}
    }
    \end{subfigure}
\hfill
    \begin{subfigure}[\small Task $\mu$]
    {\includegraphics[width=0.23\linewidth]{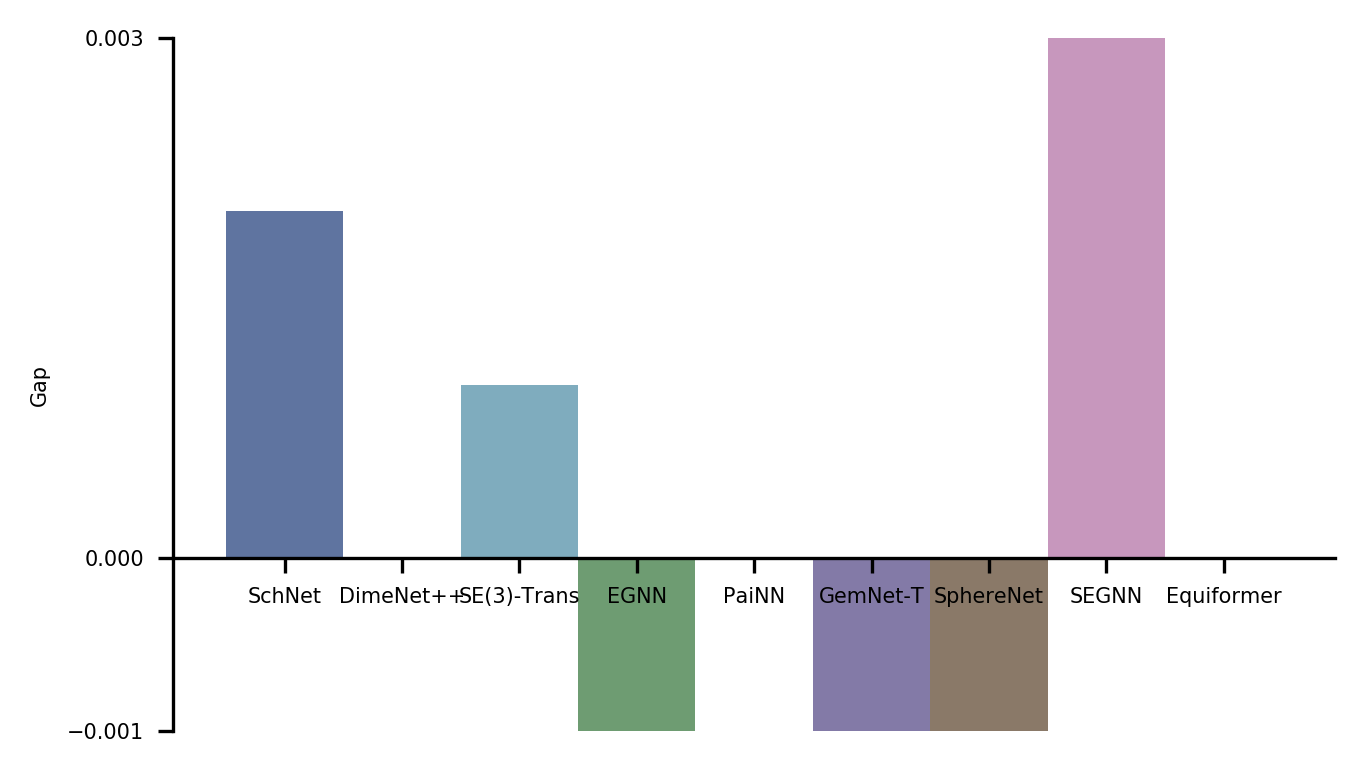}
    }
    \end{subfigure}
\hfill
    \begin{subfigure}[\small Task $C_v$]
    {\includegraphics[width=0.23\linewidth]{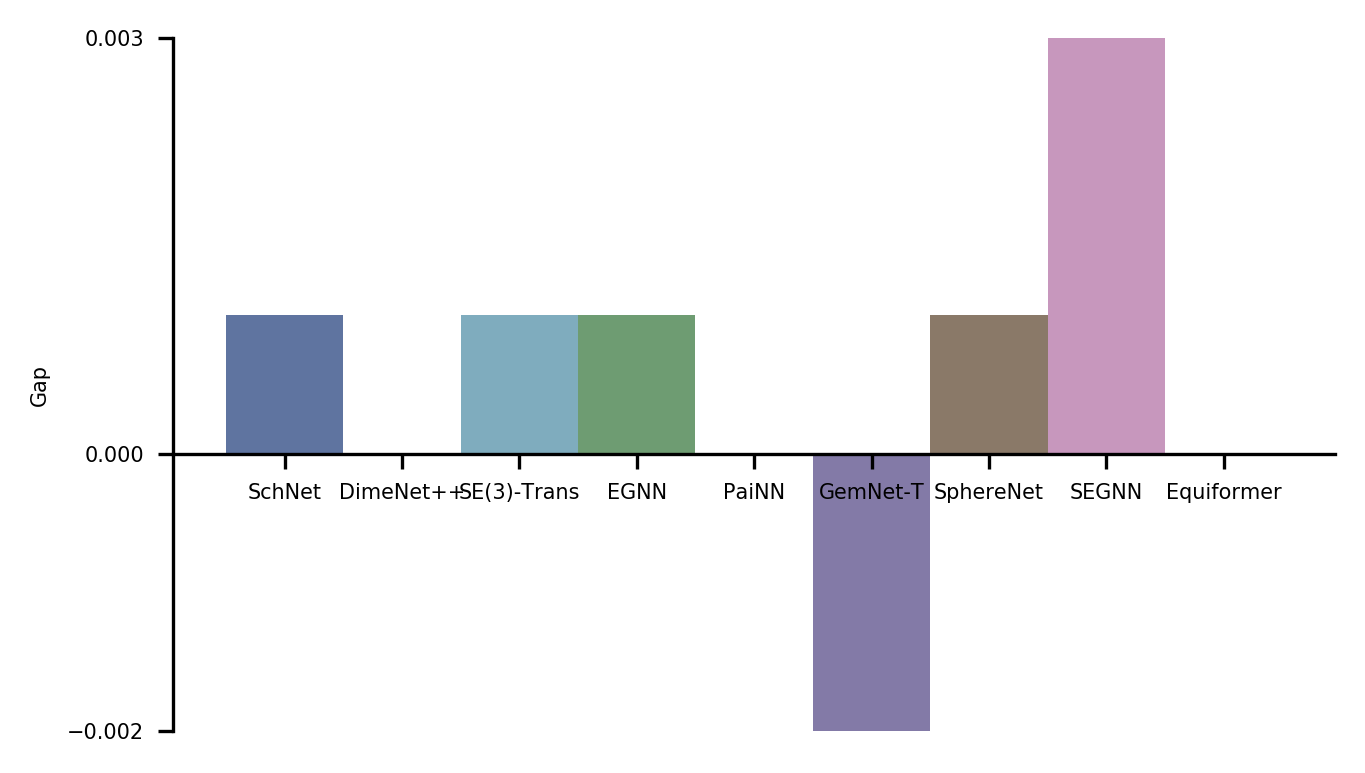}
    }
    \end{subfigure}
\hfill
    \begin{subfigure}[\small Task $G$]
    {\includegraphics[width=0.23\linewidth]{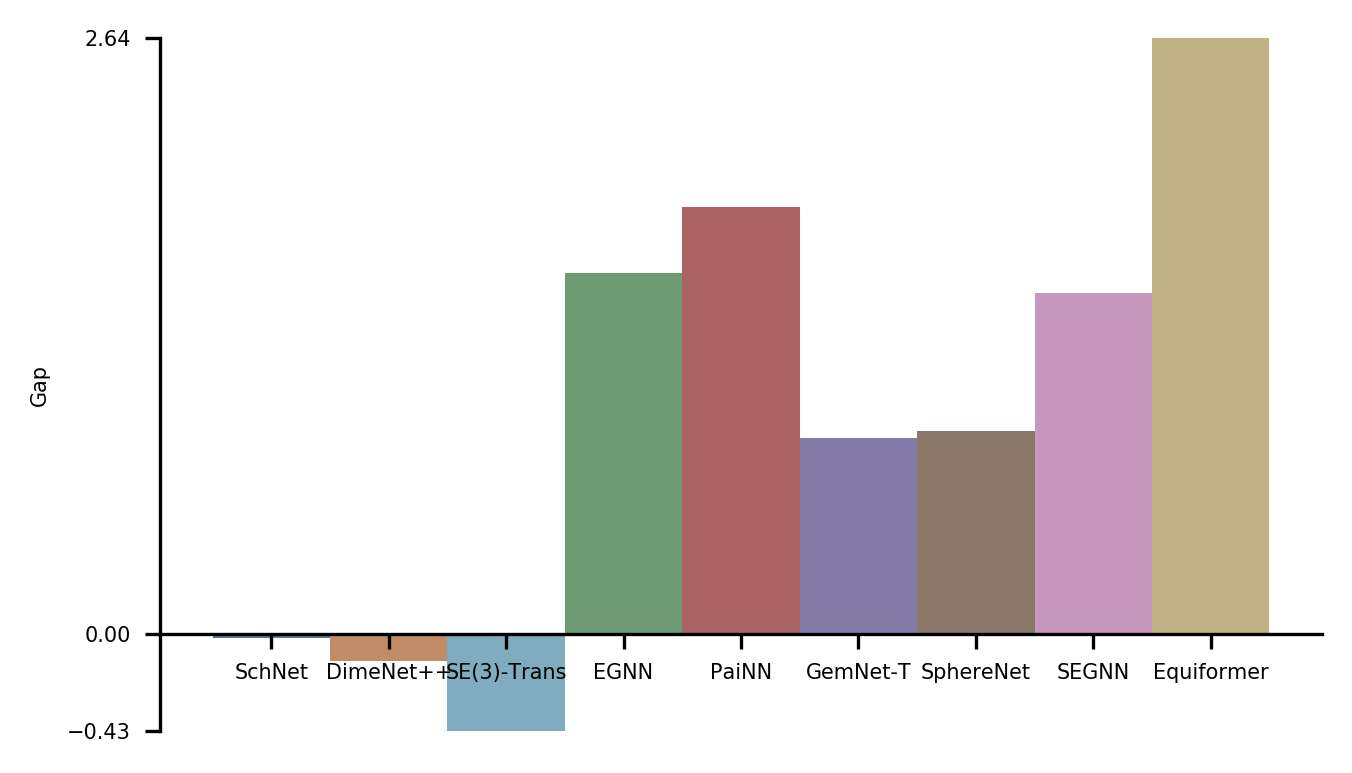}
    }
    \end{subfigure}
\hfill
    \begin{subfigure}[\small Task $H$]
    {\includegraphics[width=0.23\linewidth]{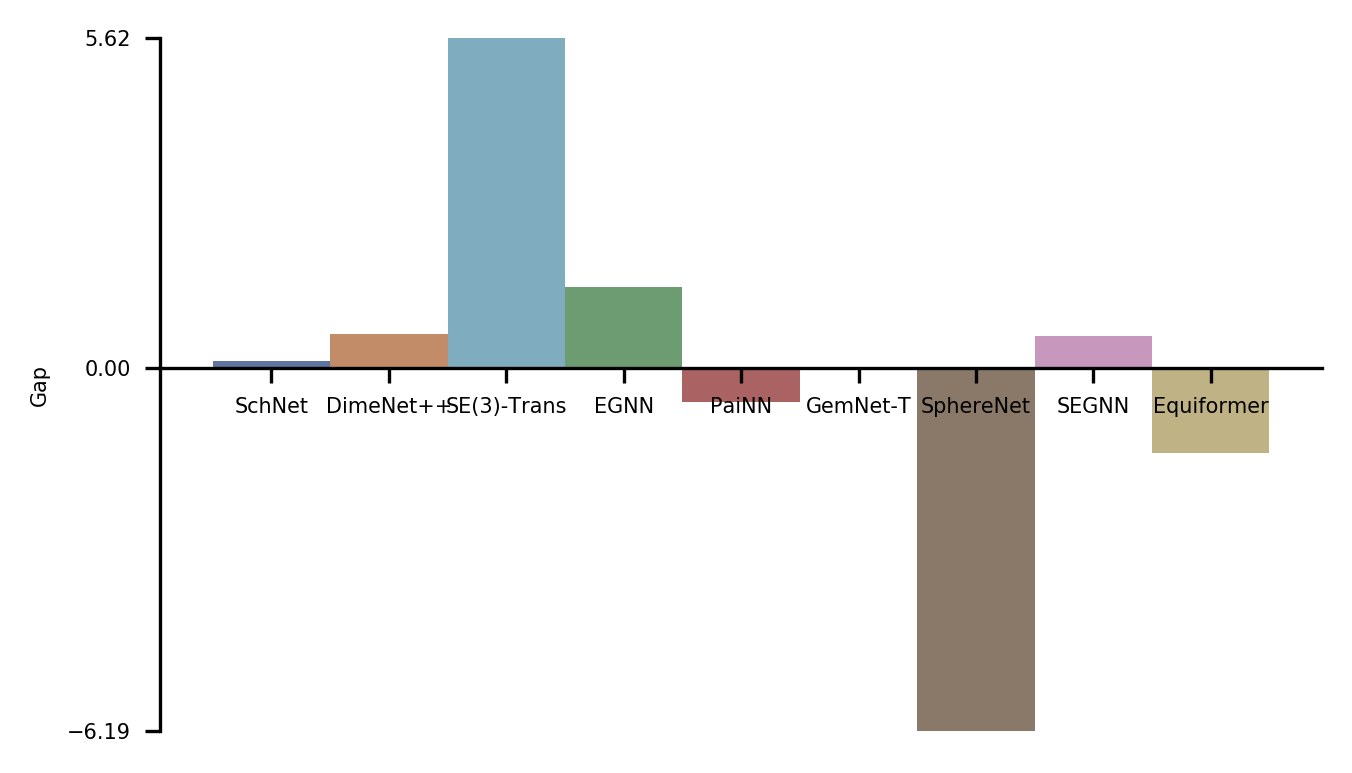}
    }
    \end{subfigure}
\hfill
    \begin{subfigure}[\small Task $R^2$]
    {\includegraphics[width=0.23\linewidth]{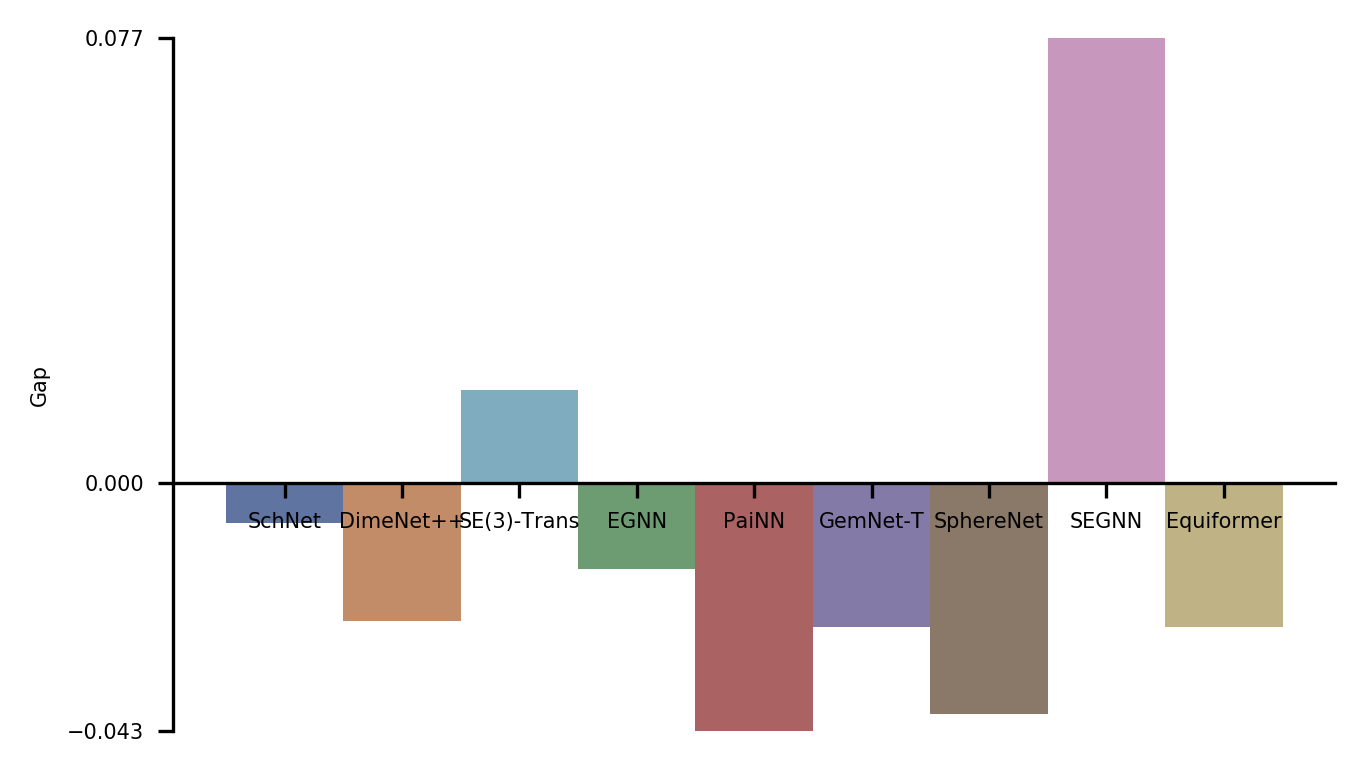}
    }
    \end{subfigure}
\hfill
    \begin{subfigure}[\small Task $U$]
    {\includegraphics[width=0.23\linewidth]{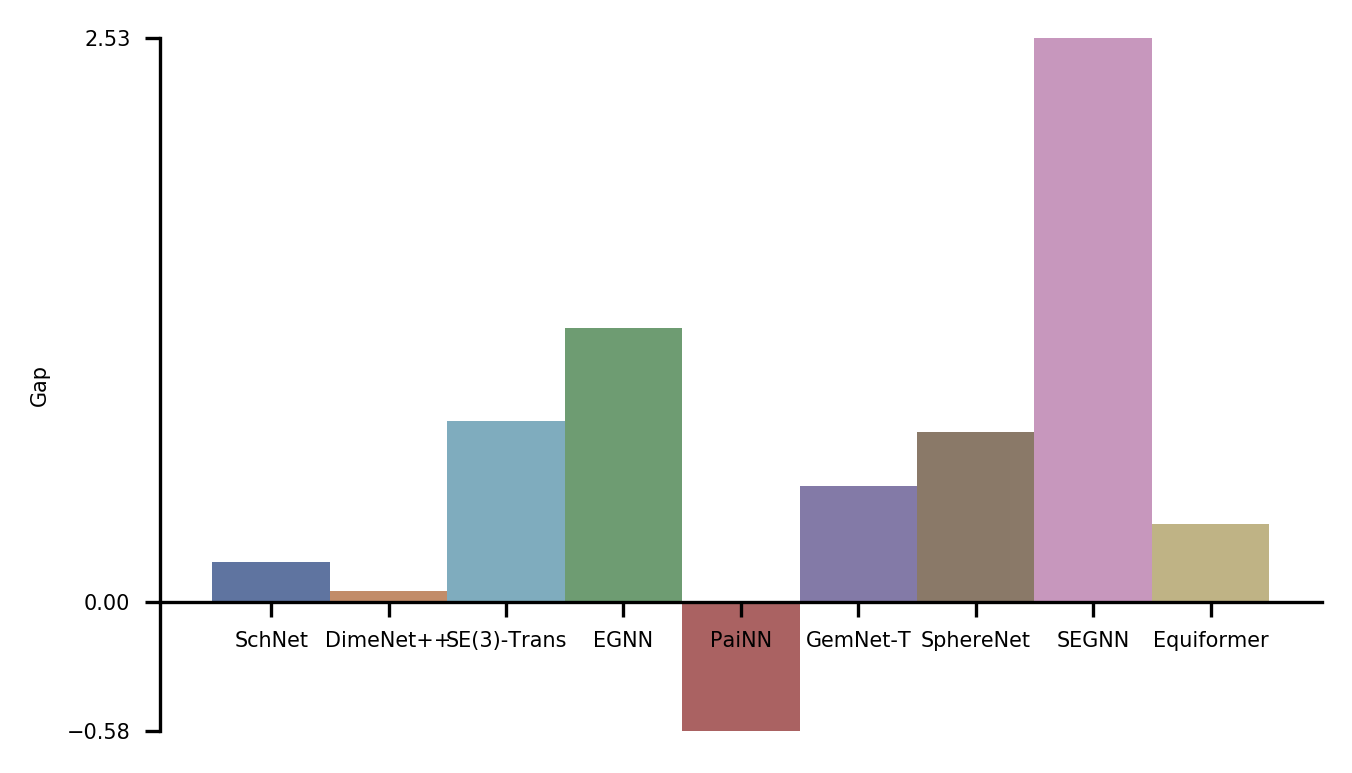}
    }
    \end{subfigure}
\hfill
    \begin{subfigure}[\small Task $U_0$]
    {\includegraphics[width=0.23\linewidth]{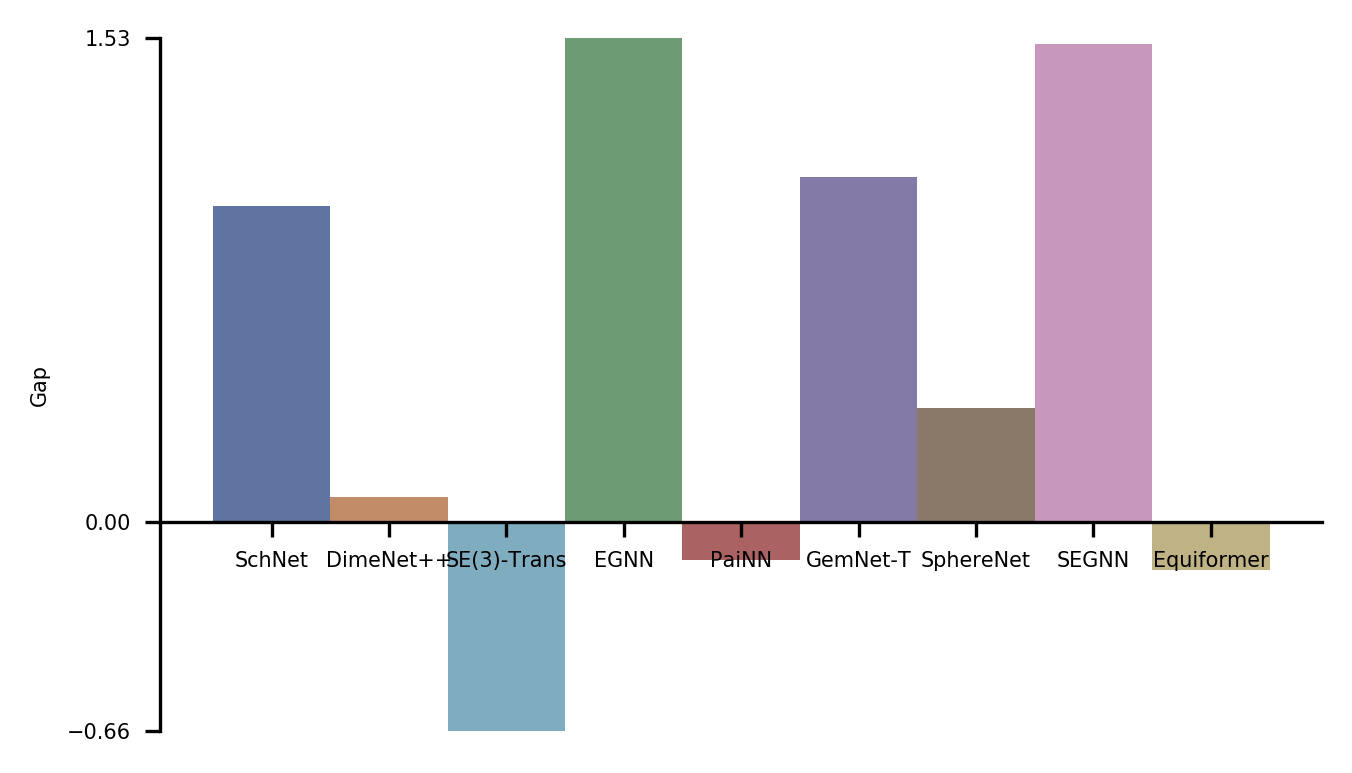}
    }
    \end{subfigure}
\hfill
    \begin{subfigure}[\small Task ZPVE]
    {\includegraphics[width=0.23\linewidth]{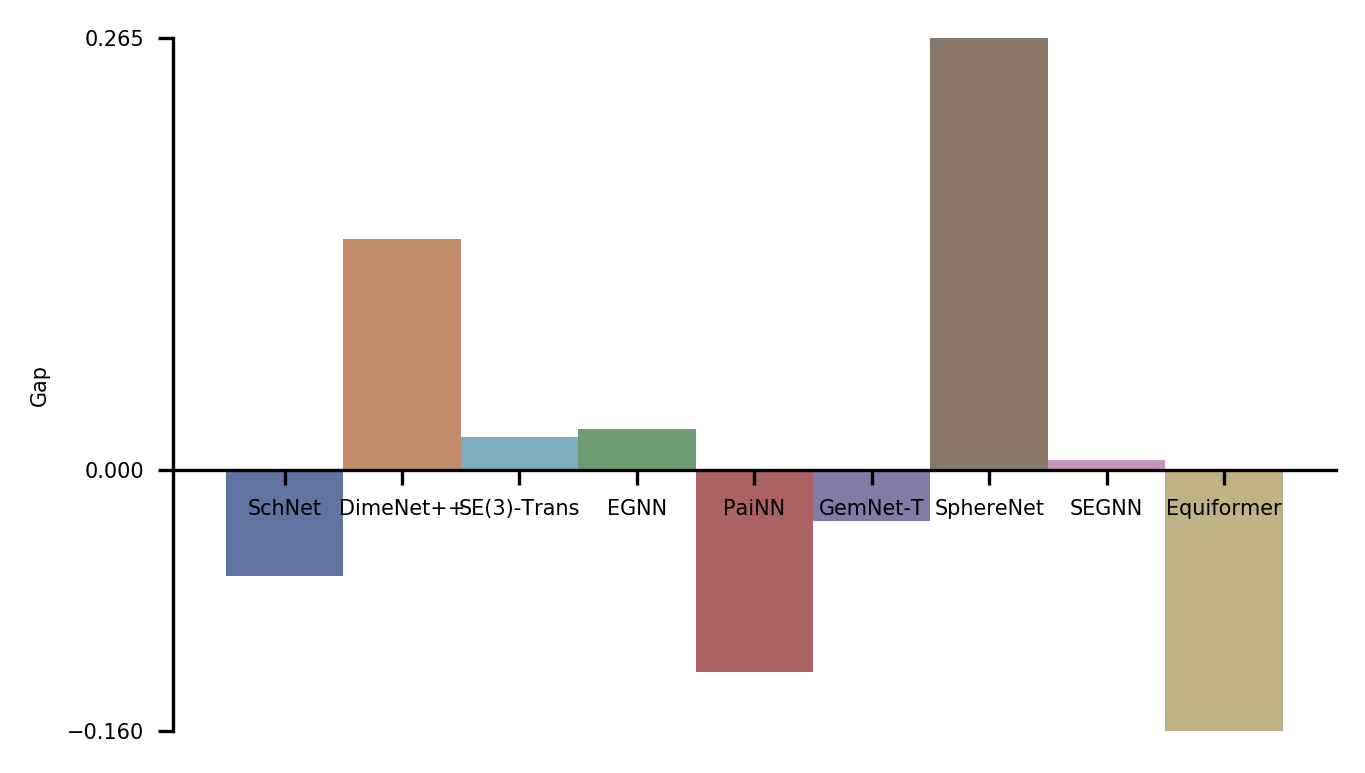}
    }
    \end{subfigure}
\vspace{-2ex}
\caption{
\small
Performance gap of MAE($d=128$) - MAE($d=300$) in QM9.
}
\label{fig:ablation_study_latent_dim_QM9}
\vspace{-2.5ex}
\end{figure}

Recent works~\cite{sun2022rethinking,wang2022evaluating} have found that the latent dimensions play an important role in molecule pretraining, and here we list the comparison between latent dimension $d=128$ and latent dimension $d=300$. 
\begin{itemize}[noitemsep,topsep=0pt]
    \item The performance comparison for QM9 is in~\Cref{tab:ablation_study_latent_dim_QM9}, and we visually plot the performance gap MAE($d=128$) - MAE($d=300$) in~\Cref{fig:ablation_study_latent_dim_QM9}. The results with $d=300$ are reported in~\Cref{tab:main_result_QM9}.
    \item The performance \textbf{(w/ normalization)} comparison for MD17 and rMD17 is in~\Cref{tab:ablation_study_latent_dim_128_MD17,tab:ablation_study_latent_dim_300_MD17,tab:ablation_study_latent_dim_128_rMD17,tab:ablation_study_latent_dim_300_rMD17}. The results with $d=300$ are reported in~\Cref{tab:main_result_MD17_rMD17,tab:main_result_MD17,tab:main_result_rMD17} except NequIP and Allegro. Their results in~\Cref{sec:app:reproduced_results_NequIP_Allegro} \textbf{(w/ normalization)} are reported~\Cref{tab:main_result_MD17_rMD17}.
    \item The performance comparison for COLL is in~\Cref{tab:ablation_study_latent_dim_COLL}, and results with $d=300$ are reported in~\Cref{tab:main_result_COLL}.
    \item The performance comparison for LBA \& LEP is in~\Cref{tab:ablation_study_latent_dim_LBA_LEP_128,tab:ablation_study_latent_dim_LBA_LEP_300}, and results with $d=300$ are reported in~\Cref{tab:main_results_LBA_LEP}.
\end{itemize}

\begin{table}[htb!]
\setlength{\tabcolsep}{10pt}
\fontsize{9}{9}\selectfont
\centering
\caption{
\small
Ablation studies of latent dimension $d$ on QM9.
110K for training, 10K for validation, and 11K for testing. The evaluation metric is the mean absolute error (MAE).
}
\label{tab:ablation_study_latent_dim_QM9}
\vspace{-1.5ex}
\begin{adjustbox}{max width=\textwidth}
\begin{tabular}{l l rrrrrrrrrrrrrrr}
\toprule
Model & $d$ & $\alpha$ $\downarrow$ & $\nabla \mathcal{E}$ $\downarrow$ & $\mathcal{E}_\text{HOMO}$ $\downarrow$ & $\mathcal{E}_\text{LUMO}$ $\downarrow$ & $\mu$ $\downarrow$ & $C_v$ $\downarrow$ & $G$ $\downarrow$ & $H$ $\downarrow$ & $R^2$ $\downarrow$ & $U$ $\downarrow$ & $U_0$ $\downarrow$ & ZPVE $\downarrow$\\
\midrule

\multirow{2}{*}{SchNet}
% 5e-4_CosineAnnealingLR_128_128_1000
& 128 & 0.068 & 49.66 & 31.91 & 26.09 & 0.030 & 0.032 & 14.17 & 14.16 & 0.126 & 14.11 & 14.27 & 1.684\\
% 5e-4_CosineAnnealingLR_300_128_1000
& 300 & 0.060 & 44.13 & 27.64 & 22.55 & 0.028 & 0.031 & 14.19 & 14.05 & 0.133 & 13.93 & 13.27 & 1.749\\
\midrule

\multirow{2}{*}{DimeNet++}
% 5e-4_CosineAnnealingLR_128_5_500
& 128 & 0.046 & 37.93 & 20.99 & 17.50 & 0.028 & 0.022 & 7.33 & 6.72 & 0.299 & 6.38 & 7.26 & 1.260\\
% 5e-4_CosineAnnealingLR_300_5_500
& 300 & 0.044 & 36.22 & 20.01 & 16.66 & 0.028 & 0.022 & 7.45 & 6.14 & 0.323 & 6.33 & 7.18 & 1.118\\
\midrule

\multirow{2}{*}{SE(3)-Trans}
% 1e-3_CosineAnnealingWarmRestarts_128_100_7_32_4_2_8_96_use_rotation_transform
& 128 & 0.144 & 55.36 & 34.59 & 34.05 & 0.051 & 0.064 & 64.85 & 76.32 & 1.763 & 69.73 & 68.22 & 5.448\\
% 1e-3_CosineAnnealingWarmRestarts_300_100_7_32_4_2_8_96_use_rotation_transform
& 300 & 0.137 & 56.52 & 34.65 & 34.41 & 0.050 & 0.063 & 65.28 & 70.70 & 1.747 & 68.92 & 68.88 & 5.428\\
\midrule

\multirow{2}{*}{EGNN}
% 5e-4_CosineAnnealingLR_128_1000
& 128 & 0.065 & 49.07 & 29.19 & 25.00 & 0.028 & 0.031 & 11.61 & 10.52 & 0.074 & 10.51 & 10.61 & 1.544\\
% 5e-4_CosineAnnealingLR_300_1000
& 300 & 0.062 & 49.56 & 30.08 & 24.98 & 0.029 & 0.030 & 10.01 & 9.14 & 0.089 & 9.28 & 9.08 & 1.519\\
\midrule

\multirow{2}{*}{PaiNN}
% 5e-4_CosineAnnealingLR_128_128_1000
& 128 & 0.049 & 44.02 & 25.92 & 20.87 & 0.016 & 0.025 & 10.32 & 7.30 & 0.126 & 7.60 & 7.51 & 1.295\\
% 5e-4_CosineAnnealingLR_300_128_1000
& 300 & 0.049 & 42.73 & 24.46 & 20.16 & 0.016 & 0.025 & 8.43 & 7.88 & 0.169 & 8.18 & 7.63 & 1.419\\
\midrule

\multirow{2}{*}{GemNet-T}
% 5e-4_CosineAnnealingLR_128_128_1000
& 128 & 0.042 & 34.49 & 17.82 & 14.80 & 0.020 & 0.021 & 8.48 & 7.05 & 0.246 & 6.94 & 6.97 & 1.201\\
% 5e-4_CosineAnnealingLR_300_128_1000
& 300 & 0.041 & 35.46 & 17.85 & 15.86 & 0.021 & 0.023 & 7.61 & 7.08 & 0.271 & 6.42 & 5.88 & 1.232\\
\midrule

\multirow{2}{*}{SphereNet}
% 5e-4_CosineAnnealingLR_128_1000
& 128 & 0.050 & 40.36 & 22.49 & 19.29 & 0.026 & 0.026 & 9.06 & 7.49 & 0.248 & 7.53 & 7.79 & 1.560\\
% 5e-4_CosineAnnealingLR_300_1000
& 300 & 0.047 & 38.93 & 21.45 & 18.25 & 0.027 & 0.025 & 8.16 & 13.68 & 0.288 & 6.77 & 7.43 & 1.295\\
\midrule

\multirow{2}{*}{SEGNN}
% 1e-4_CosineAnnealingLR_128_500
& 128 & 0.056 & 41.40 & 22.40 & 20.77 & 0.024 & 0.029 & 13.11 & 12.99 & 0.481 & 13.82 & 13.71 & 1.596\\
% 1e-4_CosineAnnealingLR_300_500
& 300 & 0.048 & 33.61 & 17.66 & 17.01 & 0.021 & 0.026 & 11.60 & 12.45 & 0.404 & 11.29 & 12.20 & 1.590\\
\midrule

\multirow{2}{*}{Equiformer}
% 5e-4_CosineAnnealingLR_128_128_300
& 128 & 0.051 & 33.52 & 17.58 & 16.83 & 0.015 & 0.023 & 17.13 & 13.14 & 0.408 & 15.23 & 13.63 & 2.182\\
% 5e-4_CosineAnnealingLR_300_128_300
& 300 & 0.051 & 33.46 & 17.93 & 16.85 & 0.015 & 0.023 & 14.49 & 14.60 & 0.433 & 14.88 & 13.78 & 2.342\\
\bottomrule
\end{tabular}
\end{adjustbox}
\vspace{-3ex}
\end{table}

%%%%%%%%%%%%%%%%%%%%%%%%%%%%%%%%%%%%%%%%%%%%%%%%%%

\clearpage
\begin{table}[htb!]
\setlength{\tabcolsep}{5pt}
\fontsize{9}{9}\selectfont
\centering
\caption{
\small
Ablation studies of latent dimension ($d=128$) on MD17.
The evaluation is the mean absolute error.
No data normalization is used.
}
\label{tab:ablation_study_latent_dim_128_MD17}
\vspace{-1.5ex}
\begin{adjustbox}{max width=\textwidth}
\begin{tabular}{ll c c c c c c c c}
\toprule
Model & Energy / Force & Aspirin $\downarrow$ & Benzene $\downarrow$ & Ethanol $\downarrow$ & Malonaldehyde $\downarrow$ & Naphthalene $\downarrow$ & Salicylic $\downarrow$ & Toluene $\downarrow$ & Uracil $\downarrow$ \\
\midrule

\multirow{2}{*}{SchNet}
% 5e-4_CosineAnnealingLR_128_5_1000_energy_force_no_normalization
& Energy & 0.695 & 0.118 & 0.182 & 0.338 & 0.210 & 0.232 & 0.157 & 0.192\\
& Force & 1.500 & 0.399 & 0.663 & 1.157 & 0.759 & 0.896 & 0.594 & 0.906\\
\midrule

\multirow{2}{*}{DimeNet++}
% 5e-4_CosineAnnealingLR_128_800_energy_force_no_normalization
& Energy & 2.104 & 1.053 & 0.971 & 1.180 & 1.472 & 1.901 & 1.988 & 1.754\\
& Force & 2.209 & 0.476 & 0.636 & 1.420 & 1.293 & 1.071 & 0.924 & 1.070\\
\midrule

\multirow{2}{*}{EGNN}
% 1e-4_CosineAnnealingLR_128_5_1_1000_energy_force_no_normalization
& Energy & 91.490 & 0.663 & 1.439 & 1.385 & 17.064 & 31.006 & 7.190 & 1.409\\
& Force & 19.211 & 1.049 & 1.983 & 2.380 & 2.185 & 3.957 & 2.453 & 2.172\\
\midrule

\multirow{2}{*}{PaiNN}
% 5e-4_CosineAnnealingLR_128_5_1000_energy_force_no_normalization
& Energy & 0.209 & 0.097 & 0.070 & 0.093 & 0.235 & 0.127 & 0.133 & 0.107\\
& Force & 0.549 & 0.053 & 0.198 & 0.328 & 0.134 & 0.284 & 0.146 & 0.180\\
\midrule

\multirow{2}{*}{GemNet-T}
% 5e-4_CosineAnnealingLR_128_5_1000_energy_force_no_normalization
& Energy & 1.299 & 0.096 & 8.418 & 0.101 & 0.116 & 0.141 & 0.095 & 11.270\\
& Force & 0.518 & 0.050 & 0.226 & 0.380 & 0.107 & 0.259 & 0.118 & 542.330\\
\midrule

\multirow{2}{*}{SphereNet}
% 1e-4_CosineAnnealingLR_128_1_1000_energy_force_no_normalization
& Energy & 0.235 & 0.104 & 0.327 & 0.136 & 0.183 & 0.771 & 0.116 & 0.147\\
& Force & 0.500 & 0.114 & 0.199 & 0.377 & 0.416 & 2.033 & 0.198 & 0.303\\
\midrule

\multirow{2}{*}{SEGNN}
% 1e-4_CosineAnnealingLR_128_5_5_0_800_energy_force_no_normalization
& Energy & 10.030 & 0.081 & 0.088 & 0.191 & 0.678 & 1.699 & 0.541 & 0.260\\
& Force & 6.793 & 0.193 & 0.456 & 0.832 & 0.734 & 1.828 & 0.957 & 0.654\\
\midrule

\multirow{2}{*}{Allegro}
% 1e-3_CosineAnnealingLR_4.0_128_1000_energy_force_no_normalization
& Energy & 2.380 & 0.278 & 0.386 & 0.583 & 0.732 & 1.131 & 0.615 & 1.357\\
& Force & 6.537 & 1.777 & 1.916 & 2.572 & 3.359 & 5.063 & 3.022 & 6.974\\
\midrule

\multirow{2}{*}{Equiformer}
% 1e-4_CosineAnnealingLR_128_5_1_1000_energy_force_no_normalization
& Energy & 0.708 & 0.076 & 0.056 & 0.102 & 0.097 & 0.191 & 0.094 & 0.103\\
& Force & 0.282 & 0.044 & 0.142 & 0.229 & 0.068 & 0.202 & 0.080 & 0.140\\
\bottomrule
\end{tabular}
\end{adjustbox}
\end{table}

\begin{table}[htb!]
\setlength{\tabcolsep}{5pt}
\fontsize{9}{9}\selectfont
\centering
\caption{
\small
Ablation studies of latent dimension ($d=300$) on MD17.
The evaluation is the mean absolute error.
No data normalization is used.
}
\label{tab:ablation_study_latent_dim_300_MD17}
\vspace{-1.5ex}
\begin{adjustbox}{max width=\textwidth}
\begin{tabular}{ll c c c c c c c c}
\toprule
Model & Energy/Force & Aspirin $\downarrow$ & Benzene $\downarrow$ & Ethanol $\downarrow$ & Malonaldehyde $\downarrow$ & Naphthalene $\downarrow$ & Salicylic $\downarrow$ & Toluene $\downarrow$ & Uracil $\downarrow$ \\
\midrule

\multirow{2}{*}{SchNet}
% 5e-4_CosineAnnealingLR_300_5_1000_energy_force_no_normalization
& Energy & 0.475 & 0.117 & 0.109 & 0.300 & 0.167 & 0.212 & 0.149 & 0.170\\
& Force & 1.203 & 0.380 & 0.386 & 0.794 & 0.587 & 0.826 & 0.568 & 0.773\\
\midrule

\multirow{2}{*}{DimeNet++}
% 5e-4_CosineAnnealingLR_300_800_energy_force_no_normalization
& Energy & 4.168 & 0.893 & 1.238 & 1.385 & 1.846 & 2.445 & 1.484 & 1.522\\
& Force & 7.212 & 0.603 & 0.753 & 1.842 & 8.515 & 1.752 & 1.037 & 1.632\\
\midrule

\multirow{2}{*}{EGNN}
% 1e-4_CosineAnnealingLR_300_5_1_1000_energy_force_no_normalization
& Energy & 17.892 & 1.142 & 0.436 & 0.896 & 12.177 & 6.964 & 4.051 & 0.854\\
& Force & 3.042 & 0.736 & 0.924 & 1.566 & 1.136 & 1.177 & 1.202 & 1.367\\
\midrule

\multirow{2}{*}{PaiNN}
% 5e-4_CosineAnnealingLR_300_5_1000_energy_force_no_normalization
& Energy & 27.626 & 0.095 & 0.063 & 0.102 & 0.622 & 0.371 & 0.165 & 0.111\\
& Force & 0.572 & 0.053 & 0.230 & 0.338 & 0.132 & 0.288 & 0.141 & 0.201\\
\midrule

\multirow{2}{*}{GemNet-T}
% 5e-4_CosineAnnealingLR_300_5_1000_energy_force_no_normalization
& Energy & 0.684 & 0.097 & 4.598 & 4.966 & 0.482 & 0.128 & 0.098 & 1.349\\
& Force & 0.558 & 0.089 & 0.219 & 0.433 & 0.212 & 0.326 & 0.174 & 486.612\\
\midrule

\multirow{2}{*}{SphereNet}
% 1e-4_CosineAnnealingLR_300_1_1000_energy_force_no_normalization
& Energy & 0.244 & 0.107 & 1.603 & 1.559 & 0.167 & 0.188 & 0.113 & 7.115\\
& Force & 0.546 & 0.135 & 0.168 & 0.667 & 0.315 & 0.479 & 0.194 & 0.442\\
\midrule

\multirow{2}{*}{SEGNN}
% 1e-4_CosineAnnealingLR_300_5_5_0_800_energy_force_no_normalization
& Energy & 17.774 & 0.086 & 0.151 & 0.247 & 0.655 & 2.173 & 0.624 & 0.259\\
& Force & 9.003 & 0.265 & 0.893 & 1.249 & 0.895 & 2.220 & 1.138 & 0.948\\
\midrule

\multirow{2}{*}{Allegro}
% 1e-3_CosineAnnealingLR_4.0_300_1000_energy_force_no_normalization
& Energy & 1.577 & 0.117 & 0.308 & 0.481 & 0.899 & 1.088 & 0.406 & 0.490\\
& Force & 4.328 & 0.358 & 1.613 & 2.185 & 3.841 & 4.731 & 1.866 & 2.627\\
\midrule

\multirow{2}{*}{Equiformer}
% 1e-4_CosineAnnealingLR_300_5_1_1000_energy_force_no_normalization
& Energy & 0.308 & 0.075 & 0.096 & 0.183 & 0.097 & 0.189 & 0.209 & 0.106\\
& Force & 0.286 & 0.045 & 0.142 & 0.230 & 0.068 & 0.200 & 0.080 & 0.141\\
\bottomrule
\end{tabular}
\end{adjustbox}
\end{table}

\begin{figure}[htb!]
\centering
    \begin{subfigure}[\small Task Aspirin]
    {\includegraphics[width=0.24\linewidth]{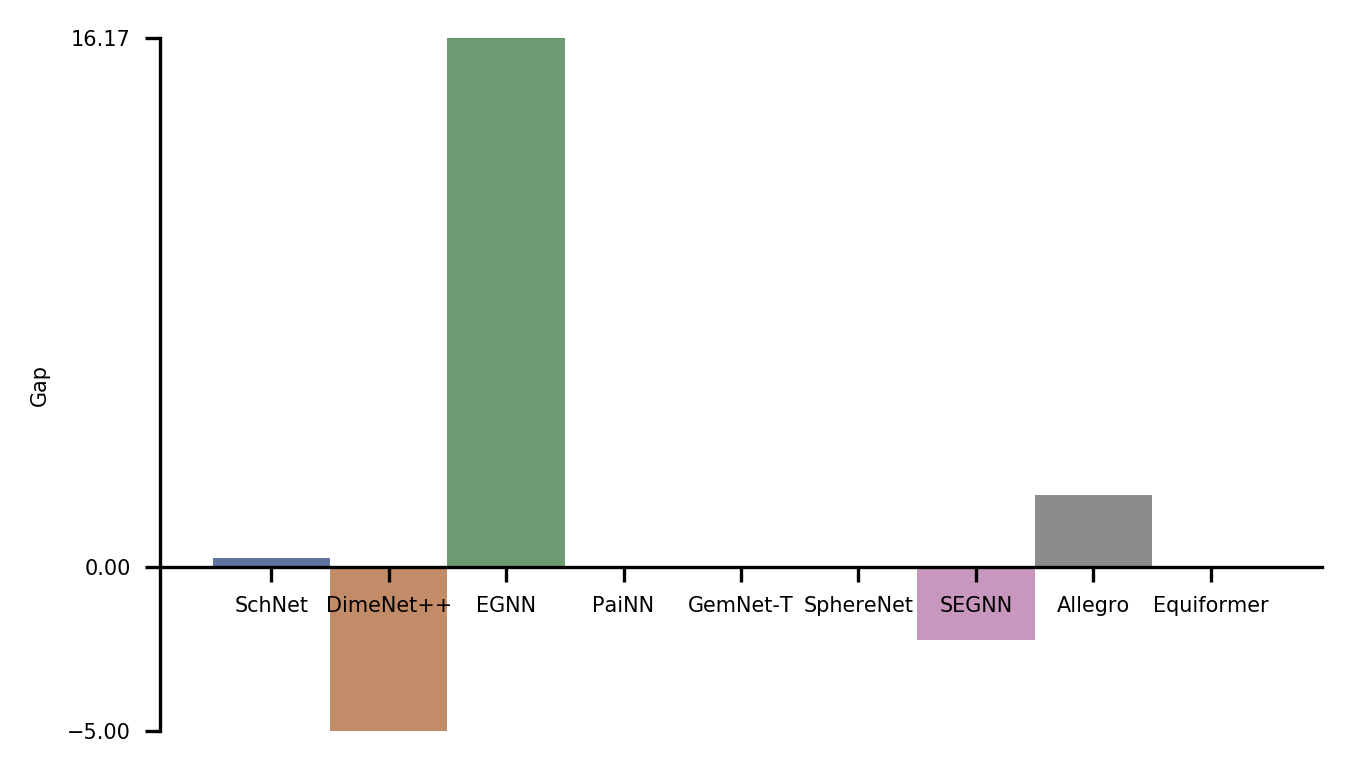}}
    \end{subfigure}
\hfill
    \begin{subfigure}[\small Task Benzene]
    {\includegraphics[width=0.24\linewidth]{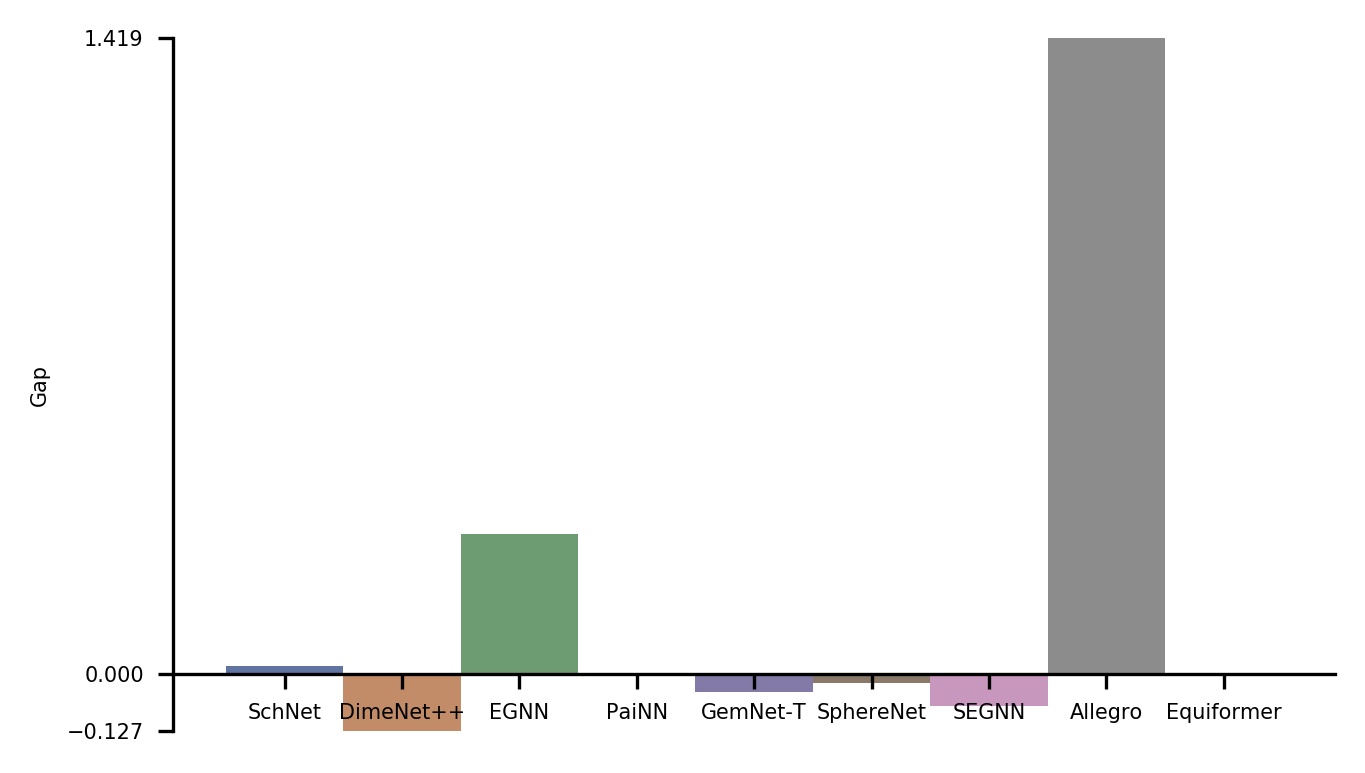}}
    \end{subfigure}
\hfill
    \begin{subfigure}[\small Task Ethanol]
    {\includegraphics[width=0.24\linewidth]{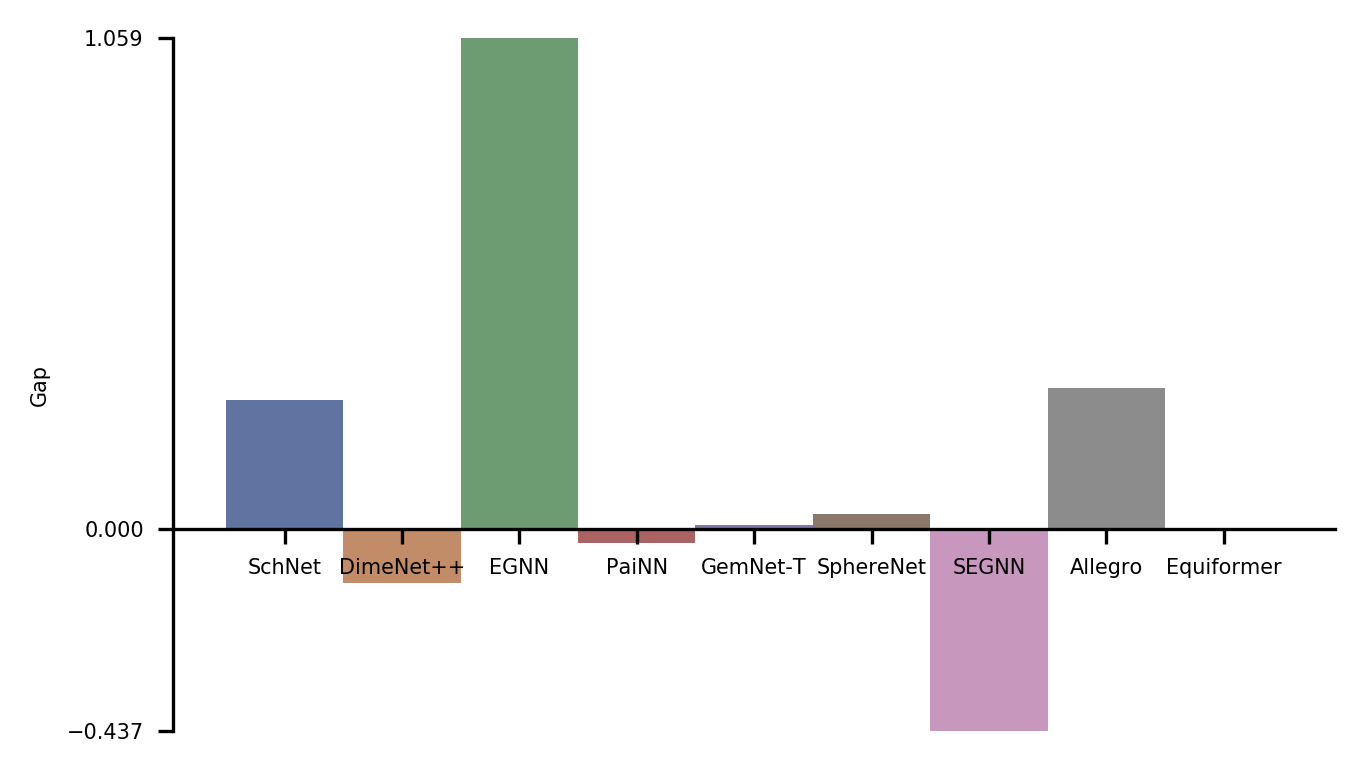}}
    \end{subfigure}
\hfill
    \begin{subfigure}[\small Task Malonaldehyde]
    {\includegraphics[width=0.24\linewidth]{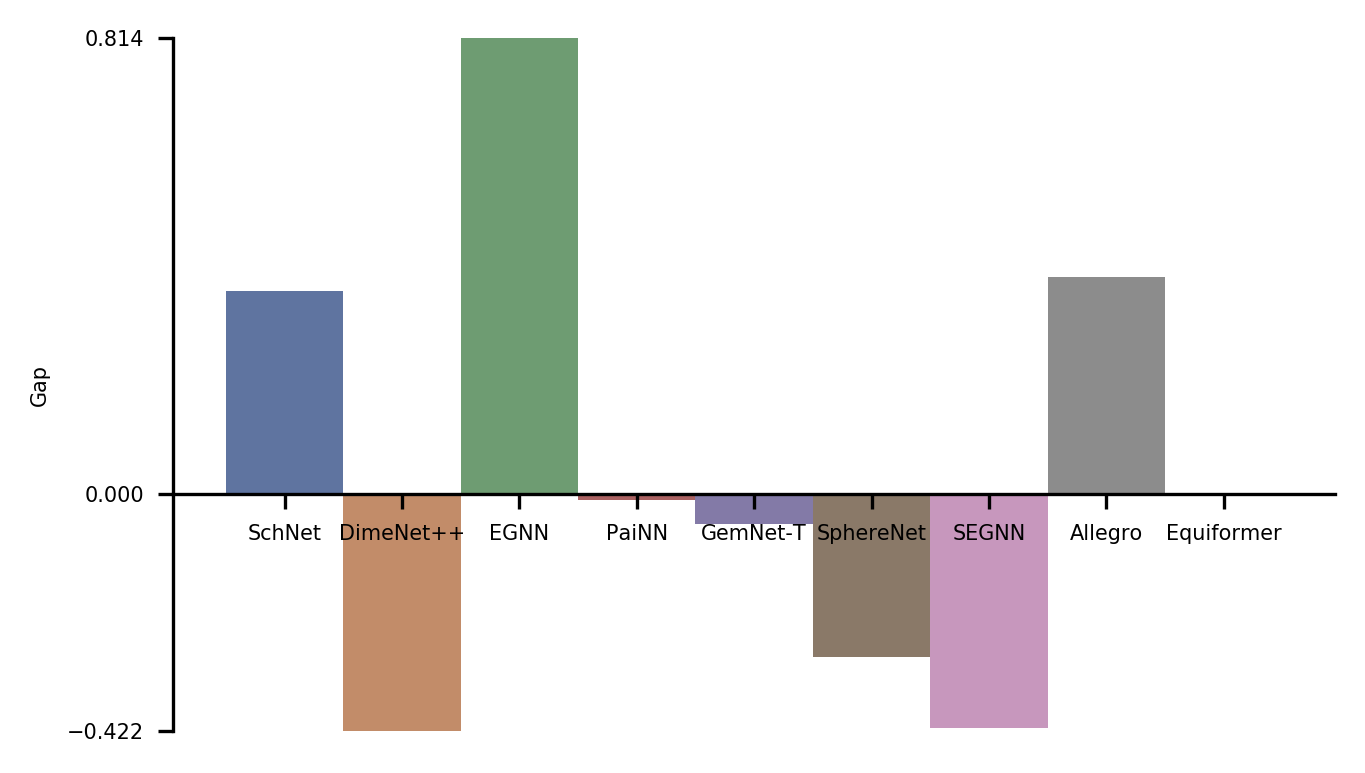}}
    \end{subfigure}
\hfill
    \begin{subfigure}[\small Task Naphthalene]
    {\includegraphics[width=0.24\linewidth]{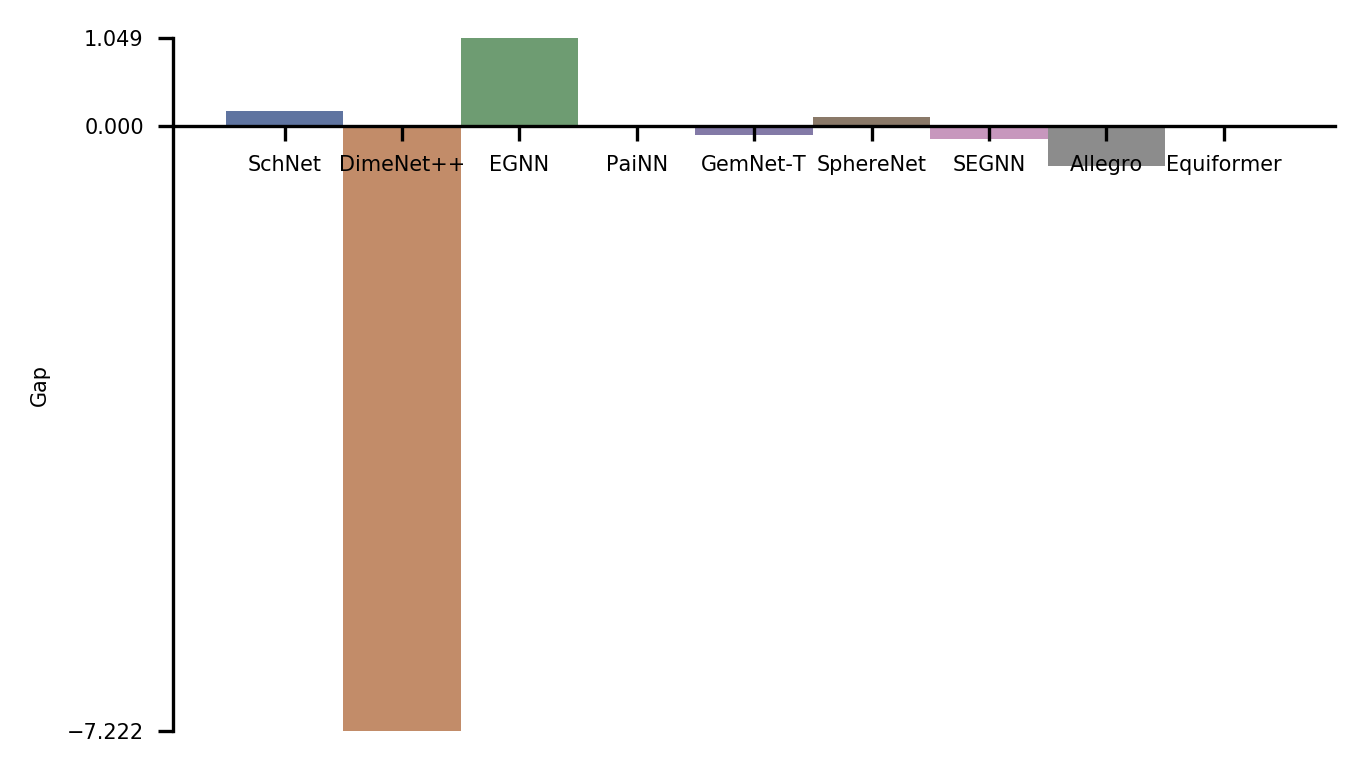}}
    \end{subfigure}
\hfill
    \begin{subfigure}[\small Task Salicylic]
    {\includegraphics[width=0.24\linewidth]{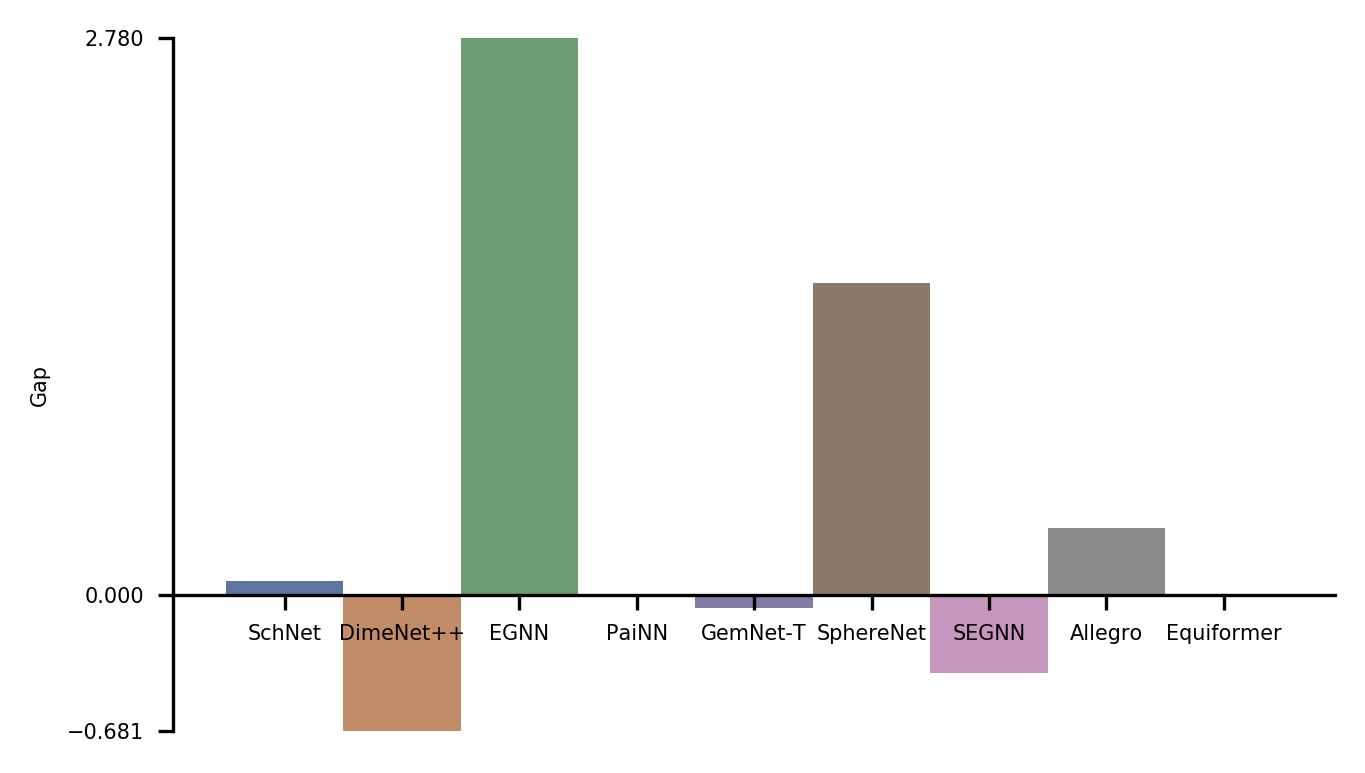}}
    \end{subfigure}
\hfill
    \begin{subfigure}[\small Task Toluene]
    {\includegraphics[width=0.24\linewidth]{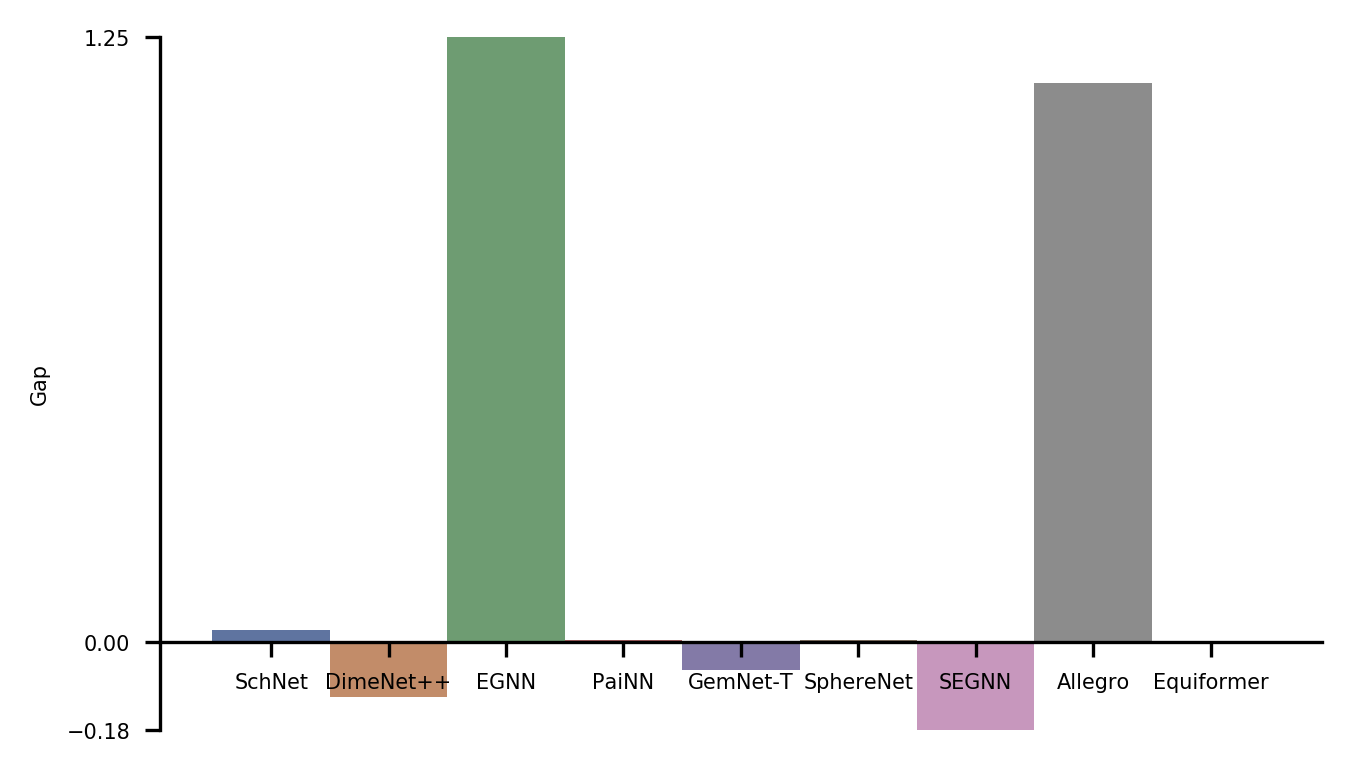}}
    \end{subfigure}
\hfill
    \begin{subfigure}[\small Task Uracil]
    {\includegraphics[width=0.24\linewidth]{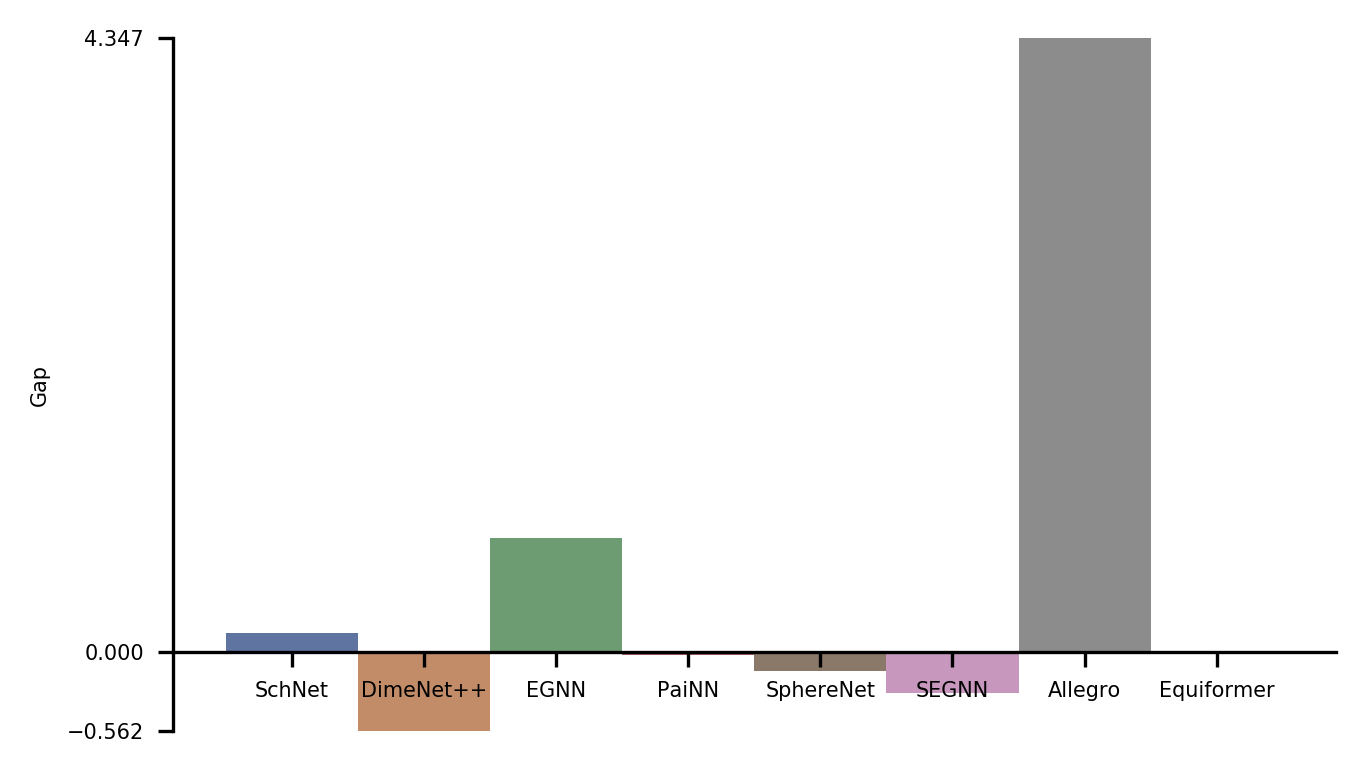}}
    \end{subfigure}
\vspace{-2ex}
\caption{
\small
Performance gap of MAE($d=128$) - MAE($d=300$) in MD17.
}
\label{fig:ablation_study_latent_dim_MD17}
\vspace{-2.5ex}
\end{figure}

\clearpage

\begin{table}[htb!]
\setlength{\tabcolsep}{5pt}
\fontsize{9}{9}\selectfont
\centering
\caption{
\small
Ablation studies of latent dimension (dim=128) on rMD17.
The evaluation is the mean absolute error.
No data normalization is used.
}
\label{tab:ablation_study_latent_dim_128_rMD17}
\vspace{-1.5ex}
\begin{adjustbox}{max width=\textwidth}
\begin{tabular}{ll c c c c c c c c c c}
\toprule
model & Energy / Force & Aspirin $\downarrow$ & Azobenzene $\downarrow$ & Benzene $\downarrow$ & Ethanol $\downarrow$ & Malonaldehyde $\downarrow$ & Naphthalene $\downarrow$ & Paracetamol $\downarrow$ & Salicylic $\downarrow$ & Toluene $\downarrow$ & Uracil $\downarrow$ \\
\midrule

\multirow{2}{*}{SchNet}
% 5e-4_CosineAnnealingLR_128_5_1000_energy_force_no_normalization
& Energy & 0.764 & 1.332 & 0.388 & 0.226 & 0.380 & 0.205 & 1.298 & 1.759 & 0.166 & 0.173\\
& Force & 1.662 & 3.071 & 0.318 & 0.782 & 1.109 & 0.805 & 2.344 & 0.950 & 0.693 & 0.943\\
\midrule

\multirow{2}{*}{DimeNet++}
% 5e-4_CosineAnnealingLR_128_800_energy_force_no_normalization
& Energy & 1.719 & 4.806 & 0.506 & 10.867 & 0.845 & 1.209 & 10.876 & 2.020 & 1.519 & 4227.668\\
& Force & 1.253 & 1.033 & 0.307 & 20.860 & 0.632 & 0.602 & 1.123 & 1.022 & 0.991 & 33549.676\\
\midrule

\multirow{2}{*}{EGNN}
% 1e-4_CosineAnnealingLR_128_5_1_1000_energy_force_no_normalization
& Energy & 89.661 & 51.554 & 4.893 & 1.065 & 9.339 & 32.901 & 77.996 & 27.114 & 12.766 & 4.519\\
& Force & 20.531 & 4.436 & 0.912 & 2.305 & 3.056 & 2.287 & 9.484 & 13.117 & 2.567 & 2.482\\
\midrule

\multirow{2}{*}{PaiNN}
% 1e-4_CosineAnnealingLR_128_5_1000_energy_force_no_normalization
& Energy & 1.949 & 5.733 & 0.036 & 0.606 & 1.626 & 2.610 & 0.541 & 0.831 & 0.158 & 0.181\\
& Force & 3.189 & 0.940 & 0.143 & 0.727 & 1.158 & 0.851 & 1.636 & 1.450 & 0.682 & 0.875\\
\midrule

\multirow{2}{*}{GemNet-T}
% 5e-4_CosineAnnealingLR_300_5_1000_energy_force_no_normalization
& Energy & 1.546 & 0.073 & 0.006 & 1.060 & 6.610 & 0.025 & 1.972 & 14.837 & 0.023 & 36.966\\
& Force & 0.555 & 0.265 & 0.026 & 0.211 & 0.425 & 0.112 & 0.368 & 0.308 & 0.120 & 0.233\\
\midrule

\multirow{2}{*}{SphereNet}
% 1e-4_CosineAnnealingLR_128_1_1000_energy_force_no_normalization
& Energy & 21.142 & 0.542 & 0.678 & 1.226 & 0.423 & 0.176 & 0.255 & 6.218 & 0.119 & 0.143\\
& Force & 0.666 & 0.781 & 0.102 & 0.313 & 0.419 & 0.500 & 0.659 & 2.244 & 0.334 & 0.425\\
\midrule

\multirow{2}{*}{SEGNN}
% 1e-4_CosineAnnealingLR_128_5_5_0_700_energy_force_no_normalization
& Energy & 11.828 & 2.729 & 0.018 & 0.081 & 0.161 & 1.333 & 3.982 & 1.476 & 1.443 & 0.221\\
& Force & 7.543 & 2.014 & 0.139 & 0.509 & 0.934 & 0.845 & 3.338 & 1.934 & 1.028 & 0.723\\
\midrule

\multirow{2}{*}{Allegro}
% 1e-3_CosineAnnealingLR_4.0_128_1000_energy_force_no_normalization
& Energy & 6.142 & 2.221 & 0.094 & 0.465 & 0.592 & 1.320 & 2.196 & 1.239 & 0.584 & 1.739\\
& Force & 4.891 & 5.727 & 0.960 & 2.166 & 2.630 & 3.546 & 4.571 & 5.949 & 2.885 & 6.610\\
\midrule

\multirow{2}{*}{Equiformer}
% 1e-4_CosineAnnealingLR_128_5_1_1000_energy_force_no_normalization
& Energy & 0.480 & 0.119 & 0.031 & 0.085 & 0.098 & 0.065 & 0.848 & 0.261 & 0.082 & 0.214\\
& Force & 0.303 & 0.132 & 0.020 & 0.163 & 0.242 & 0.069 & 0.260 & 0.217 & 0.077 & 0.150\\
\bottomrule
\end{tabular}
\end{adjustbox}
\end{table}

\begin{table}[htb!]
\setlength{\tabcolsep}{5pt}
\fontsize{9}{9}\selectfont
\centering
\caption{
Ablation studies of latent dimension ($d=300$) on rMD17.
The evaluation is the mean absolute error.
No data normalization is used.
}
\label{tab:ablation_study_latent_dim_300_rMD17}
\vspace{-1.5ex}
\begin{adjustbox}{max width=\textwidth}
\begin{tabular}{ll c c c c c c c c c c}
\toprule
Model & Energy/Force & Aspirin $\downarrow$ & Azobenzene $\downarrow$ & Benzene $\downarrow$ & Ethanol $\downarrow$ & Malonaldehyde $\downarrow$ & Naphthalene $\downarrow$ & Paracetamol $\downarrow$ & Salicylic $\downarrow$ & Toluene $\downarrow$ & Uracil $\downarrow$ \\
\midrule

\multirow{2}{*}{SchNet}
% 5e-4_CosineAnnealingLR_300_5_1000_energy_force_no_normalization
& Energy & 0.534 & 1.818 & 0.111 & 1.757 & 0.260 & 0.124 & 8.138 & 2.618 & 0.119 & 7.029\\
& Force & 1.243 & 3.596 & 0.233 & 0.449 & 0.862 & 0.587 & 2.320 & 0.878 & 0.574 & 0.762\\
\midrule

\multirow{2}{*}{DimeNet++}
% 5e-4_CosineAnnealingLR_300_800_energy_force_no_normalization
& Energy & 2.438 & 3.955 & 0.741 & 1.456 & 2.317 & 1.648 & 2.261 & 1.555 & 1.210 & 2.320\\
& Force & 2.009 & 1.243 & 0.340 & 1.213 & 7.029 & 0.629 & 1.047 & 0.934 & 0.921 & 3.181\\
\midrule

\multirow{2}{*}{EGNN}
% 1e-4_CosineAnnealingLR_300_5_1_1000_energy_force_no_normalization
& Energy & 17.350 & 21.333 & 0.315 & 0.402 & 0.534 & 12.164 & 26.902 & 7.794 & 15.021 & 1.669\\
& Force & 3.825 & 2.330 & 0.529 & 0.989 & 1.334 & 1.183 & 2.313 & 1.571 & 1.165 & 1.323\\
\midrule

\multirow{2}{*}{PaiNN}
% 1e-4_CosineAnnealingLR_300_5_1000_energy_force_no_normalization
& Energy & 30.156 & 0.107 & 0.010 & 1.170 & 0.070 & 5.297 & 0.117 & 5.219 & 0.045 & 2.478\\
& Force & 0.573 & 0.326 & 0.032 & 0.316 & 0.377 & 0.161 & 0.440 & 0.321 & 0.231 & 0.235\\
\midrule

\multirow{2}{*}{GemNet-T}
% 5e-4_CosineAnnealingLR_300_5_1000_energy_force_no_normalization
& Energy & 5.389 & 7.770 & 0.007 & 1.615 & 9.496 & 0.031 & 2.173 & 21.411 & 959.745 & 994.036\\
& Force & 0.555 & 0.347 & 0.033 & 0.233 & 0.337 & 0.154 & 0.388 & 0.371 & 0.400 & 1.165\\
\midrule

\multirow{2}{*}{SphereNet}
% 1e-4_CosineAnnealingLR_300_1_1000_energy_force_no_normalization
& Energy & 0.304 & 0.257 & 0.052 & 0.072 & 0.138 & 0.093 & 0.183 & 0.771 & 20.479 & 12.211\\
& Force & 0.622 & 0.532 & 0.076 & 0.217 & 0.500 & 0.279 & 0.482 & 2.088 & 0.254 & 0.959\\
\midrule

\multirow{2}{*}{SEGNN}
% 1e-4_CosineAnnealingLR_300_5_5_0_700_energy_force_no_normalization
& Energy & 15.721 & 3.474 & 0.270 & 0.130 & 0.182 & 1.110 & 4.197 & 1.494 & 0.814 & 1.115\\
& Force & 8.549 & 2.579 & 0.174 & 0.846 & 1.185 & 0.926 & 3.191 & 2.056 & 1.241 & 0.966\\
\midrule

\multirow{2}{*}{Allegro}
% 1e-3_CosineAnnealingLR_4.0_300_1000_energy_force_no_normalization
& Energy & 1.339 & 2.441 & 0.049 & 0.339 & 0.651 & 3.781 & 0.978 & 1.356 & 0.451 & 2.497\\
& Force & 3.861 & 4.609 & 0.467 & 1.579 & 1.816 & 3.428 & 3.693 & 5.086 & 2.241 & 5.183\\
\midrule

\multirow{2}{*}{Equiformer}
% 1e-4_CosineAnnealingLR_300_5_1_1000_energy_force_no_normalization
& Energy & 0.375 & 0.127 & 0.027 & 0.064 & 0.085 & 0.069 & 0.215 & 0.143 & 0.104 & 0.200\\
& Force & 0.305 & 0.132 & 0.020 & 0.162 & 0.240 & 0.070 & 0.258 & 0.218 & 0.077 & 0.149\\
\bottomrule
\end{tabular}
\end{adjustbox}
\end{table}

\begin{figure}[htb!]
\centering
    \begin{subfigure}[\small Task Aspirin]
    {\includegraphics[width=0.24\linewidth]{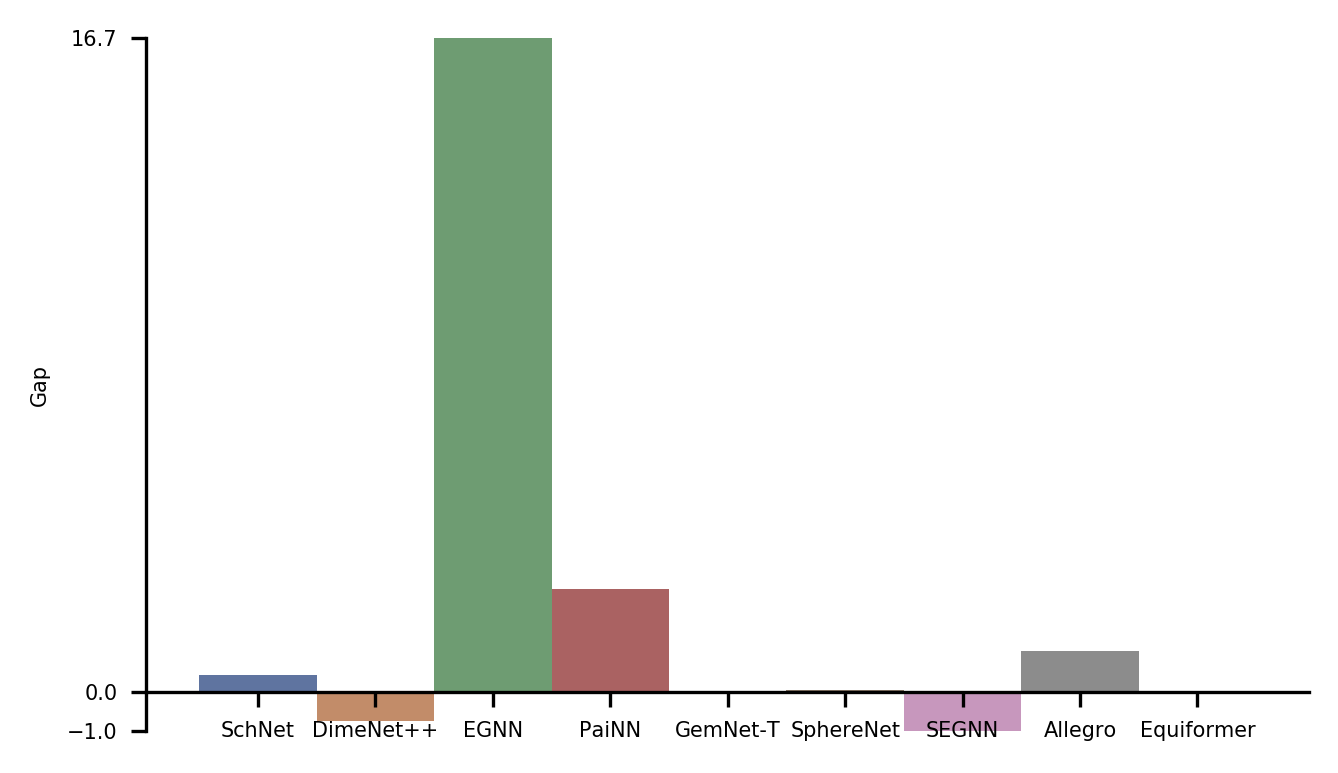}}
    \end{subfigure}
% \hfill
    \begin{subfigure}[\small Task Azobenzene]
    {\includegraphics[width=0.24\linewidth]{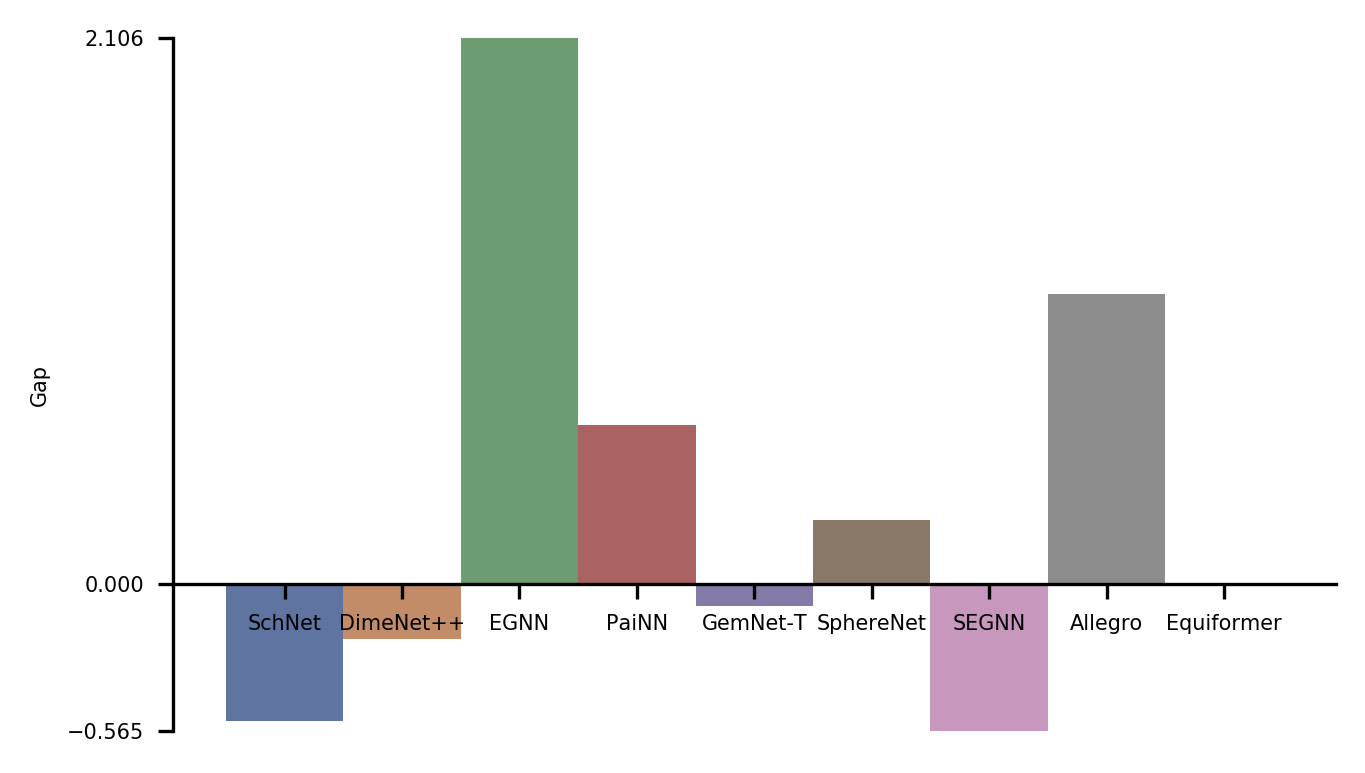}}
    \end{subfigure}
% \hfill
    \begin{subfigure}[\small Task Benzene]
    {\includegraphics[width=0.24\linewidth]{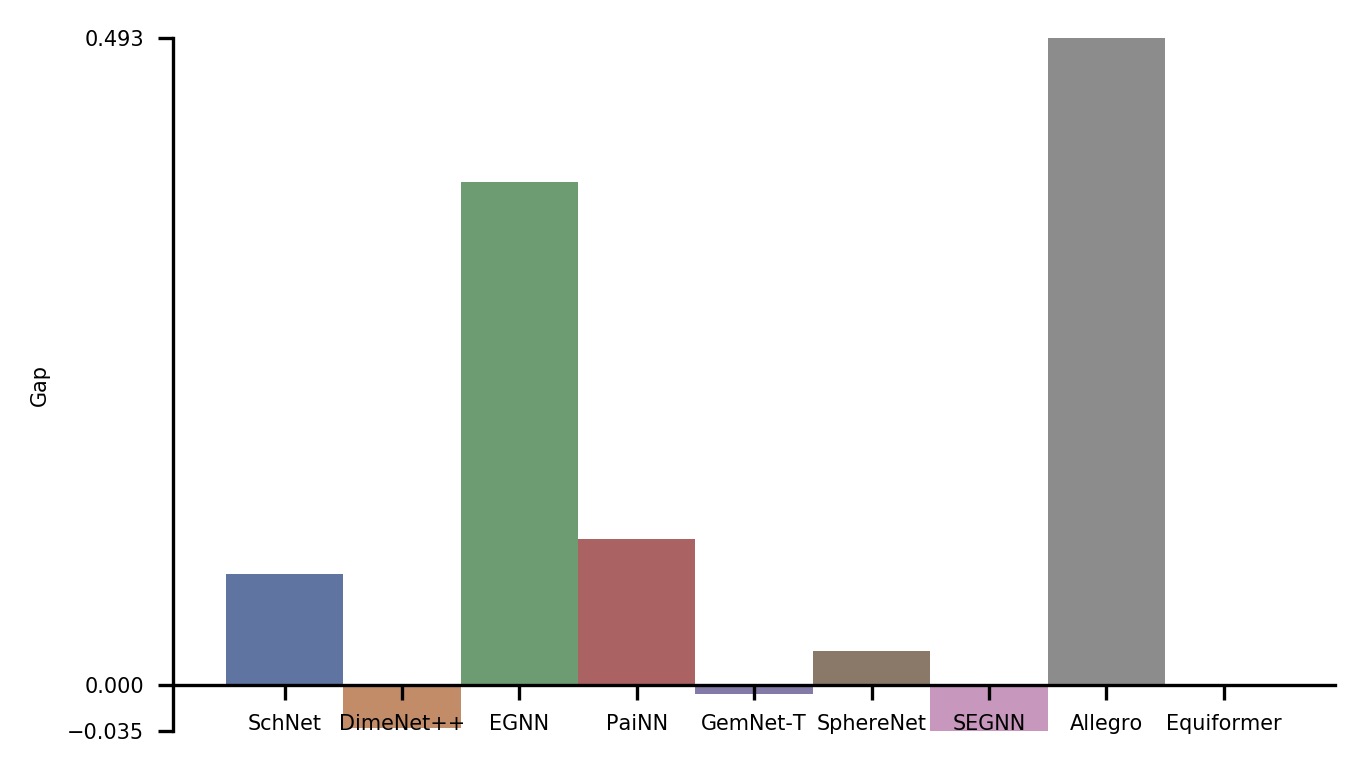}}
    \end{subfigure}
% \hfill
    \begin{subfigure}[\small Task Ethanol]
    {\includegraphics[width=0.24\linewidth]{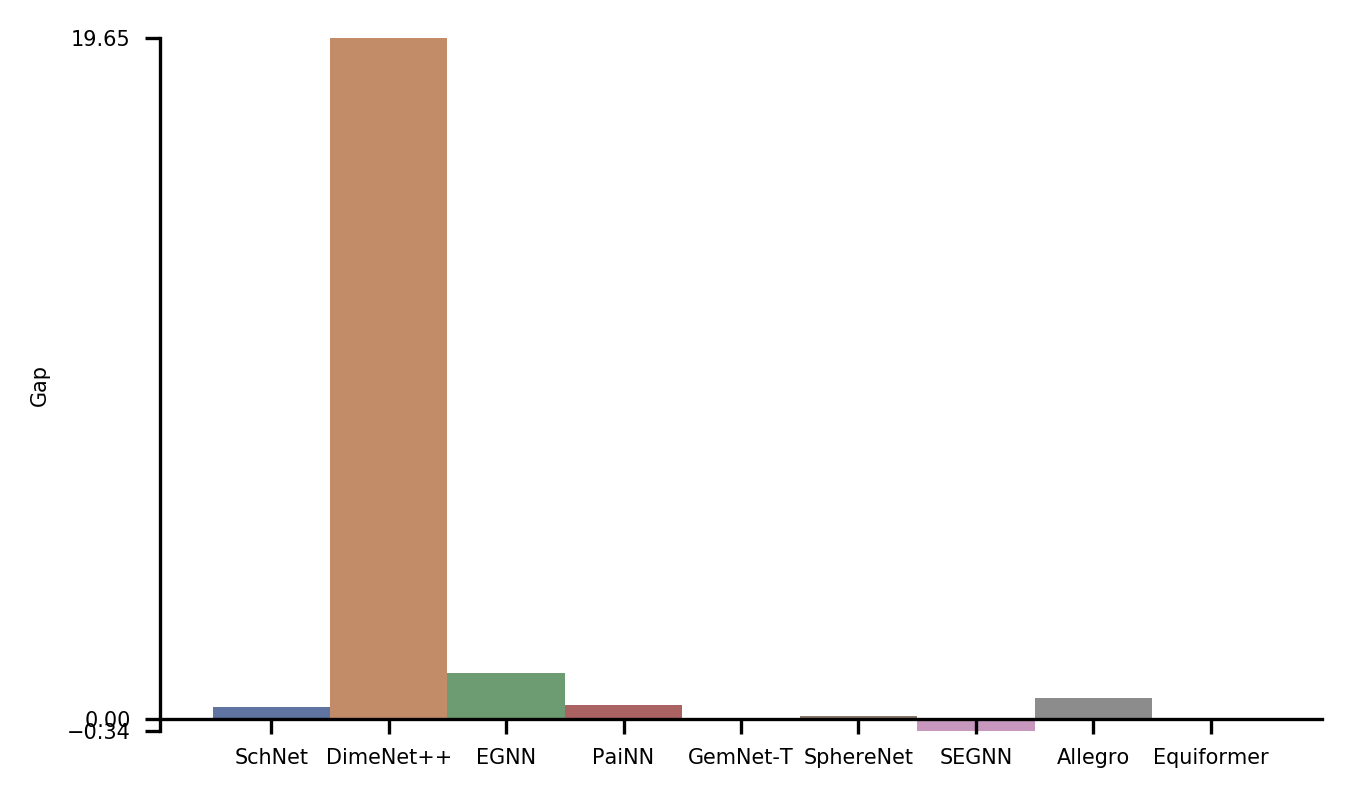}}
    \end{subfigure}
% \hfill
    \begin{subfigure}[\small Task Malonaldehyde]
    {\includegraphics[width=0.24\linewidth]{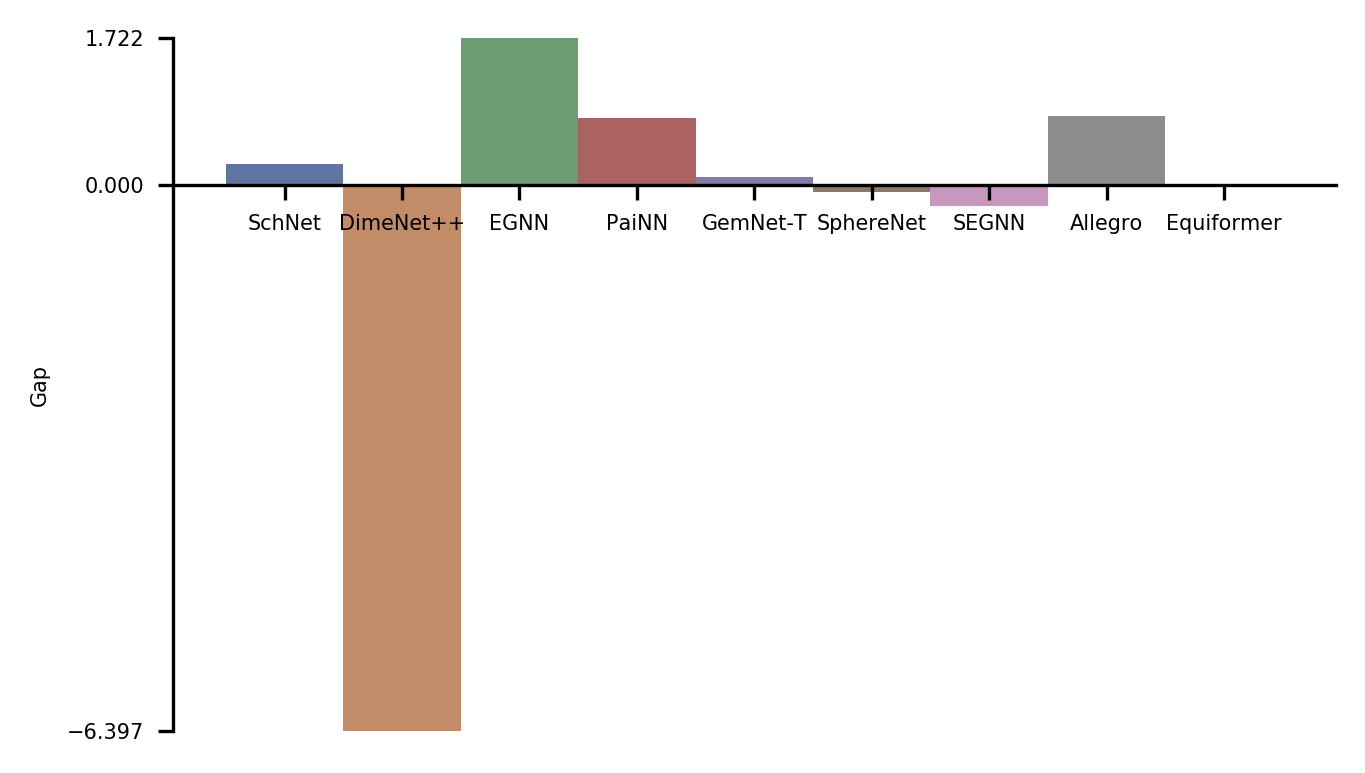}}
    \end{subfigure}
% \hfill
    \begin{subfigure}[\small Task Naphthalene]
    {\includegraphics[width=0.24\linewidth]{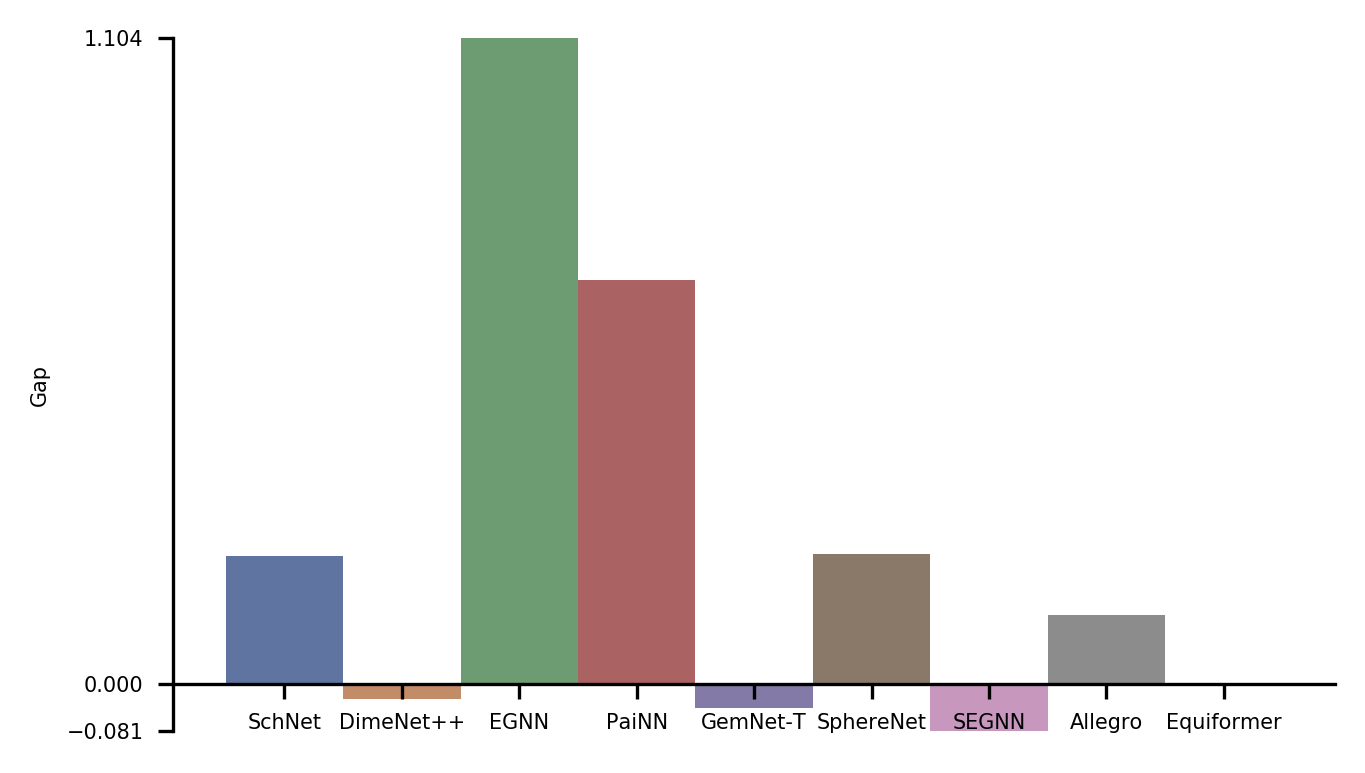}}
    \end{subfigure}
% \hfill
    \begin{subfigure}[\small Task Paracetamol]
    {\includegraphics[width=0.24\linewidth]{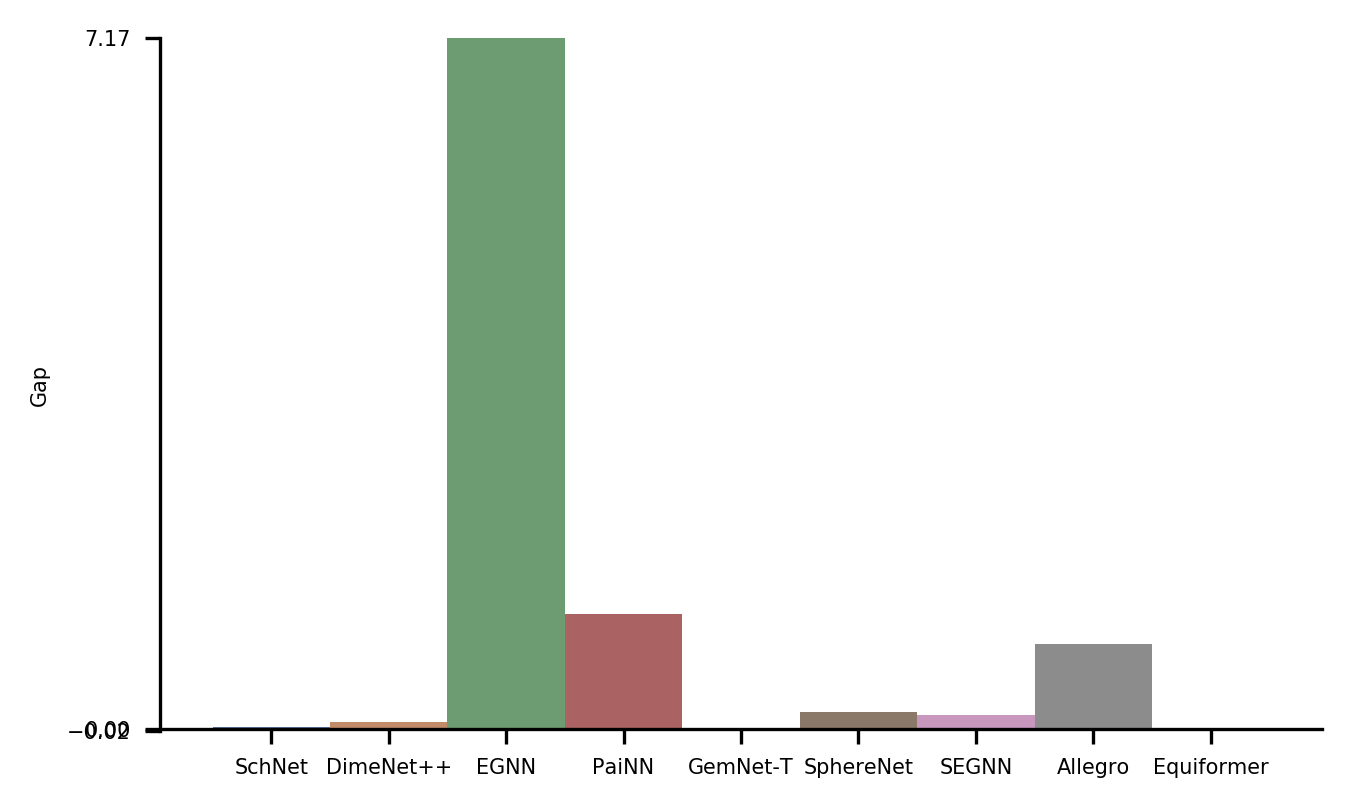}}
    \end{subfigure}
% \hfill
    \begin{subfigure}[\small Task Salicylic]
    {\includegraphics[width=0.24\linewidth]{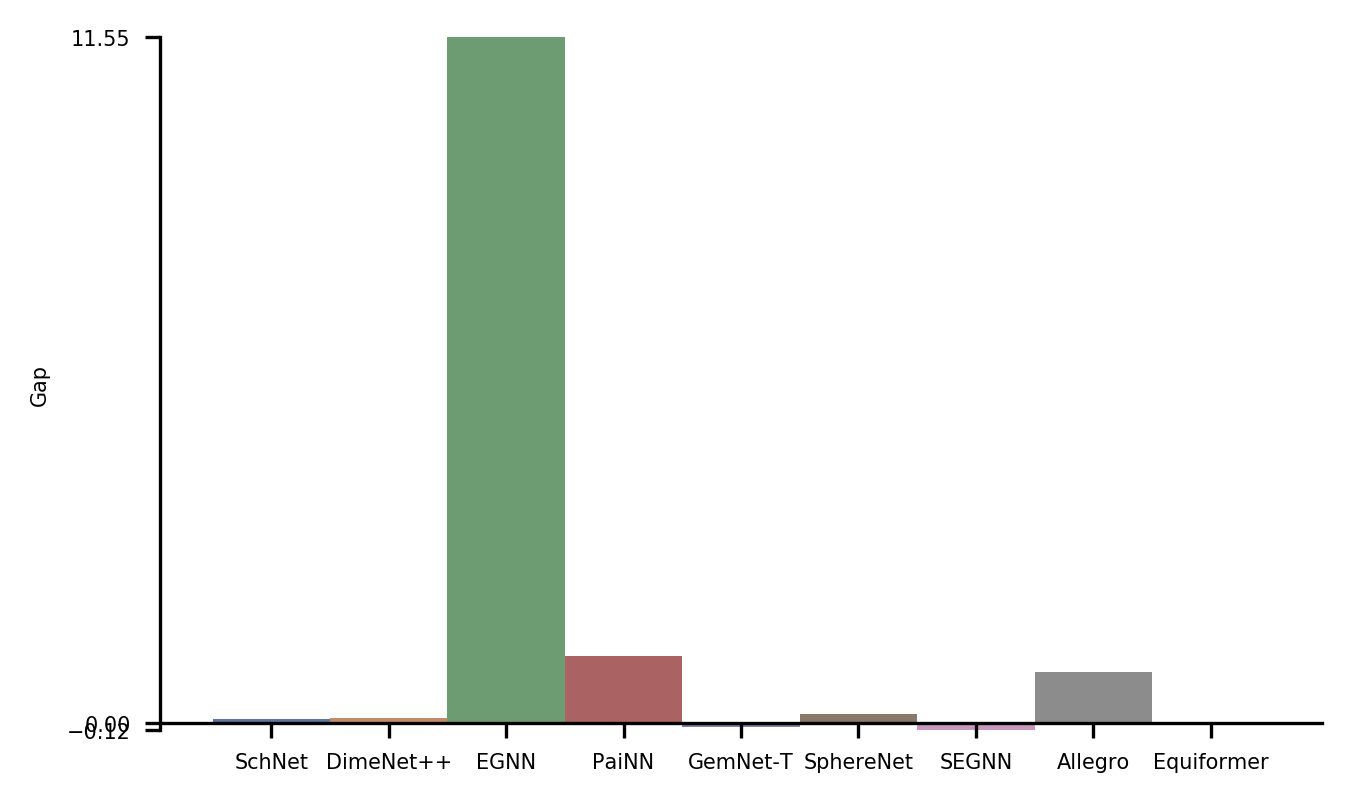}}
    \end{subfigure}
% \hfill
    \begin{subfigure}[\small Task Toluene]
    {\includegraphics[width=0.24\linewidth]{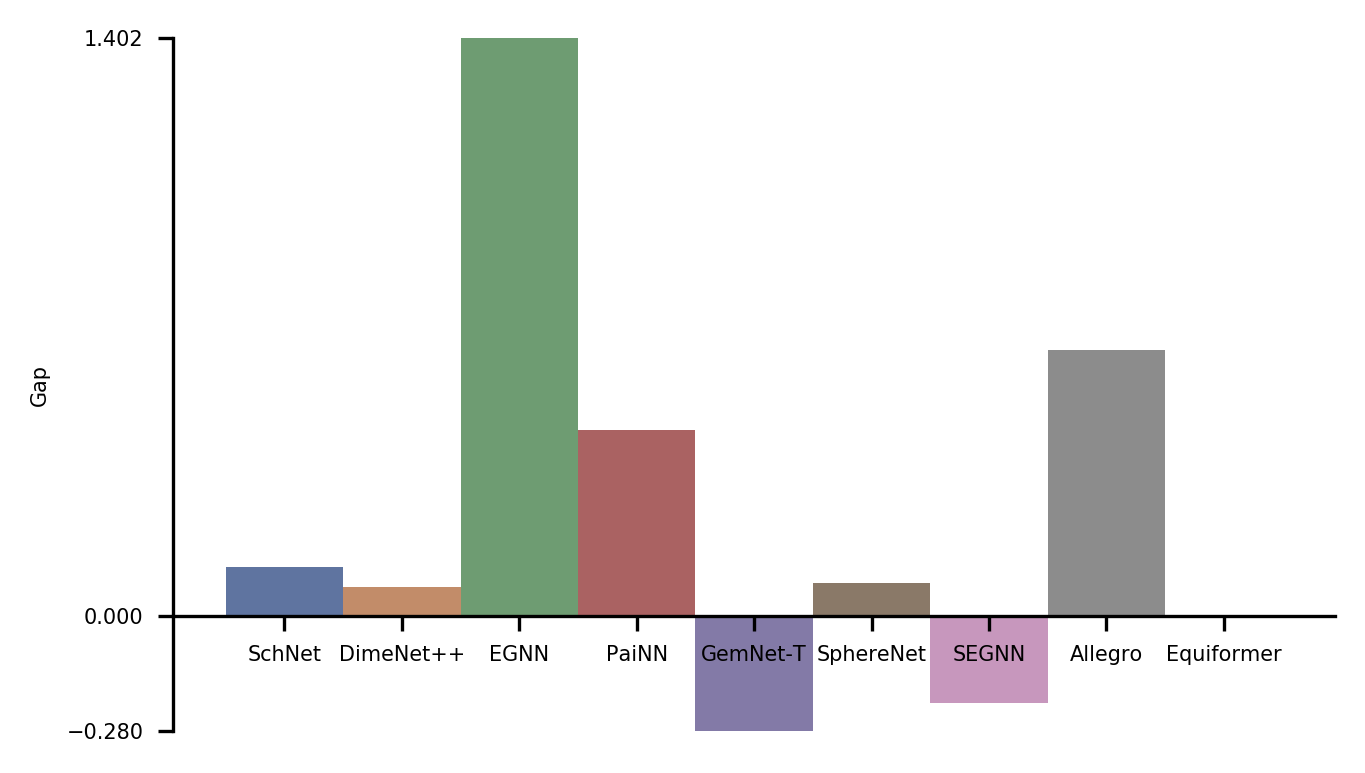}}
    \end{subfigure}
% \hfill
    \begin{subfigure}[\small Task Uracil]
    {\includegraphics[width=0.24\linewidth]{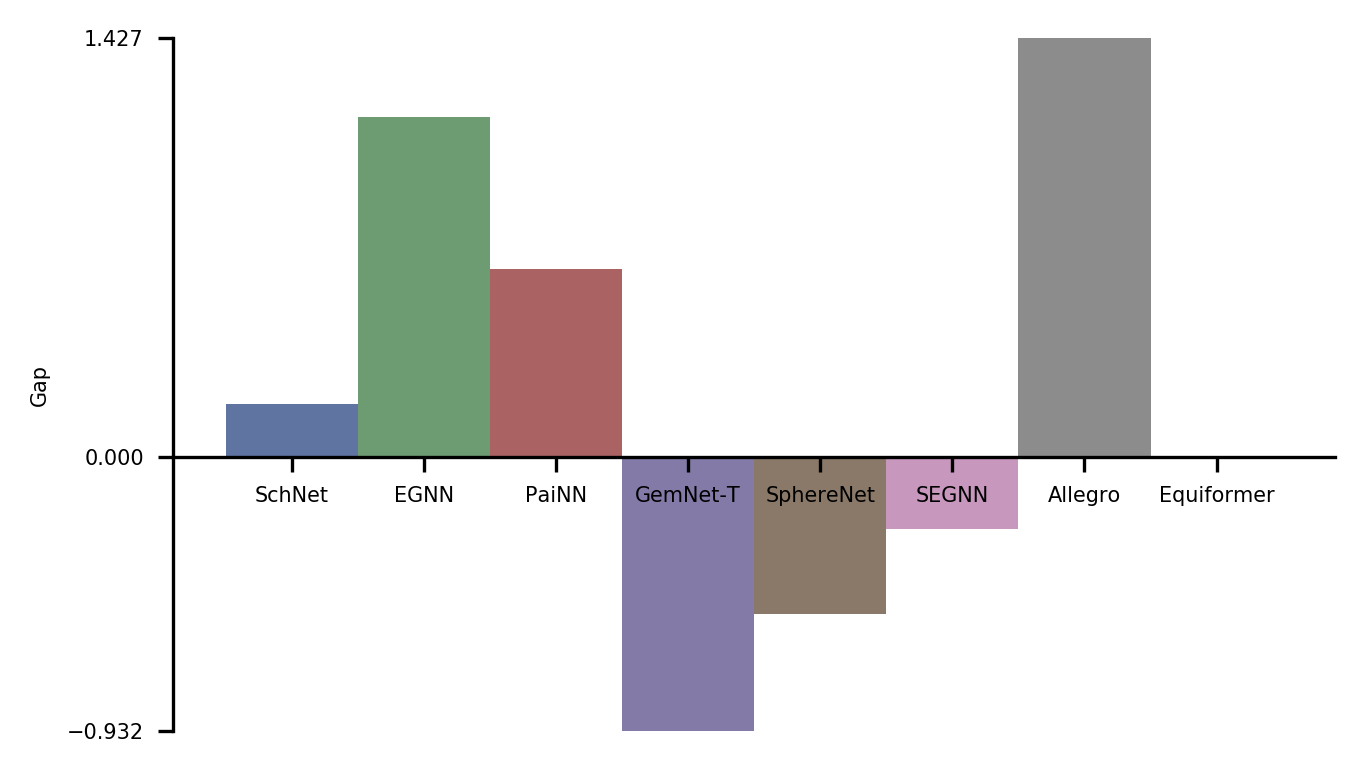}}
    \end{subfigure}
\vspace{-2ex}
\caption{
\small
Performance gap of MAE($d=128$) - MAE($d=300$) in rMD17.
}
\label{fig:ablation_study_latent_dim_rMD17}
\vspace{-2.5ex}
\end{figure}

\clearpage
\begin{table}[htb!]
\setlength{\tabcolsep}{5pt}
\fontsize{9}{9}\selectfont
\centering
\caption{
Ablation studies of latent dimension on COLL.
120k for training, 10k for val, 9.48k for test.
The evaluation metric is the mean absolute error (MAE).
}
\label{tab:ablation_study_latent_dim_COLL}
\vspace{-1.5ex}
\begin{adjustbox}{max width=\textwidth}
\begin{tabular}{l l c c}
\toprule
Model & $d$ & energy $\downarrow$ & force $\downarrow$\\
\midrule

\multirow{2}{*}{SchNet}
% 5e-4_CosineAnnealingLR_128_1000
& 128 & 0.171 & 0.135\\
% 5e-4_CosineAnnealingLR_300_1000
& 300 & 0.178 & 0.130\\
\midrule

\multirow{2}{*}{DimeNet++ }
% 5e-4_CosineAnnealingLR_128_1000
& 128 & 0.049 & 0.058\\
% 5e-4_CosineAnnealingLR_300_1000
& 300 & 0.036 & 0.049\\
\midrule

\multirow{2}{*}{EGNN}
% 5e-4_CosineAnnealingLR_128_1000
& 128 & 0.786 & 0.151\\
% 5e-4_CosineAnnealingLR_300_1000
& 300 & 1.808 & 0.234\\
\midrule

\multirow{2}{*}{PaiNN}
% 5e-4_CosineAnnealingLR_128_1000
& 128 & 0.047 & 0.066\\
% 5e-4_CosineAnnealingLR_300_1000
& 300 & 0.030 & 0.052\\
\midrule

\multirow{2}{*}{GemNet-T}
% 5e-4_CosineAnnealingLR_128_500
& 128 & 0.022 & 0.035\\
% 5e-4_CosineAnnealingLR_300_500
& 300 & 0.017 & 0.028\\
\midrule

\multirow{2}{*}{SphereNet}
% % 5e-4_CosineAnnealingLR_128_50_2_300
% SphereNet & 0.040 & 0.056\\
% 5e-4_CosineAnnealingLR_128_50_6_300
& 128 & 0.039 & 0.049\\
% % 5e-4_CosineAnnealingLR_300_50_2_300
% SphereNet & 0.075 & 0.072\\
% 1e-4_CosineAnnealingLR_300_50_6_300
& 300 & 0.032 & 0.047\\
\midrule

\multirow{2}{*}{SEGNN}
% 1e-4_CosineAnnealingLR_2_128_100
& 128 & 7.054 & 0.511\\
% 1e-4_CosineAnnealingLR_2_300_100
& 300 & 7.085 & 0.642\\
\midrule

\multirow{2}{*}{Equiformer}
% 5e-4_CosineAnnealingLR_5_128_200
& 128 & 0.034 & 0.030\\
% 5e-4_CosineAnnealingLR_5_300_200
& 300 & 0.036 & 0.030\\
\bottomrule
\end{tabular}
\end{adjustbox}
\end{table}

\begin{table}[htb!]
\setlength{\tabcolsep}{20pt}
\fontsize{9}{10}\selectfont
\caption{
Ablation studies of latent dimension ($d=128$) on 2 binding affinity prediction tasks.
We select three evaluation metrics for LBA: the root mean squared error (RMSD), the Pearson correlation ($R_p$) and the Spearman correlation ($R_S$). LEP is a binary classification task, and we use the area under the curve for receiver operating characteristics (ROC) and precision-recall (PR) for evaluation. We run cross validation with 5 seeds and report the mean and std.
}
\label{tab:ablation_study_latent_dim_LBA_LEP_128}
\centering
\vspace{-1.5ex}
\begin{adjustbox}{max width=\textwidth}
\begin{tabular}{l c c c c c}
\toprule
\multirow{2}{*}{\makecell{Model}} & \multicolumn{3}{c}{LBA} & \multicolumn{2}{c}{LEP}\\
\cmidrule(lr){2-4}
\cmidrule(lr){5-6}
 & RMSD $\downarrow$ & $R_P$ $\uparrow$ & $R_C$ $\uparrow$ & ROC $\uparrow$ & PR $\uparrow$\\
\midrule
% LBA 1e-4_CosineAnnealingLR_128_64_300
SchNet & 1.509 $\pm$ 0.05 & 0.510 $\pm$ 0.02 & 0.487 $\pm$ 0.01
% LEP 1e-4_CosineAnnealingLR_128_16_300
 & 0.444 $\pm$ 0.03 & 0.391 $\pm$ 0.02\\
 
% LBA 1e-4_CosineAnnealingLR_128_32_300
DimeNet++ & 1.808 $\pm$ 0.46 & 0.557 $\pm$ 0.01 & 0.566 $\pm$ 0.01
% LEP 1e-4_CosineAnnealingLR_128_4_300
 & 0.582 $\pm$ 0.06 & 0.494 $\pm$ 0.03\\

% LBA 1e-4_CosineAnnealingLR_128_128_300
EGNN & 1.531 $\pm$ 0.02 & 0.452 $\pm$ 0.01 & 0.419 $\pm$ 0.01
% LEP 1e-4_CosineAnnealingLR_128_16_300
 & 0.702 $\pm$ 0.05 & 0.603 $\pm$ 0.07\\

% LBA 1e-3_CosineAnnealingLR_128_128_300
PaiNN & 1.460 $\pm$ 0.03 & 0.569 $\pm$ 0.01 & 0.564 $\pm$ 0.01
% LEP 1e-4_CosineAnnealingLR_128_16_300
 & 0.627 $\pm$ 0.07 & 0.499 $\pm$ 0.09\\

% LBA 1e-4_CosineAnnealingLR_128_16_300
GemNet & 130.621 $\pm$ 13.90 & -0.114 $\pm$ 0.54 & -0.116 $\pm$ 0.55
% LEP 1e-4_CosineAnnealingLR_128_4_300
 & 0.623 $\pm$ 0.05 & 0.552 $\pm$ 0.05\\

% LBA 1e-4_CosineAnnealingLR_128_3_128_300
SphereNet & 1.605 $\pm$ 0.02 & 0.533 $\pm$ 0.00 & 0.527 $\pm$ 0.00
% LEP 1e-4_CosineAnnealingLR_128_3_16_300
 & 0.556 $\pm$ 0.05 & 0.471 $\pm$ 0.05\\

% LBA 1e-4_CosineAnnealingLR_128_5_16_300
SEGNN & 1.422 $\pm$ 0.04 & 0.560 $\pm$ 0.02 & 0.537 $\pm$ 0.03
% LEP 1e-4_CosineAnnealingLR_128_5_2_300
 & 0.582 $\pm$ 0.08 & 0.517 $\pm$ 0.09\\
\bottomrule
\end{tabular}
\end{adjustbox}
\end{table}

\begin{table}[htb!]
\setlength{\tabcolsep}{20pt}
\fontsize{9}{10}\selectfont
\caption{
Ablation studies of latent dimension ($d=300$) on 2 binding affinity prediction tasks.
We select three evaluation metrics for LBA: the root mean squared error (RMSD), the Pearson correlation ($R_p$) and the Spearman correlation ($R_S$). LEP is a binary classification task, and we use the area under the curve for receiver operating characteristics (ROC) and precision-recall (PR) for evaluation. We run cross validation with 5 seeds and report the mean and std.
}
\label{tab:ablation_study_latent_dim_LBA_LEP_300}
\centering
\vspace{-1.5ex}
\begin{adjustbox}{max width=\textwidth}
\begin{tabular}{l c c c c c}
\toprule
\multirow{2}{*}{\makecell{Model}} & \multicolumn{3}{c}{LBA} & \multicolumn{2}{c}{LEP}\\
\cmidrule(lr){2-4}
\cmidrule(lr){5-6}
 & RMSD $\downarrow$ & $R_P$ $\uparrow$ & $R_C$ $\uparrow$ & ROC $\uparrow$ & PR $\uparrow$\\
\midrule
% LBA 1e-4_CosineAnnealingLR_300_64_300
SchNet & 1.521 $\pm$ 0.02 & 0.474 $\pm$ 0.01 & 0.452 $\pm$ 0.01
% LEP 1e-4_CosineAnnealingLR_300_16_300
 & 0.450 $\pm$ 0.03 & 0.379 $\pm$ 0.03\\

% LBA 1e-4_CosineAnnealingLR_300_32_300
DimeNet++ & 1.672 $\pm$ 0.09 & 0.550 $\pm$ 0.01 & 0.556 $\pm$ 0.01
% LEP 1e-4_CosineAnnealingLR_300_4_300
 & 0.590 $\pm$ 0.06 & 0.496 $\pm$ 0.05\\

% LBA 1e-4_CosineAnnealingLR_300_128_300
EGNN & 1.494 $\pm$ 0.04 & 0.503 $\pm$ 0.04 & 0.483 $\pm$ 0.05
% LEP 1e-4_CosineAnnealingLR_300_16_300
 & 0.657 $\pm$ 0.05 & 0.559 $\pm$ 0.05\\
 
% LBA 1e-3_CosineAnnealingLR_300_128_300
PaiNN & 1.434 $\pm$ 0.02 & 0.583 $\pm$ 0.02 & 0.580 $\pm$ 0.02
% LEP 1e-4_CosineAnnealingLR_300_16_300
 & 0.585 $\pm$ 0.02 & 0.432 $\pm$ 0.03\\

% LBA 1e-4_CosineAnnealingLR_300_16_300
GemNet & 269.427 $\pm$ 148.62 & 0.029 $\pm$ 0.50 & 0.036 $\pm$ 0.51
% LEP 1e-4_CosineAnnealingLR_128_8_300
 & 0.674 $\pm$ 0.04 & 0.565 $\pm$ 0.05\\

% LBA 1e-4_CosineAnnealingLR_300_3_128_300
SphereNet & 1.581 $\pm$ 0.02 & 0.538 $\pm$ 0.01 & 0.529 $\pm$ 0.01
% LEP 1e-4_CosineAnnealingLR_300_3_8_300
 & 0.523 $\pm$ 0.04 & 0.432 $\pm$ 0.05\\

% LBA 1e-4_CosineAnnealingLR_300_5_16_300
SEGNN & 1.416 $\pm$ 0.03 & 0.566 $\pm$ 0.02 & 0.550 $\pm$ 0.02
% LEP 1e-4_CosineAnnealingLR_300_5_2_300
 & 0.574 $\pm$ 0.03 & 0.485 $\pm$ 0.03\\
\bottomrule
\end{tabular}
\end{adjustbox}
\end{table}

%%%%%%%%%%%%%%%%%%%%%%%%%%%%%%%%%%%%%%%%%%%%%%%%%%
\subsection{Ablation Study on Data Normalization for Molecular Dynamics Prediction} \label{sec:app:ablation_study_data_normalization}
Allegro~\cite{musaelian2022learning} and NequIP~\cite{batzner20223} introduce a normalization strategy for molecular dynamics (energy and force) prediction on MD17 and rMD17 datasets:
\begin{equation}
\small
\hat y_E = y_E * \text{Force Mean} + \text{Energy Mean} * \text{\# Atom},
\end{equation}
where $y_E$ is the original predicted energy, and $\hat y_E$ is the normalized prediction.
We find this trick important and would like to systematically test it here.
Notice that as shown in~\Cref{sec:app:ablation_latent_dimension}, the latent dimension is an important factor, and here we would like to conduct the ablation studies on both factors.\looseness=-1
\begin{itemize}[noitemsep,topsep=0pt]
    \item MD17 w/o normalization and $d=128$ in~\Cref{tab:ablation_study_latent_dim_128_MD17}, $d=300$ in~\Cref{tab:ablation_study_latent_dim_300_MD17}. rMD17 w/o normalization and $d=128$ in~\Cref{tab:ablation_study_latent_dim_128_rMD17}, $d=300$ in~\Cref{tab:ablation_study_latent_dim_300_rMD17}.
    \item In the following tables, we test:
    MD17 w/ normalization and $d=128$ in~\Cref{tab:ablation_study_latent_dim_128_MD17_with_normalization}, $d=300$ in~\Cref{tab:ablation_study_latent_dim_300_MD17_with_normalization}. rMD17 w/ normalization and $d=128$ in~\Cref{tab:ablation_study_latent_dim_128_rMD17_with_normalization}, $d=300$ in~\Cref{tab:ablation_study_latent_dim_300_rMD17_with_normalization}.
\end{itemize}

\begin{table}[htb!]
\setlength{\tabcolsep}{5pt}
\fontsize{9}{9}\selectfont
\centering
\caption{
\small
Ablation studies of latent dimension ($d=128$) on MD17.
The evaluation is the mean absolute error.
Data normalization is used.
}
\label{tab:ablation_study_latent_dim_128_MD17_with_normalization}
\vspace{-1.5ex}
\begin{adjustbox}{max width=\textwidth}
\begin{tabular}{ll c c c c c c c c}
\toprule
Model & Energy / Force & Aspirin $\downarrow$ & Benzene $\downarrow$ & Ethanol $\downarrow$ & Malonaldehyde $\downarrow$ & Naphthalene $\downarrow$ & Salicylic $\downarrow$ & Toluene $\downarrow$ & Uracil $\downarrow$ \\
\midrule

\multirow{2}{*}{SchNet}
% 5e-4_CosineAnnealingLR_128_5_1000_energy_force_with_normalization
& Energy & 0.588 & 0.099 & 0.072 & 0.111 & 0.125 & 0.207 & 0.110 & 0.118\\
& Force & 1.008 & 0.200 & 0.297 & 0.491 & 0.299 & 0.547 & 0.346 & 0.383\\
\midrule

\multirow{2}{*}{DimeNet++}
% 5e-4_CosineAnnealingLR_128_800_energy_force_with_normalization
& Energy & 0.370 & 0.154 & 28.604 & 57144.066 & 0.289 & 15.497 & 0.206 & 0.317\\
& Force & 0.578 & 0.110 & 89.512 & 2119653.000 & 0.930 & 90.846 & 0.540 & 0.535\\
\midrule

\multirow{2}{*}{EGNN}
% 1e-4_CosineAnnealingLR_128_5_1_1000_energy_force_with_normalization
& Energy & 0.668 & 0.144 & 0.470 & 0.238 & 0.481 & 0.462 & 0.234 & 0.429\\
& Force & 1.249 & 0.461 & 1.042 & 0.827 & 0.913 & 0.927 & 0.631 & 1.227\\
\midrule

\multirow{2}{*}{PaiNN}
% 5e-4_CosineAnnealingLR_128_5_1000_energy_force_with_normalization
& Energy & 0.146 & 0.095 & 0.057 & 0.083 & 0.113 & 0.110 & 0.095 & 0.104\\
& Force & 0.315 & 0.034 & 0.157 & 0.244 & 0.074 & 0.177 & 0.093 & 0.120\\
\midrule

\multirow{2}{*}{GemNet-T}
% 5e-4_CosineAnnealingLR_128_5_1000_energy_force_with_normalization
& Energy & 0.175 & 0.097 & 0.055 & 0.080 & 0.130 & 0.112 & 0.093 & 0.105\\
& Force & 0.284 & 0.042 & 0.141 & 0.191 & 0.082 & 0.167 & 0.080 & 0.120\\
\midrule

\multirow{2}{*}{SphereNet}
% 5e-4_CosineAnnealingLR_128_2_1000_energy_force_with_normalization
& Energy & 0.168 & 0.095 & 0.061 & 0.110 & 0.115 & 0.120 & 0.095 & 0.113\\
& Force & 0.305 & 0.042 & 0.173 & 0.280 & 0.083 & 0.219 & 0.088 & 0.189\\
\midrule

\multirow{2}{*}{SEGNN}
% 1e-4_CosineAnnealingLR_128_5_5_0_800_energy_force_with_normalization
& Energy & 0.337 & 0.069 & 0.060 & 0.092 & 0.101 & 0.151 & 0.092 & 0.104\\
& Force & 0.879 & 0.077 & 0.236 & 0.365 & 0.251 & 0.564 & 0.307 & 0.281\\
\midrule

\multirow{2}{*}{Allegro}
% 1e-3_CosineAnnealingLR_4.0_128_1000_energy_force_with_normalization
& Energy & 0.290 & 0.096 & 0.064 & 0.105 & 0.143 & 0.151 & 0.123 & 0.112\\
& Force & 0.646 & 0.073 & 0.228 & 0.346 & 0.285 & 0.407 & 0.265 & 0.245\\
\midrule

\multirow{2}{*}{Equiformer}
% 1e-4_CosineAnnealingLR_128_5_1_1000_energy_force_with_normalization
& Energy & 0.140 & 0.072 & 0.056 & 0.085 & 0.090 & 0.112 & 0.078 & 0.101\\
& Force & 0.315 & 0.057 & 0.159 & 0.250 & 0.069 & 0.204 & 0.083 & 0.156\\
\bottomrule
\end{tabular}
\end{adjustbox}
\end{table}

\begin{table}[htb!]
\setlength{\tabcolsep}{5pt}
\fontsize{9}{9}\selectfont
\centering
\caption{
Ablation studies of latent dimension ($d=300$) on MD17.
The evaluation is the mean absolute error.
Data normalization is used.
}
\label{tab:ablation_study_latent_dim_300_MD17_with_normalization}
\vspace{-1.5ex}
\begin{adjustbox}{max width=\textwidth}
\begin{tabular}{ll c c c c c c c c}
\toprule
Model & Energy / Force & Aspirin $\downarrow$ & Benzene $\downarrow$ & Ethanol $\downarrow$ & Malonaldehyde $\downarrow$ & Naphthalene $\downarrow$ & Salicylic $\downarrow$ & Toluene $\downarrow$ & Uracil $\downarrow$ \\
\midrule

\multirow{2}{*}{SchNet}
% 5e-4_CosineAnnealingLR_300_5_1000_energy_force_with_normalization
& Energy & 0.321 & 0.099 & 0.074 & 0.125 & 0.129 & 0.155 & 0.130 & 0.171\\
& Force & 1.055 & 0.191 & 0.318 & 0.522 & 0.328 & 0.597 & 0.387 & 0.401\\
\midrule

\multirow{2}{*}{DimeNet++}
% 5e-4_CosineAnnealingLR_300_800_energy_force_with_normalization
& Energy & 0.628 & 56451512.000 & 0.192 & 0.480 & 14056.564 & 0.421 & 27644.078 & 7522.200\\
& Force & 2.632 & 688219840.000 & 1.029 & 1.703 & 173932.344 & 0.621 & 972773.375 & 16002.980\\
\midrule

\multirow{2}{*}{EGNN}
% 1e-4_CosineAnnealingLR_300_5_1_1000_energy_force_with_normalization
& Energy & 0.393 & 0.125 & 0.072 & 0.112 & 0.249 & 0.257 & 0.158 & 0.164\\
& Force & 0.695 & 0.442 & 0.269 & 0.415 & 0.439 & 0.641 & 0.447 & 0.536\\
\midrule

\multirow{2}{*}{PaiNN}
% 5e-4_CosineAnnealingLR_300_5_1000_energy_force_with_normalization
& Energy & 0.149 & 0.102 & 0.056 & 0.083 & 0.118 & 0.113 & 0.093 & 0.104\\
& Force & 0.331 & 0.037 & 0.163 & 0.252 & 0.082 & 0.187 & 0.097 & 0.122\\
\midrule

\multirow{2}{*}{GemNet-T}
% 5e-4_CosineAnnealingLR_300_5_1000_energy_force_with_normalization
& Energy & 0.162 & 0.142 & 0.068 & 0.089 & 0.136 & 0.115 & 0.095 & 0.106\\
& Force & 0.329 & 0.052 & 0.206 & 0.262 & 0.101 & 0.234 & 0.091 & 0.146\\
\midrule

\multirow{2}{*}{SphereNet}
% 1e-4_CosineAnnealingLR_300_1_1000_energy_force_with_normalization
& Energy & 0.212 & 0.096 & 0.081 & 0.101 & 0.116 & 0.145 & 0.099 & 0.120\\
& Force & 0.334 & 0.047 & 0.177 & 0.309 & 0.087 & 0.238 & 0.097 & 0.212\\
\midrule

\multirow{2}{*}{SEGNN}
% 1e-4_CosineAnnealingLR_300_5_5_0_800_energy_force_with_normalization
& Energy & 0.345 & 0.069 & 0.072 & 0.097 & 0.096 & 0.354 & 0.093 & 0.110\\
& Force & 1.023 & 0.080 & 0.331 & 0.452 & 0.227 & 0.803 & 0.314 & 0.327\\
\midrule

\multirow{2}{*}{Allegro}
% 1e-3_CosineAnnealingLR_4.0_300_1000_energy_force_with_normalization
& Energy & 0.256 & 0.096 & 0.060 & 0.088 & 0.131 & 0.139 & 0.114 & 0.110\\
& Force & 0.579 & 0.064 & 0.198 & 0.292 & 0.233 & 0.349 & 0.233 & 0.216\\
\midrule

\multirow{2}{*}{Equiformer}
% 1e-4_CosineAnnealingLR_300_5_1_1000_energy_force_with_normalization
& Energy & 0.143 & 0.073 & 0.061 & 0.085 & 0.090 & 0.107 & 0.077 & 0.100\\
& Force & 0.315 & 0.058 & 0.158 & 0.251 & 0.069 & 0.204 & 0.083 & 0.156\\
\bottomrule
\end{tabular}
\end{adjustbox}
\end{table}

\begin{figure}[htb!]
\centering
    \begin{subfigure}[\small Task Aspirin]
    {\includegraphics[width=0.24\linewidth]{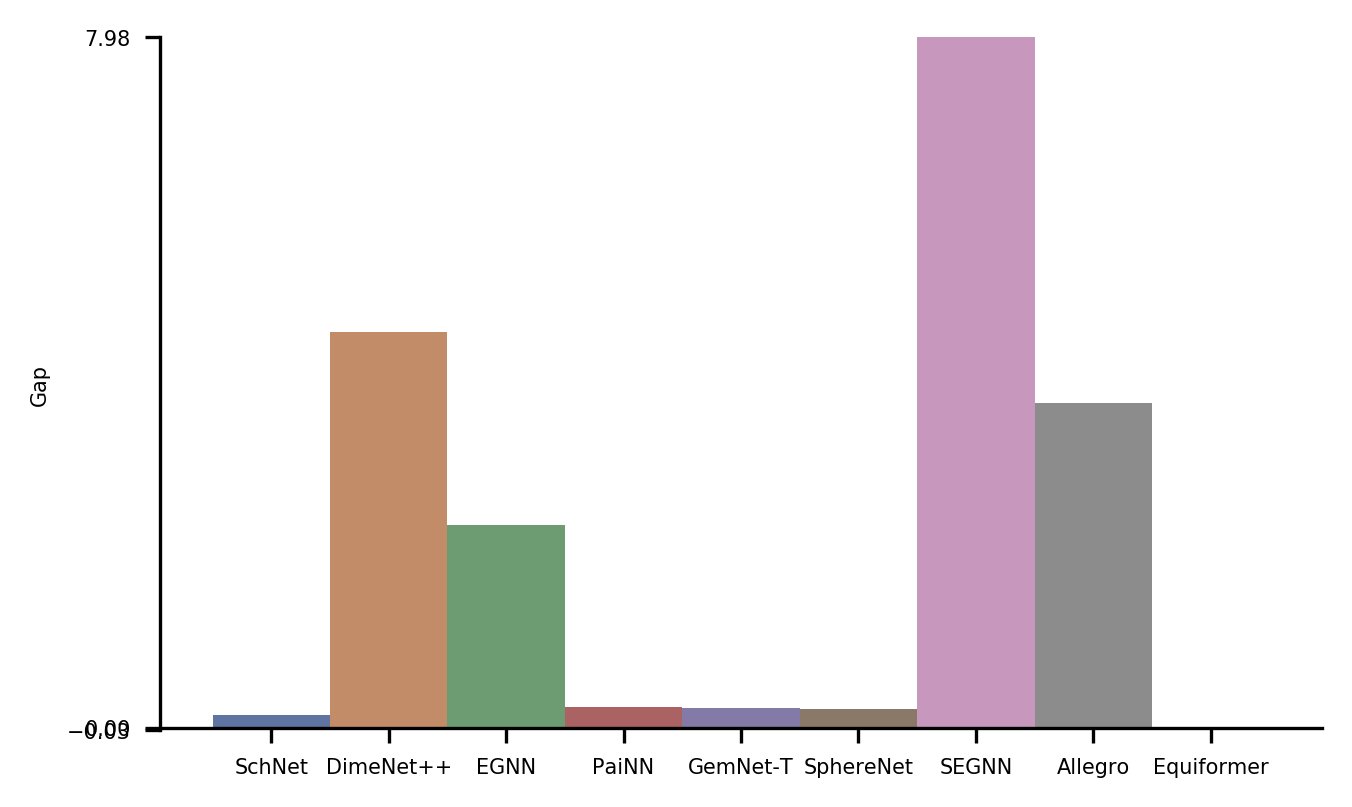}}
    \end{subfigure}
\hfill
    \begin{subfigure}[\small Task Benzene]
    {\includegraphics[width=0.24\linewidth]{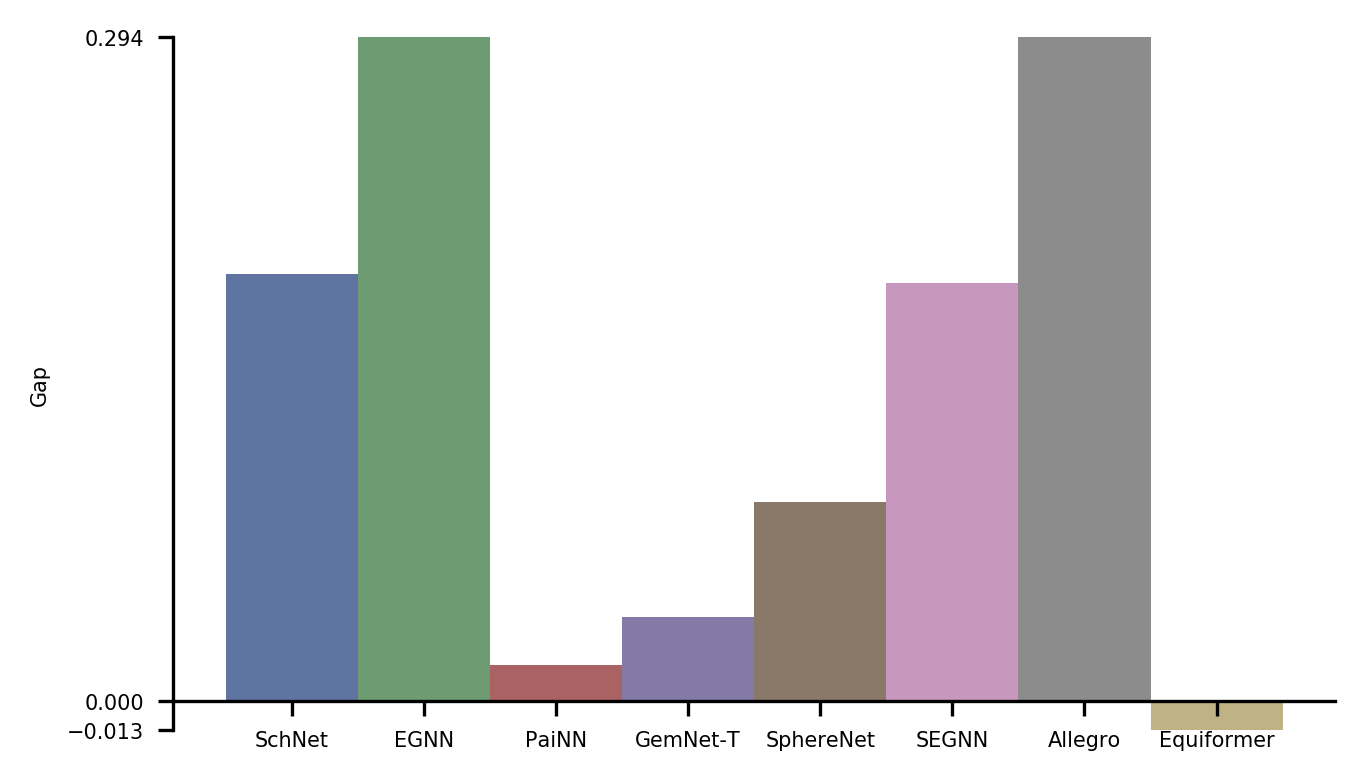}}
    \end{subfigure}
\hfill
    \begin{subfigure}[\small Task Ethanol]
    {\includegraphics[width=0.24\linewidth]{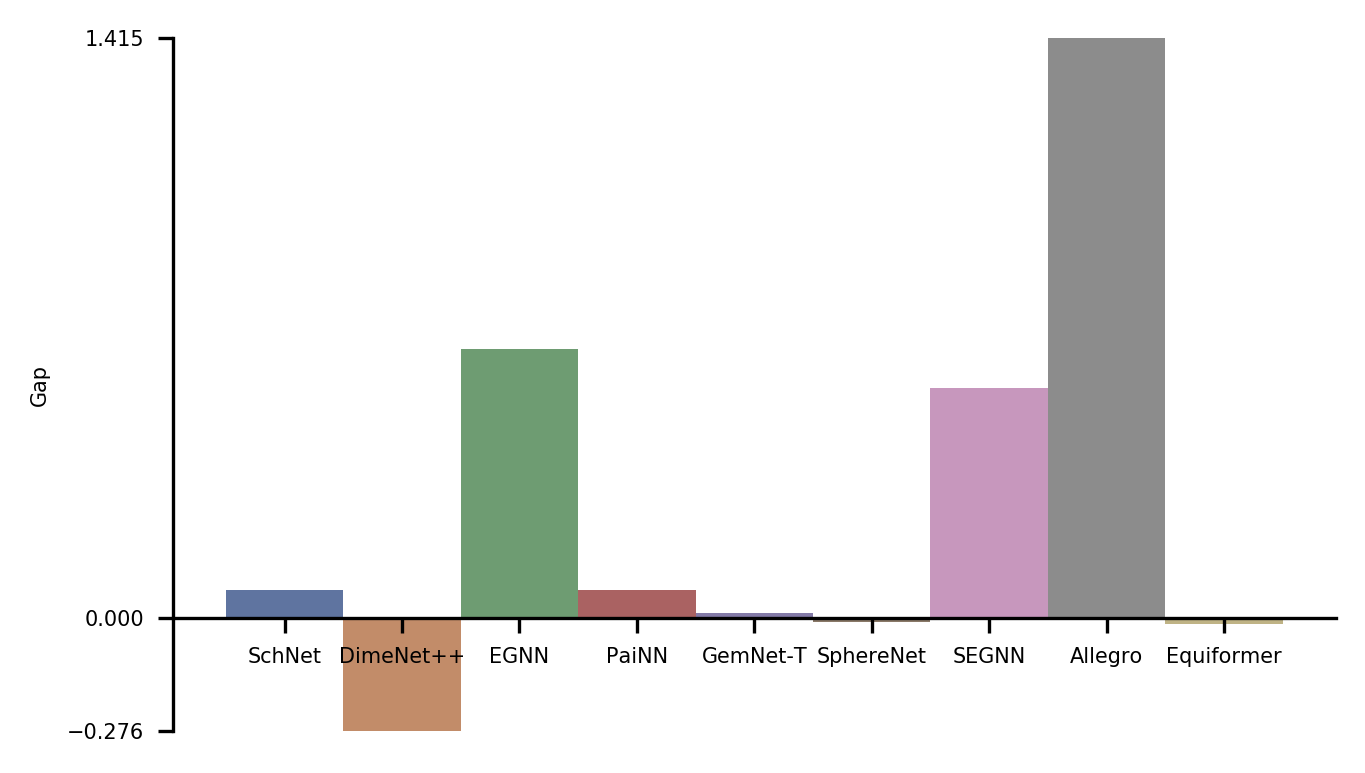}}
    \end{subfigure}
\hfill
    \begin{subfigure}[\small Task Malonaldehyde]
    {\includegraphics[width=0.24\linewidth]{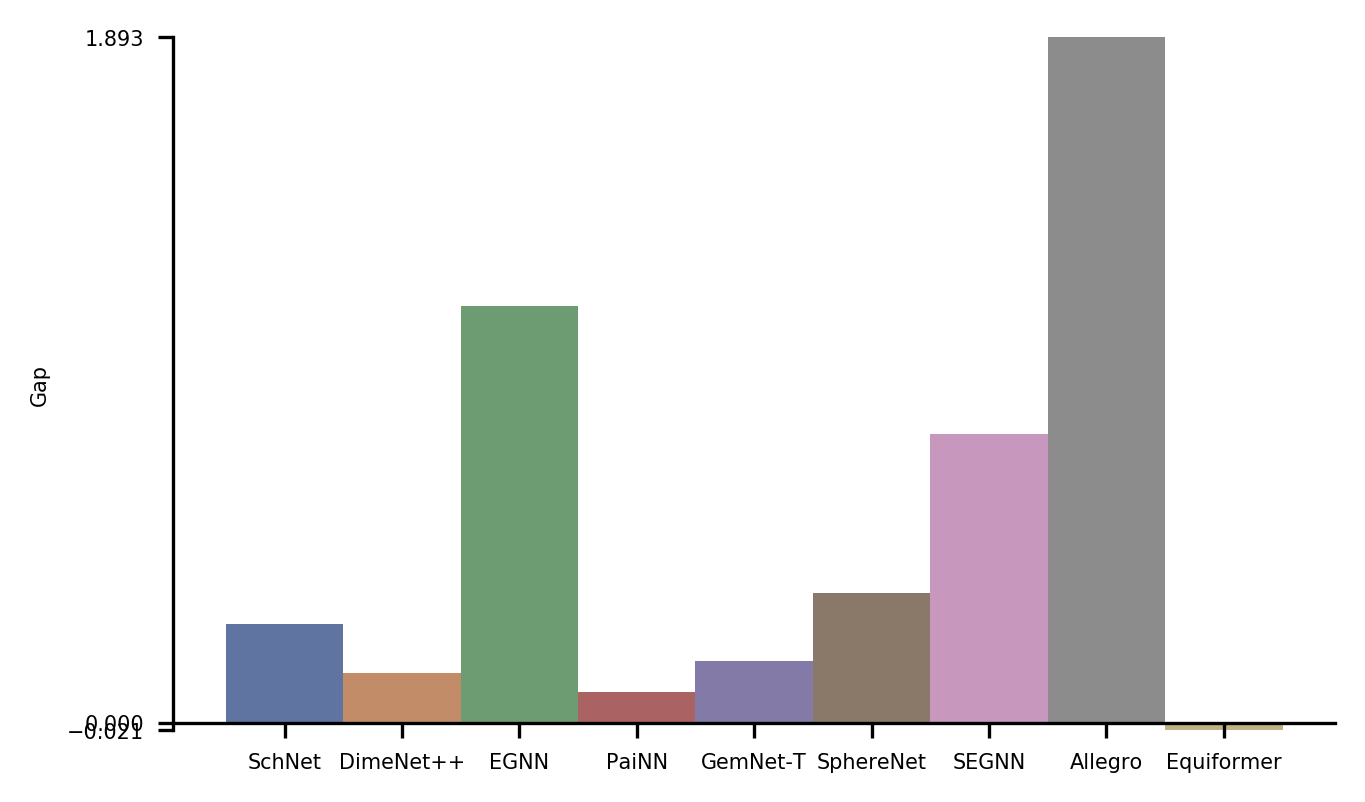}}
    \end{subfigure}
\hfill
    \begin{subfigure}[\small Task Naphthalene]
    {\includegraphics[width=0.24\linewidth]{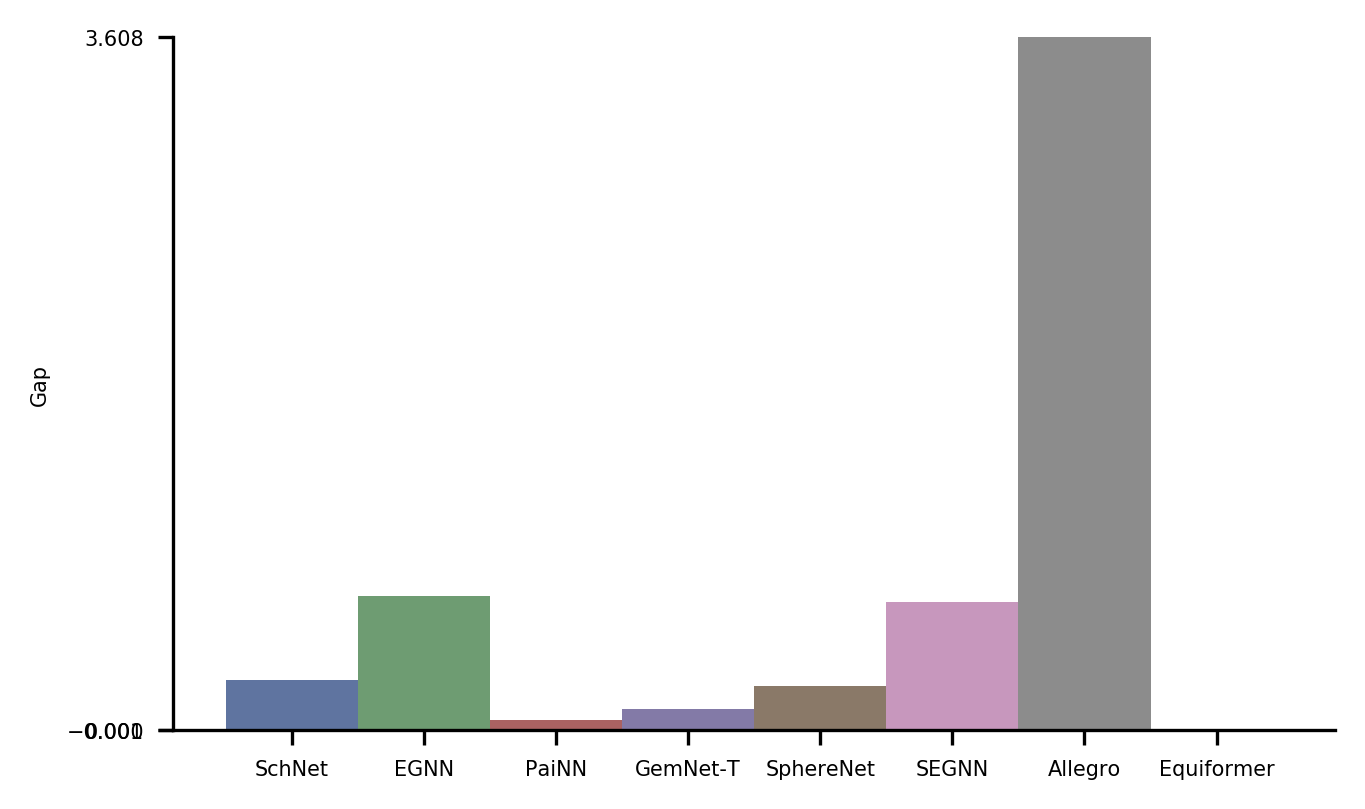}}
    \end{subfigure}
\hfill
    \begin{subfigure}[\small Task Salicylic]
    {\includegraphics[width=0.24\linewidth]{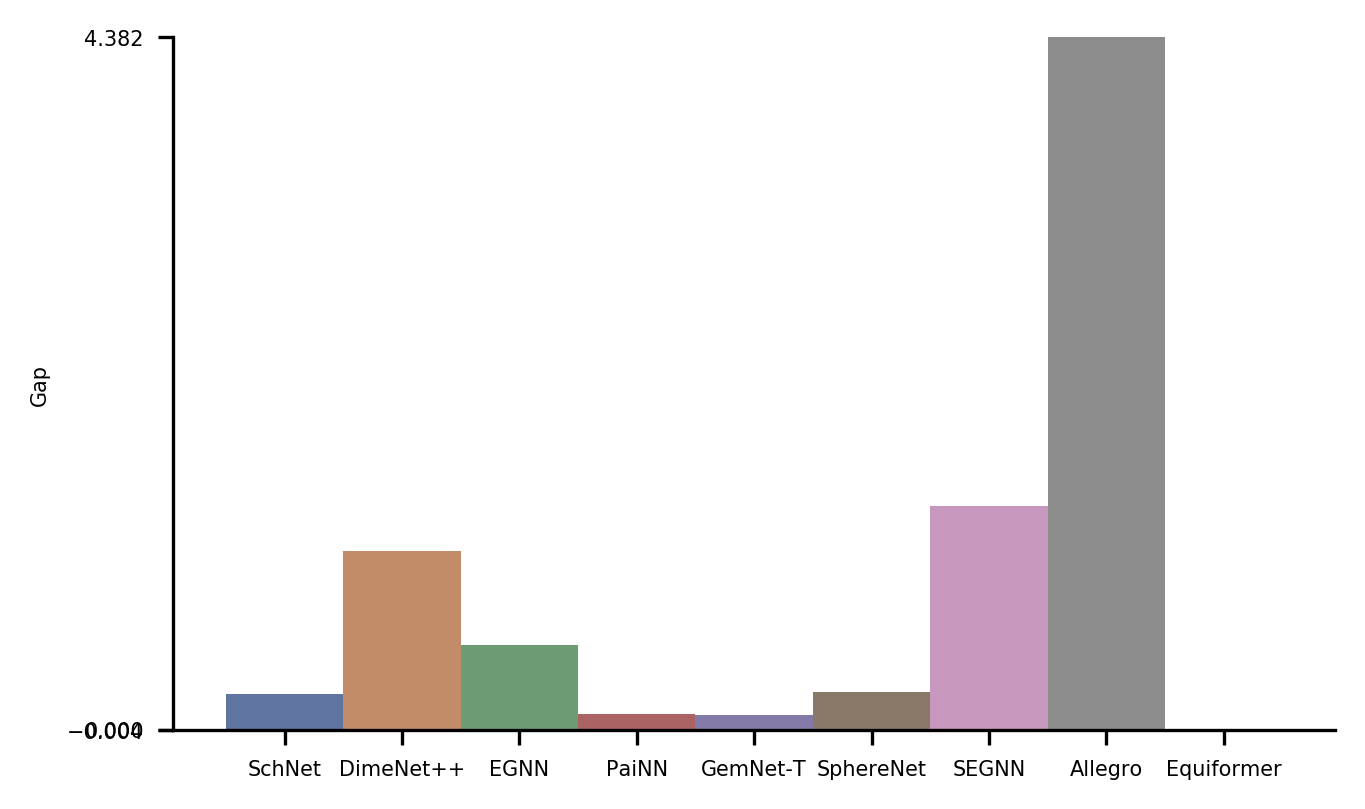}}
    \end{subfigure}
\hfill
    \begin{subfigure}[\small Task Toluene]
    {\includegraphics[width=0.24\linewidth]{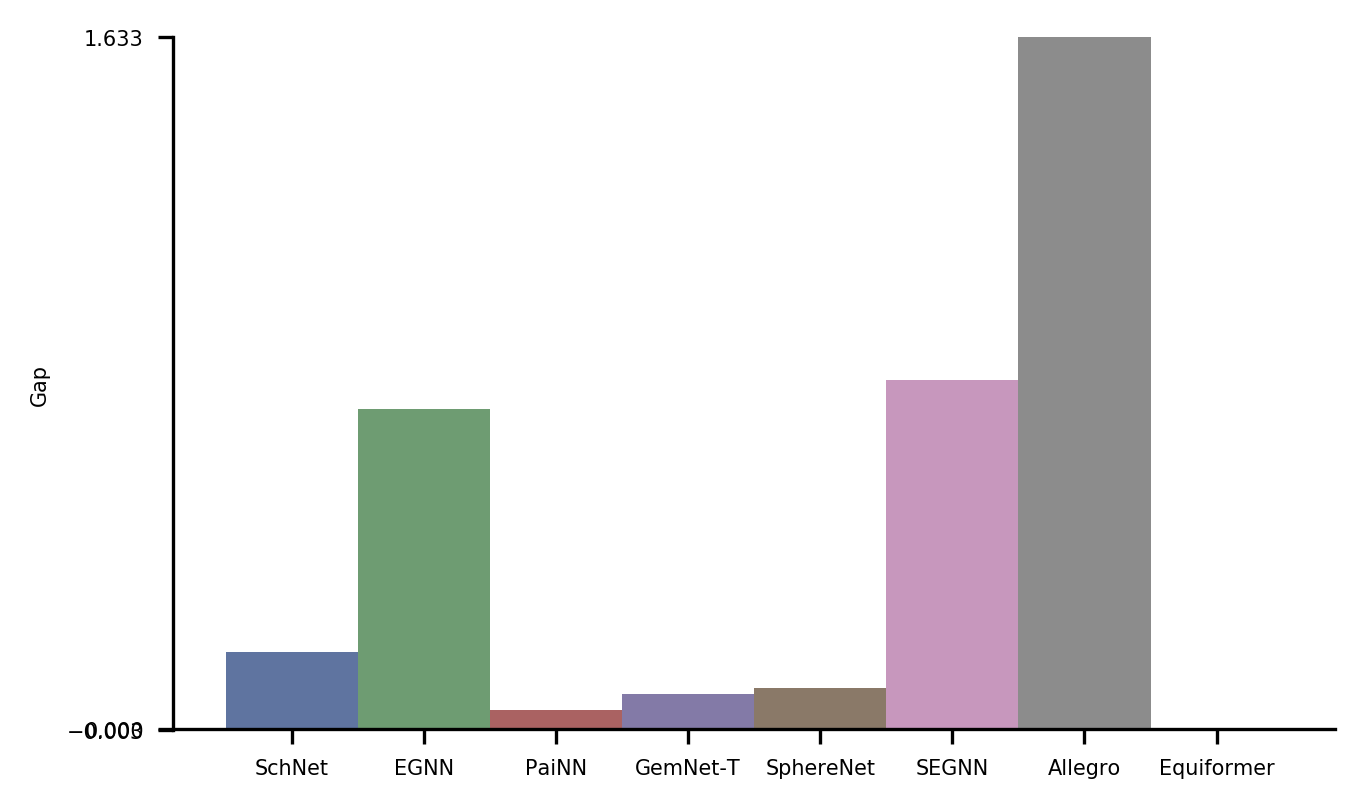}}
    \end{subfigure}
\hfill
    \begin{subfigure}[\small Task Uracil]
    {\includegraphics[width=0.24\linewidth]{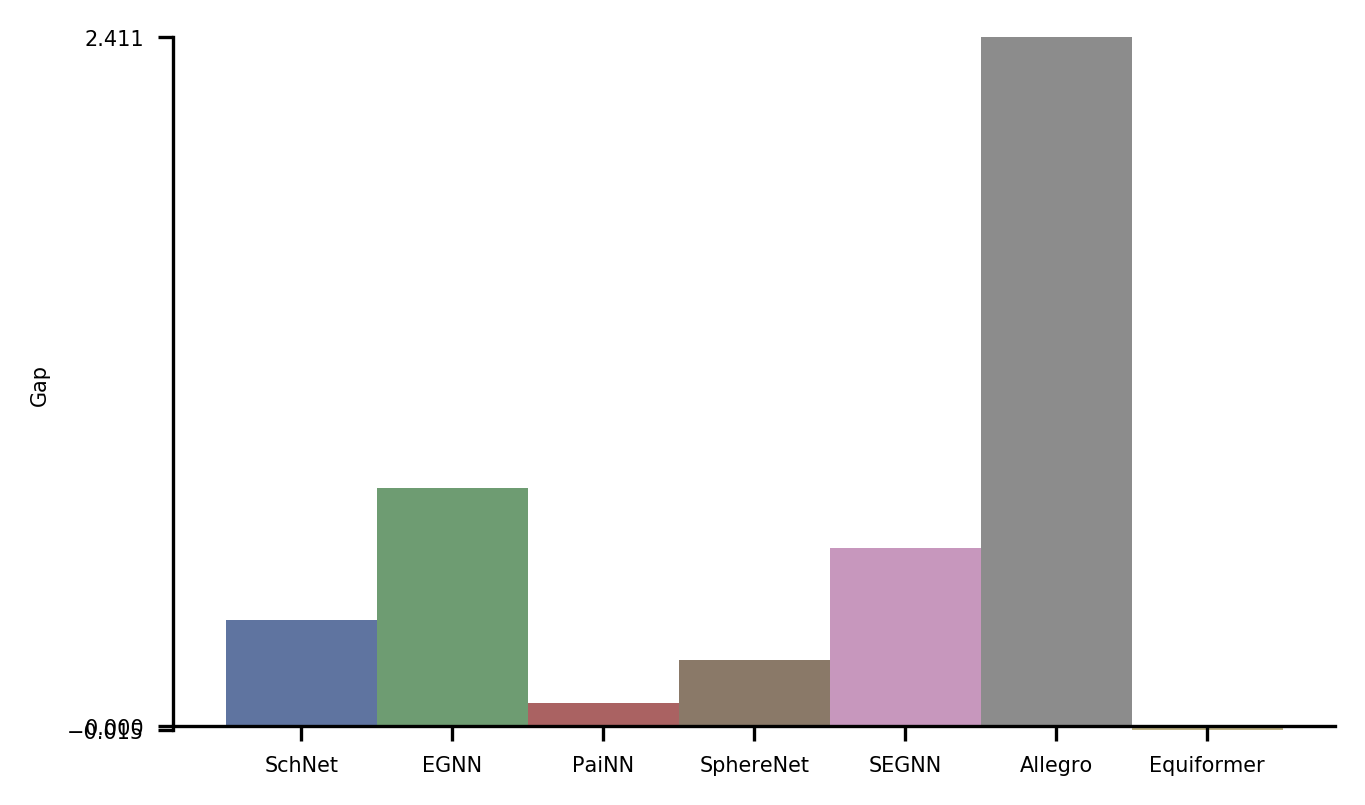}}
    \end{subfigure}
\vspace{-2ex}
\caption{
\small
Performance gap of MAE(force prediction, $d=300$ and w/o normalization) - MAE(force prediction, $d=300$ and w/ normalization) in MD17.
}
\vspace{-2.5ex}
\end{figure}

\clearpage
\null
\begin{table}[htb!]
\setlength{\tabcolsep}{5pt}
\fontsize{9}{9}\selectfont
\centering
\caption{
\small
Ablation studies of latent dimension (dim=128) on rMD17.
The evaluation is the mean absolute error.
Data normalization is used.
}
\label{tab:ablation_study_latent_dim_128_rMD17_with_normalization}
\vspace{-1.5ex}
\begin{adjustbox}{max width=\textwidth}
\begin{tabular}{ll c c c c c c c c c c}
\toprule
Model & Energy / Force & Aspirin $\downarrow$ & Azobenzene $\downarrow$ & Benzene $\downarrow$ & Ethanol $\downarrow$ & Malonaldehyde $\downarrow$ & Naphthalene $\downarrow$ & Paracetamol $\downarrow$ & Salicylic $\downarrow$ & Toluene $\downarrow$ & Uracil $\downarrow$ \\
\midrule

\multirow{2}{*}{SchNet}
% 5e-4_CosineAnnealingLR_128_5_1000_energy_force_with_normalization
& Energy & 0.702 & 0.583 & 0.013 & 0.055 & 0.079 & 0.058 & 0.179 & 0.103 & 0.064 & 0.054\\
& Force & 1.028 & 0.780 & 0.137 & 0.311 & 0.488 & 0.314 & 0.757 & 0.578 & 0.371 & 0.373\\
\midrule

\multirow{2}{*}{DimeNet++}
% 5e-4_CosineAnnealingLR_128_800_energy_force_with_normalization
& Energy & 0.321 & 0.560 & 0.050 & 0.093 & 0.142 & 0.157 & 0.299 & 0.353 & 0.143 & 0.318\\
& Force & 0.536 & 0.424 & 0.102 & 0.284 & 0.420 & 0.239 & 0.447 & 0.639 & 0.231 & 0.341\\
\midrule

\multirow{2}{*}{EGNN}
% 1e-4_CosineAnnealingLR_128_5_1_1000_energy_force_with_normalization
& Energy & 0.653 & 0.684 & 0.056 & 0.275 & 0.238 & 0.440 & 0.476 & 0.514 & 0.233 & 0.395\\
& Force & 1.102 & 1.003 & 0.275 & 0.939 & 0.955 & 0.826 & 0.971 & 0.911 & 0.560 & 1.031\\
\midrule

\multirow{2}{*}{PaiNN}
% 1e-4_CosineAnnealingLR_128_5_1000_energy_force_with_normalization
& Energy & 0.187 & 0.076 & 0.006 & 0.046 & 0.076 & 0.048 & 0.109 & 0.063 & 0.033 & 0.040\\
& Force & 0.551 & 0.260 & 0.035 & 0.282 & 0.396 & 0.151 & 0.407 & 0.328 & 0.177 & 0.238\\
\midrule

\multirow{2}{*}{GemNet-T}
% 5e-4_CosineAnnealingLR_300_5_1000_energy_force_with_normalization
& Energy & 0.116 & 0.058 & 0.002 & 0.038 & 0.078 & 0.018 & 0.082 & 0.047 & 0.017 & 0.023\\
& Force & 0.329 & 0.198 & 0.020 & 0.179 & 0.328 & 0.094 & 0.267 & 0.226 & 0.090 & 0.155\\
\midrule

\multirow{2}{*}{SphereNet}
% 1e-4_CosineAnnealingLR_128_1_1000_energy_force_with_normalization
& Energy & 0.124 & 0.069 & 0.019 & 0.039 & 0.074 & 0.040 & 0.096 & 0.063 & 0.042 & 0.061\\
& Force & 0.325 & 0.189 & 0.028 & 0.174 & 0.282 & 0.091 & 0.265 & 0.226 & 0.095 & 0.191\\
\midrule

\multirow{2}{*}{SEGNN}
% 1e-4_CosineAnnealingLR_128_5_5_0_700_energy_force_with_normalization
& Energy & 0.509 & 0.171 & 0.005 & 0.039 & 0.056 & 0.052 & 0.194 & 0.150 & 0.080 & 0.045\\
& Force & 1.129 & 0.603 & 0.054 & 0.279 & 0.394 & 0.254 & 0.792 & 0.682 & 0.365 & 0.327\\
\midrule

\multirow{2}{*}{Allegro}
% 1e-3_CosineAnnealingLR_4.0_128_1000_energy_force_with_normalization
& Energy & 0.348 & 0.183 & 0.005 & 0.046 & 0.081 & 0.094 & 0.190 & 0.131 & 0.080 & 0.046\\
& Force & 0.673 & 0.385 & 0.039 & 0.249 & 0.371 & 0.289 & 0.476 & 0.430 & 0.270 & 0.254\\
\midrule

\multirow{2}{*}{Equiformer}
% 1e-4_CosineAnnealingLR_128_5_1_1000_energy_force_with_normalization
& Energy & 0.106 & 0.044 & 0.002 & 0.030 & 0.038 & 0.016 & 0.112 & 0.050 & 0.021 & 0.025\\
& Force & 0.321 & 0.134 & 0.026 & 0.183 & 0.264 & 0.070 & 0.284 & 0.222 & 0.079 & 0.164\\
\bottomrule
\end{tabular}
\end{adjustbox}
\end{table}

\begin{table}[htb!]
\setlength{\tabcolsep}{5pt}
\fontsize{9}{9}\selectfont
\centering
\caption{\small
Ablation studies of latent dimension (dim=300) on rMD17.
The evaluation is the mean absolute error.
Data normalization is used.
}
\label{tab:ablation_study_latent_dim_300_rMD17_with_normalization}
\vspace{-1.5ex}
\begin{adjustbox}{max width=\textwidth}
\begin{tabular}{ll c c c c c c c c c c}
\toprule
Model & Energy / Force & Aspirin $\downarrow$ & Azobenzene $\downarrow$ & Benzene $\downarrow$ & Ethanol $\downarrow$ & Malonaldehyde $\downarrow$ & Naphthalene $\downarrow$ & Paracetamol $\downarrow$ & Salicylic $\downarrow$ & Toluene $\downarrow$ & Uracil $\downarrow$ \\
\midrule

\multirow{2}{*}{SchNet}
% 5e-4_CosineAnnealingLR_300_5_1000_energy_force_with_normalization
& Energy & 0.556 & 0.482 & 0.013 & 0.059 & 0.107 & 0.067 & 0.218 & 0.122 & 0.119 & 0.064\\
& Force & 1.115 & 0.824 & 0.094 & 0.338 & 0.536 & 0.349 & 0.783 & 0.636 & 0.397 & 0.391\\
\midrule

\multirow{2}{*}{DimeNet++}
% 5e-4_CosineAnnealingLR_300_800_energy_force_with_normalization
& Energy & 0.339 & 0.257 & 10.026 & 0.118 & 0.201 & 0.135 & 0.550 & 0.213 & 0.156 & 1.382\\
& Force & 0.588 & 0.456 & 378.561 & 0.313 & 0.453 & 0.263 & 0.493 & 0.601 & 0.262 & 4.510\\
\midrule

\multirow{2}{*}{EGNN}
% 1e-4_CosineAnnealingLR_300_5_1_1000_energy_force_with_normalization
& Energy & 0.455 & 0.522 & 0.048 & 0.070 & 0.068 & 0.212 & 0.313 & 0.233 & 0.359 & 0.150\\
& Force & 0.738 & 0.720 & 0.234 & 0.314 & 0.391 & 0.515 & 0.684 & 0.618 & 0.682 & 0.603\\
\midrule

\multirow{2}{*}{PaiNN}
% 1e-4_CosineAnnealingLR_300_5_1000_energy_force_with_normalization
& Energy & 0.127 & 0.056 & 0.002 & 0.037 & 0.056 & 0.017 & 0.078 & 0.044 & 0.022 & 0.024\\
& Force & 0.443 & 0.183 & 0.019 & 0.237 & 0.331 & 0.095 & 0.331 & 0.248 & 0.126 & 0.171\\
\midrule

\multirow{2}{*}{GemNet-T}
% 5e-4_CosineAnnealingLR_300_5_1000_energy_force_with_normalization
& Energy & 0.116 & 0.058 & 0.002 & 0.038 & 0.078 & 0.018 & 0.082 & 0.047 & 0.017 & 0.023\\
& Force & 0.329 & 0.198 & 0.020 & 0.179 & 0.328 & 0.094 & 0.267 & 0.226 & 0.090 & 0.155\\
\midrule

\multirow{2}{*}{SphereNet}
% 1e-4_CosineAnnealingLR_300_1_1000_energy_force_with_normalization
& Energy & 0.132 & 0.087 & 0.010 & 0.048 & 0.123 & 0.027 & 0.101 & 0.079 & 0.027 & 0.066\\
& Force & 0.348 & 0.203 & 0.023 & 0.194 & 0.315 & 0.090 & 0.283 & 0.248 & 0.094 & 0.206\\
\midrule

\multirow{2}{*}{SEGNN}
% 1e-4_CosineAnnealingLR_300_5_5_0_700_energy_force_with_normalization
& Energy & 0.570 & 0.300 & 0.005 & 0.037 & 0.064 & 0.061 & 0.283 & 0.210 & 0.096 & 0.062\\
& Force & 1.313 & 0.732 & 0.055 & 0.264 & 0.499 & 0.285 & 1.003 & 0.782 & 0.346 & 0.410\\
\midrule

\multirow{2}{*}{Allegro}
% 1e-3_CosineAnnealingLR_4.0_300_1000_energy_force_with_normalization
& Energy & 0.294 & 0.167 & 0.004 & 0.043 & 0.056 & 0.070 & 0.170 & 0.093 & 0.063 & 0.037\\
& Force & 0.597 & 0.347 & 0.034 & 0.212 & 0.312 & 0.237 & 0.435 & 0.367 & 0.233 & 0.217\\
\midrule

\multirow{2}{*}{Equiformer}
% 1e-4_CosineAnnealingLR_300_5_1_1000_energy_force_with_normalization
& Energy & 0.101 & 0.044 & 0.002 & 0.030 & 0.041 & 0.016 & 0.090 & 0.045 & 0.020 & 0.024\\
& Force & 0.321 & 0.134 & 0.026 & 0.180 & 0.265 & 0.070 & 0.284 & 0.223 & 0.079 & 0.164\\
\bottomrule
\end{tabular}
\end{adjustbox}
\end{table}

\begin{figure}[htb!]
\centering
    \begin{subfigure}[\small Task Aspirin]
    {\includegraphics[width=0.24\linewidth]{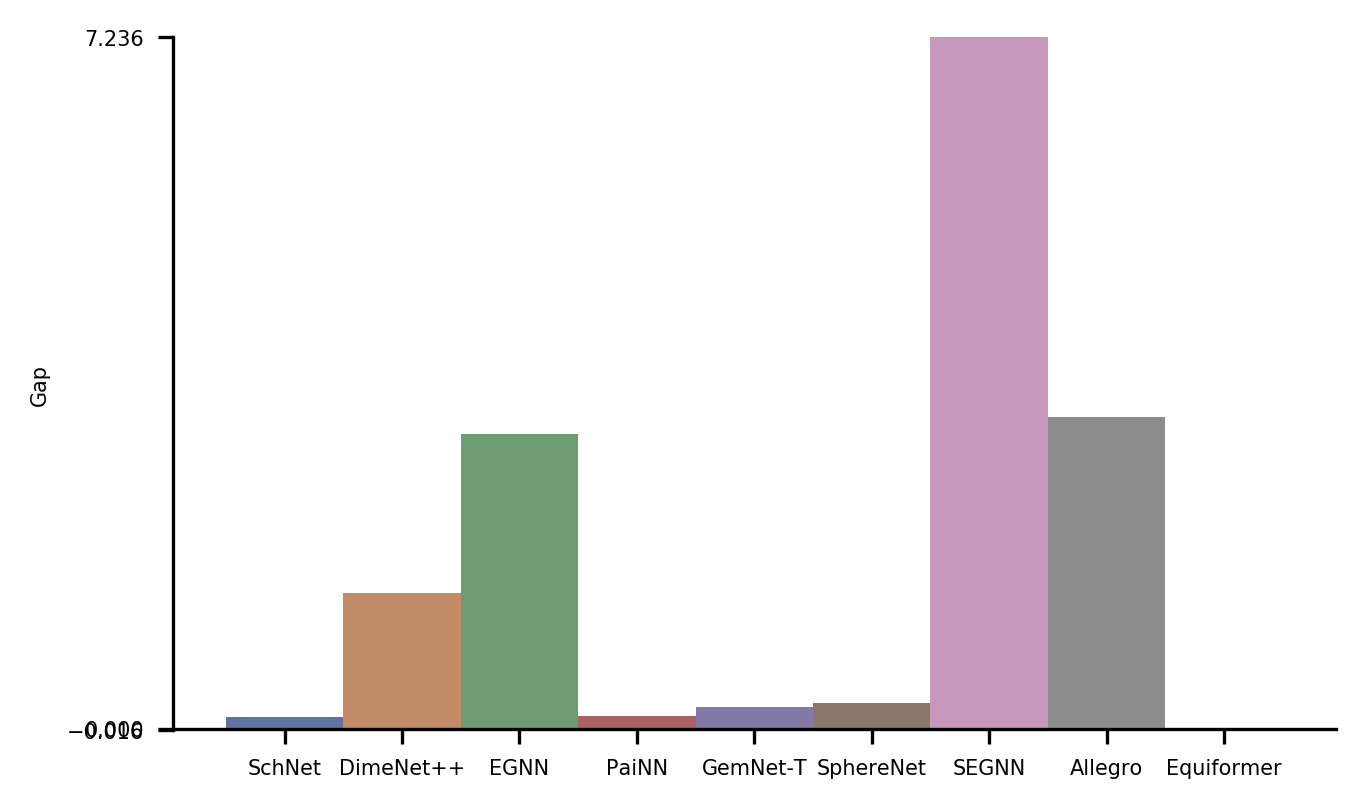}}
    \end{subfigure}
% \hfill
    \begin{subfigure}[\small Task Azobenzene]
    {\includegraphics[width=0.24\linewidth]{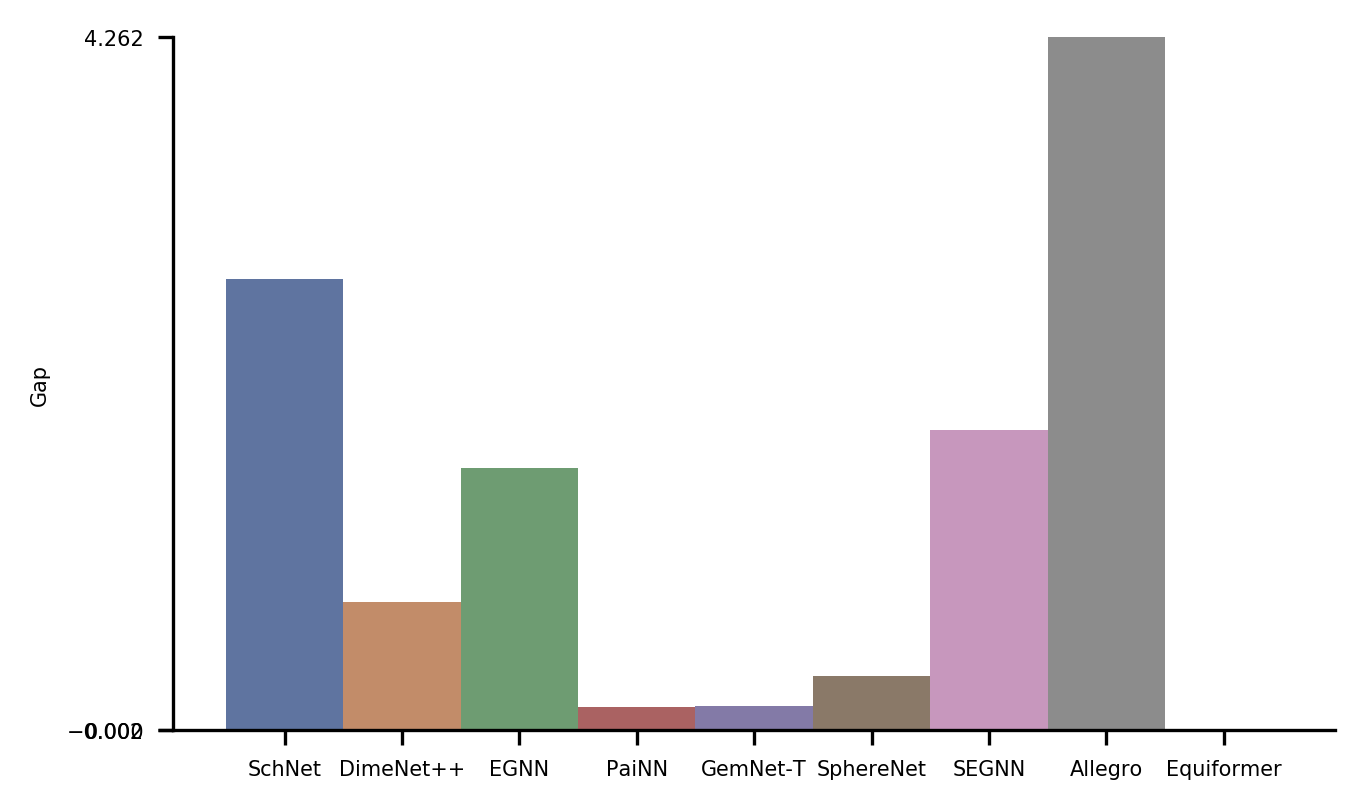}}
    \end{subfigure}
% \hfill
    \begin{subfigure}[\small Task Benzene]
    {\includegraphics[width=0.24\linewidth]{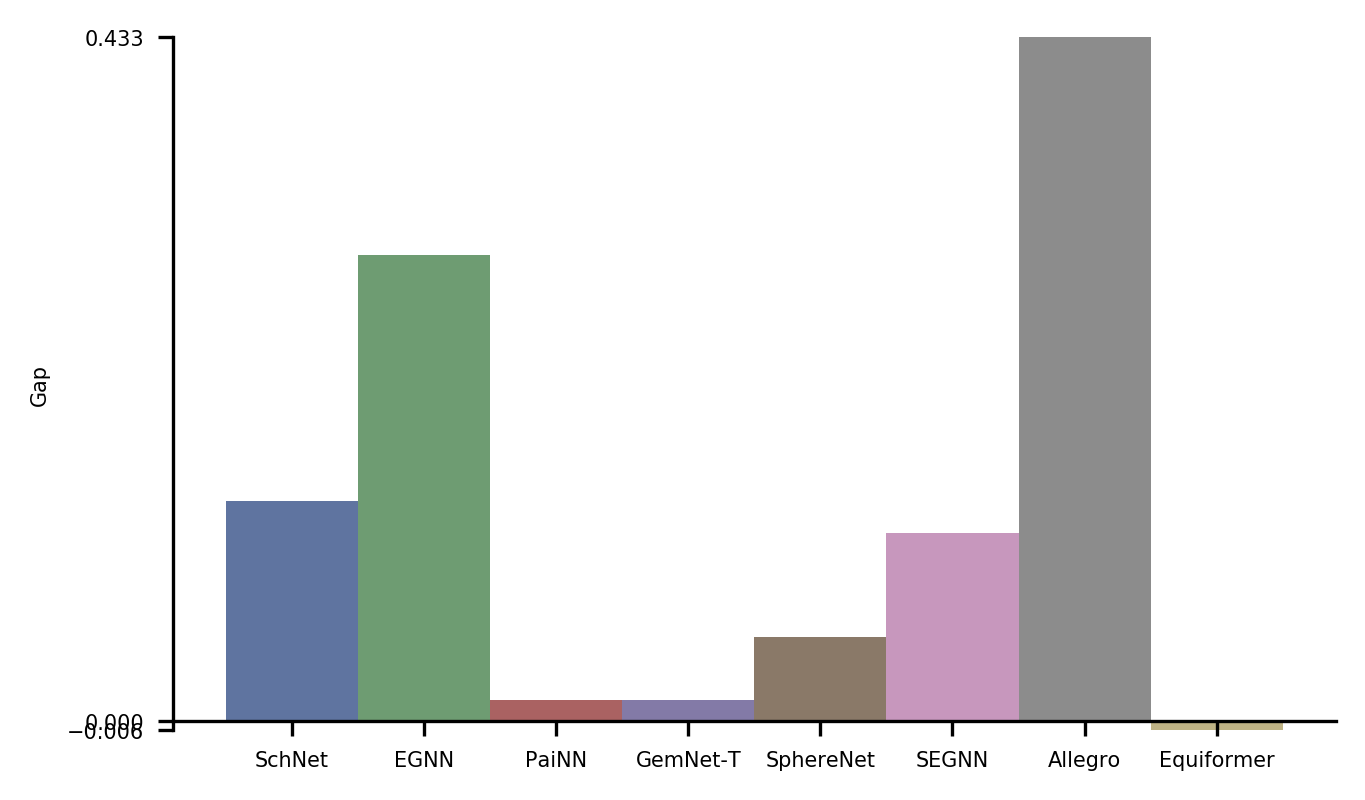}}
    \end{subfigure}
% \hfill
    \begin{subfigure}[\small Task Ethanol]
    {\includegraphics[width=0.24\linewidth]{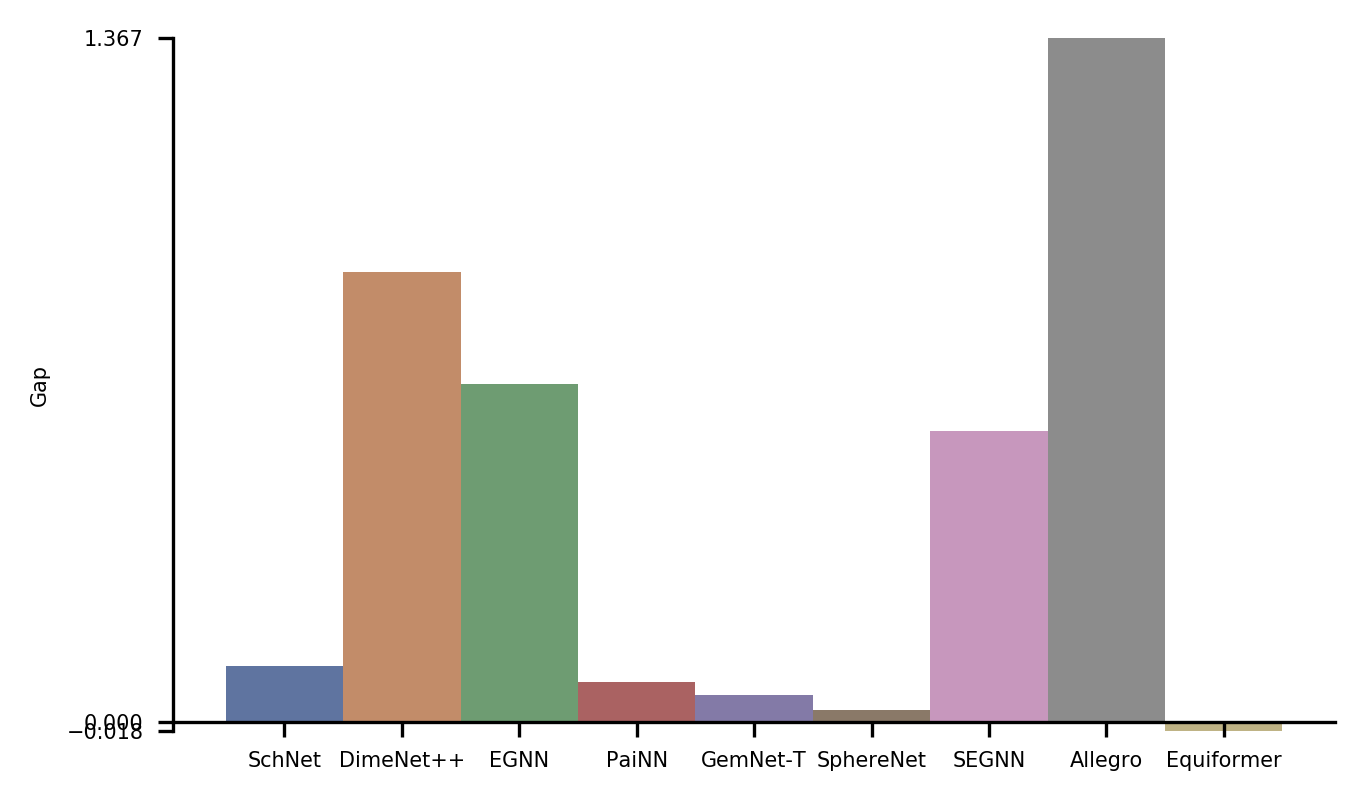}}
    \end{subfigure}
% \hfill
    \begin{subfigure}[\small Task Malonaldehyde]
    {\includegraphics[width=0.24\linewidth]{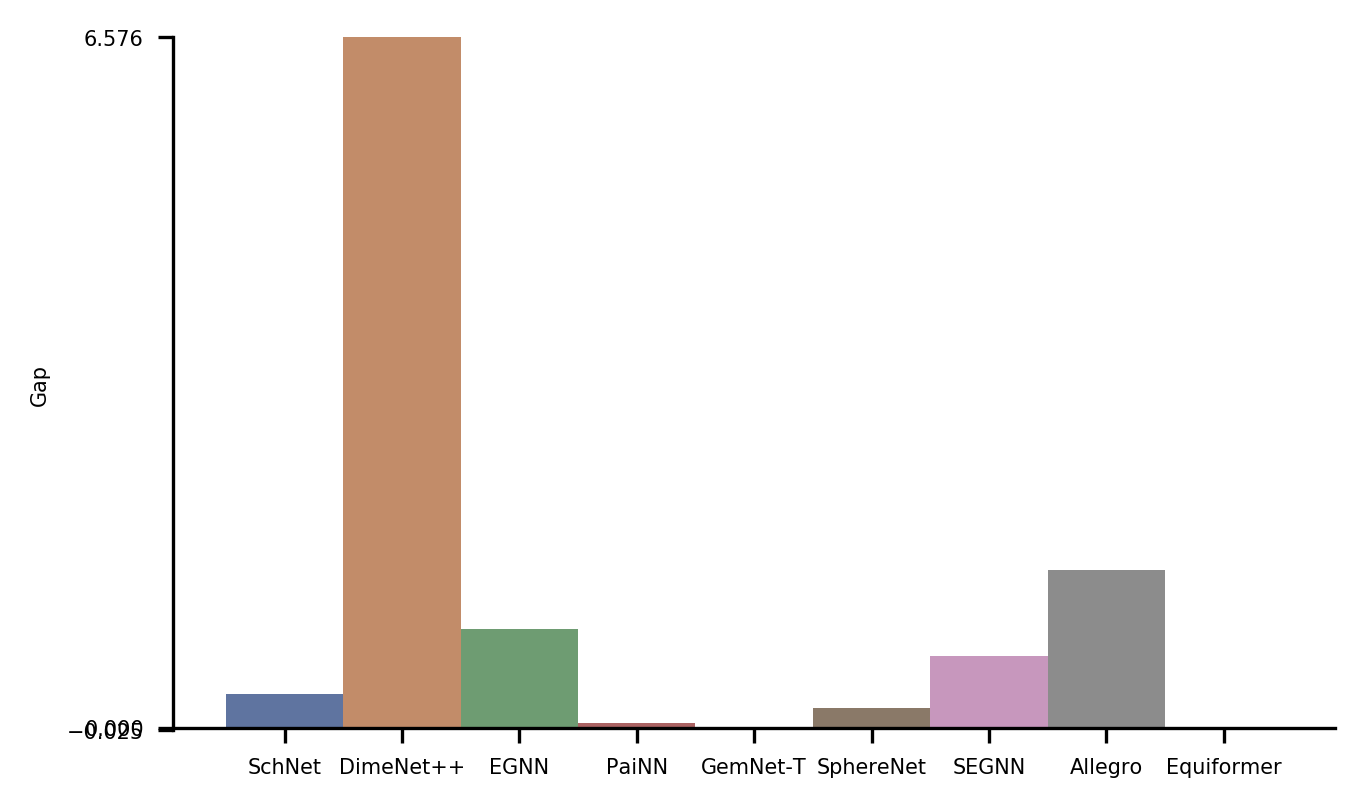}}
    \end{subfigure}
% \hfill
    \begin{subfigure}[\small Task Naphthalene]
    {\includegraphics[width=0.24\linewidth]{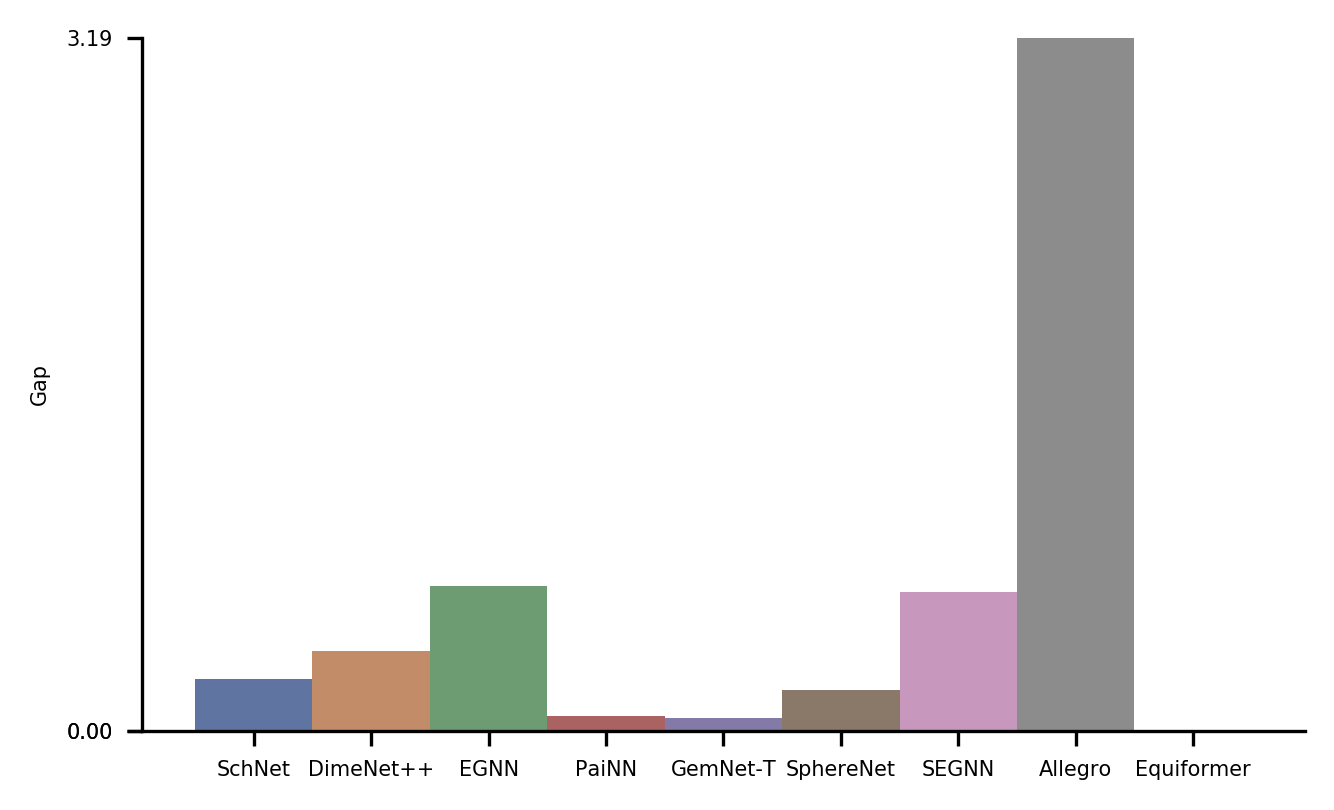}}
    \end{subfigure}
% \hfill
    \begin{subfigure}[\small Task Paracetamol]
    {\includegraphics[width=0.24\linewidth]{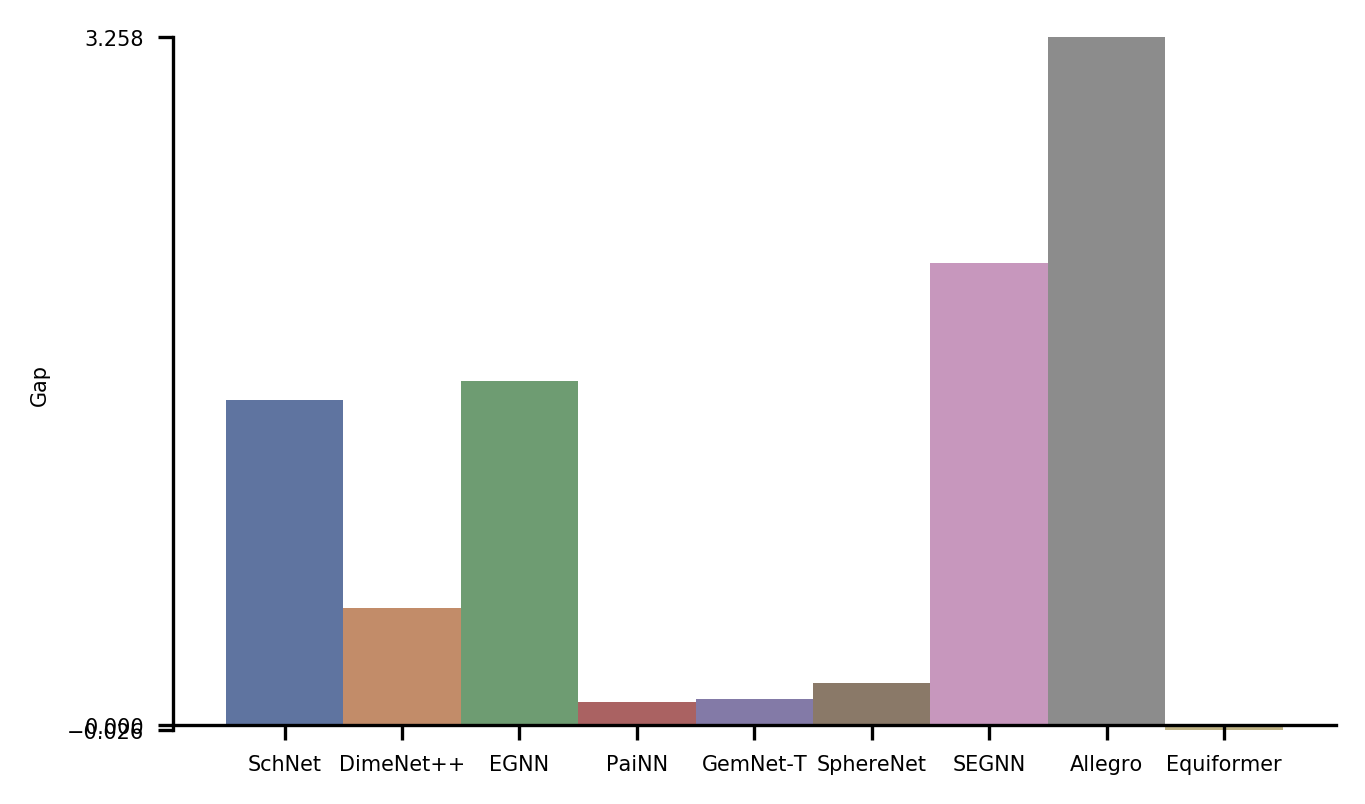}}
    \end{subfigure}
% \hfill
    \begin{subfigure}[\small Task Salicylic]
    {\includegraphics[width=0.24\linewidth]{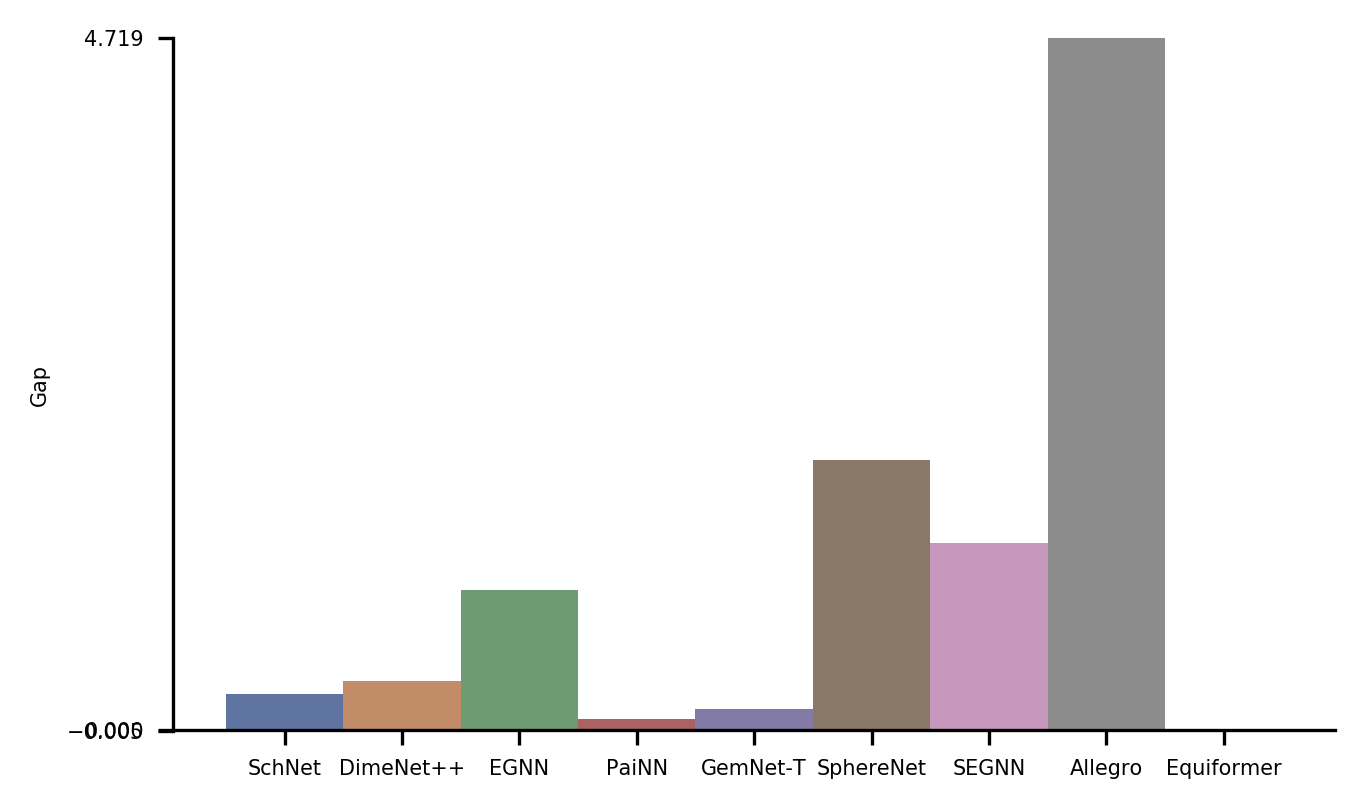}}
    \end{subfigure}
% \hfill
    \begin{subfigure}[\small Task Toluene]
    {\includegraphics[width=0.24\linewidth]{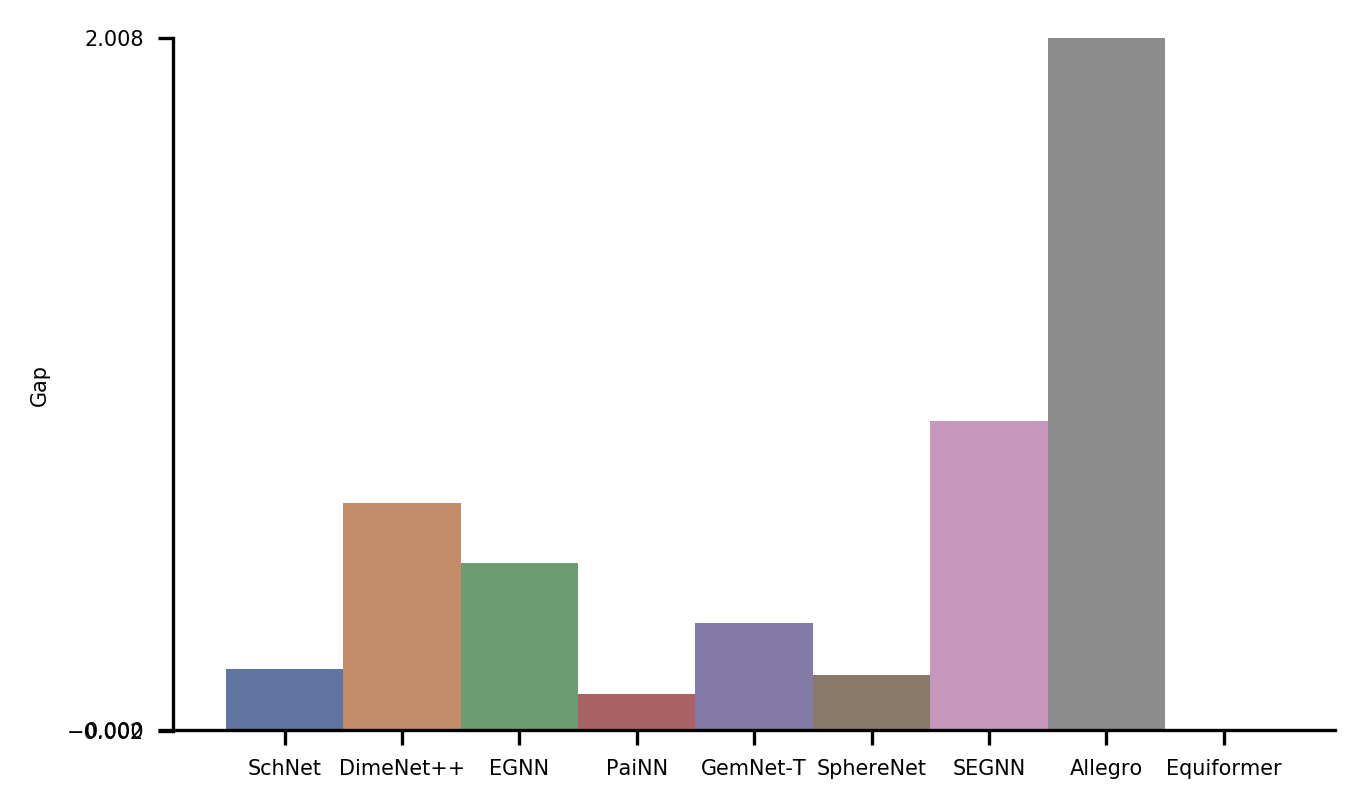}}
    \end{subfigure}
% \hfill
    \begin{subfigure}[\small Task Uracil]
    {\includegraphics[width=0.24\linewidth]{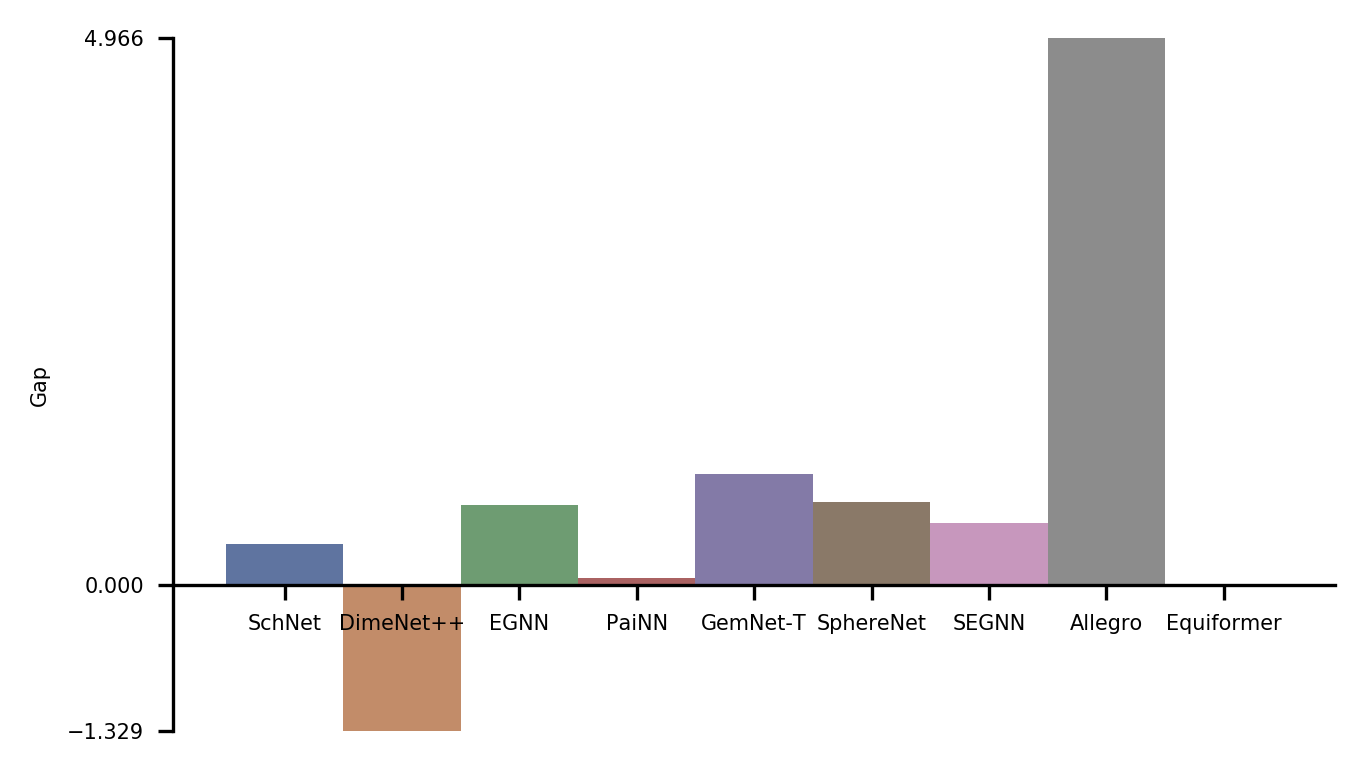}}
    \end{subfigure}
\vspace{-2ex}
\caption{
\small
Performance gap of MAE(force prediction, $d=300$ and w/o normalization) - MAE(force prediction, $d=300$ and w/ normalization) in rMD17.
}
\vspace{-2.5ex}
\end{figure}

\clearpage
%%%%%%%%%%%%%%%%%%%%%%%%%%%%%%%%%%%%%%%%%%%%%%%%%%
\subsection{Ablation Studies on Reproduced Results of NequIP and Allegro} \label{sec:app:reproduced_results_NequIP_Allegro}

Here we would like to further discuss NequIP and Allegro.

\begin{itemize}[noitemsep,topsep=0pt]
    \item NequIP has no explicit molecule-level representation, and we directly put its results below.
    \item Allegro adopts $d=512$ by default (by far we are mainly checking $d=128$ and $d=300$).
    \item We can reproduce NequIP and Allegro results w/ data normalization, as shown below.
\end{itemize}

\begin{table}[htb!]
\setlength{\tabcolsep}{5pt}
\fontsize{9}{9}\selectfont
\centering
\caption{
\small
Ablation study of data normalization on NequIP and Allegro on MD17.
The evaluation is the mean absolute error.
Here Allegro uses $d=512$, and both NequIP and Allegro can match the reported results~\cite{batzner20223,musaelian2022learning} w/ normalization.
}
\vspace{-1.5ex}
\begin{adjustbox}{max width=\textwidth}
\begin{tabular}{lll c c c c c c c c}
\toprule
Model & Normalization & Energy / Force & Aspirin $\downarrow$ & Benzene $\downarrow$ & Ethanol $\downarrow$ & Malonaldehyde $\downarrow$ & Naphthalene $\downarrow$ & Salicylic $\downarrow$ & Toluene $\downarrow$ & Uracil $\downarrow$ \\
\midrule

\multirow{4}{*}{NequIP}
& \multirow{2}{*}{w/o Normalization}
% 1e-3_CosineAnnealingLR__4.0_1000_energy_force_no_normalization
& Energy & 8.333 & 0.355 & 0.971 & 2.293 & 1.032 & 2.952 & 1.303 & 1.266\\
& & Force & 23.769 & 2.383 & 5.832 & 12.099 & 5.247 & 14.048 & 6.800 & 8.060\\
\cmidrule(lr){2-11} 

& \multirow{2}{*}{w/ Normalization}
% 1e-3_CosineAnnealingLR__4.0_1000_energy_force_with_normalization
& Energy & 0.175 & 0.095 & 0.058 & 0.089 & 0.114 & 0.114 & 0.094 & 0.105\\
& & Force & 0.383 & 0.039 & 0.195 & 0.294 & 0.091 & 0.212 & 0.106 & 0.136\\
\midrule

\multirow{4}{*}{Allegro}
& \multirow{2}{*}{w/o Normalization}
% 1e-3_CosineAnnealingLR_4.0_512_1000_energy_force_no_normalization
& Energy & 1.138 & 0.154 & 0.258 & 1.330 & 0.824 & 1.114 & 0.441 & 0.613\\
& & Force & 3.405 & 0.823 & 1.412 & 4.191 & 3.743 & 4.934 & 1.968 & 3.544\\
\cmidrule(lr){2-11} 

& \multirow{2}{*}{w/ Normalization}
% 1e-3_CosineAnnealingLR_4.0_512_1000_energy_force_with_normalization
& Energy & 0.240 & 0.096 & 0.058 & 0.085 & 0.128 & 0.130 & 0.107 & 0.107\\
& & Force & 0.553 & 0.058 & 0.179 & 0.259 & 0.207 & 0.311 & 0.203 & 0.184\\
\bottomrule
\end{tabular}
\end{adjustbox}
\end{table}

\begin{table}[htb!]
\setlength{\tabcolsep}{5pt}
\fontsize{9}{9}\selectfont
\centering
\caption{
\small
Ablation study of data normalization on NequIP and Allegro on rMD17.
The evaluation is the mean absolute error.
Here Allegro uses $d=512$.
}
\vspace{-1.5ex}
\begin{adjustbox}{max width=\textwidth}
\begin{tabular}{lll c c c c c c c c c c}
\toprule
Model & Normalization & Energy / Force & Aspirin $\downarrow$ & Azobenzene $\downarrow$ & Benzene $\downarrow$ & Ethanol $\downarrow$ & Malonaldehyde $\downarrow$ & Naphthalene $\downarrow$ & Paracetamol $\downarrow$ & Salicylic $\downarrow$ & Toluene $\downarrow$ & Uracil $\downarrow$ \\
\midrule

\multirow{4}{*}{NequIP}
& \multirow{2}{*}{w/o Normalization}
% 1e-3_CosineAnnealingLR__4.0_1000_energy_force_no_normalization
& Energy & 9.618 & 1.993 & 3.048 & 0.936 & 2.313 & 2.089 & 5.136 & 3.302 & 1.306 & 1.738\\
& & Force & 22.904 & 6.406 & 1.523 & 6.027 & 12.372 & 5.529 & 17.574 & 15.693 & 7.094 & 10.220\\
\cmidrule(lr){2-13}

& \multirow{2}{*}{w/ Normalization}
% 1e-3_CosineAnnealingLR__4.0_1000_energy_force_with_normalization
& Energy & 0.147 & 0.049 & 0.003 & 0.034 & 0.061 & 0.018 & 0.078 & 0.047 & 0.020 & 0.021\\
& & Force & 0.407 & 0.176 & 0.019 & 0.218 & 0.310 & 0.092 & 0.308 & 0.230 & 0.113 & 0.142\\
\midrule

\multirow{4}{*}{Allegro}
& \multirow{2}{*}{w/o Normalization}
% 1e-3_CosineAnnealingLR_4.0_512_1000_energy_force_no_normalization
& Energy & 1.366 & 0.872 & 0.029 & 1.002 & 0.417 & 1.756 & 0.944 & 1.035 & 0.437 & 0.387\\
& & Force & 3.186 & 2.763 & 0.237 & 2.799 & 2.125 & 3.815 & 3.081 & 4.781 & 2.048 & 1.939\\
\cmidrule(lr){2-13}

& \multirow{2}{*}{w/ Normalization}
% 1e-3_CosineAnnealingLR_4.0_512_1000_energy_force_with_normalization
& Energy & 0.223 & 0.146 & 0.003 & 0.033 & 0.053 & 0.060 & 0.156 & 0.079 & 0.054 & 0.031\\
& & Force & 0.558 & 0.308 & 0.029 & 0.198 & 0.264 & 0.207 & 0.409 & 0.331 & 0.210 & 0.187\\
\bottomrule
\end{tabular}
\end{adjustbox}
\end{table}

%%%%%%%%%%%%%%%%%%%%%%%%%%%%%%%%%%%%%%%%%%%%%%%%%%
\subsection{Ablation Study on the Data Split of Crystalline Material}
In the main paper, we report the results on MatBench with 60\%-20\%-20\% for train-valid-test split. To verify the reproducibility correctness of \framework{}, we carry on an ablation study with the same setting as MatBench~\cite{dunn2020benchmarking}. Notice that MatBench adopts the setting in KGCNN~\cite{REISER2021100095}: with seed 18012019 and 80\% for training and 20\% for the test. The reproduced results are in~\Cref{tab:MatBench_reproduced_results}.

The mean evaluation metrics of SchNet and DimeNet++ with cross-validation are reported in \href{https://matbench.materialsproject.org/Benchmark%20Info/matbench_v0.1/}{MatBench leaderboard} and \href{https://github.com/aimat-lab/gcnn_keras/tree/master/training/results}{KGCNN leaderboard}, and evaluation metrics of the PaiNN are reported in \href{https://github.com/aimat-lab/gcnn_keras/tree/master/training/results}{KGCNN leaderboard}.

\begin{table}[htb!]
\setlength{\tabcolsep}{5pt}
\fontsize{9}{9}\selectfont
\centering
\caption{
\small
Reproduced results on 8 MatBench tasks.
}
\label{tab:MatBench_reproduced_results}
\vspace{-1.5ex}
\begin{adjustbox}{max width=\textwidth}
\begin{tabular}{l rrrrrrrrr}
\toprule
\multirow{2}{*}{Model} & Per. $E_{\text{form}}$ $\downarrow$ & Dielectric $\downarrow$ & $log_{10} G$ $\downarrow$ & $log_{10} K$ $\downarrow$ & $E_{\text{exfo}}$ $\downarrow$ & Phonons $\downarrow$ & $E_{\text{form}}$ $\downarrow$ & Band Gap $\downarrow$\\
& 18,928 & 4,764 & 10,987 & 10,987 & 636 & 1,265 & 106,113 & 106,113\\
\midrule

SchNet (MatBench) & 0.0342 & 0.3277 & 0.0796 & 0.0590 & 42.6637 & 38.9636 & 0.0218 & 0.2352 \\
SchNet (KGCNN) & 0.0347 & 0.3241 & 0.0798 & 0.0584 & 48.0629 & 40.2982 & 0.0215 & 0.9351\\

% 5e-4_CosineAnnealingLR_300_64_800_5_5_image_gathered
SchNet (\framework{}, ours) & 0.035 & 0.334 & 0.080 & 0.060 & 49.363 & 35.172 & 0.023 & 0.226\\
\midrule

DimeNet++ (MatBench) & 0.0376 & 0.3400 & 0.0792 & 0.0572 & 49.0243 & 37.4619 & 0.0235 & 0.1993\\
DimeNet++ (KGCNN) & 0.0373 & 0.3337 & 0.0805 & 0.0579 & 49.2113 & 36.7288 & 0.0233 & 0.2089 \\

% 1e-4_CosineAnnealingLR_128_32_5_17_mean_image_gathered_800
DimeNet++ (\framework{}, ours) & 0.033 & 0.340 & 0.080 & 0.060 & 47.700 & 33.564 &
% 1e-4_CosineAnnealingLR_128_32_5_17_mean_image_gathered_600
0.022 & 0.207\\
\midrule

PaiNN (KGCNN) & 0.0456 & 0.3587 & 0.0851 & 0.0646 & 50.5886 & 47.2212 & 0.0244 & 0.2220\\

% 5e-4_CosineAnnealingLR_128_32_5_3.25_image_gathered_mean_1000
PaiNN (\framework{}, ours) & 0.033 & 0.323 & 0.081 & 0.053 & 42.325 & 38.859 &
% 1e-4_CosineAnnealingLR_128_32_5_3.25_image_gathered_mean_800
0.022 & 0.192\\
\bottomrule

\end{tabular}
\end{adjustbox}
\end{table}

\clearpage
%%%%%%%%%%%%%%%%%%%%%%%%%%%%%%%%%%%%%%%%%%%%%%%%%%
\subsection{Ablation Study on the Data Augmentation of Crystalline Material}

The default latent dimension $d=300$ for most of the models, except for EGNN and SEGNN, which lead to the out-of-memory exception. Besides, SEGNN may collapse with gathered DA, so we skip that in the comparison.

\begin{table}[htb!]
\setlength{\tabcolsep}{5pt}
\fontsize{9}{9}\selectfont
\centering
\caption{
\small
Ablation study on data augmentation (DA) on MatBench and QMOF.
}
\vspace{-2ex}
\begin{adjustbox}{max width=\textwidth}
\begin{tabular}{l l rrrrrrrrrrr}
\toprule
\multirow{4}{*}{Model} & \multirow{4}{*}{DA} & \multicolumn{8}{c}{MatBench} & \multicolumn{1}{c}{QMOF}\\
\cmidrule(lr){3-10} \cmidrule(lr){11-11}
& & Per. $E_{\text{form}}$ $\downarrow$ & Dielectric $\downarrow$ & $log_{10} G$ $\downarrow$ & $log_{10} K$ $\downarrow$ & $E_{\text{exfo}}$ $\downarrow$ & Phonons $\downarrow$ & $E_{\text{form}}$ $\downarrow$ & Band Gap $\downarrow$ & Band Gap $\downarrow$\\
& & 18,928 & 4,764 & 10,987 & 10,987 & 636 & 1,265 & 106,113 & 106,113 & 20,425\\
\midrule

\multirow{2}{*}{SchNet}
% 5e-4_CosineAnnealingLR_300_32_800_5_3.25_image_gathered
& gathered & 0.040 & 0.334 & 0.081 & 0.060 & 65.201 & 42.586 & 0.026 & 0.327
% 5e-4_CosineAnnealingLR_300_32_5_5_image_gathered_300
& 0.236\\

% 5e-4_CosineAnnealingLR_300_32_800_5_3.25_image_expanded
& expanded & 0.048 & 0.338 & 0.086 & 0.066 & 62.991 & 46.301 & 0.042 & 0.253
% 5e-4_CosineAnnealingLR_300_32_5_5_image_expanded_300
& 0.278\\
\midrule

\multirow{2}{*}{DimeNet++}
& gathered
% 1e-4_CosineAnnealingLR_300_16_5_17_mean_image_gathered_800
& 0.037 & 0.357 & 0.081 & 0.058 & 68.685 & 38.339 &
% 1e-4_CosineAnnealingLR_300_16_5_17_mean_image_gathered_300
0.025 & 0.208
% 5e-4_CosineAnnealingLR_128_32_5_17_mean_image_gathered_300
& 0.235\\

& expanded
% 1e-4_CosineAnnealingLR_300_16_5_17_mean_image_expanded_800
& 0.042 & 0.334 & 0.088 & 0.064 & 69.579 & 45.223 &
% 1e-4_CosineAnnealingLR_300_16_5_17_mean_image_expanded_300
0.041 & 0.235
% 5e-4_CosineAnnealingLR_128_32_5_17_mean_image_expanded_300
& 0.245\\
\midrule

\multirow{2}{*}{EGNN}
% 1e-4_CosineAnnealingLR_300_32_5_17_image_gathered_800
& gathered & 0.407 & 0.329 & 0.128 & 0.087 & 76.245 & 86.922 & 0.097 & 0.296
% 5e-4_CosineAnnealingLR_128_16_300_5_image_gathered
& 0.493\\

% 1e-4_CosineAnnealingLR_300_32_5_17_image_expanded_800
& expanded & 0.039 & 0.306 & 0.089 & 0.064 & 78.205 & 76.143 & 0.026 & 0.211
% 5e-4_CosineAnnealingLR_128_16_300_5_image_expanded
& 0.256\\
\midrule

\multirow{2}{*}{PaiNN}
& gathered
% 1e-4_CosineAnnealingLR_300_32_5_3.25_17_image_gathered_1000
& 0.038 & 0.317 & 0.080 & 0.053 & 67.752 & 44.602 & 0.022 & 0.190
% 5e-4_CosineAnnealingLR_128_32_5_3.25_mean_image_gathered_300
& 0.240\\

& expanded
% 1e-4_CosineAnnealingLR_300_32_5_3.25_17_image_expanded_1000
& 0.038 & 0.327 & 0.083 & 0.056 & 73.224 & 59.930 & 0.029 & 0.203
% 5e-4_CosineAnnealingLR_128_64_5_3.25_mean_image_gathered_300
& 0.221\\
\midrule

\multirow{2}{*}{GemNet-T}
& gathered
% 5e-4_CosineAnnealingLR_300_32_5_17_0_image_gathered_800
& 0.042 & 0.325 & 0.088 & 0.061 & 68.425 & 48.986 &
% 5e-4_CosineAnnealingLR_300_32_5_17_0_image_gathered_150
0.026 & 0.186
% 5e-4_CosineAnnealingLR_300_16_5_17_0_image_gathered_200
& 0.207\\

& expanded
% 5e-4_CosineAnnealingLR_300_32_5_17_0_image_expanded_800
& 0.042 & 0.364 & 0.090 & 0.063 & 68.376 & 57.316 &
% 5e-4_CosineAnnealingLR_300_32_5_17_0_image_expanded_150
0.042 & 0.212
% 5e-4_CosineAnnealingLR_300_16_5_17_0_image_expanded_200
& 0.236\\
\midrule

\multirow{2}{*}{SphereNet}
& gathered
% 5e-4_CosineAnnealingLR_300_32_800_5_17_image_gathered
& 0.043 & 0.388 & 0.087 & 0.061 & 72.987 & 36.300 &
% 5e-4_CosineAnnealingLR_300_32_300_5_17_image_gathered
0.029 & 0.217
% 5e-4_CosineAnnealingLR_128_16_5_17_image_gathered_300
& 0.251\\

& expanded
% 5e-4_CosineAnnealingLR_300_32_800_5_17_image_expanded
& 0.047 & 0.359 & 0.090 & 0.062 & 69.267 & 49.401 &
% 5e-4_CosineAnnealingLR_300_32_300_5_17_image_expanded
0.039 & 0.233
% 5e-4_CosineAnnealingLR_128_16_5_17_image_expanded_300
& 0.268\\
\midrule

% % 1e-4_CosineAnnealingLR_128_4_60_5_17_image_gathered
% gathered SEGNN QMOF & 0.492\\

\multirow{1}{*}{SEGNN}
& expanded
% 5e-4_CosineAnnealingLR_128_16_300_5_17_image_expanded
& 0.046 & 0.360 & 0.087 & 0.059 & 65.052 & 43.638 &
% 5e-4_CosineAnnealingLR_128_16_40_5_17_image_expanded
0.047 & 0.330
% 1e-4_CosineAnnealingLR_128_4_60_5_17_image_expanded
& 0.330\\
\midrule

\multirow{2}{*}{Equiformer}
& gathered
% 1e-4_CosineAnnealingLR_300_32_5_17_image_gathered_500
& 0.046 & 0.280 & 0.087 & 0.057 & 62.977 & 37.381 &
% 1e-4_CosineAnnealingLR_300_32_5_17_image_gathered_100
0.031 & 0.216
% 1e-4_CosineAnnealingLR_128_8_5_17_image_gathered_150
& 0.238\\

& expanded
% 1e-4_CosineAnnealingLR_300_32_5_17_image_expanded_500
& 0.047 & 0.314 & 0.086 & 0.061 & 69.845 & 54.087 &
% 1e-4_CosineAnnealingLR_300_32_5_17_image_expanded_100
0.040 & 0.242
% 1e-4_CosineAnnealingLR_128_8_5_17_image_expanded_150
& 0.271\\
\bottomrule

\end{tabular}
\end{adjustbox}
\end{table}

\begin{figure}[htb!]
\centering
    \begin{subfigure}[\small MatBench Task Per. $E_{\text{form}}$]
    {\includegraphics[width=0.3\linewidth]{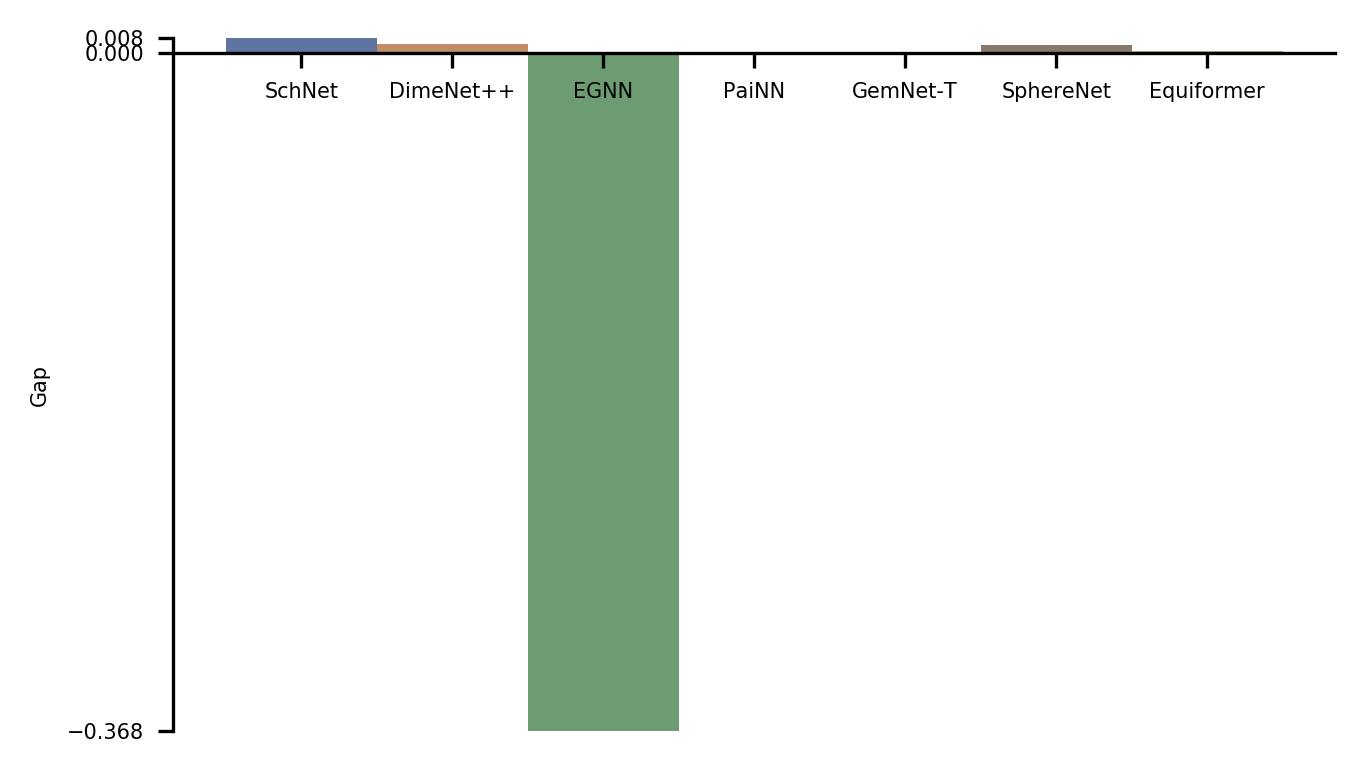}}
    \end{subfigure}
\hfill
    \begin{subfigure}[\small MatBench Task Dielectric]
    {\includegraphics[width=0.3\linewidth]{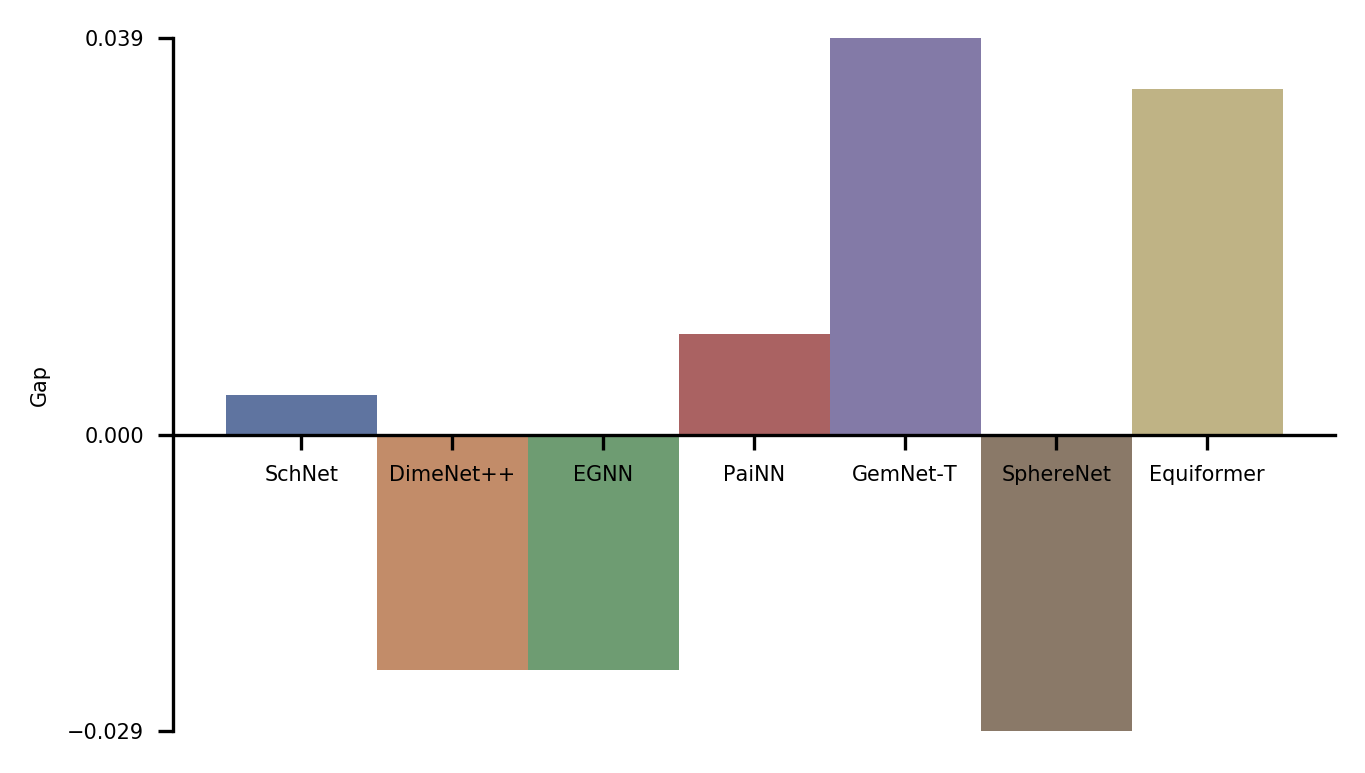}}
    \end{subfigure}
\hfill
    \begin{subfigure}[\small MatBench Task $log_{10} G$]
    {\includegraphics[width=0.3\linewidth]{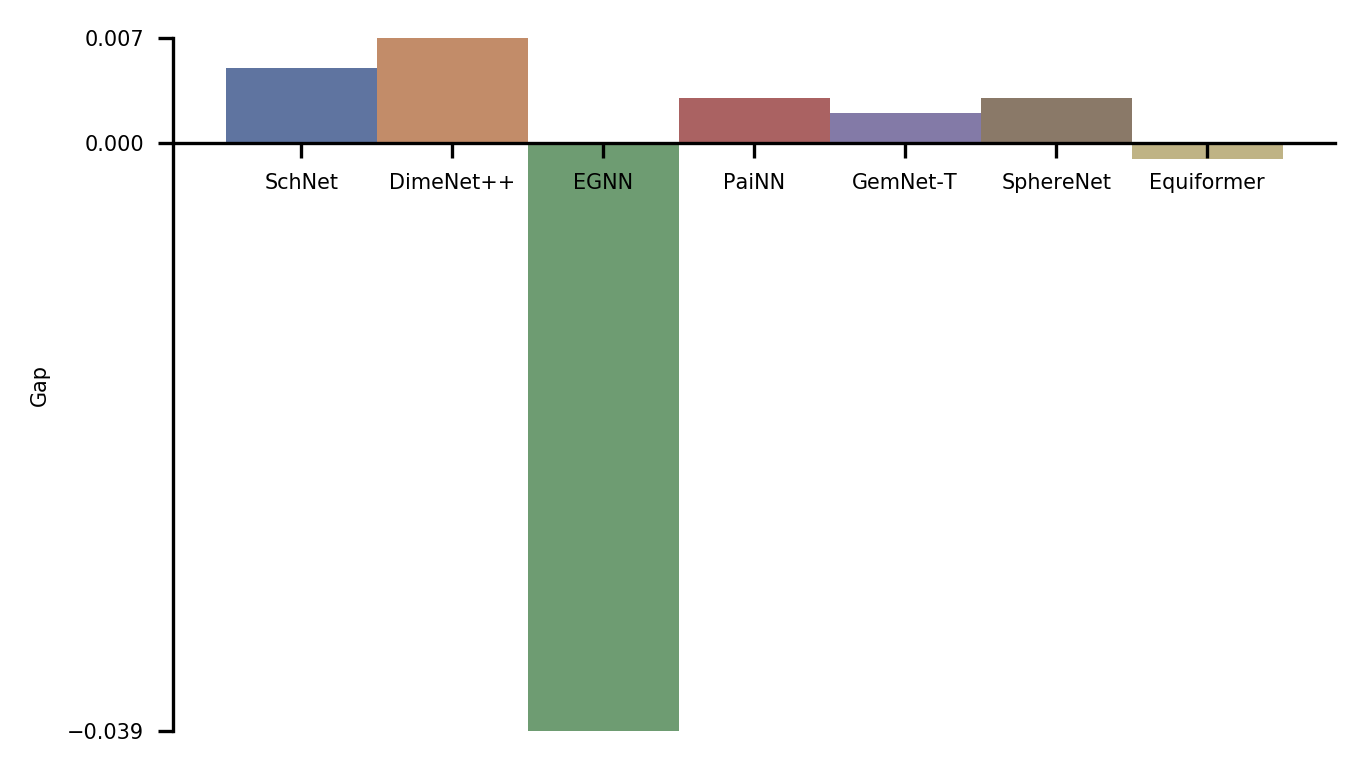}}
    \end{subfigure}
\hfill
    \begin{subfigure}[\small MatBench Task $log_{10} K$]
    {\includegraphics[width=0.3\linewidth]{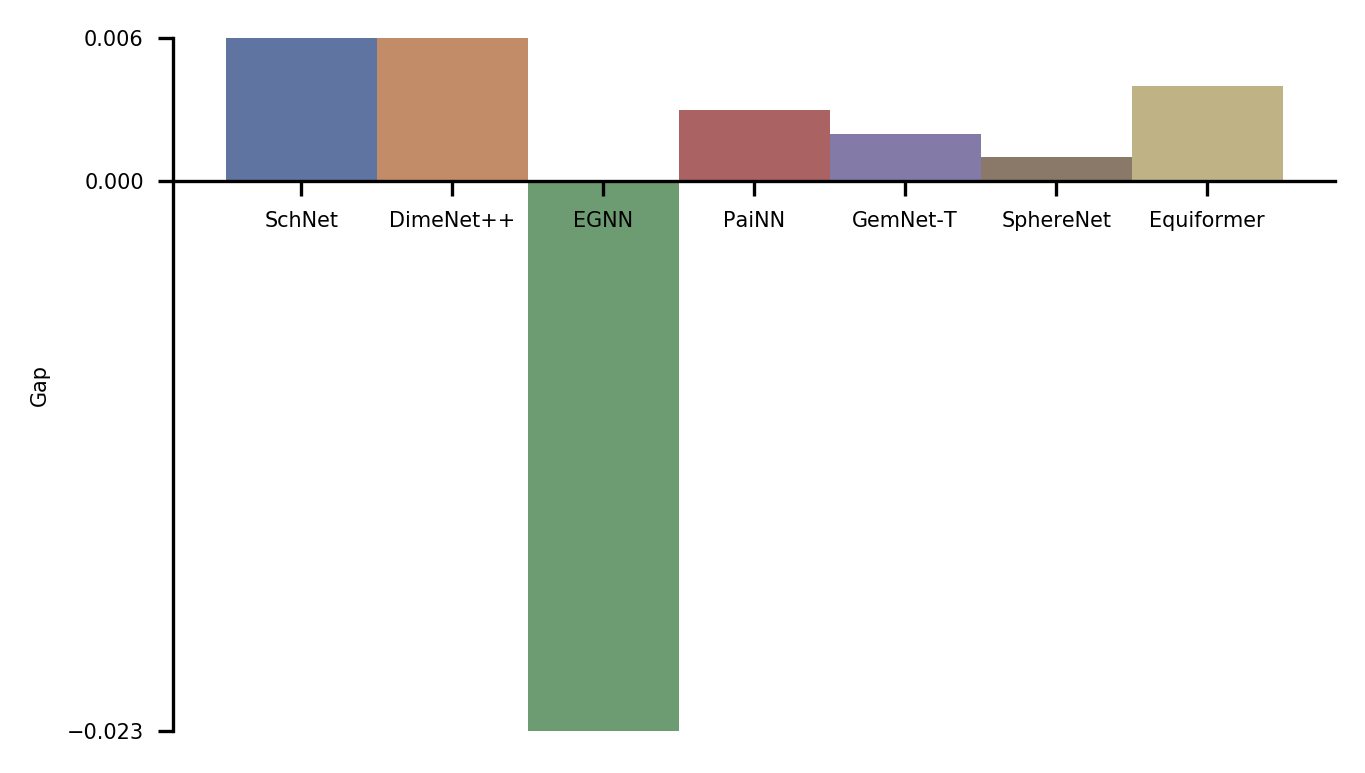}}
    \end{subfigure}
\hfill
    \begin{subfigure}[\small MatBench Task $E_{\text{exfo}}$]
    {\includegraphics[width=0.3\linewidth]{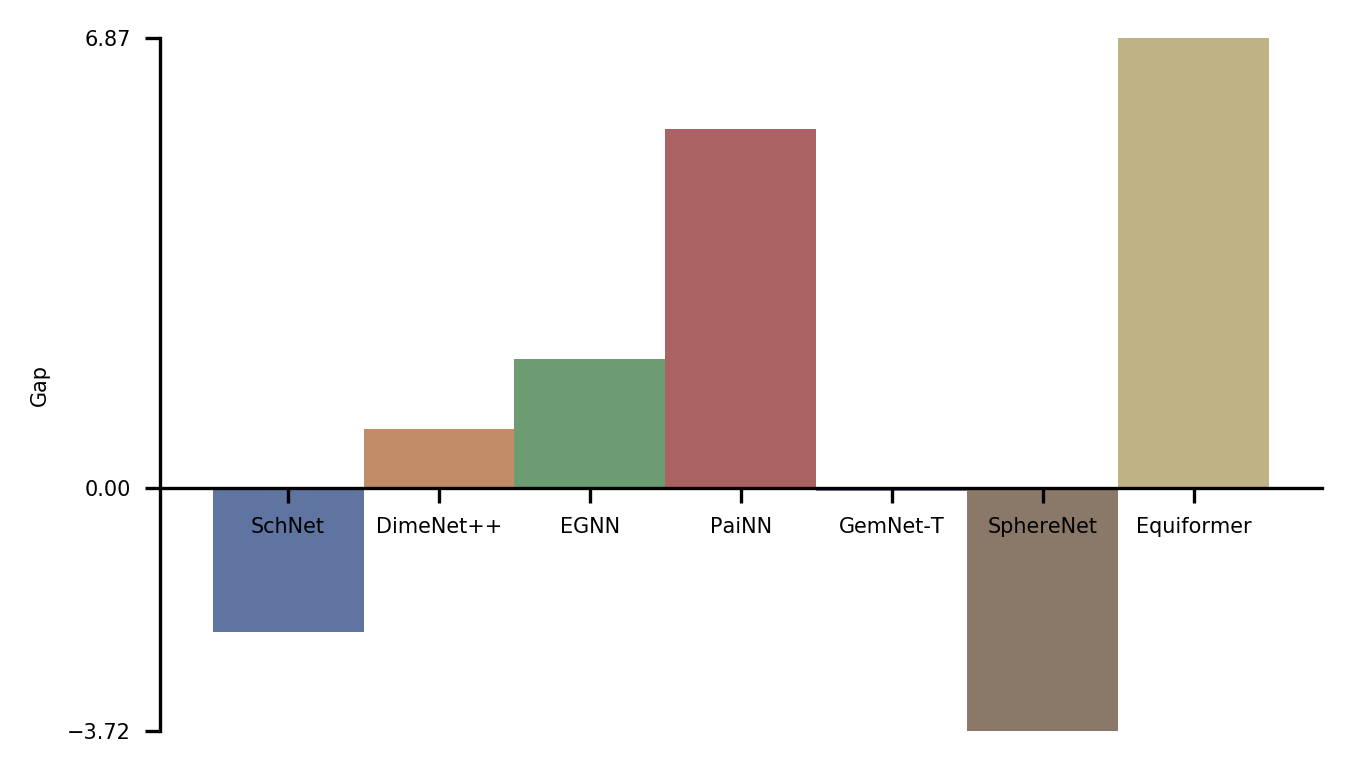}}
    \end{subfigure}
\hfill
    \begin{subfigure}[\small MatBench Task Phonons]
    {\includegraphics[width=0.3\linewidth]{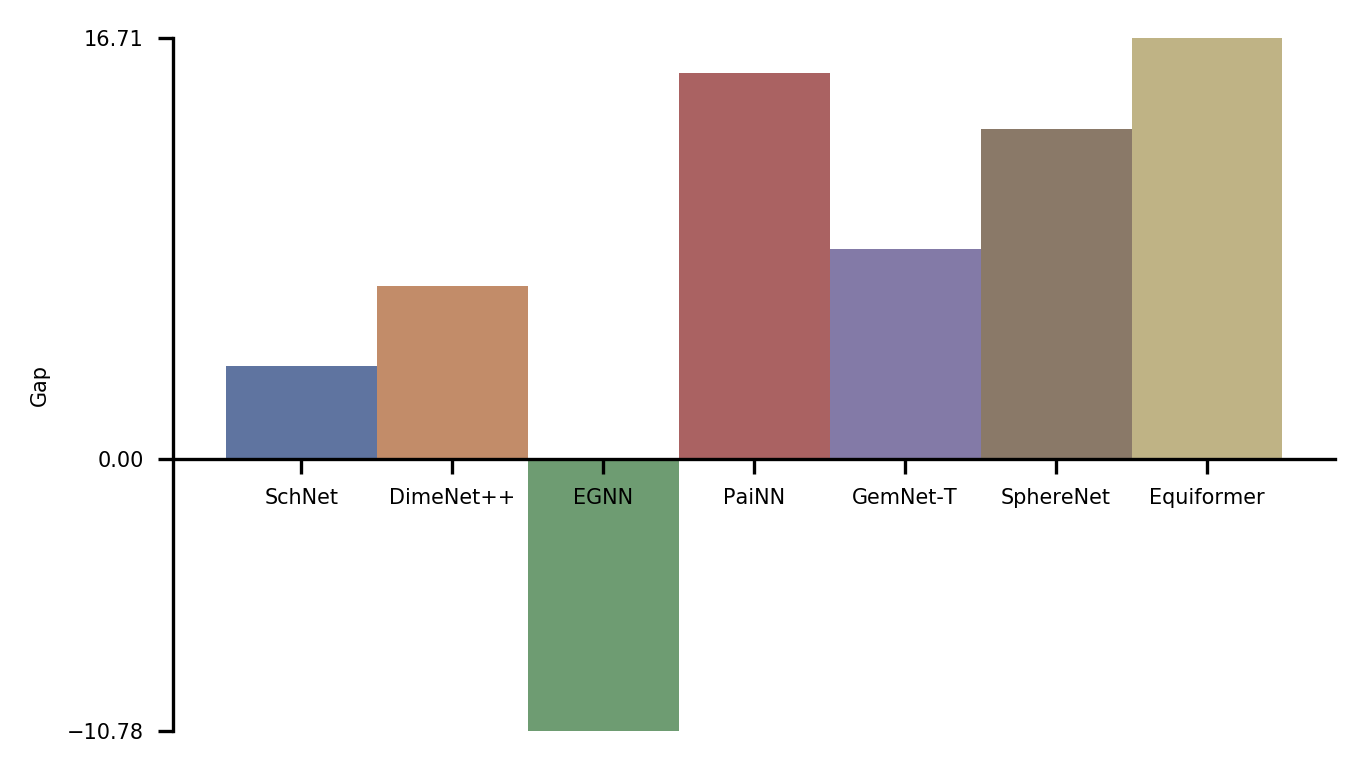}}
    \end{subfigure}
\hfill
    \begin{subfigure}[\small MatBench Task $E_{\text{form}}$]
    {\includegraphics[width=0.3\linewidth]{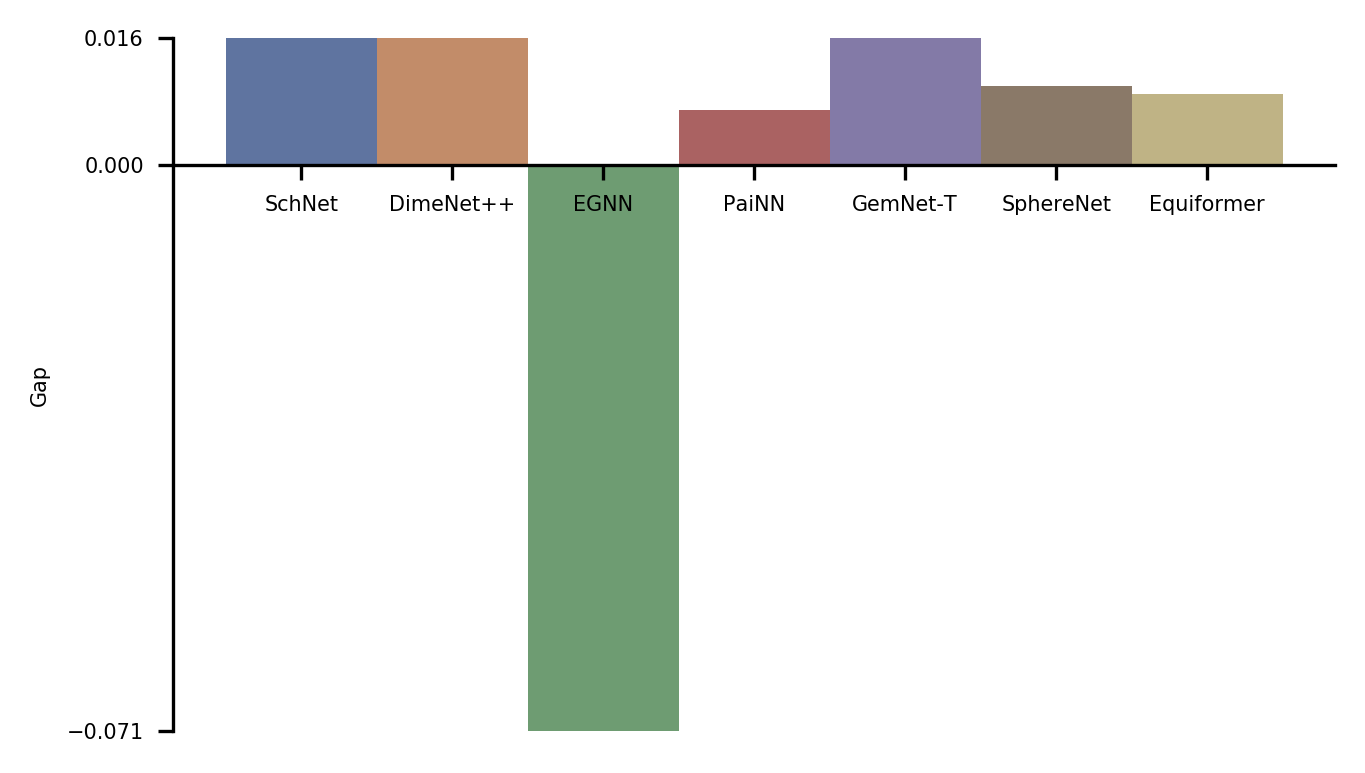}}
    \end{subfigure}
\hfill
    \begin{subfigure}[\small MatBench Task Band Gap]
    {\includegraphics[width=0.3\linewidth]{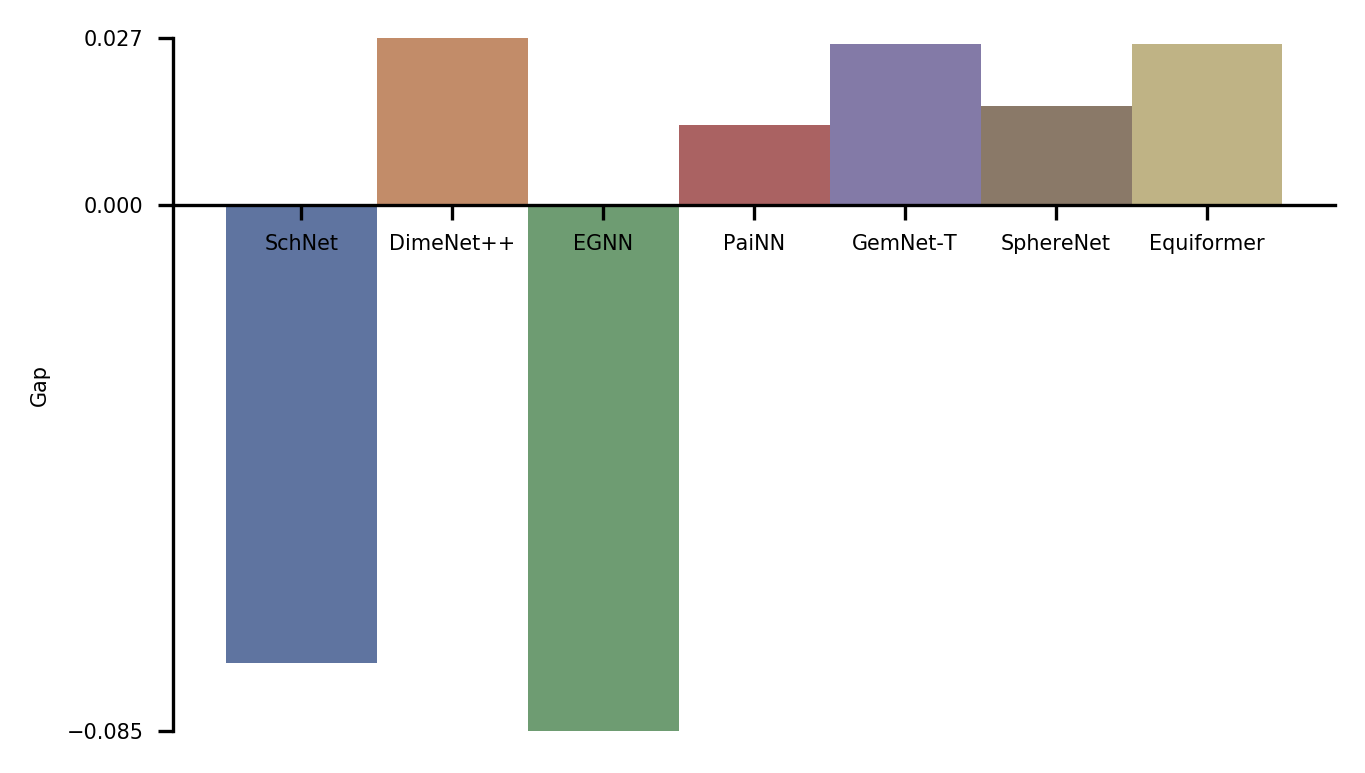}}
    \end{subfigure}
\hfill
    \begin{subfigure}[\small QMOF Task Band Gap]
    {\includegraphics[width=0.3\linewidth]{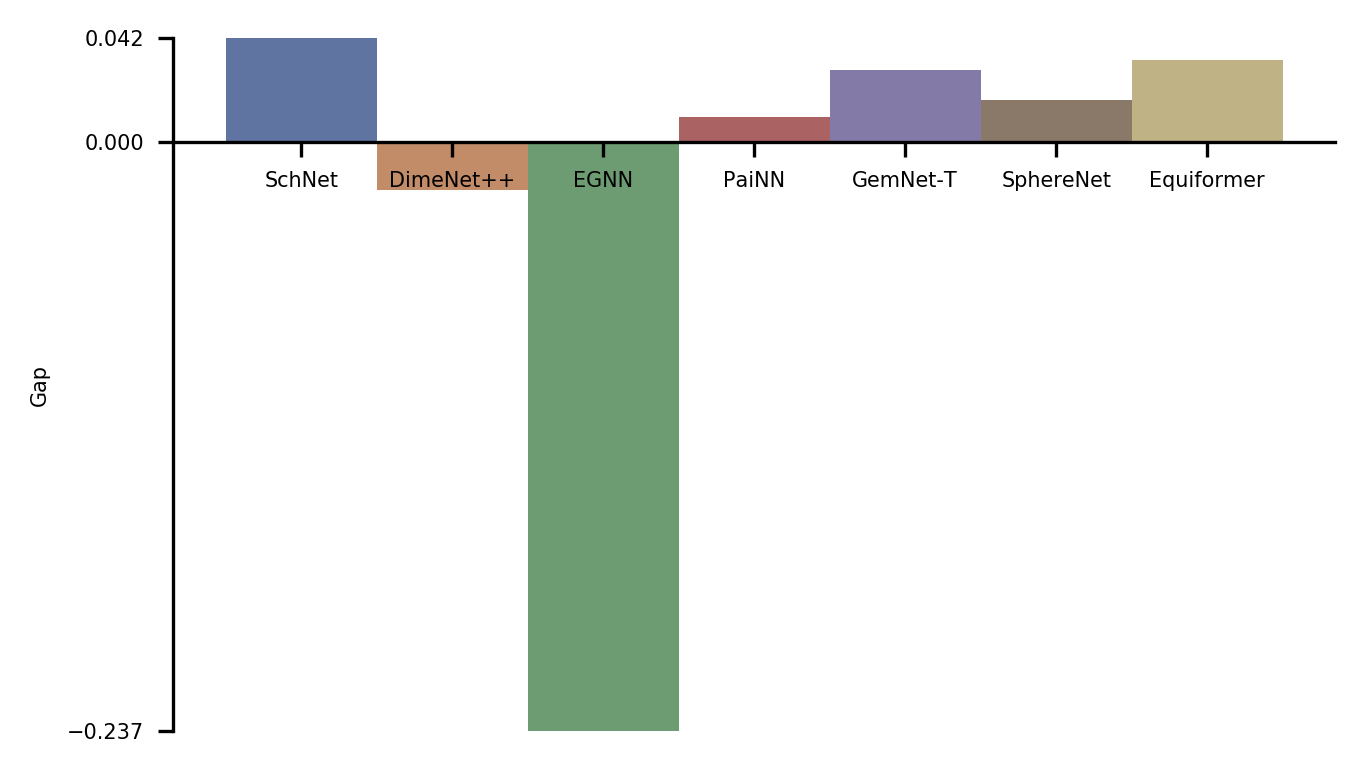}}
    \end{subfigure}
\hfill
\vspace{-2ex}
\caption{
\small
Performance gap of DA: MAE(expanded DA) - MAE(gathered DA), in MatBench and QMOF.
}
\vspace{-2.5ex}
\end{figure}

\clearpage
\newpage
%%%%%%%%%%%%%%%%%%%%%%%%%%%%%%%%%%%%%%%%%%%%%%%%%%
\subsection{An Evidence Example On The Importance of Atom Types and Atom Coordinates} \label{sec:ablation_atom_types_atom_coordinates}

First, it has been widely acknowledged~\cite{engel2018applied} that the atom positions or molecule shapes are important factors to the quantum properties. Here we carry out an evidence example to empirically verify this. The goal here is to make predictions on 12 quantum properties in QM9.

The molecule geometric data includes two main components as input features: the atom types and atom coordinates. Other key information can be inferred accordingly, including the pairwise distances and torsion angles. We consider corruption on each of the component to empirically test their importance accordingly.
\begin{itemize}[noitemsep,topsep=0pt]
    \item Atom type corruption. There are in total 118 types of atom types, and the standard embedding option is to apply the one-hot encoding. In the corruption case, we replace all the atom types with a hold-out index, {\ie}, index 119.
    \item Atom coordinate corruption. Originally QM9 includes atom coordinates that are in the stable state, and now we replace them with the coordinates generated with MMFF~\cite{halgren1996merck} from RDKit~\cite{landrum2013rdkit}.
\end{itemize}

\begin{table}[htb!]
\setlength{\tabcolsep}{5pt}
\fontsize{9}{9}\selectfont
\centering
\caption{\small
An evidence example of molecular data.
The goal is to predict 12 quantum properties (regression tasks) of 3D molecules (with 3D coordinates on each atom).
The evaluation metric is MAE.
}
\vspace{-1.5ex}
\label{tab:app:evidence_example}
\begin{adjustbox}{max width=\textwidth}
\begin{tabular}{l l c c c c c c c c c c c c c c}
\toprule
Model & Mode & $\alpha$ $\downarrow$ & $\nabla \mathcal{E}$ $\downarrow$ & $\mathcal{E}_\text{HOMO}$ $\downarrow$ & $\mathcal{E}_\text{LUMO}$ $\downarrow$ & $\mu$ $\downarrow$ & $C_v$ $\downarrow$ & $G$ $\downarrow$ & $H$ $\downarrow$ & $R^2$ $\downarrow$ & $U$ $\downarrow$ & $U_0$ $\downarrow$ & ZPVE $\downarrow$\\
\midrule
\multirow{3}{*}{SchNet}
& Stable Geometry & 0.070 & 50.59 & 32.53 & 26.33 & 0.029 & 0.032 & 14.68 & 14.85 & 0.122 & 14.70 & 14.44 & 1.698\\
& Type Corruption & 0.074 & 52.07 & 33.64 & 26.75 & 0.032 & 0.032 & 21.68 & 22.93 & 0.231 & 23.01 & 22.99 & 1.677\\
& Coordinate Corruption & 0.265 & 110.59 & 79.92 & 78.59 & 0.422 & 0.113 & 57.07 & 58.92 & 18.649 & 60.71 & 59.32 & 5.151\\

\midrule
\multirow{3}{*}{DimeNet++}
& Stable Geometry & 0.046 & 37.41 & 20.89 & 17.54 & 0.030 & 0.023 & 7.89 & 6.71 & 0.310 & 6.74 & 6.94 & 1.193\\
& Type Corruption & 0.052 & 40.05 & 24.42 & 19.33 & 0.031 & 0.024 & 9.57 & 8.53 & 0.322 & 8.84 & 8.34 & 1.299\\
& Coordinate Corruption & 0.257 & 202.34 & 88.33 & 167.63 & 0.514 & 0.115 & 77.95 & 628.73 & 19.923 & 72.92 & 804.56 & 5.950\\

\midrule
\multirow{3}{*}{SphereNet}
& Stable Geometry & 0.048 & 39.98 & 22.69 & 18.98 & 0.026 & 0.027 & 8.94 & 6.95 & 0.234 & 7.33 & 7.34 & 1.620\\
& Type Corruption & 0.049 & 41.09 & 23.56 & 20.08 & 0.028 & 0.028 & 13.21 & 14.63 & 0.287 & 16.35 & 13.74 & 2.063\\
& Coordinate Corruption & 0.228 & 100.25 & 69.89 & 70.12 & 0.379 & 0.094 & 52.04 & 56.86 & 17.539 & 55.61 & 55.12 & 4.684\\

\midrule
\multirow{3}{*}{PaiNN}
& Stable Geometry & 0.048 & 44.50 & 26.00 & 21.11 & 0.016 & 0.025 & 8.31 & 7.67 & 0.132 & 7.77 & 7.89 & 1.322\\
& Type Corruption & 0.057 & 45.61 & 27.22 & 22.16 & 0.016 & 0.025 & 11.48 & 11.60 & 0.181 & 11.15 & 10.89 & 1.339\\
& Coordinate Corruption & 0.223 & 108.31 & 73.43 & 72.35 & 0.391 & 0.095 & 48.40 & 51.82 & 16.828 & 51.43 & 48.95 & 4.395\\
\bottomrule
\end{tabular}
\end{adjustbox}
% \vspace{-2ex}
\end{table}

We take SchNet and PaiNN as the backbone 3D GNN models, and the results are in~\Cref{tab:app:evidence_example}. We can observe that 
(1) Both corruption examples lead to performance decrease.
(2) The atom coordinate corruption may lead to more severe performance decrease than the atom type corruption.
To put this into another way  is that, when we corrupt the atom types with the same hold-out type, it is equivalently to removing the atom type information. Thus, this can be viewed as using the equilibrium atom coordinates alone, and the property prediction is comparatively robust. This observation can also be supported from the domain perspective. According to the valence bond theory, the atom type information can be implicitly and roughly inferred from the atom coordinates.

Therefore, by combining all the above observations and analysis, one can draw the conclusion that, \textit{for molecule geometry data, the atom coordinates reveal more fundamental information for representation learning}.

%%%%%%%%%%%%%%%%%%%%%%%%%%%%%%%%%%%%%%%%%%%%%%%%%%
\subsection{Ablation on the Effect of Residue Type}
As discussed in~\Cref{sec:data_structure,sec:data_preprocessing}, proteins have four levels of backbone structures. In~\Cref{sec:ablation_atom_types_atom_coordinates}, we carefully check the effect of atom types and atom coordinates in small molecules, and here we would like to check the effect of side residue type in protein geometry-related tasks.

For experiments, we take one of the most recent works, CDConv~\cite{fan2023cdconv}, as the backbone geometric model. The ablation study results are as in~\Cref{tab:ablation_study_residue_type}. We observe that the performance drops on all the tasks, and the performance drops on Sup and Fam are much more significant. This reveals that the effect of residue type may differ for different tasks, yet it is preferred to have them encoded for geometric modeling.

\begin{table}[htb!]
\setlength{\tabcolsep}{5pt}
\fontsize{9}{9}\selectfont
\centering
\caption{\small 
The effect of residue type on the performance of CDConv.
}
\label{tab:ablation_study_residue_type}
\vspace{-1.5ex}
\begin{adjustbox}{max width=\textwidth}
\begin{tabular}{l l c cccc}
\toprule
\multirow{2}{*}{Model} & \multirow{2}{*}{Residue Type} & \multirow{2}{*}{EC} & \multicolumn{4}{c}{Fold}\\
\cmidrule(lr){4-7}
& & & Fold & Sup & Fam & Avg \\
\midrule
CDConv & w/ residue type & 86.887 & 60.028 & 79.904 & 99.528 & 79.820\\
CDConv & w/o residue type & 86.144 & 41.783 & 61.164 & 95.598 & 66.182\\
\bottomrule
\end{tabular}
\end{adjustbox}
\end{table}

\newpage
\section{Resources}

We use a single GPU (V100 or A100) for each task. Note that we try to run all the models with the same epoch numbers, yet some models are too large in terms of computational memory and time, so we have to reduce the computational time. Thus, we list the running time for the main tasks below for readers to check.

In total, it takes over 639 GPU days (without any hyperparameter tuning, random seeds, or ablation studies). It takes around 1370 GPU days if we include ablation studies discussed in~\Cref{sec:app:ablation_studies}.

\begin{table}[h]
\centering
\caption{\small Running time for each (model, task, epoch) per epoch on small molecules and crystal materials. There are eight tasks in MatBench with various dataset sizes, and we take 2 times $E_{form}$ for illustration here.
For NequIP and Allegro, as you can find in \href{https://github.com/chao1224/Geom3D}{the GitHub repository}, we do tune their hyperparameters on QM9, yet not being able to reproduce the results. So we may as well report their numbers here.}
\vspace{-1.5ex}
\begin{adjustbox}{max width=\textwidth}
\begin{tabular}{l l rrrrrrrrr r}
\toprule
Model & & QM9 & MD17 & rMD17 & COLL & LBA & LEP & MatBench & QMOF & Total\\
\midrule
\multirow{2}{*}{SchNet~\cite{schutt2018schnet}}
& epochs & 1,000 & 1,000 & 1,000 & 1000 & 300 & 300 & 1000 & 300 
& \multirow{2}{*}{9.3 days}\\
& time & 36s & 9s & 8s & 46s & 7s & 5s & 77s & 53s\\
\midrule

\multirow{2}{*}{DimeNet++~\cite{klicpera2020fast}}
& epochs & 500 & 800 & 800 & 1000 & 300 & 300 & 300 & 300
& \multirow{2}{*}{53.3 days}\\
& time & 185s & 200s & 200s & 288s & 58s & 52s & 470s & 45s\\
\midrule

\multirow{2}{*}{SE(3)-Trans~\cite{fuchs2020se}}
& epochs & 100 & -- & -- & -- & -- & -- & -- & -- 
& \multirow{2}{*}{24.2 days}\\
& time & 1740s & -- & -- & -- & -- & -- & -- & -- \\
\midrule

\multirow{2}{*}{EGNN~\cite{satorras2021n}}
& epochs & 1000 & 1000 & 1000 & 1000 & 300 & 300 & 800 & 300
& \multirow{2}{*}{22.5 days}\\
& time & 85s & 12s & 12s & 100s & 18s & 14s & 319s & 300s\\
\midrule

\multirow{2}{*}{PaiNN~\cite{schutt2021equivariant}}
& epochs & 1000 & 1000 & 1000 & 1000 & 300 & 300 & 1000 & 300
& \multirow{2}{*}{13.3 days}\\
& time & 46s & 8s & 7s & 61s & 12s & 8s & 176s & 150s\\
\midrule

\multirow{2}{*}{GemNet-T~\cite{klicpera_gemnet_2021}}
& epochs & 1000 & 1000 & 1000 & 1000s & 300 & 300 & 150 & 200
& \multirow{2}{*}{56.8 days}\\
& time & 273s & 52s & 48s & 412s & 75s & 82s & 600s & 480s\\
\midrule

\multirow{2}{*}{SphereNet~\cite{liu2021spherical}}
& epochs & 1000 & 1000 & 1000 & 300 & 300 & 300 & 300 & 300
& \multirow{2}{*}{78.7 days}\\
& time & 250s & 185s & 180s & 418s & 14s & 14s & 480s & 340s\\
\midrule

\multirow{2}{*}{SEGNN~\cite{brandstetter2021geometric}}
& epochs & 500 & 800 & 800 & 100 & 300 & 300 & 40 & 60
& \multirow{2}{*}{81.7 days}\\
& time & 470s & 245s & 234s & 1450s & 370s & 324s & 3500s & 2750s \\
\midrule

\multirow{2}{*}{NequIP~\cite{batzner20223}}
& epochs & 1000 & 1000 & 1000 & -- & -- & -- & -- & -- 
& \multirow{2}{*}{20.5 days}\\
& time & 106s & 29s & 27s & -- & -- & -- & -- & -- \\
\midrule

\multirow{2}{*}{Allegro~\cite{musaelian2022learning}}
& epochs & 1000 & 1000 & 1000 & -- & -- & -- & -- & -- 
& \multirow{2}{*}{22.0 days}\\
& time & 133s & 17s & 17s & -- & -- & -- & -- & -- \\
\midrule

\multirow{2}{*}{Equiformer~\cite{liao2022equiformer}}
& epochs & 300 & 1000 & 1000 & 100 & -- & -- & 100 & 150
& \multirow{2}{*}{57.7 days}\\
& time & 739s & 87s & 126s & 660s  & -- & -- & 1130s & 936s\\
\bottomrule
\end{tabular}
\end{adjustbox}
\end{table}

\begin{table}[h]
\centering
\caption{\small Running time for each (model, task, epoch) per epoch on proteins.}
\vspace{-1.5ex}
\begin{adjustbox}{max width=\textwidth}
\begin{tabular}{l l rrr}
\toprule
Model & & EC & FOLD\\
\midrule

\multirow{2}{*}{GVP-GNN~\cite{JingBowen2020LfPS}}
& epochs & 300 & 400
& \multirow{2}{*}{0.62 days}\\
& time & 150s & 21s \\
\midrule

\multirow{2}{*}{IEConv~\cite{HermosillaPedro2020ICaP}}
& epochs & -- & 200
& \multirow{2}{*}{0.85 days}\\
& time & -- & 368s\\
\midrule

\multirow{2}{*}{GearNet~\cite{ZhangZuobai2022PRLb}}
& epochs & 300 & 400
& \multirow{2}{*}{0.31 days}\\
& time & 61s & 21s\\
\midrule

\multirow{2}{*}{ProNet~\cite{WangLimei2022LHPR}}
& epochs & 400 & 1000
& \multirow{2}{*}{0.50 days}\\
& time & 60s & 19s\\
\midrule

\multirow{2}{*}{CDConv~\cite{fan2023cdconv}}
& epochs & 150 & 400
& \multirow{2}{*}{0.79 days}\\
& time & 175s & 104s \\

\bottomrule
\end{tabular}
\end{adjustbox}
\end{table}
\clearpage

\begin{table}[h]
\centering
\caption{\small Running time for each (pretraining algorithm, dataset, backbone model) per epoch.}
\vspace{-1.5ex}
\begin{adjustbox}{max width=\textwidth}
\begin{tabular}{ll rr r}
\toprule
Dataset & & \makecell[r]{PCQM4Mv2\\(w/ SchNet)} & \makecell[r]{PCQM4Mv2\\(w/ PaiNN)} & Total\\
\midrule

\multirow{2}{*}{Supervised}
& epochs & 100 & 100
& \multirow{2}{*}{13.4 days}\\
& time & 426s & 560s\\
\midrule

\multirow{2}{*}{Type Prediction}
& epochs & 100 & 100
& \multirow{2}{*}{13.5 days}\\
& time & 433s & 572s \\
\midrule

\multirow{2}{*}{Distance Prediction}
& epochs & 100 & 100
& \multirow{2}{*}{13.4 days}\\
& time & 403s & 530s\\
\midrule

\multirow{2}{*}{Angle Prediction}
& epochs & 100 & --
& \multirow{2}{*}{6.4 days}\\
& time & 479s & --\\
\midrule

\multirow{2}{*}{3D InfoGraph~\cite{liu2023molecular}}
& epochs & 100 & 100
& \multirow{2}{*}{13.5 days}\\
& time & 448s & 592s\\
\midrule

\multirow{2}{*}{GraphMVP~\cite{liu2022pretraining}}
& epochs & 100 & 100
& \multirow{2}{*}{14.0 days}\\
& time & 701s & 754s\\
\midrule

\multirow{2}{*}{3D InfoMax~\cite{liu2022pretraining,stark20223d}}
& epochs & 100 & 100
& \multirow{2}{*}{13.5 days}\\
& time & 493s & 584s\\
\midrule

\multirow{2}{*}{GeoSSL-RR~\cite{liu2023molecular}}
& epochs & 100 & 100
& \multirow{2}{*}{14.2 days}\\
& time & 680s & 924s\\
\midrule

\multirow{2}{*}{GeoSSL-EBM-NCE~\cite{liu2023molecular}}
& epochs & 100 & 100
& \multirow{2}{*}{14.2 days}\\
& time & 630s & 980s\\
\midrule

\multirow{2}{*}{GeoSSL-InfoNCE~\cite{liu2023molecular}}
& epochs & 100 & 100
& \multirow{2}{*}{14.1 days}\\
& time & 598s & 952s\\
\midrule

\multirow{2}{*}{GeoSSL-DDM~\cite{liu2023molecular}}
& epochs & 100 & 100
& \multirow{2}{*}{15.0 days}\\
& time & 1100s & 1200s\\
\midrule

\multirow{2}{*}{GeoSSL-DDM-1L~\cite{zaidi2022pre}}
& epochs & 100 & 100
& \multirow{2}{*}{14.4 days}\\
& time & 780s & 1010s\\
\midrule

\multirow{2}{*}{3D-EMGP~\cite{jiao20223d}}
& epochs & -- & 100
& \multirow{2}{*}{7.6 days}\\
& time & -- & 980s\\
\midrule

\multirow{2}{*}{MoleculeSDE-VE~\cite{liu2023moleculeSDE}}
& epochs & 50 & 50
& \multirow{2}{*}{14.5 days}\\
& time & 1906s & 1933s\\
\midrule

\multirow{2}{*}{MoleculeSDE-VP~\cite{liu2023moleculeSDE}}
& epochs & 50 & 50
& \multirow{2}{*}{14.5 days}\\
& time & 1906s & 1933s\\

\bottomrule
\end{tabular}
\end{adjustbox}
\end{table}

We would also like to acknowledge the following nice implementations and tutorials of geometric models:
\begin{itemize}[noitemsep,topsep=0pt]
    \item e3nn: Euclidean Neural Networks, by Tess \cite{geiger2022e3nn}
    \item TFN~\cite{thomas2018tensor}
    \item MaterialProject~\cite{jain2013commentary} and MatBench~\cite{DunnAlexander2020Bmpp}
    \item Keras Graph Convolution Neural Networks (KGCNN)~\cite{REISER2021100095}
    \item DIG~\cite{REISER2021100095}
    \item TorchDrug~\cite{zhu2022torchdrug}
\end{itemize}

\end{document}